\documentclass[lettersize,journal]{IEEEtran}
\usepackage{amsmath,amsfonts}
\usepackage{algorithmic}
\usepackage{algorithm}
\usepackage{array}
\usepackage[caption=false,font=normalsize,labelfont=sf,textfont=sf]{subfig}
\usepackage{textcomp}
\usepackage{stfloats}
\usepackage{url}
\usepackage{verbatim}
\usepackage{graphicx}
\usepackage{cite}
\usepackage{subcaption}
\usepackage{caption}
\usepackage{hyperref} 
\usepackage{booktabs}
\usepackage{multirow}
\usepackage{amssymb}
\usepackage{xcolor} 
\usepackage{colortbl}
\usepackage{float} 
\definecolor{lightgrayopacity}{gray}{0.9}
\usepackage{lipsum}  
\usepackage{rotating}
\usepackage{parskip} 
\usepackage{url}
\begin{document}

\title{Motion Segmentation for Neuromorphic Aerial Surveillance}

\author{Sami Arja, Alexandre Marcireau, Saeed Afshar, Bharath Ramesh, Gregory Cohen %
\thanks{S. Arja, A. Marcireau, S. Afshar, B. Ramesh, G. Cohen are with the International Center for Neuromorphic Systems, Western Sydney University. Email: s.elarja@westernsydney.edu.au}%
\thanks{Manuscript received April 19, 2021; revised August 16, 2021.}}

\maketitle

\begin{abstract}

Aerial surveillance demands rapid and precise detection of moving objects in dynamic environments. Event cameras, which draw inspiration from biological vision systems, present a promising alternative to frame-based sensors due to their exceptional temporal resolution, superior dynamic range, and minimal power requirements. Unlike traditional frame-based sensors that capture redundant information at fixed intervals, event cameras asynchronously record pixel-level brightness changes, providing a continuous and efficient data stream ideal for fast motion segmentation. While these sensors are ideal for fast motion segmentation, existing event-based motion segmentation methods often suffer from limitations such as the need for per-scene parameter tuning or reliance on manual labelling, hindering their scalability and practical deployment. In this paper, we address these challenges by introducing a novel motion segmentation method that leverages self-supervised vision transformers on both event data and optical flow information. Our approach eliminates the need for human annotations and reduces dependency on scene-specific parameters. In this paper, we used the EVK4-HD Prophesee event camera onboard a highly dynamic aerial platform in urban settings. We conduct extensive evaluations of our framework across multiple datasets, demonstrating state-of-the-art performance compared to existing benchmarks. Our method can effectively handle various types of motion and an arbitrary number of moving objects. Code and dataset are available at: \url{https://samiarja.github.io/evairborne/}

\end{abstract}

\begin{IEEEkeywords}
Event Camera, Motion Segmentation, Saliency Detection, Motion Compensation, Aerial Surveillance.
\end{IEEEkeywords}

\section{Introduction}
\IEEEPARstart{A}{erial} platforms such as drones, helicopters, and aircrafts equipped with video cameras provide a versatile and effective means of enhancing safety and security in both civilian and military contexts, offering high mobility and a large surveillance scope~\cite{nguyen_state_2022}. However, the task of interpreting data from aerial surveillance videos presents significant challenges for human operators, primarily due to the overwhelming volume of visual information, which can lead to fatigue and decreased effectiveness \cite{helldin2012automation,erlandsson2011modeling,erlandsson2012calculating,erlandsson2013comparing}. To minimize the workload for operators or analysts, this paper emphasizes the need for an automated scene understanding technique, with motion segmentation being identified as the key task to achieve this goal.

In scene understanding~\cite{aarthi2017scene}, accurately detecting and estimating the motion of objects is essential. Yet, challenges such as altitude changes, platform instability, motion parallax, camera movement, lighting variations, and dense ground structures can hinder reliable detection and tracking of key targets or activities~\cite{nguyen_state_2022}.

\begin{figure}[H]
  \centering
  \begin{minipage}[b]{0.49\textwidth}
    \centering
    \hspace*{-0.25cm}\includegraphics[width=\textwidth]{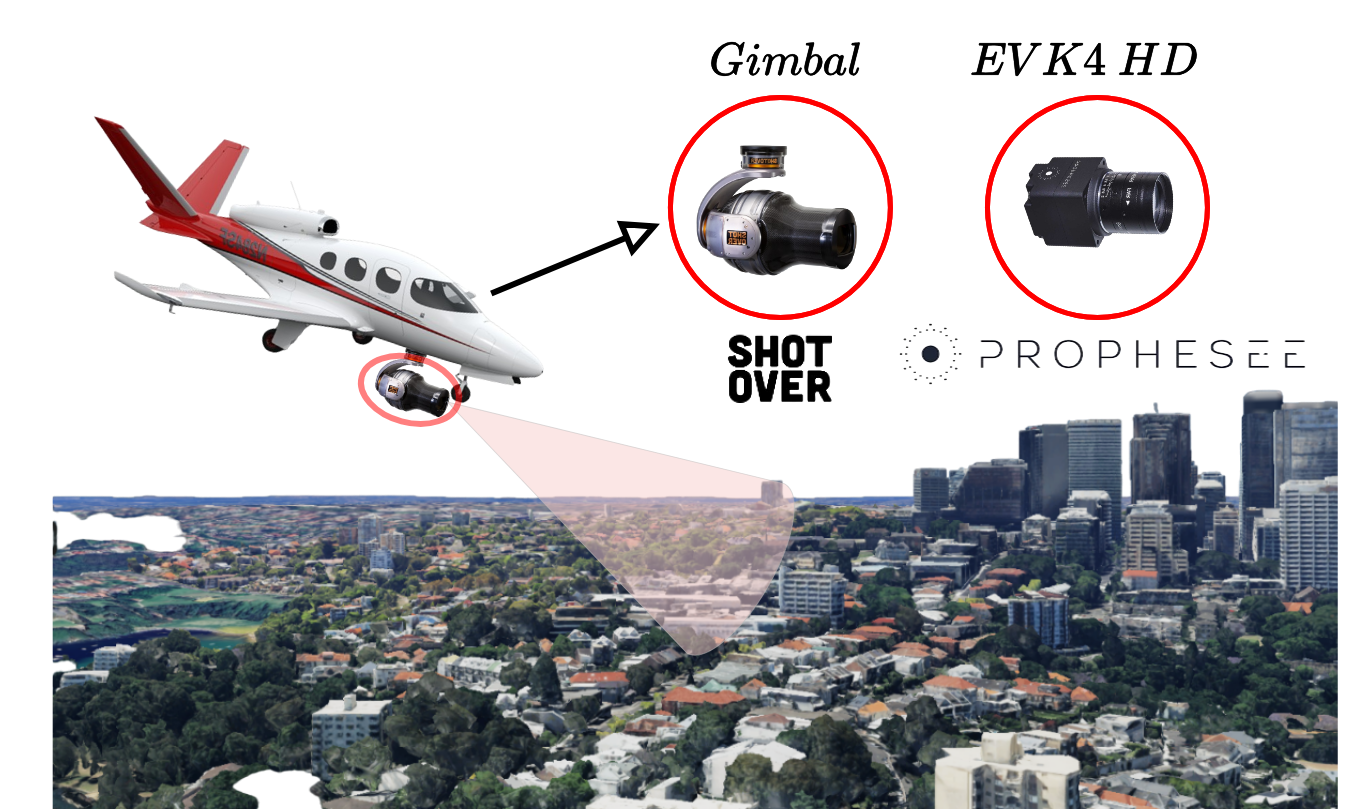}
    \vspace{-8pt}
    \caption*{(a)}
  \end{minipage}

  \begin{minipage}[b]{0.47\textwidth}
    \centering
    \begin{minipage}[b]{0.32\textwidth}
      \centering
      Input events\\
      \includegraphics[width=\textwidth]{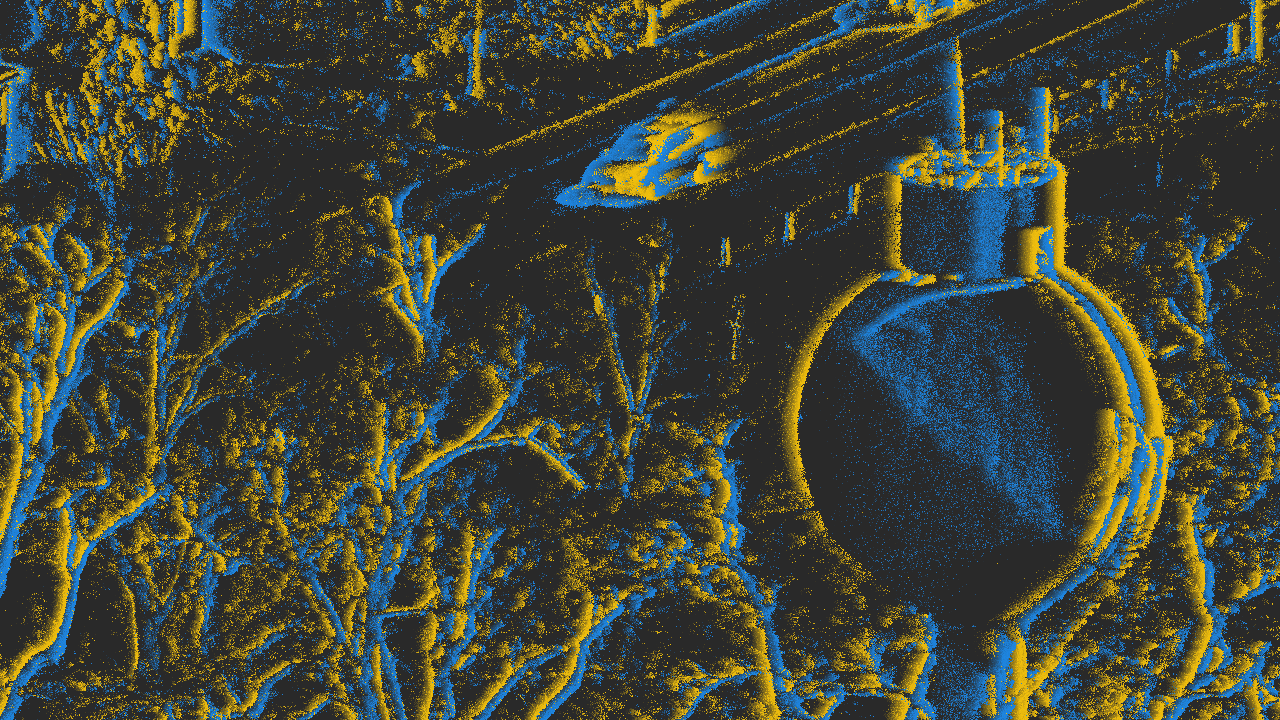}
      \vspace{-15pt}
    \end{minipage}
    \hfill
    \begin{minipage}[b]{0.32\textwidth}
      \centering
      EMSGC \cite{zhou_event-based_2021}\\
      \includegraphics[width=\textwidth]{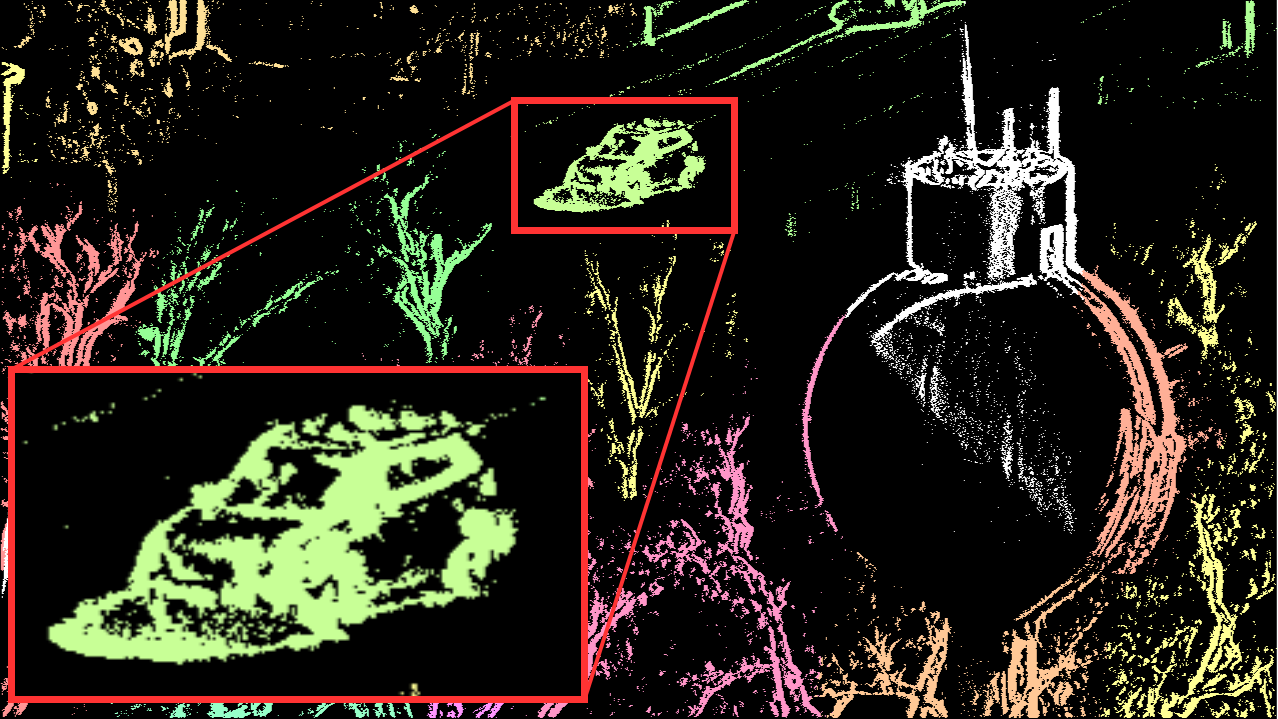}
      \vspace{-15pt}
    \end{minipage}
    \hfill
    \begin{minipage}[b]{0.32\textwidth}
      \centering
      \textbf{Ours}\\
      \includegraphics[width=\textwidth]{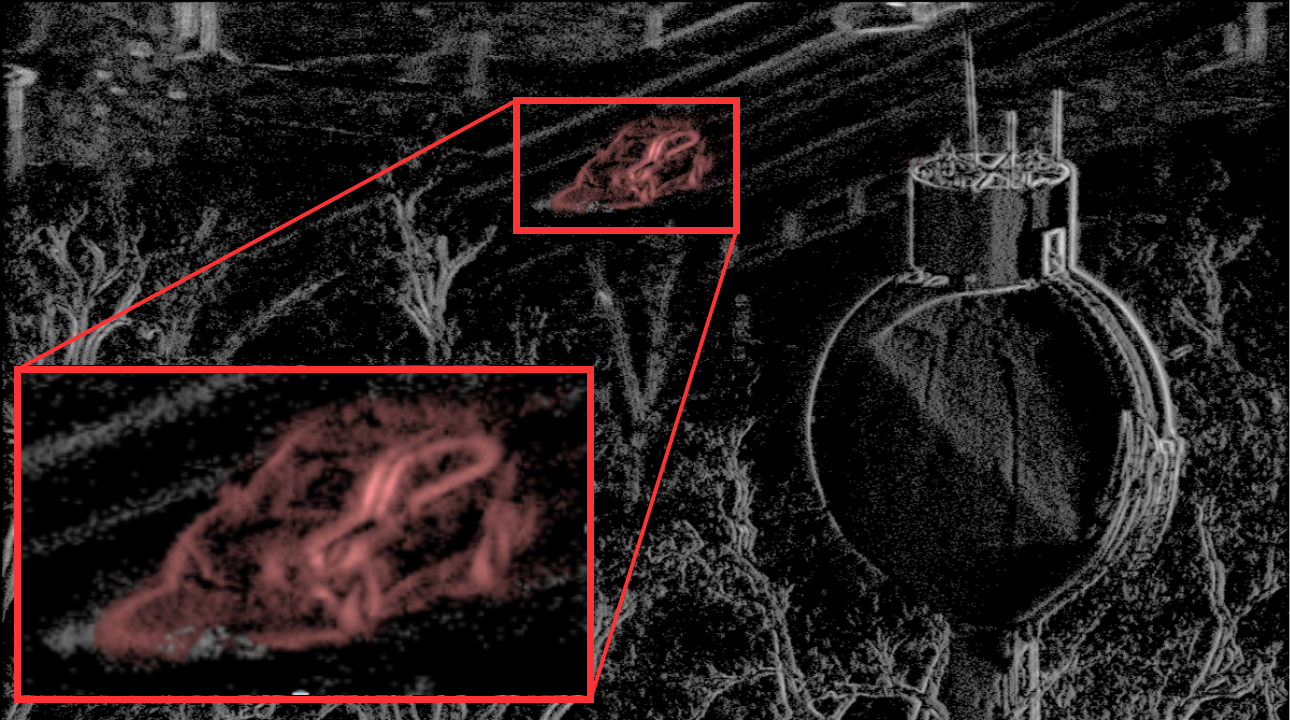}
      \vspace{-15pt}
    \end{minipage}

    \vspace{8pt} %
    \begin{minipage}[b]{0.32\textwidth}
      \centering
      \includegraphics[width=\textwidth]{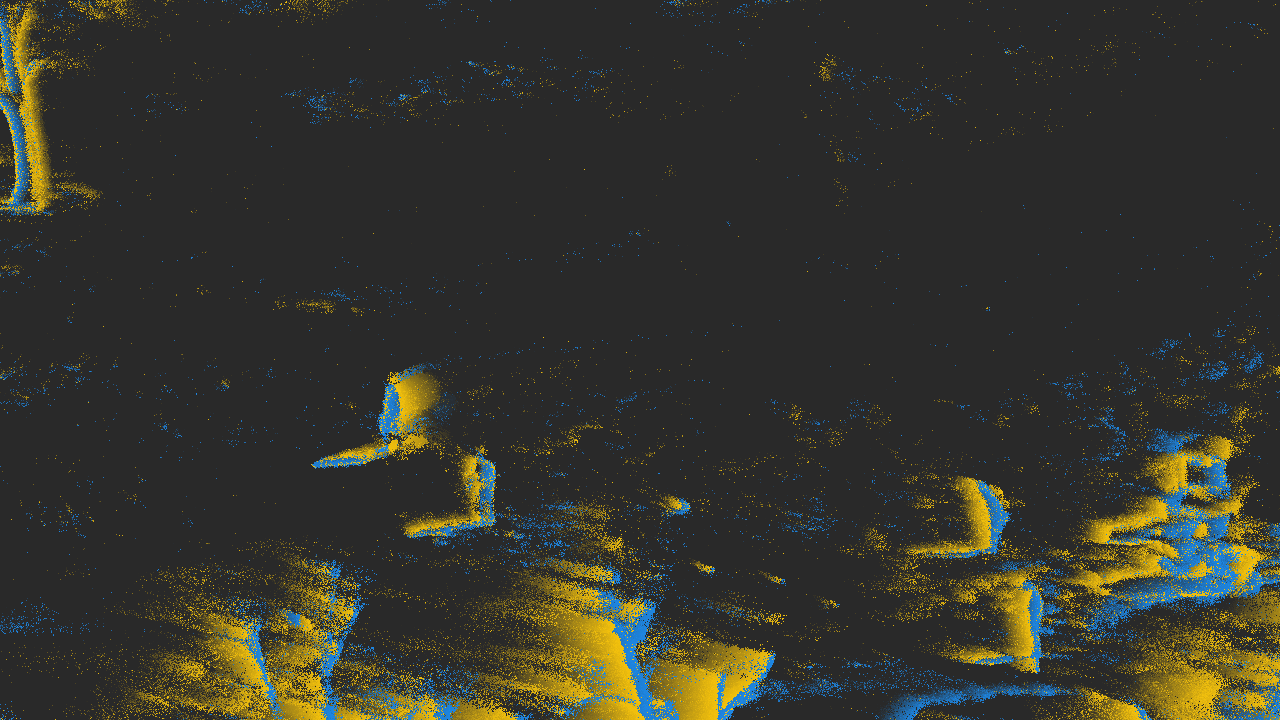}
      \vspace{-18pt}
      \caption*{(b)}
    \end{minipage}
    \hfill
    \begin{minipage}[b]{0.32\textwidth}
      \centering
      \includegraphics[width=\textwidth]{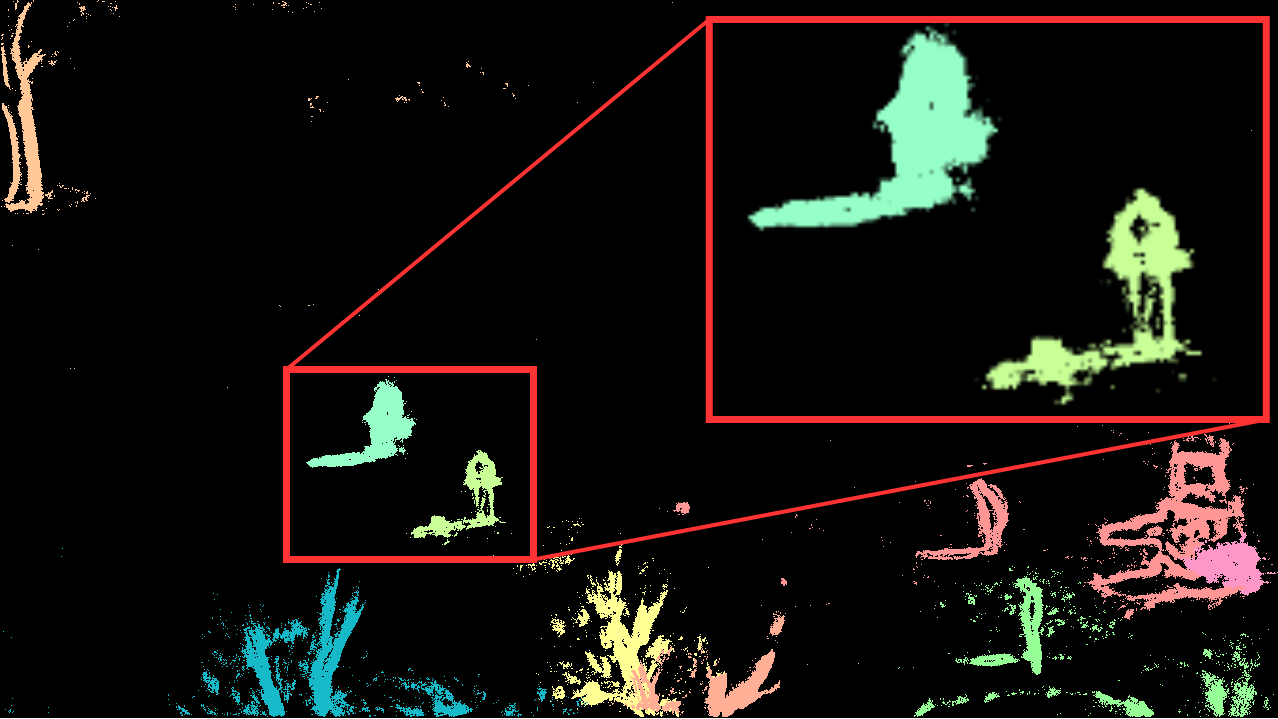}
      \vspace{-18pt}
      \caption*{(c)}
    \end{minipage}
    \hfill
    \begin{minipage}[b]{0.32\textwidth}
      \centering
      \includegraphics[width=\textwidth]{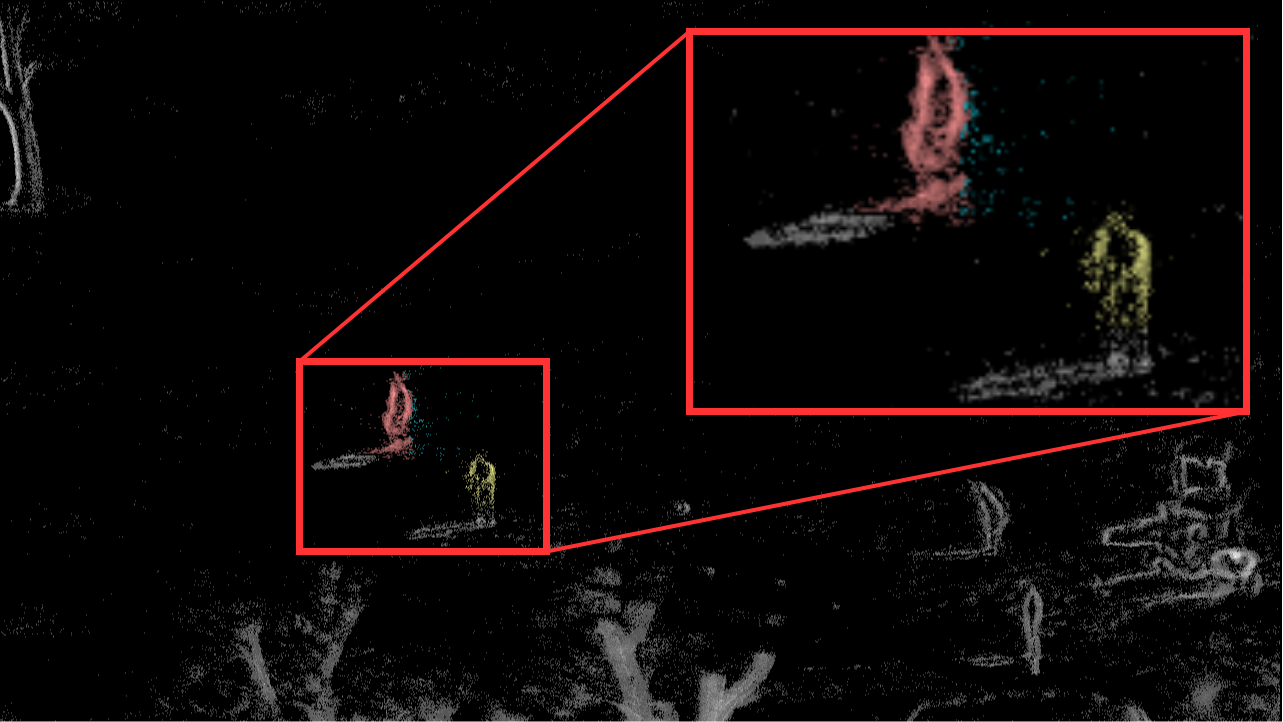}
      \vspace{-18pt}
      \caption*{(d)}
    \end{minipage}
  \end{minipage}

  \vspace{-6pt}
  \caption{(a) An overview of the neuromorphic aerial surveillance setup using a stabilized gimbal from \href{https://shotover.com/products/m1}{Shotover} on a small airplane and the EVK4-HD event camera from \href{https://www.prophesee.ai/event-camera-evk4}{Prophesee}. (b) Event data of a car in motion (top row), and pedestrians (bottom row). (c) and (d) Comparative motion segmentation results from the existing EMSGC~\cite{zhou_event-based_2021} and our method, with a zoom-in window to show the most salient objects. Our method creates sharper edges for moving objects and effectively compensates for background motion. \textit{It avoids over-segmentation, provides segmentation outputs that are consistent over time, accurately describes the number of moving objects, and produces per-pixel continuous and discrete labels per moving object.}}
  \label{fig:solution}
\end{figure}

Since aerial platforms are almost always moving, aerial cameras observe two types of motion: \textit{actual} motion from objects moving on or over the ground and \textit{ego} motion. Telling the two apart is required for scene analysis, hence it is critical to detect and understand motion cues. This paper primarily focuses on identifying local movement within a scene, regardless of the object's shape and appearance based only on their motion cues.

Aerial surveillance systems using frame-based sensors like RGB cameras or LiDARs suffer from bandwidth-latency tradeoffs, often resulting in significant delays. Their limited dynamic range reduces effectiveness in diverse lighting, and they are susceptible to motion blur and exposure issues in fast-changing scenes. Neuromorphic event cameras \cite{lichtsteiner_128times128_2008,Finateu2020510A1} overcome such limitations by asynchronously capturing changes in pixel intensity, which enables them to acquire visual information at the same rate as the scene dynamics with very high temporal resolution and a high dynamic range. Additionally, they only consume power in response to motion and only for pixels that undergo changes. A comprehensive review of event cameras, including their algorithms and applications, is provided in the recent survey~\cite{gallego_event-based_2022}.

This paper introduces an event-based motion segmentation algorithm for aerial surveillance, employing a high-resolution Prophesee Gen4 HD event camera onboard a small airplane as shown in Fig.~\ref{fig:solution}. The process is split into two primary phases. Initially, we identify the most salient objects from events within a specific time frame using an unsupervised approach. Subsequently, we estimate the local motion of each identified object by integrating the Contrast Maximization (CMax) \cite{gallego_unifying_2018} framework combined with blur detection algorithm \cite{golestaneh2017spatially}.

In summary, the main contributions of this paper are as follows:
\begin{itemize}
    \item An unsupervised method for discovering and segmenting moving objects using motion cues from an aerial platform, eliminating the need for manual annotations.
    \item A dynamic mask refinement process~\cite{random_walk_neurips2020,xie2022segmenting} that integrates appearance information from a self-supervised DINO~\cite{caron_emerging_2021} model to improve the accuracy of the saliency masks.
    \item The introduction of the Ev-Airborne dataset, providing high-resolution data from an aerial platform, complete with ground truth annotations.
    \item Superior segmentation performance on major benchmarks, showcasing strong generalization capabilities compared to existing event-based motion segmentation methods.
\end{itemize}

\section{Related Work}

\subsection{Event-based Motion Segmentation}

Given a group of events during a time interval, motion segmentation algorithms seek to locate and group events that are triggered by the same objects based only on their motion cues. This field has witnessed significant advancements recently, with several new datasets~\cite{EVIMO2,mitrokhin_ev-imo_2019,mitrokhin_event-based_2018,almatrafi_distance_2020,zhou_event-based_2021} playing a crucial role in enabling the development of more sophisticated algorithms. Some researchers \cite{Stoffregen19iccv,zhou_event-based_2021,parameshwara_0-mms_2021,lu_event-based_2021,chen_progressivemotionseg_2022,Stoffregen_segmentation_acra17,liu_motion_2024} formulated motion segmentation as a joint optimisation problem by leveraging the idea of motion compensation framework~\cite{gallego_unifying_2018}. Other researchers \cite{mitrokhin_ev-imo_2019,sanket_evdodgenet_2020,mitrokhin_learning_2020,parameshwara_spikems_2021,wang_-evmoseg_2023,zhang_multi-scale_2023,jiang_event-based_2024,georgoulis_out_2024,alkendi2024neuromorphic} formulated it as a detection and/or tracking problem leveraging the recent advancements in deep learning. Most recently,~\cite {wang_-evmoseg_2023} demonstrated an unsupervised technique for segmenting the motion of Independent Moving Objects (IMOs) without the need for human annotations.

\subsection{Vision Transformers}

Attention-based transformer models \cite{NIPS2017_Attention} have become the standard architecture in natural language processing (NLP) and have recently been extended to visual tasks \cite{dosovitskiy2020vit}. Vision Transformers (ViT) \cite{dosovitskiy2020vit} is almost identical to the original Transformers, where ViT tokenize non-overlapping patches of the input image and then feeds them into the transformer encoders. Recently, ViT has aroused extensive interest in computer vision due to their competitive performance compared with CNNs \cite{pmlr-v139-touvron21a,liu2021Swin,chu2021Twins,arnab_vivit_2021,gberta_2021_ICML,tong2022videomae} and has been extended to many challenging tasks including object detection \cite{dai_dynamic_2021,zhu2021deformable,10.1007/978-3-030-58452-8_13}, segmentation \cite{cheng2021perpixel}, image enhancement \cite{strudel_segmenter_2021,chen_pre-trained_2021} and video understanding \cite{liu2022video,arnab_vivit_2021}. In event-based vision, transformers components have been used in classification \cite{Sabater_2022_CVPR,Wang2022ExploitingSS}, image reconstruction \cite{weng_event-based_2021}, monocular depth estimation \cite{liu2022event}, and object detection \cite{Gehrig_2023_CVPR}. Many variants of ViT with self-supervised learning have been proposed like DINO \cite{caron_emerging_2021} which can automatically segment the background pixels of an image, even though they were never trained using pixel-level supervision, MoCo-V3 \cite{chen_empirical_2021} which demonstrates that using contrastive learning can achieve strong results, and DeiT \cite{pmlr-v139-touvron21a} which enable ViT to be trained on smaller ImageNet-1k datasets.

\subsection{Unsupervised Saliency Detection}

The task aims at identifying the visually interesting objects in an image which involves detecting the most salient object and segmenting the accurate region of that object. Prior to the deep learning revolution, saliency methods mainly relied on different priors and handcrafted features \cite{zhu2014saliency,cheng2013efficient,cheng2014global,goferman2011context}. More recently, unsupervised deep models \cite{nguyen2019deepusps,zhang2018deep,Yasarla_2024_WACV}, propose to learn saliency maps with noisy pseudo-labels from multiple noisy unsupervised saliency methods. \cite{wang_tokencut_2022} proposed a novel method to detect and segment salient objects in images and videos using features obtained by a self-supervised transformer, where the image patches are organised into a fully connected graph. In event-based vision, data is naturally captured in the form of edges and contours, removing the need for some early stages of processing. \cite{afshar_event-based_2020,ralph_real-time_2022} developed a method for learning key features in an unsupervised, event-per-event and online manner without relying on an intermediate frame representation. However, in moving-camera scenarios like aerial surveillance, every edge becomes salient, making it difficult to distinguish between the most relevant features in an event-per-event manner. In this paper, we demonstrate that "TokenCut"~\cite{wang_tokencut_2022} can be applied directly to event data as an initial processing step, without the need for event labels or extra training.

Our method, similar to \cite{wang_-evmoseg_2023}, eliminates the need for expensive annotated training data and performs per-event segmentation in an unsupervised manner like \cite{zhou_event-based_2021,Stoffregen19iccv,lu_event-based_2021}. However, our method differs significantly by addressing key limitations: (1) by formulating the problem as saliency detection followed by per-object motion compensation, producing sharp motion-compensated images, (2) unlike \cite{Stoffregen19iccv}, we do not cluster events, which can be ineffective in noisy and highly dense scenes, (3) our method uses only events as input, without needing additional information like depth \cite{wang_-evmoseg_2023}. Additionally, we avoid dividing the accumulated image into uniform subvolumes, thus preventing motion estimation in low-activity or empty regions as in \cite{zhou_event-based_2021}.

\section{Method}

\begin{figure*}[h] %
  \centering
  \includegraphics[width=0.75\textwidth]{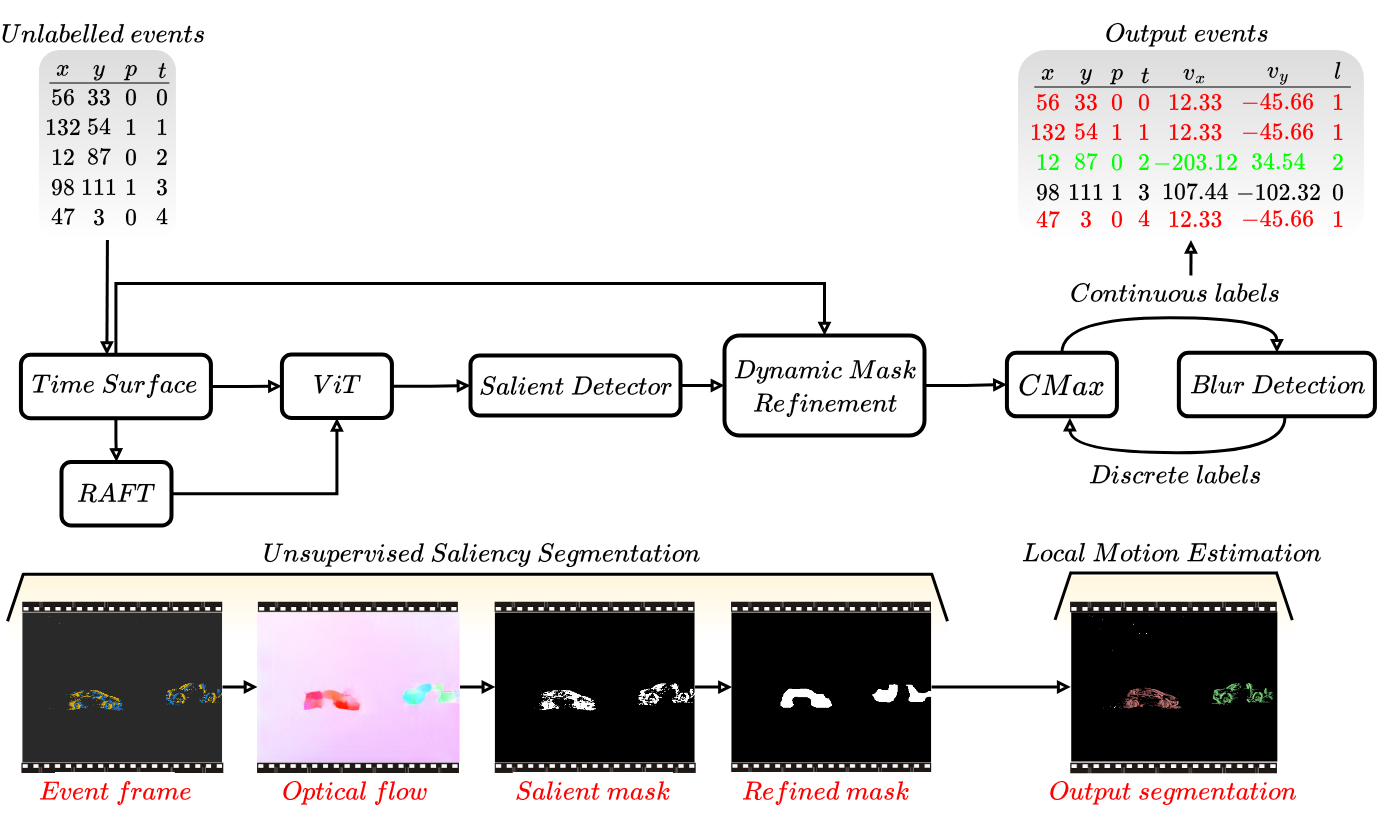}
  \caption{Overview of the event-based motion segmentation architecture. Our method performs a pixel-wise motion segmentation in two consecutive stages. Firstly, the incoming event stream is transformed into a frame, which, along with the generated optical flow from RAFT \cite{teed2020raft}, feeds into a self-supervised DINO \cite{caron_emerging_2021}. DINO extracts a series of feature vectors, after which a dynamic refinement strategy fine-tunes the mask predictions. Secondly, events within the predicted mask are isolated, using CMax algorithm \cite{gallego_unifying_2018} combined with a blur detection \cite{golestaneh2017spatially} to assign both continuous motion label and a discrete label to each event.}
   \label{fig:networkarchitecture}
\end{figure*}

Our method for event-based motion segmentation built upon recent work from \cite{wang_tokencut_2022,caron_emerging_2021,gallego_unifying_2018}, isolates moving objects from the background and then estimates the motion of each object independently. This technique is crucial for aerial surveillance, as it offers two types of information: where are the most salient objects? and how are they moving?

First, a coarse binary mask for every moving object is generated using self-supervised DINO \cite{caron_emerging_2021}, the coarse masks are then subsequently refined to fully cover the object using the dynamic mask refinement technique~\cite{random_walk_neurips2020,xie2022segmenting}, this refinement is crucial for correcting issues such as irregular coverage, poor temporal consistency, and size variations found in the initial masks. Second, we apply CMax \cite{gallego_unifying_2018} on the events within the mask to estimate the local motion of the salient objects and recover their sharp edges, and a blur detection technique \cite{golestaneh2017spatially} to identify and mask out the blurry part of the image and assign discrete labels for each object, this eliminates the need of empirical thresholding for selecting sharp edges. 

To break it down simply, the algorithm's input is an event, represented as $e_{i}=(\boldsymbol{u_{i}},p_{i},t_{i})$, where $i$ is the index of the event. Here, $\boldsymbol{u_{i}}=(x_{i},y_{i})$ denotes the pixel coordinates of the event, $p_{i
}$ $\in$ \{0, 1\} indicates the polarity of the contrast change, and $t_{i}$ is the timestamp of this change, measured in $\mu s$. The output is $e_{i}=(\boldsymbol{u_{i}},p_{i},t_{i},\boldsymbol{\mathcal{M}}_i,l_{i})$ where $\boldsymbol{\mathcal{M}}_i$ is a continuous motion label for each event, describing the nature of the motion (be it translation, rotation, or scaling), and $l_{i}$ is a per-event discrete label for each IMO. Fig.~\ref{fig:networkarchitecture} illustrates every step of the event-based motion segmentation algorithm.

\subsection{Event Processing}

Event cameras feature pixel-level circuitry that asynchronously captures events triggered by changes in logarithmic brightness. Given the impracticality of processing millions of such events per second individually in aerial surveillance on standard computing units, a preprocessing method is employed that groups events while preserving temporal details, facilitating integration with the ViT architecture. In this work, we process the events using linear decaying kernels to generate a 2D time-based representation~\cite{afshar_investigation_2019}. To do this, we define two distinct functions: $T_{i} \in \mathbb{R}^{H\times W}$ which maps a pixel position to a timestamp, expressed as $\boldsymbol{u} : t \mapsto T_i(\boldsymbol{u})$, and $P_{i} \in \mathbb{R}^{H\times W}$
which represents the polarity of the most recent event generate by pixel $u$, expressed as $\boldsymbol{u} : p \mapsto P_i(\boldsymbol{u})$. The process described in Eq.~\ref{eq:timesurface} encapsulates both the temporal dynamics and polarity information at each point in the observed space.

\begin{equation}
    I_i(\boldsymbol{u}, t) = P_i(\boldsymbol{u}) \cdot \left(1 + \frac{T_i(\boldsymbol{u}) - t}{2\tau_e}\right) \label{eq:timesurface}
\end{equation}

where $\tau_e$ is the time constant parameter. In this work, we generate a frame at time intervals ranging from 10 to 50 $ms$ which leads to a video sequence $v=$\{$I_1,...,I_N$\}, $v \in \mathbb{R}^{N\times H\times W \times 3}$ with $N$ frames for any given event sequence.

\subsection{Unsupervised Saliency Segmentation}
\label{sec:eventcamproc}

Given video $v$, the model predicts the object mask as $mask=$\{$M_1,...,M_N$\}, $mask \in \mathbb{R}^{N\times H\times W \times 1}$ following the work in~\cite{wang_tokencut_2022}. The process begins with the construction of the optical flow with RAFT~\cite{teed2020raft} (trained on synthetic data~\cite{butler2012naturalistic,geiger_vision_2013}), and then extracting feature vectors for each image and its optical flow using a self-supervised ViT model, specifically ViT-S16, with a self-distillation loss DINO ~\cite{caron_emerging_2021}. This step divides the image into non-overlapping patches of size $P\times P$ patches, resulting in $K=\frac{H\times W}{P^2}$ patches, each transformed into a token with a feature vector and positional encoding. For notation, $\boldsymbol{k}_i^I$ and $\boldsymbol{k}_i^F$ denote the feature of $i_{th}$ image patch and flow patch respectively, where the image feature provides segmentation of semantically similar objects while flow features focus on moving objects. For segmentation, the Normalized Cuts (NCut)~\cite{jianbo_shi_normalized_2000} algorithm is applied, treating segmentation as a graph partitioning task. A fully connected graph $\mathcal{G} = (\mathcal{V}, \mathcal{E})$ is created where nodes $\mathcal{V}$ represent image patches and edges $\mathcal{E}$ denote the similarity between these patches, combining both image and flow features. This similarity is quantified using cosine similarity~\cite{wang_tokencut_2022,LOST,vangansbeke2022discovering,wang2023cut} from DINO's last attention layer described as $W_{ij} = \frac{\mathbf{k}_i \cdot \mathbf{k}_j}{\|\mathbf{k}_i\|_2 \|\mathbf{k}_j\|_2}$, and edges represents the average over the similarities between image feature and flow features as in Eq.~\ref{eq:imgsimilarity}.

\begin{equation}
\mathcal{E}_{i,j} = 
\begin{cases} 
1, & \text{if } \frac{W_{ij}^I + W_{ij}^F}{2} \geq \tau \\
\epsilon, & \text{otherwise}.
\end{cases}
\label{eq:imgsimilarity}
\end{equation}

where $\tau$ is a hyper-parameter and $\epsilon$ is a small value ($1\mathrm{e}{-5}$) to ensure a fully connected graph. NCut minimizes the cost of partitioning the graph into two sub-graphs, i.e., a bipartition, by solving a generalized eigensystem $(D - W)x = \lambda Dx$ by finding the eigenvector $x$ that corresponds to the second smallest eigenvalue $\lambda$, where $D$ is a diagonal matrix with $d(i) = \sum_j W_{ij}$ and $W$ is the similarity matrix. After obtaining the bipartition $x^t$ from NCut at $t$ we obtain two disjoint groups of patches and construct a binary mask where:

\begin{equation}
mask_{ij}^t = 
\begin{cases} 
1, & \text{if } mask_{ij}^t \geq \overline{x^t} \\
0, & \text{otherwise}.
\end{cases}
\label{eq:partition}
\end{equation}

The foreground is identified by selecting patches in the mask that correspond to the maximum absolute value in the second smallest eigenvector and are less connected to the overall graph. The output $mask$ can then be given as pseudo-labels for the next processing step.

\subsection{Dynamic Mask Refinement}
\label{sec:dynmaskrefin}

\begin{figure}[h] 
\centering
\setlength{\fboxrule}{1.0pt}
\setlength{\fboxsep}{0pt}
\setlength{\tabcolsep}{1pt} %
\renewcommand{\arraystretch}{0.5} %
\begin{tabular}{c c c c c c}
    & \rotatebox{0}{\small\color{gray!90} Input} 
    & \rotatebox{0}{\small\color{gray!90} RAFT} 
    & \rotatebox{0}{\small\color{gray!90} Mask} 
    & \rotatebox{0}{\small\color{gray!90} DMR}
    & \rotatebox{0}{\small\color{gray!90} Output} \\
    & \includegraphics[height=0.5in,width=0.55in]{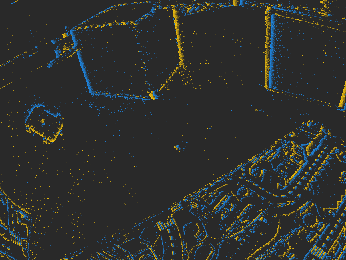}
    & \includegraphics[height=0.5in,width=0.55in]{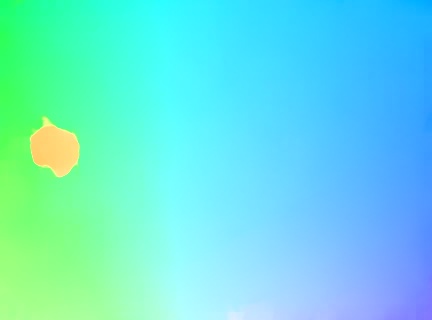}
    & \includegraphics[height=0.5in,width=0.55in]{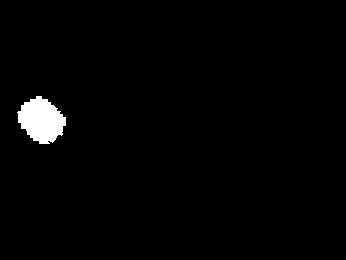}
    & \includegraphics[height=0.5in,width=0.55in]{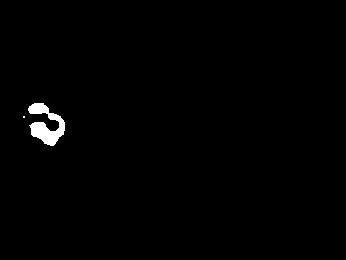}
    & \includegraphics[height=0.5in,width=0.55in]{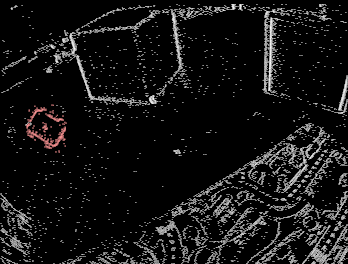} \\
    & \includegraphics[height=0.5in,width=0.55in]{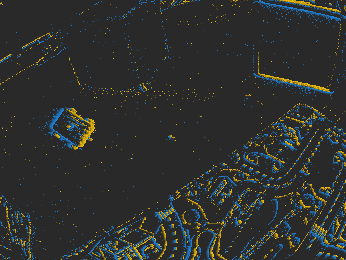}
    & \includegraphics[height=0.5in,width=0.55in]{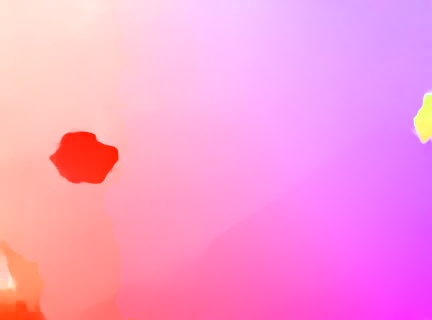}
    & \fcolorbox{red}{white}{\includegraphics[height=0.5in,width=0.55in]{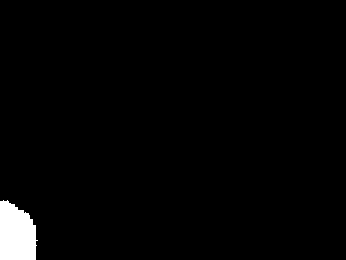}}
    & \fcolorbox{red}{white}{\includegraphics[height=0.5in,width=0.55in]{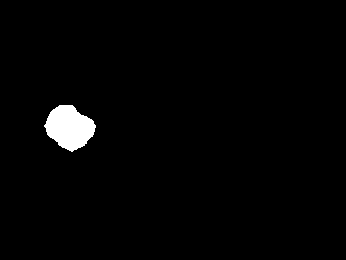}}
    & \includegraphics[height=0.5in,width=0.55in]{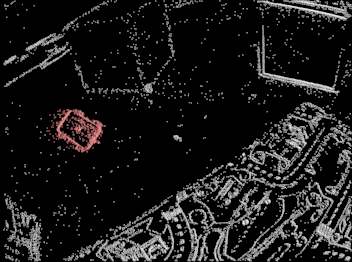} \\
    & \includegraphics[height=0.5in,width=0.55in]{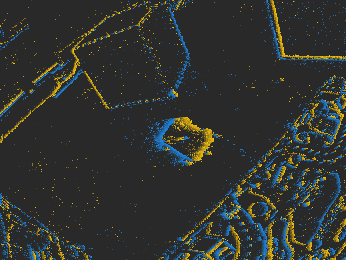}
    & \includegraphics[height=0.5in,width=0.55in]{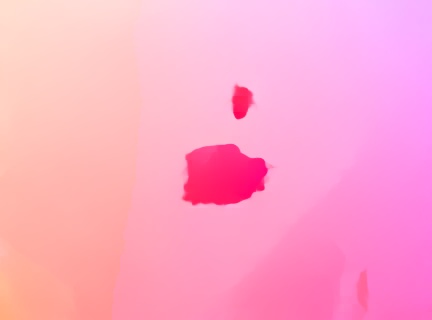}
    & \fcolorbox{red}{white}{\includegraphics[height=0.5in,width=0.55in]{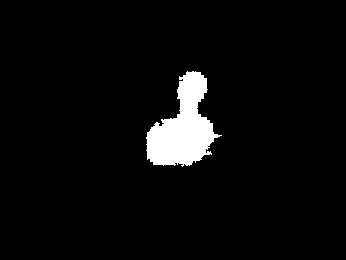}}
    & \fcolorbox{red}{white}{\includegraphics[height=0.5in,width=0.55in]{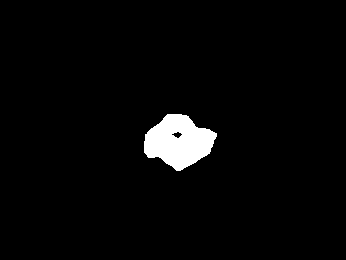}}
    & \includegraphics[height=0.5in,width=0.55in]{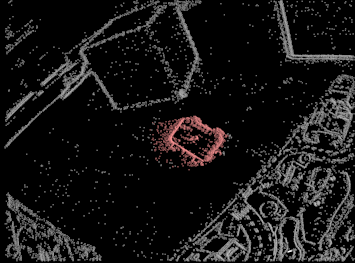} \\
    & \includegraphics[height=0.5in,width=0.55in]{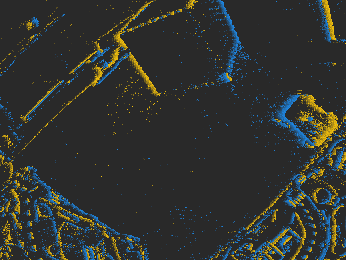}
    & \includegraphics[height=0.5in,width=0.55in]{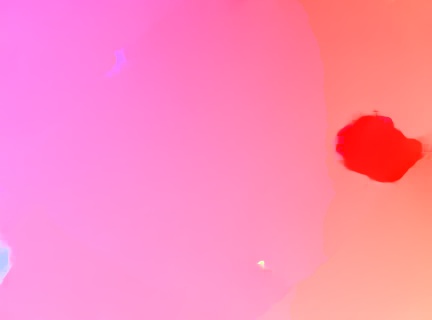}
    & \includegraphics[height=0.5in,width=0.55in]{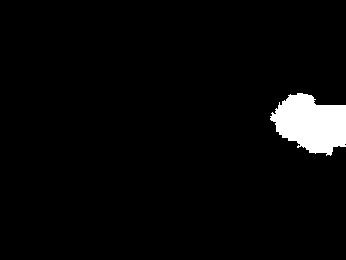}
    & \includegraphics[height=0.5in,width=0.55in]{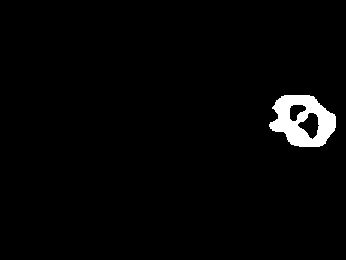}
    & \includegraphics[height=0.5in,width=0.55in]{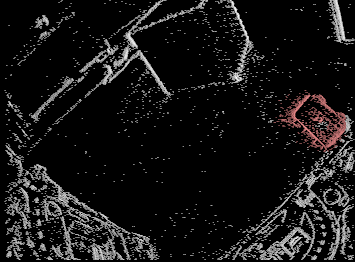} \\
\end{tabular}
\caption{Illustration of the DMR method \cite{random_walk_neurips2020,xie2022segmenting}, showing its role in maintaining temporal consistency and addressing issues with noisy optical flow. It handles scenarios where masks are either absent or irregular (in red boxes), ensuring that these inconsistencies do not disrupt the motion compensation process over time.}
\label{tb:testtimeadaptation}
\end{figure}

The output of the NCut algorithm are coarse object masks due to the large transformer patches. We refine these masks using Bilateral Solver (BS)~\cite{BarronPoole2016} or Conditional Random Field (CRF)~\cite{krahenbuhl2011efficient}. Despite refinement, issues like irregular coverage, poor temporal coherence, and size discrepancies persist as illustrated in Fig.~\ref{tb:testtimeadaptation}. To enhance appearance consistency and address previous issues, our approach incorporates a test-time adaptation process inspired by \cite{xie2022segmenting}. We refer to this process as dynamic mask refinement (DMR). This process, akin to mask propagation in self-supervised tracking as detailed in \cite{jabri2020space,lai_mast_2020,Vondrick_2018_ECCV}, unfolds in three steps. Initially, the frame features are refined using a DINO-pretrained ViT encoder \cite{caron_emerging_2021}. Then, keyframes are selected for mask propagation. In the final step, the object masks are bi-directionally propagated from these keyframes, dynamically refining them.

Given a video sequence $v \in \mathbb{R}^{N\times H\times W \times 3}$ with $N$ frames, we use DINO-ViT-Small with patch sizes $16\times 16$ to extract features: $\{f_1, \ldots, f_N\} = \{\Phi(I_1), \ldots, \Phi(I_N)\}$, where $f_N \in \mathbb{R}^{H \times W \times 384}$. We use the predicted masks from our model as reference $mask=$\{$M_1,...,M_N$\}, where $mask \in \mathbb{R}^{N\times H\times W \times 1}$ (no ground truth) are passed as noisy annotations to fine-tune the last two layers of the DINO model. This process aims to propagate object masks between neighbouring pixels from previous frames based on a high confidence score. Specifically, the mask propagation process is performed by measuring the temporal coherence of predicted object masks for keyframe selection at each frame $t$, and propagating it to the next frame using $\hat{M}_t = \text{Mask-prop}(\hat{M}_{t-1}^f)$. Temporal coherence is assessed by bi-bidirectionally propagating the mask from a keyframe at $t_k$, selected based on minimum $L1$ loss within a set of temporally consistent frames $\left\{ t_s\right\}$ with $L_{t_s}<L_{mean}$, over a temporal window of size $n$, summarized as:

\begin{equation}
L_{\text{mean}} = \frac{1}{T} \sum_{t=1}^{T} \left| \hat{M}_{t}^{f} - \hat{M}_t \right|, \quad t_k = \underset{t_s \in \{t_s | L_{t_s} < L_{\text{mean}}\}}{\text{argmin}}\, L_{t_s}
\label{eq:l1loss}
\end{equation}

This process has shown to be effective in recovering and smoothing the predicted mask as shown in Fig.~\ref{tb:testtimeadaptation}, enhancing the quality of saliency masks.

\subsection{Continuous-Discrete Labelling Update}
\label{sec:motionlabelupd}

The method, termed B-CMax, addresses event-based multi-motion segmentation by jointly estimating motion parameters $\boldsymbol{\mathcal{M}}$ and discrete labels $l$ per-event. This approach integrates CMax~\cite{gallego_unifying_2018} for motion estimation with blur detection~\cite{golestaneh2017spatially} to solve the event-object associations problem~\cite{stoffregen_event_2019,Zhou21tnnls}. CMax estimates the motion of the IMOs by producing sharp edges. Concurrently, blur detection isolates these sharp-edged regions, eliminating the need for a fixed threshold.

\textbf{CMax.} Introduced by \cite{gallego_unifying_2018}, it works by estimating the camera's relative motion vector to a group of events $\mathcal{E} = \{e_i\}_{i=1}^{N_e}$. Specifically, events are geometrically transformed $\mathcal{E} \rightarrow \acute{\mathcal{E}}$ according to a warping function $\boldsymbol{W}$ based on a motion candidate $\boldsymbol{\mathcal{M}}$ to a reference time $t_{ref}$, according to \cite{shiba_event_2022,mitrokhin_event-based_2018}, the full warping function can be described as,

\begin{equation}
\boldsymbol{\acute{u}_i}=\boldsymbol{W}(\boldsymbol{u}_i, t_{ref}; \boldsymbol{\mathcal{M}})= \boldsymbol{u_i} - t_{ref} \left( \mathbf{v} + (h_z + 1)R(\phi)\boldsymbol{u_i} - \boldsymbol{u_i} \right),
\label{eq:4dwarping}
\end{equation}

where $\boldsymbol{\mathcal{M}}=\left[ \mathbf{v}, h_z, \phi \right]$ describes the 4DoF motion parameters, with $\mathbf{v}=\left[ v_x,v_y \right]$ as the linear velocity for the 2DoF translation motion across x and y-axis, $h_z$ is the 1DoF scaling motion (zoom in/out) and $\phi$ is the angular velocity for the 1DoF rotation motion (roll) across the z-axis. $R$ is the $3\times 3$ rodrigues rotation matrix \cite{robotmani,gallego_compact_2015}. The warped events are then accumulated into an image $H$ referred to as Image of Warped Events (IWE) described as $H(\boldsymbol{\acute{u}};\boldsymbol{\mathcal{M}}) \dot{=}\sum_{i=1}^{N}b_{k}\delta(\boldsymbol{u}-\boldsymbol{\acute{u}_{i}})$, where each pixel sums the values of the warped events $\boldsymbol{\acute{u}_i}$ that fall within it. The contrast is then calculated as a function of $\boldsymbol{\mathcal{M}}$ using the variance objective function \cite{gallego_focus_2019,gallego_unifying_2018}, described as $\sigma^2(H(\boldsymbol{\acute{u}};\boldsymbol{\mathcal{M}}))$.

Our strategy focuses on identifying the correct $\boldsymbol{\mathcal{M}}$ that maximizes variance. In this work, our objective is to estimate $\boldsymbol{\mathcal{M}}$ for each IMO as well as for the camera's ego-motion. Across all benchmarks, including our Ev-Airborne dataset, we observed that translational motion was predominant. Consequently, there was no need to estimate rotational and scaling motions. This simplification accelerated the process and reduced the search space for the appropriate motion candidates.

\textbf{Blur Detection.} Introduced by \cite{golestaneh2017spatially}, it estimates blur in an image without prior information about the blur's nature. It utilizes Discrete Cosine Transform (DCT) coefficients to capture the degree of blurriness at various resolutions. By aggregating high-frequency coefficients across all resolutions, it can effectively distinguish between blurred and sharp regions. Its ability to estimate blur in both small-scale and large-scale structures makes it a great candidate for detecting sharp edges in objects of different sizes and resolutions.

B-CMax initially takes events from the refined mask, denoted as $e_{mask}$ over a time window $\Delta t$, which covers multiple IMOs undergoing motion in different directions and speeds. The primary step involves estimating the dominant motion among $e_{mask}$ with CMax. Following this, blur detection is applied to segregate and assign a discrete label for the object with sharp edges (i.e. with high contrast), while leaving the blurry edges due to events misalignment for the next loop. However, we observed that the masks generated through blur detection can be fragmented, outlining the object's periphery and not its entirety. To resolve this, we incorporated a dilation step to maximize the coverage of the mask. The final stage involves updating the motion model $\boldsymbol{\mathcal{M}}_{mask}$ only for the events within the masked area with sharp edges and assigning a discrete label $l_{mask}$ to each event in $e_{mask}$. The process iterates until it reaches a termination criterion, set to trigger when fewer than $10\%$ of the events remain in $e_{mask}$. This threshold is designed to avoid inaccurate motion estimations that can occur when too few events are available. Additional details about the B-CMax algorithm are provided in Supplementary Materials Section B.

\section{Experiments}
\label{sec:experiments}

\subsection{Experimental Setup}
\label{sec:experimentsetup}

\textbf{Datasets and Evaluation Metrics}. Our model was rigorously tested on five public benchmarks tailored for motion segmentation, along with our Ev-Airborne dataset. The characteristics of each dataset are detailed in Tab.~\ref{tab:datasummary}. For the EED~\cite{mitrokhin_event-based_2018}, EV-IMO~\cite{mitrokhin_ev-imo_2019}, and EV-IMO2~\cite{EVIMO2} datasets, we carried out both quantitative and qualitative evaluations, leveraging their high-quality ground truth data that includes IMOs and camera ego-motion. We also conducted qualitative analyses on the DistSurf~\cite{almatrafi_distance_2020} and EMSGC~\cite{zhou_event-based_2021} datasets, which feature sequences with non-rigid motions across various environments. Our Ev-Airborne dataset underwent both quantitative and qualitative assessments, offering a novel benchmark for motion segmentation recorded with a high-resolution Prophesee Gen4 camera. This dataset stands out for its high-resolution, HDR sequences that showcase a wide range of non-rigid motions and numerous small IMOs. Given the aerial platform's constant movement, camera ego-motion presents a significant challenge. The dataset also captures scenes from varying oblique angles, adding complexities like parallax effects and motion induced by platform shaking. Additional details about the Ev-Airborne dataset are provided in Supplementary Materials Section A.

\begin{table}[h]
\centering
\caption{Summary and characteristics of the datasets.}
\label{tab:datasummary}
\resizebox{\columnwidth}{!}{%
\begin{tabular}{cccccccc}
\hline
Dataset & Camera & Camera Status & \#IMO & Env. & HDR & Non-rigid & Small IMO \\ \hline
EED \cite{mitrokhin_event-based_2018} & DAVIS240 & Moving & 1-3 & Indoor & $\Large\color{green}{\checkmark}$ & $\Large\color{red}{\pmb\times}$ & $\Large\color{green}{\checkmark}$ \\ \hline
EV-IMO \cite{mitrokhin_ev-imo_2019} & DAVIS346 & Moving & 1-3 & Indoor & $\Large\color{red}{\pmb\times}$ & $\Large\color{red}{\pmb\times}$ & $\Large\color{red}{\pmb\times}$ \\ \hline
EV-IMO2 \cite{EVIMO2} & Prophesee Gen3 & Moving & 1-3 & Indoor & $\Large\color{red}{\pmb\times}$ & $\Large\color{red}{\pmb\times}$ & $\Large\color{red}{\pmb\times}$ \\ \hline
DistSurf \cite{almatrafi_distance_2020} & DAVIS346 & Static & 1-3 & In/Outdoor & $\Large\color{red}{\pmb\times}$ & $\Large\color{red}{\pmb\times}$ & $\Large\color{red}{\pmb\times}$ \\ \hline
EMSGC \cite{zhou_event-based_2021} & DAVIS346 & Moving & 1-3 & Indoor & $\Large\color{red}{\pmb\times}$ & $\Large\color{green}{\checkmark}$ & $\Large\color{red}{\pmb\times}$ \\ \hline
Ev-Airborne (Ours) & Prophesee Gen4 & Moving & 1-N & Outdoor & $\Large\color{green}{\checkmark}$ & $\Large\color{green}{\checkmark}$ & $\Large\color{green}{\checkmark}$ \\ \hline
\end{tabular}%
}
\end{table}

Following previous works~\cite{zhou_event-based_2021,lu_event-based_2021,mitrokhin_event-based_2018,stoffregen_event_2019,parameshwara_0-mms_2021,mitrokhin_ev-imo_2019,Parameshwara_evmoms} we report performance using two standard metrics: the detection rate and Intersection over Union (IoU).

\textbf{Implementation Details}. To process streams of events, we adopt a sliding window method, dividing the stream into event packets based on time intervals $\Delta t$. For sequences from DAVIS240, DAVIS346, and Prophesee Gen3, $\Delta t$ ranges between 1ms and 10ms. For the higher-resolution Ev-Airborne, $\Delta t$ is set between 10ms and 50ms. We assume motion parameters remain constant within each window. The time surface is constructed with $\tau=0.1s$. For RAFT flow, we extract data at gaps of $\pm$1 frame, and for smaller moving objects, at $\pm$7 frames to account for minimal displacement, utilizing the \texttt{raft-sintel.pth} model. BS was used as the initial mask refinement method and \texttt{ViT-S/16} model was used for the unsupervised saliency detector as well as for the DMR to extract features from patches and to enhance appearance consistency. The blur map was generated with $scale=2$ without downsampling. The CMax optimization process considers the motion space as a grid, systematically evaluating each motion candidate within this grid.

\textbf{Computational Performance}. Our method is primarily implemented in PyTorch, with the motion estimation component developed in C++ to speed up the CMax process. The algorithm operates on an RTX 3080 with 64 cores at 3.5GHz, running on an Ubuntu 20.04 LTS system. The algorithm processes event streams at various resolutions from $190\times 180$ to $1280\times 720$ as shown in Tab.~\ref{tab:runtime}. Performance is influenced by factors like noise levels, scene dynamics, and the temporal window $\Delta t$. Nonetheless, this provides a useful overview of the algorithm's overall performance. The algorithm's complexity is linearly dependent on the number of events, salient mask size, IWE pixels, and optimization, with DMR and B-CMax consuming the most processing time. Adjusting the temporal window $\Delta t$ can expedite processing, however, sufficient events are crucial for accurate motion estimation. Although motion estimation can be parallelized, this was not investigated, as it is out of the scope of this work. We believe that our proposed work stands out primarily by leveraging the key features of event cameras that record motion continuously in HDR scenarios. These attributes are critical for achieving precise motion estimation in aerial scenarios. Additionally, we address the challenge of lacking labelled segmentation masks which was accomplished by applying self-supervised learning to the realm of event cameras.

\begin{table*}[h]
\centering
\caption{Algorithm runtime on all datasets. We report the runtime in $ms$ of each processing step followed by the average (e.g. $\pm$) over multiple recordings for each dataset.}
\label{tab:runtime}
\resizebox{\textwidth}{!}{%
\begin{tabular}{lccccccc}
\hline
\textbf{Avg. Time (ms)} & \textbf{Time surface} & \textbf{RAFT} & \textbf{Frame features} & \textbf{Flow features} & \textbf{Building the graph} & \textbf{DMR} & \textbf{B-CMax/class} \\ \hline
EED & 0.49 $\pm$ 0.11 & 388.33 $\pm$ 55 & 85.92 $\pm$ 30 & 79.85 $\pm$ 23 & 104.07 $\pm$ 130 & 607.47 $\pm$ 50 & 594.23 $\pm$ 130 \\ 
EV-IMO & 1.48 $\pm$ 0.34 & 457.96 $\pm$ 150 & 102.36 $\pm$ 55  & 101.06 $\pm$ 76 & 94.58 $\pm$ 150 & 866.67 $\pm$ 90 & 894.45 $\pm$ 180 \\
EV-IMO2 & 3.29 $\pm$ 0.95 & 465.79 $\pm$ 170 & 102.21 $\pm$ 72  & 104.03 $\pm$ 89 & 173.74 $\pm$ 210 & 1240.76 $\pm$ 120 & 1502.47 $\pm$ 210 \\
EV-Airborne & 12.13 $\pm$ 5 & 2435.40 $\pm$ 200 & 1274.64 $\pm$ 120 & 1253.44 $\pm$ 110 & 2671.04 $\pm$ 240 & 1518.13 $\pm$ 125 & 5621.54 $\pm$ 350 \\ 
\hline
\end{tabular}%
}
\end{table*}

\subsection{Quantitative Analysis}
\label{sec:quant}

\setlength{\parskip}{5pt}

The performance of our algorithm was assessed on the EED~\cite{mitrokhin_event-based_2018} dataset using the detection rate metric and compared with leading methods \cite{mitrokhin_event-based_2018,stoffregen_event_2019,parameshwara_0-mms_2021,zhou_event-based_2021,lu_event-based_2021}. Baseline data was sourced from relevant publications due to the unavailability of their source codes. As Tab.~\ref{tab:eed} indicates, our algorithm consistently surpasses others across all sequences. Notably, it achieved 100\% on both the \texttt{Fast drone} and \texttt{Multiple objects} sequences, and improved detection rates in the \texttt{Lightning variation} sequence. This success is attributed to our DMR technique, which effectively propagates and connects missing/irregular masks throughout each sequence.

We also evaluate our algorithm on the EV-IMO and EV-IMO2 using the IoU metric when high-quality ground-truth segmentation masks are provided. For EV-IMO, the average IoU was calculated for the \texttt{box, wall, table,} and \texttt{tabletop} sequences. Results in Tab.~\ref{tab:evimo} place our method as the second-best, trailing by only $3.6\%$ behind the leading approach from \cite{lu_event-based_2021}. It's important to note that IoU is a more detailed metric compared to the detection rate used in Tab.~\ref{tab:eed}, hence the variations in outcomes between the two tables are expected. We faced the same challenges reported in~\cite{zhou_event-based_2021} where the ground truth masks may misalign with actual images due to inaccuracies in CAD models and motion capture, and when the motion of the IMOs moving with the speed as the camera they become excluded from IoU calculations causing a deterioration in the score. For the EV-IMO2 benchmark, our algorithm has achieved a superior success rate of $79.28\%$, significantly outperforming the baseline approach, which stands at $64.38\%$.

\begin{table}[h]
\centering
\caption{Comparison with state-of-the-art methods on the EED \cite{mitrokhin_event-based_2018} dataset using detection rate of IMOs (in \%).}
\label{tab:eed}
\resizebox{0.5\textwidth}{!}{%
\begin{tabular}{lcccccc}
\hline
Sequence name & EED~\cite{mitrokhin_event-based_2018} & EMSMC \cite{Stoffregen19iccv} & 0-MMS \cite{parameshwara_0-mms_2021} & EMSGC \cite{zhou_event-based_2021} & MSMF \cite{lu_event-based_2021} & \cellcolor{lightgrayopacity}\textbf{Ours}  \\ \hline
Lighting variation  & 84.52 & 80.51 & - & 93.51 & 92.21 & \cellcolor{lightgrayopacity} \textbf{96.00}\\
Occlusions  & 90.83 & 92.31 & - & \textbf{100.0} & \textbf{100.0} & \cellcolor{lightgrayopacity} \textbf{100.0}\\ 
Fast drone  & 92.78 & 96.30 & - & 96.30 & 96.30 & \cellcolor{lightgrayopacity} \textbf{100.0}\\ 
What is background?  & 89.21 & \textbf{100.0} & - & \textbf{100.0} & \textbf{100.0} & \cellcolor{lightgrayopacity} \textbf{100.0}\\
Multiple objects  & 87.32 & 96.77 & - & 0 & 95.67 & \cellcolor{lightgrayopacity} \textbf{100.0}\\ \hline
Average  & 89.34 & 92.28 & 94.20 & 77.96 & 96.84 & \cellcolor{lightgrayopacity} \textbf{99.20}\\ \hline
\end{tabular}
}
\end{table}

\begin{table}[h]
\centering
\caption{Comparison with state-of-the-art methods on the EVIMO~\cite{mitrokhin_ev-imo_2019} dataset using the IoU metric (in \%).}
\label{tab:evimo}
\resizebox{0.5\textwidth}{!}{%
\begin{tabular}{lcccccc}
\hline
Dataset & EV-IMO \cite{mitrokhin_ev-imo_2019} & MOMS \cite{Parameshwara_evmoms}  & MOMS-E \cite{parameshwara_0-mms_2021} & EMSGC \cite{zhou_event-based_2021} & MSMF \cite{lu_event-based_2021} & \cellcolor{lightgrayopacity}\textbf{Ours} \\ \hline
EV-IMO & 77.00 & 74.82 & 80.37 & 76.81 & \textbf{80.73} & \cellcolor{lightgrayopacity} 77.13 \\
EV-IMO2  & - & - &  - & 64.38 & - & \cellcolor{lightgrayopacity} \textbf{79.28} \\
\hline
\end{tabular}
}
\end{table}

\begin{table}[h]
\centering
\caption{Quantitative evaluation of our method on the Ev-Airborne dataset compared with EMSGC~\cite{zhou_event-based_2021} using detection rate of IMOs (in \%).}
\label{tab:ev-airborne}
\resizebox{0.5\textwidth}{!}{%
\begin{tabular}{lcc}
\hline
Sequence name & EMSGC~\cite{zhou_event-based_2021} & \cellcolor{lightgrayopacity}\textbf{Ours}  \\
\hline
Golf car high oblique   & 75.00 & \cellcolor{lightgrayopacity} \textbf{86.96} \\
Moving SUV low oblique   & \textbf{100.00} & \cellcolor{lightgrayopacity} \textbf{100.00} \\ 
Airplane takeoff 1   & 73.21 & \cellcolor{lightgrayopacity} \textbf{100.00} \\
Airplane takeoff 2   & 81.54 & \cellcolor{lightgrayopacity} \textbf{100.00} \\
Moving car low oblique   & 97.22 & \cellcolor{lightgrayopacity} \textbf{100.00} \\
Moving SUV/dome  & \textbf{100.00} & \cellcolor{lightgrayopacity} \textbf{100.00} \\
Big cars low oblique   & 71.45 & \cellcolor{lightgrayopacity} \textbf{89.03} \\ 
Pedestrians   & 65.44 & \cellcolor{lightgrayopacity} \textbf{72.14} \\ 
Small cars high oblique   & 66.97 & \cellcolor{lightgrayopacity} \textbf{81.75} \\
Small cars low oblique   & \textbf{76.47} & \cellcolor{lightgrayopacity} 71.81 \\ \hline
Average & 80.73 & \cellcolor{lightgrayopacity} \textbf{90.16} \\ \hline
\end{tabular}
}
\end{table}

For the Ev-Airborne dataset, we compared the detection rate performance of our method against EMSGC~\cite{zhou_event-based_2021}. This comparison is particularly relevant as both our method and EMSGC are unsupervised approaches, making the evaluation between both implementations valid. Our algorithm notably outperforms the baseline approach, achieving a $90.06\%$ average detection rate. In specific scenarios like \texttt{Airplane takeoff 1,2}, \texttt{Moving car low oblique}, and \texttt{Big cars low oblique}, our method attained a $100\%$ detection rate, a significant improvement over EMSGC's $70.42\%$ to $97.22\%$. Our method achieved a $100\%$ detection rate in the \texttt{Moving SUV low oblique} and \texttt{Moving SUV/dome} scenarios, but experienced a slight decrease to $71.81\%$ in \texttt{Small cars low oblique}, marginally lower than EMSGC's $76.47\%$. This reduction in performance is likely attributable to the smaller size of the IMOs and the parallax effect of the background in these scenarios.

\subsection{Qualitative Analysis}
\label{sec:quanl}

\begin{figure*}[t] 
\centering
\renewcommand*{\arraystretch}{0.3}
\setlength{\tabcolsep}{0.5pt} %

\begin{tabular}{c c c c c c c c}
    & \color{gray!90}Background & \color{gray!90}Occlusion & \color{gray!90}Fast drone & \color{gray!90}Light variation & \small\color{gray!90}Multiple objects & \color{gray!90}Hand & \color{gray!90}Cars\\
    & \includegraphics[height=0.86in,width=0.97in]{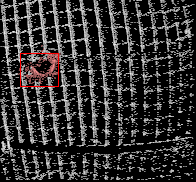}
    & \includegraphics[height=0.86in,width=0.97in]{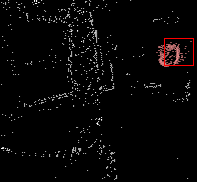}
    & \includegraphics[height=0.86in,width=0.97in]{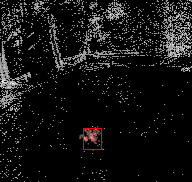}
    & \includegraphics[height=0.86in,width=0.97in]{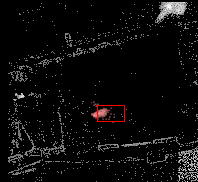}
    & \includegraphics[height=0.86in,width=0.97in]{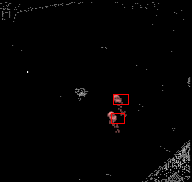} & 
    \includegraphics[height=0.86in,width=0.97in]{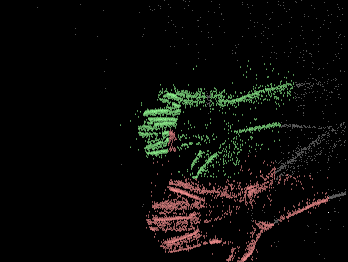} &
    \includegraphics[height=0.86in,width=0.97in]{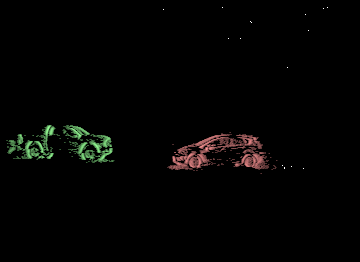} \\
    & \includegraphics[height=0.86in,width=0.97in]{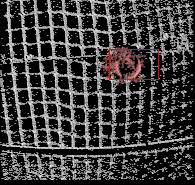}
    & \includegraphics[height=0.86in,width=0.97in]{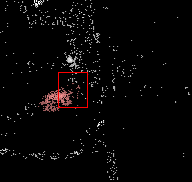}
    & \includegraphics[height=0.86in,width=0.97in]{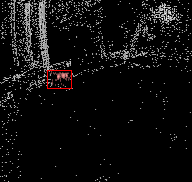}
    & \includegraphics[height=0.86in,width=0.97in]{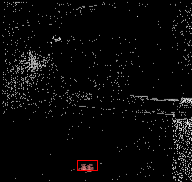}
    & \includegraphics[height=0.86in,width=0.97in]{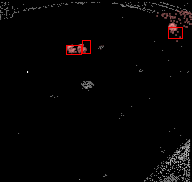} &
    \includegraphics[height=0.86in,width=0.97in]{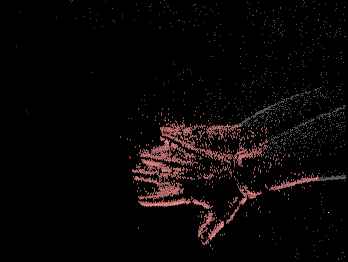}  &
    \includegraphics[height=0.86in,width=0.97in]{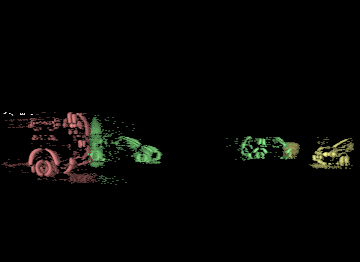} \\
\end{tabular}

\begin{tabular}{c c c c c c c c}
  \color{gray!90}Box & \color{gray!90}Table  & \color{gray!90}Wall  & \color{gray!90}Slope & \color{gray!90}Drones  & \color{gray!90}Corridor & \color{gray!90}Cast \\ 
   \includegraphics[height=0.86in,width=0.97in]{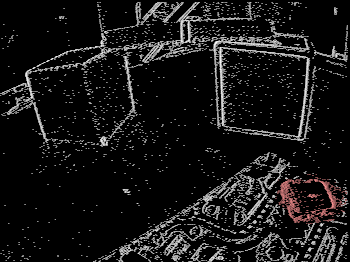} & 
  \includegraphics[height=0.86in,width=0.97in]{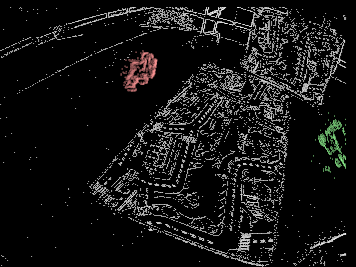} & 
  \includegraphics[height=0.86in,width=0.97in]{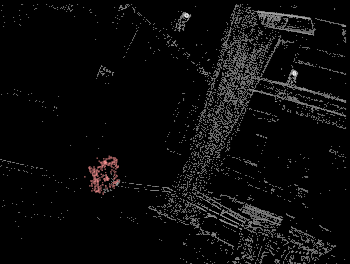} & 
  \includegraphics[height=0.86in,width=0.97in]{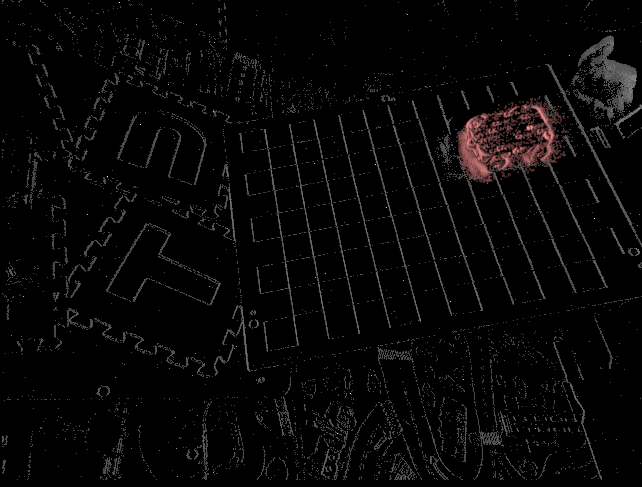} & 
  \includegraphics[height=0.86in,width=0.97in]{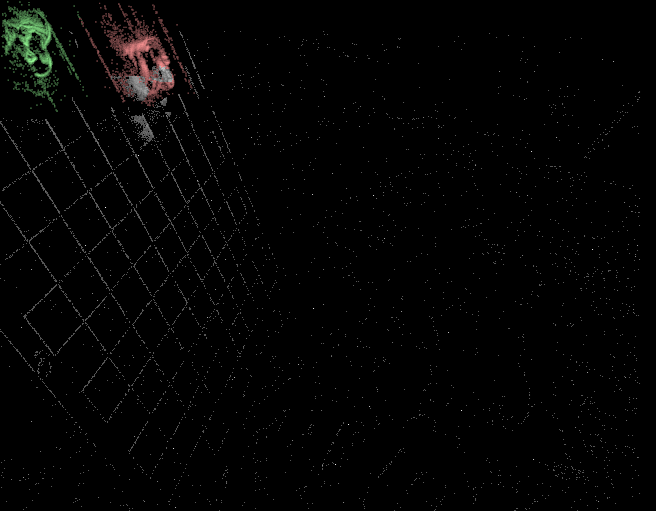} & 
  \includegraphics[height=0.86in,width=0.97in]{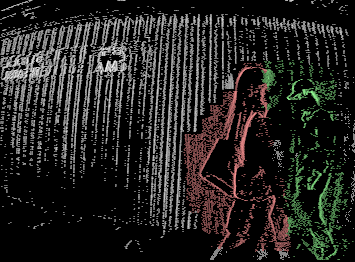} &
  \includegraphics[height=0.86in,width=0.97in]{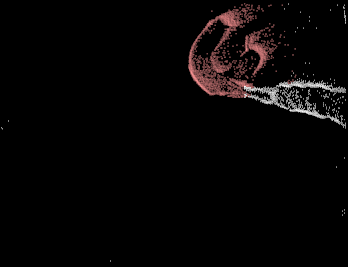}\\
  
  \includegraphics[height=0.86in,width=0.97in]{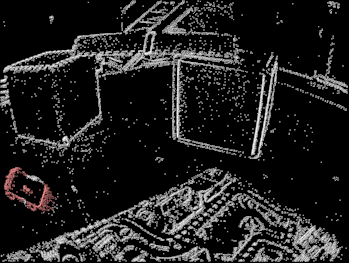} & 
  \includegraphics[height=0.86in,width=0.97in]{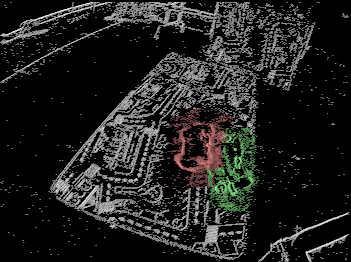} & 
  \includegraphics[height=0.86in,width=0.97in]{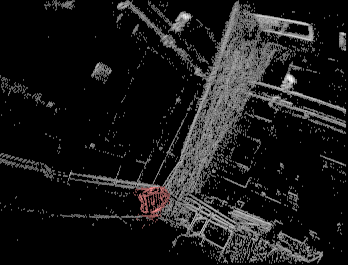} & 
  \includegraphics[height=0.86in,width=0.97in]{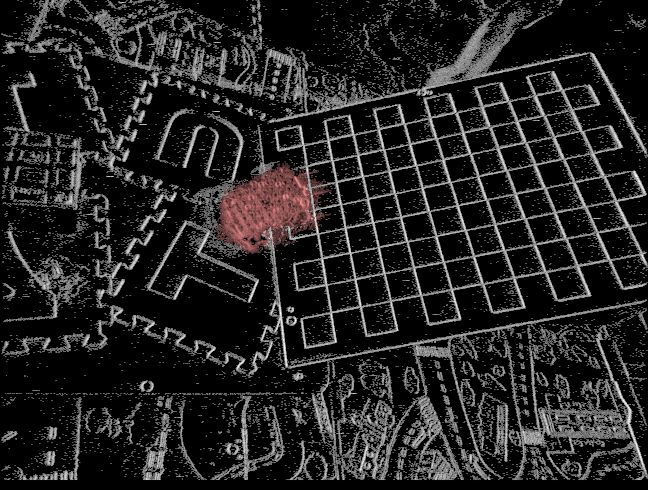} & 
  \includegraphics[height=0.86in,width=0.97in]{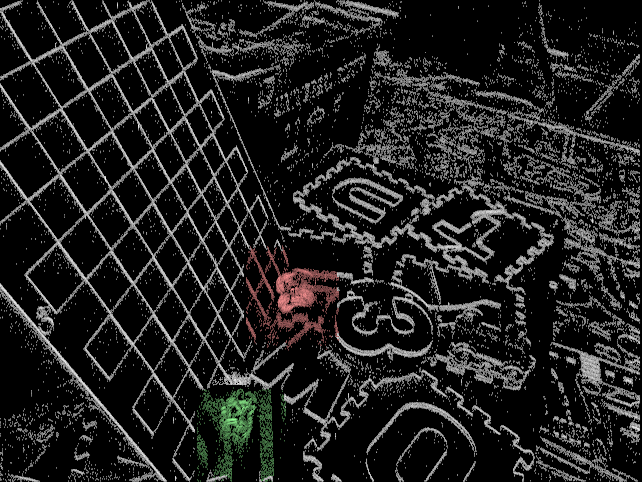} & 
  \includegraphics[height=0.86in,width=0.97in]{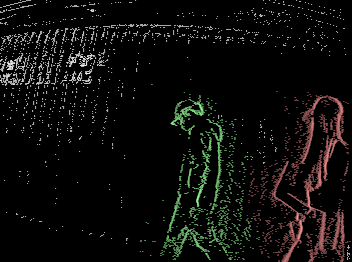} &
  \includegraphics[height=0.86in,width=0.97in]{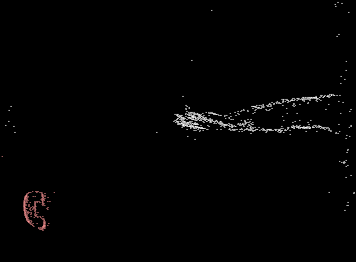} \\
\end{tabular}
\caption{Combined qualitative results on EED~\cite{mitrokhin_event-based_2018} (red bounding boxes are ground truth), EV-IMO~\cite{mitrokhin_ev-imo_2019}, EV-IMO2~\cite{EVIMO2}, DistSurf~\cite{almatrafi_distance_2020}, and HKUST-EMS~\cite{zhou_event-based_2021} datasets.}
\label{fig:combined_qualitative_results}
\end{figure*}

We provide extensive qualitative evaluations of the same benchmarks reported in Sec.~\ref{sec:quant}, and extended the evaluation to include sequences from DistSurf~\cite{almatrafi_distance_2020} and EMSGC~\cite{zhou_event-based_2021}, with results showcased in Fig.~\ref{fig:combined_qualitative_results}. Our algorithm effectively categorizes motion across all scenarios. The motion segmentation remains robust regardless of object speed, occlusion, or lighting challenges. Note that, ground truth bounding boxes on DAVIS grayscale images, manually annotated, can be misaligned with event data, leading to offsets, particularly in fast-moving objects. In the DistSurf sequences, \texttt{Hand} and \texttt{Cars}, our method accurately segments multiple overlapping IMOs even with a stationary camera. This observation is consistent in \texttt{Corridor} and \texttt{Cast} sequences, where a static camera captures moving objects. Additionally, we tested our algorithm in diverse scenarios, including scenes with appearance changes due to rotation and varying resolutions, as seen in EV-IMO~\cite{mitrokhin_ev-imo_2019} and EV-IMO2~\cite{EVIMO2} datasets. Our algorithm not only assigns the correct labels that describe the number of motions in the scenes but also recovers the sharp edges of the IMOs and background. Even though we do not exhibit the rectangular segmentation boundaries present in~\cite{parameshwara_0-mms_2021}, sometimes the predicted mask is larger than the IMO causing it to cover a small portion of the background.

\begin{figure*}[t] 
\centering
\renewcommand*{\arraystretch}{0.3}
\setlength{\tabcolsep}{0.5pt}
\begin{tabular}{c c c c c c}
    \rotatebox{90}{\hspace{0.1cm}\color{gray!90}EMSGC\cite{zhou_event-based_2021}} & 
    \includegraphics[height=0.9in,width=1.35in]{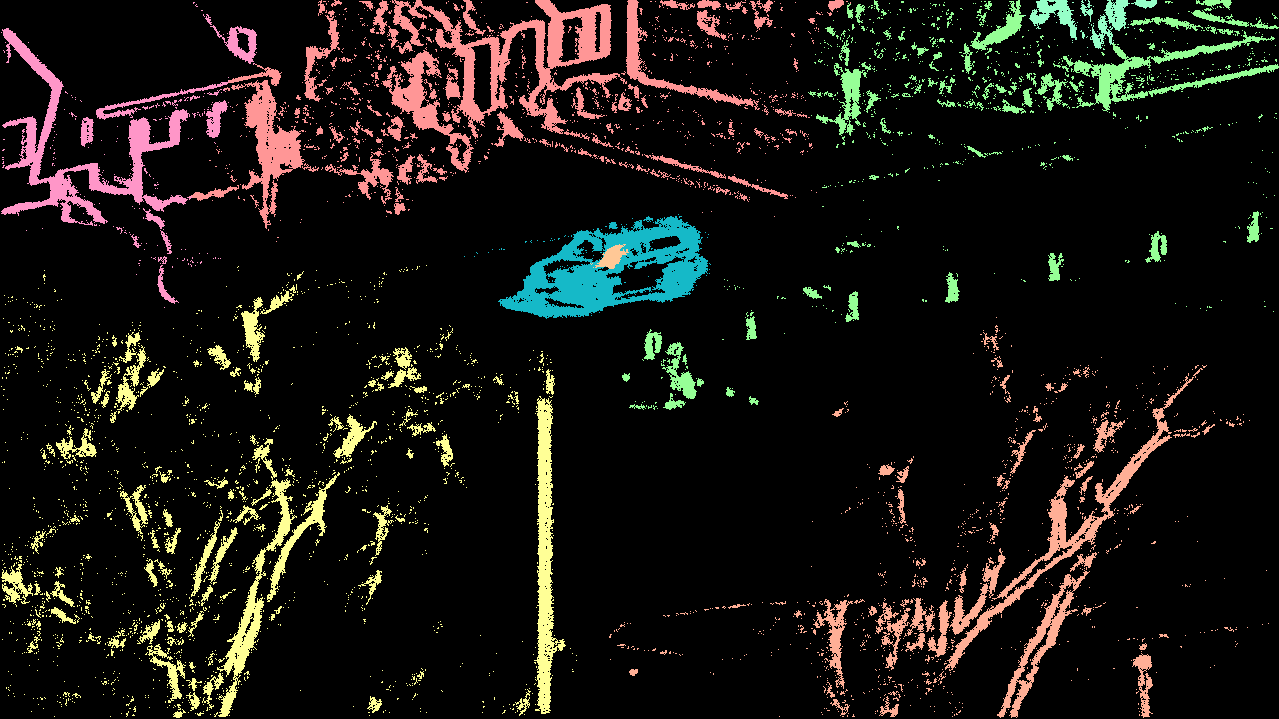} & 
    \includegraphics[height=0.9in,width=1.35in]{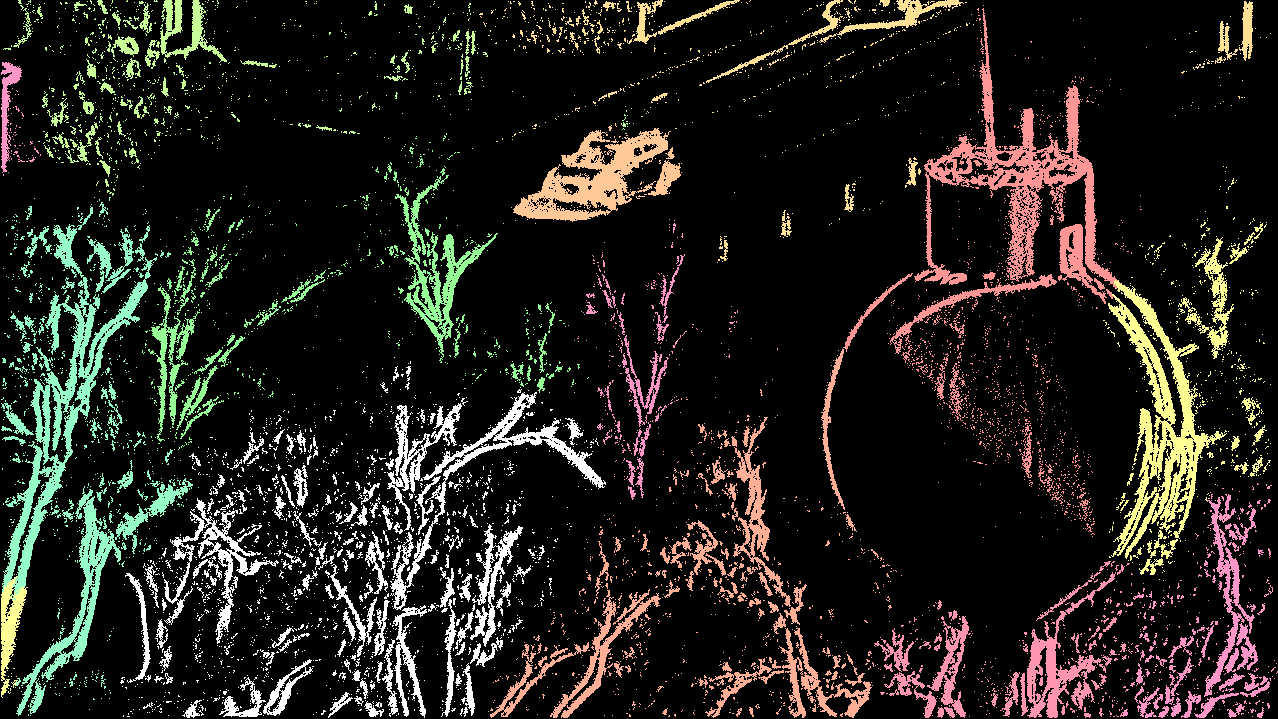} &
    \includegraphics[height=0.9in,width=1.35in]{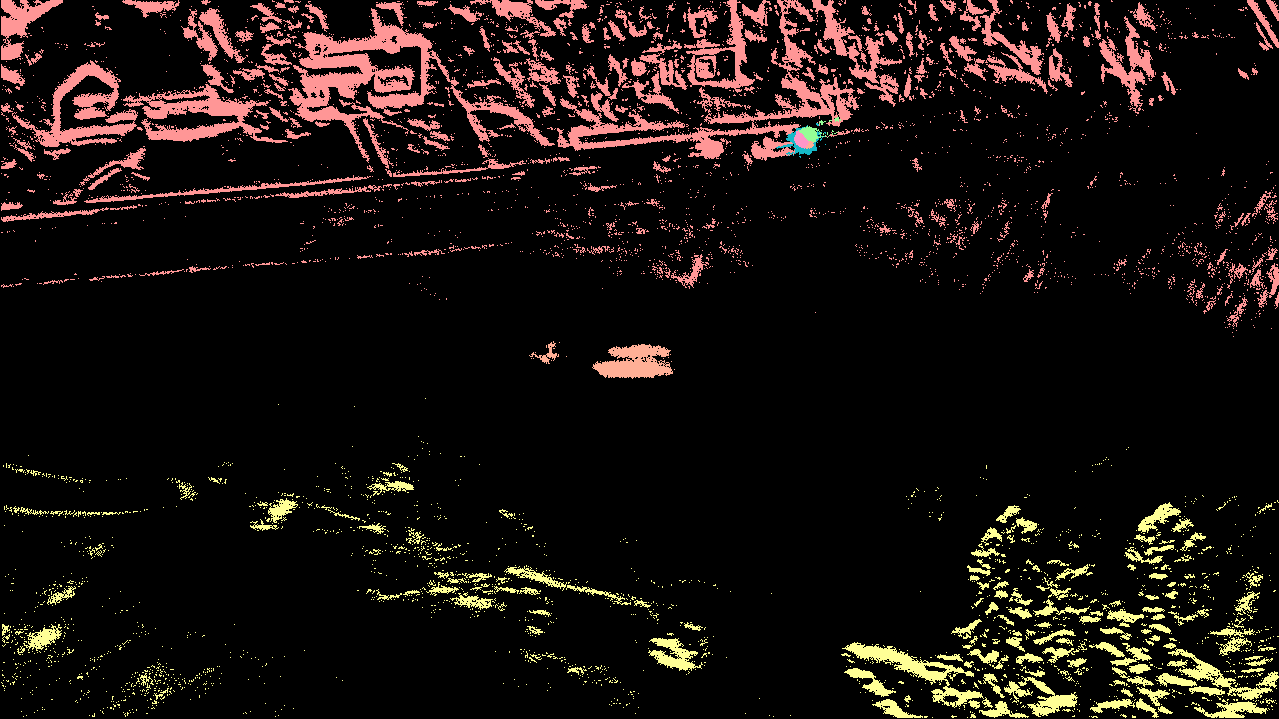} &
    \includegraphics[height=0.9in,width=1.35in]{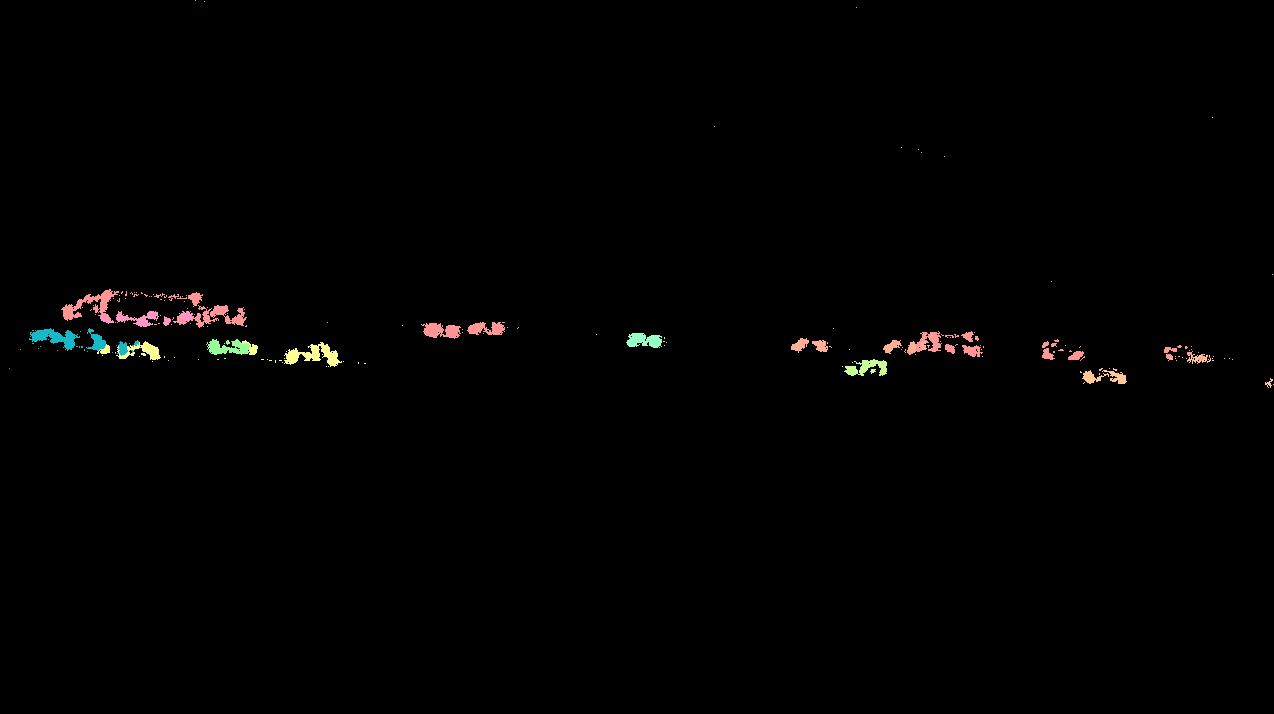} &
    \includegraphics[height=0.9in,width=1.35in]{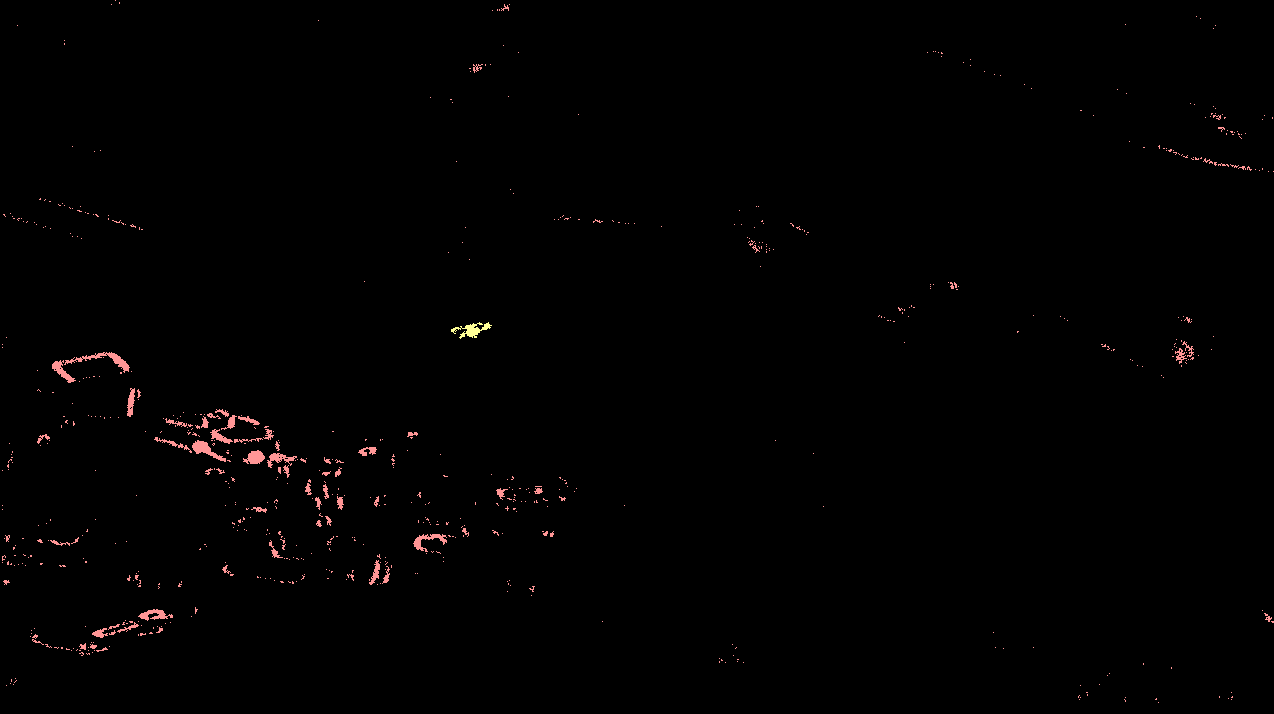} \\
    \rotatebox{90}{\hspace{0.6cm}\textbf{\color{red!100}Ours}} &
    \includegraphics[height=0.9in,width=1.35in]{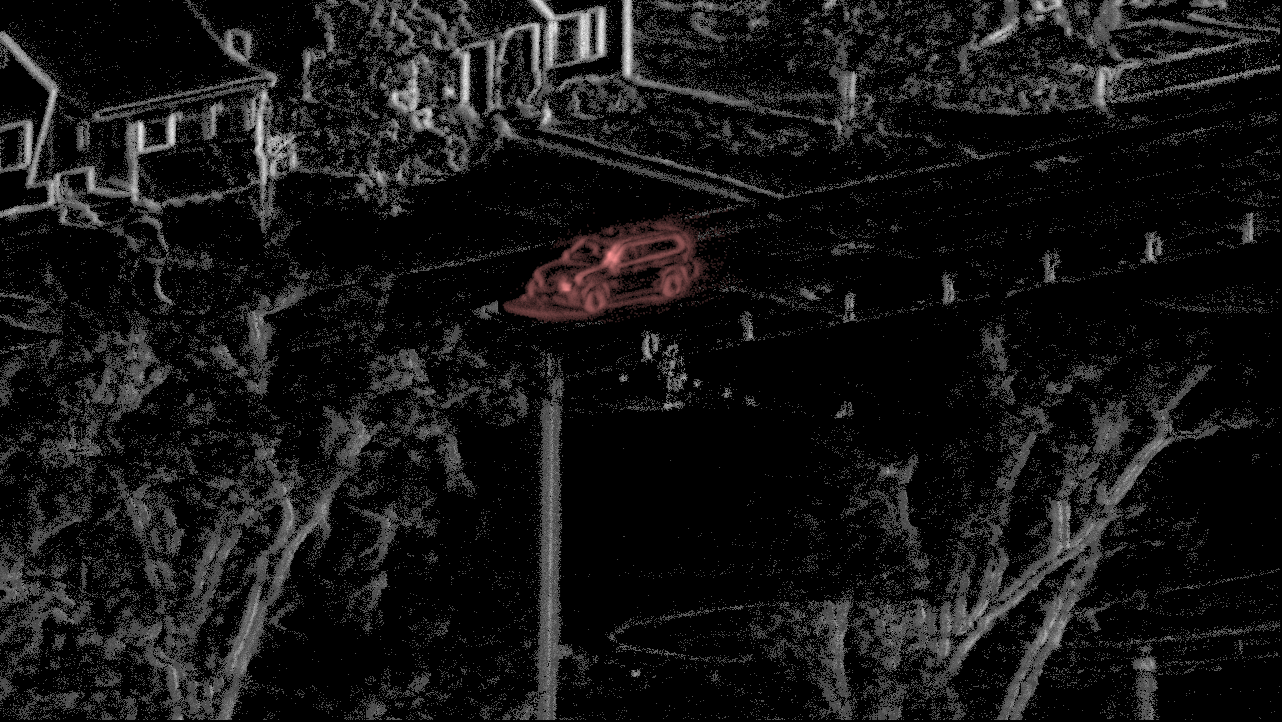} & 
    \includegraphics[height=0.9in,width=1.35in]{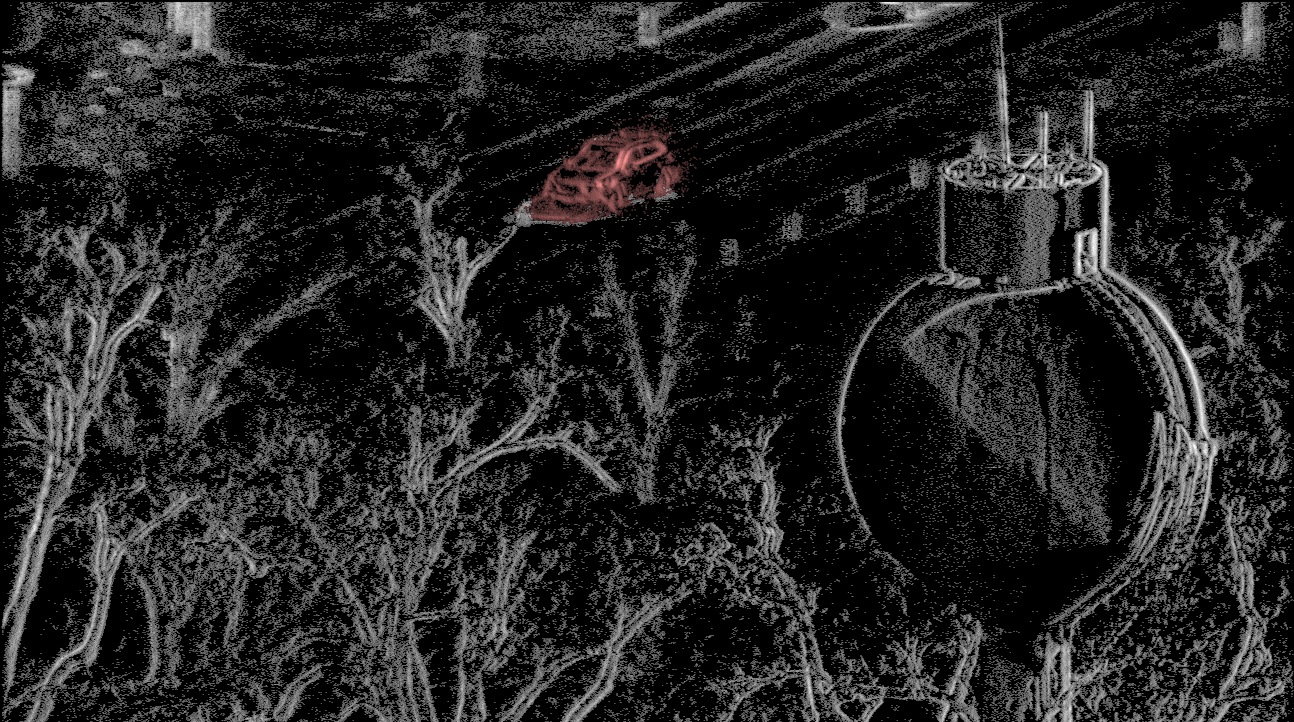} &
    \includegraphics[height=0.9in,width=1.35in]{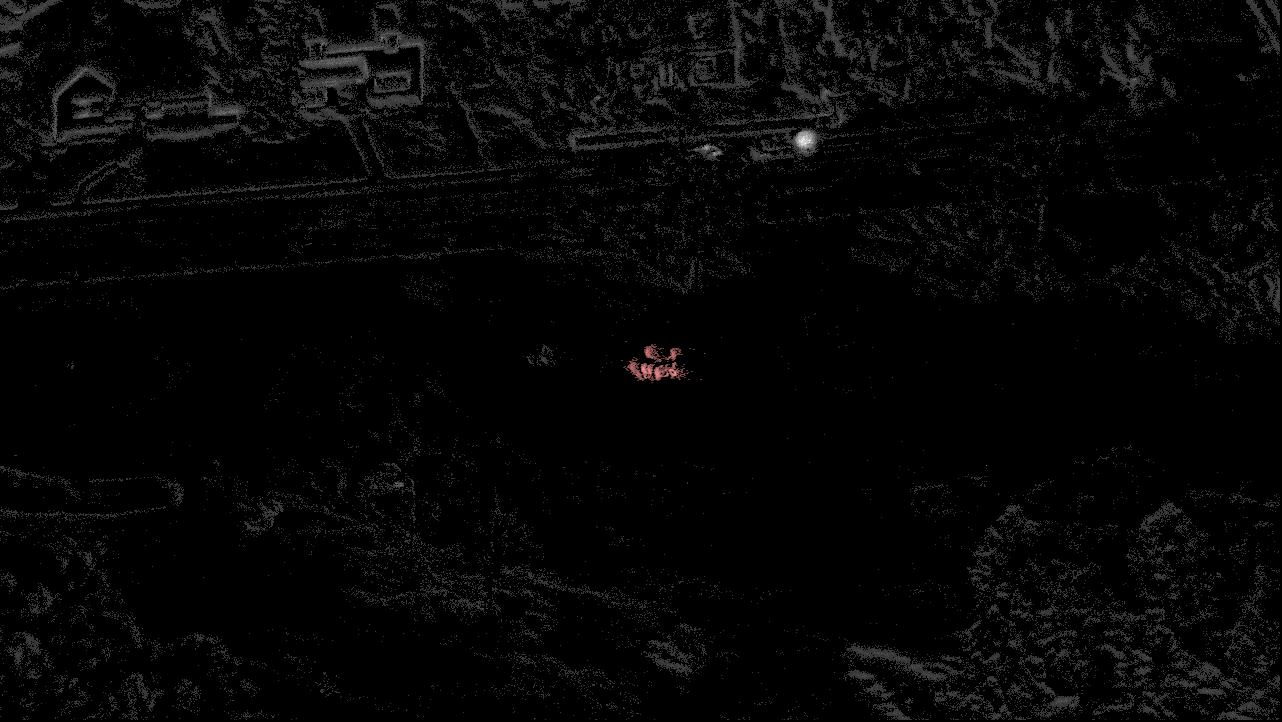} &
    \includegraphics[height=0.9in,width=1.35in]{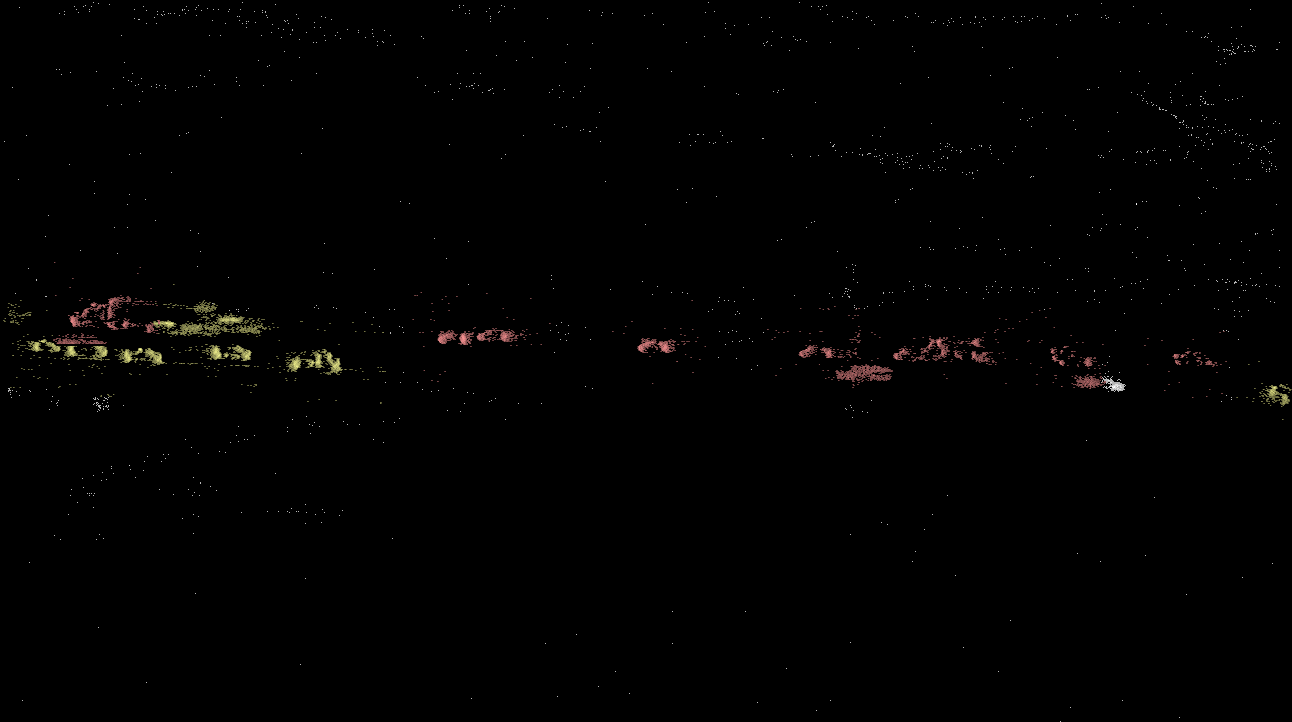} &
    \includegraphics[height=0.9in,width=1.35in]{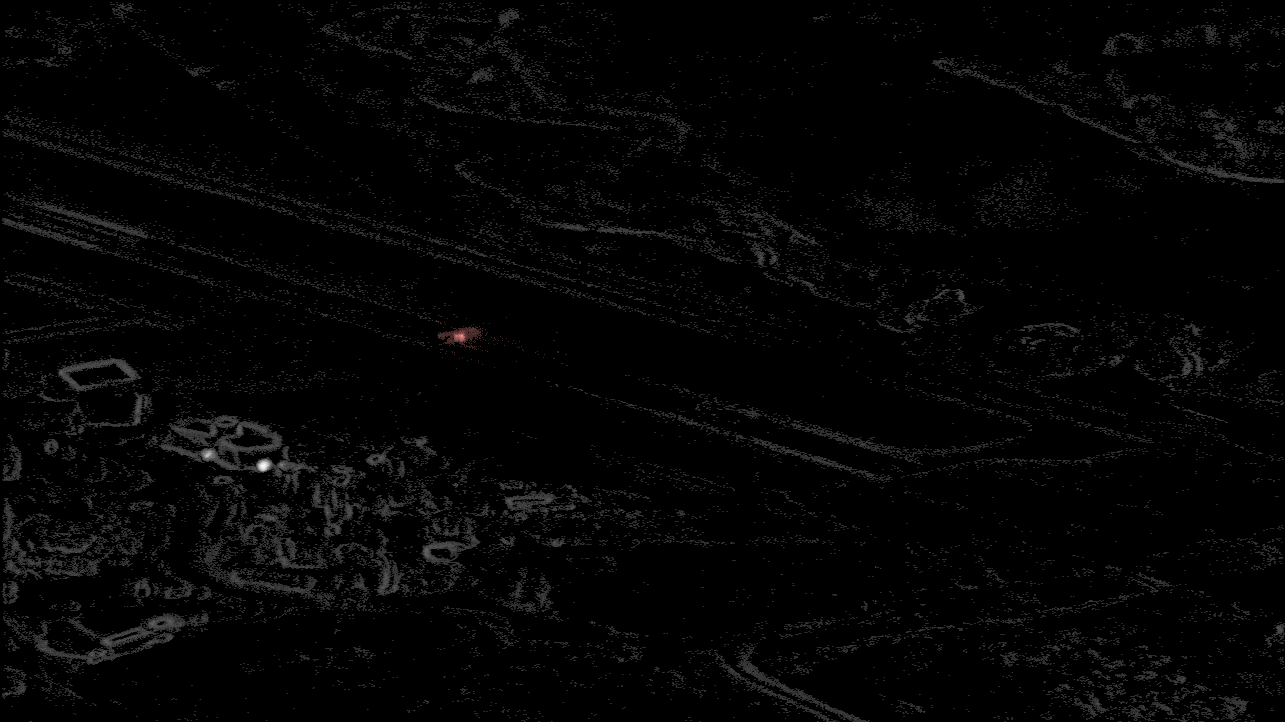} \\
    \rotatebox{90}{\hspace{0.1cm}\color{gray!90}EMSGC\cite{zhou_event-based_2021}} & 
    \includegraphics[height=0.9in,width=1.35in]{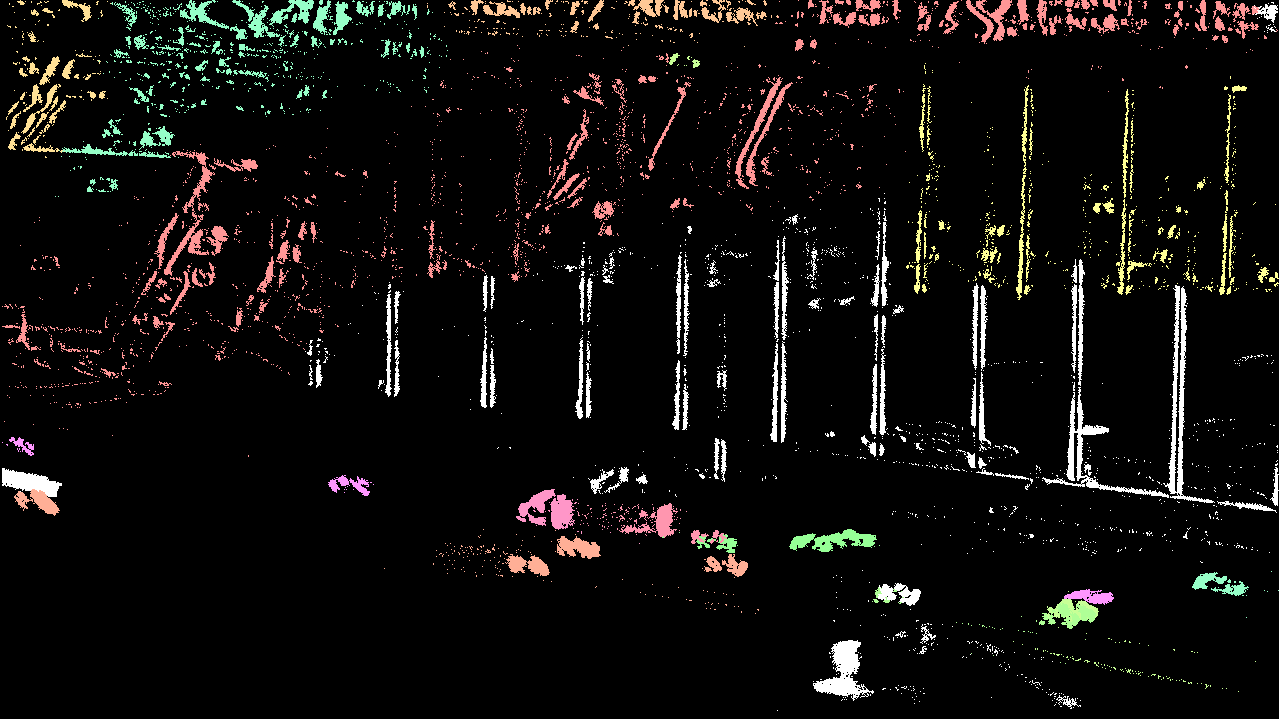} &
    \includegraphics[height=0.9in,width=1.35in]{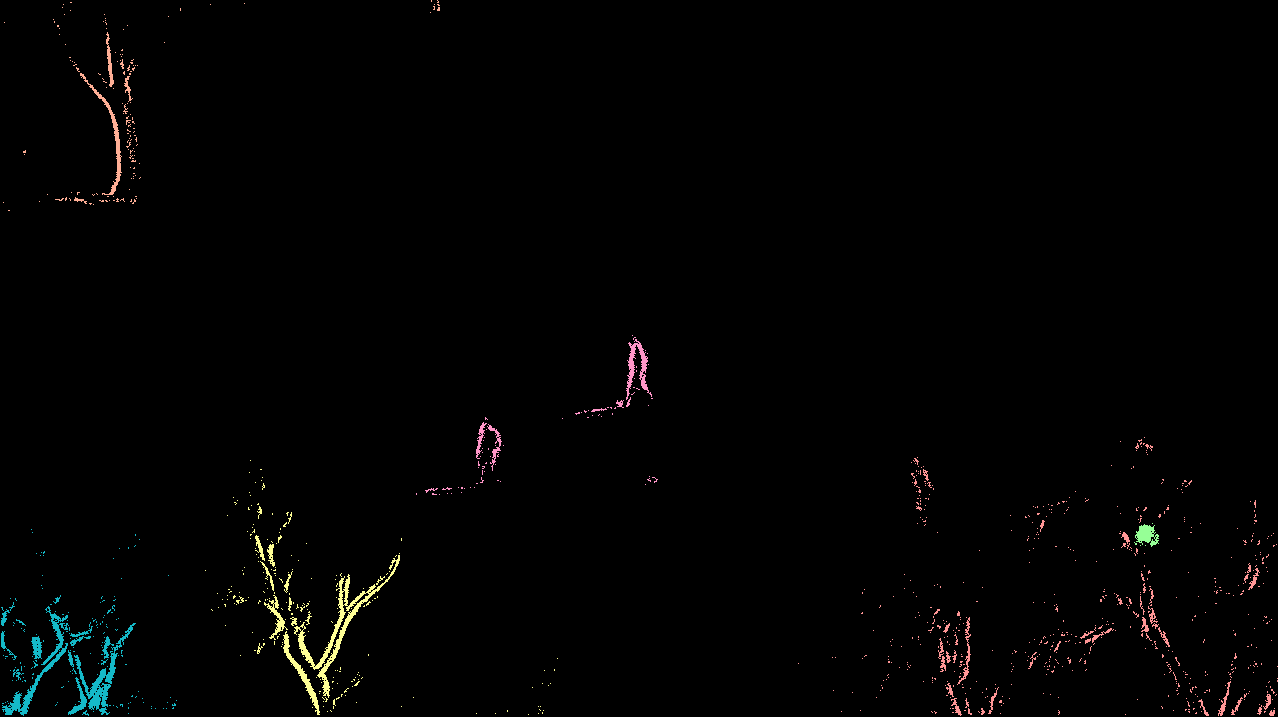} &
    \includegraphics[height=0.9in,width=1.35in]{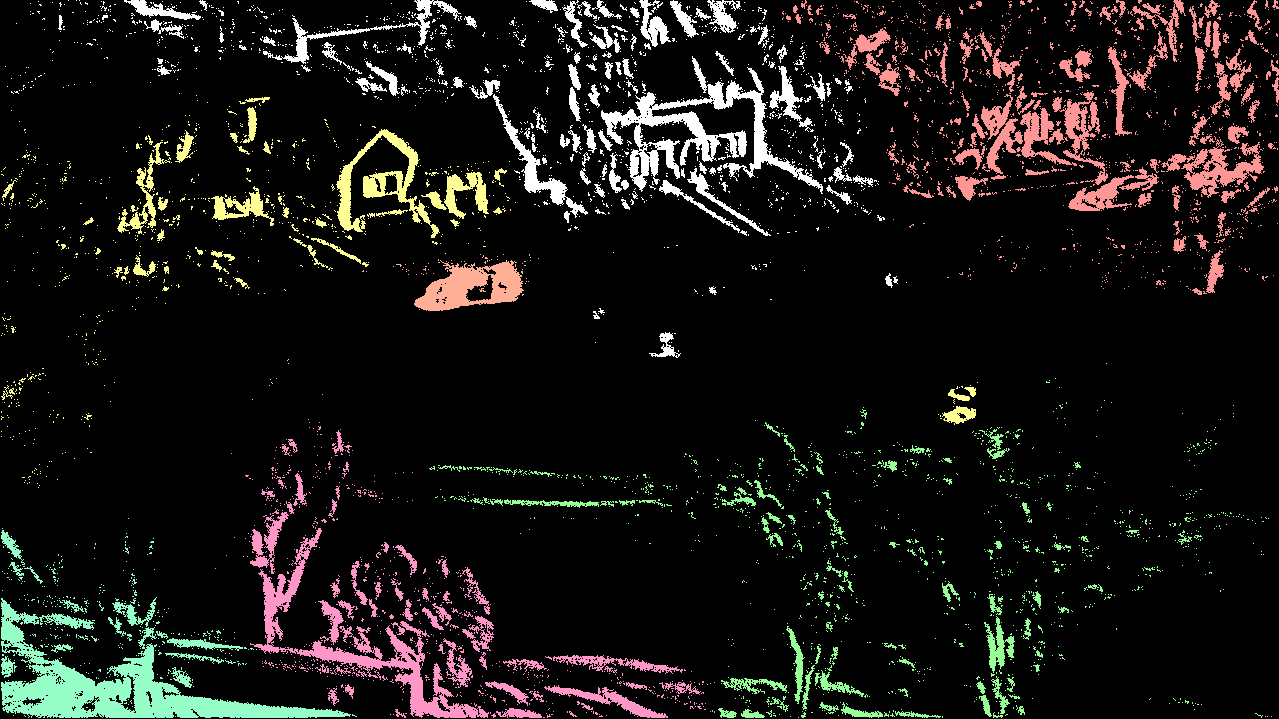} &
    \includegraphics[height=0.9in,width=1.35in]{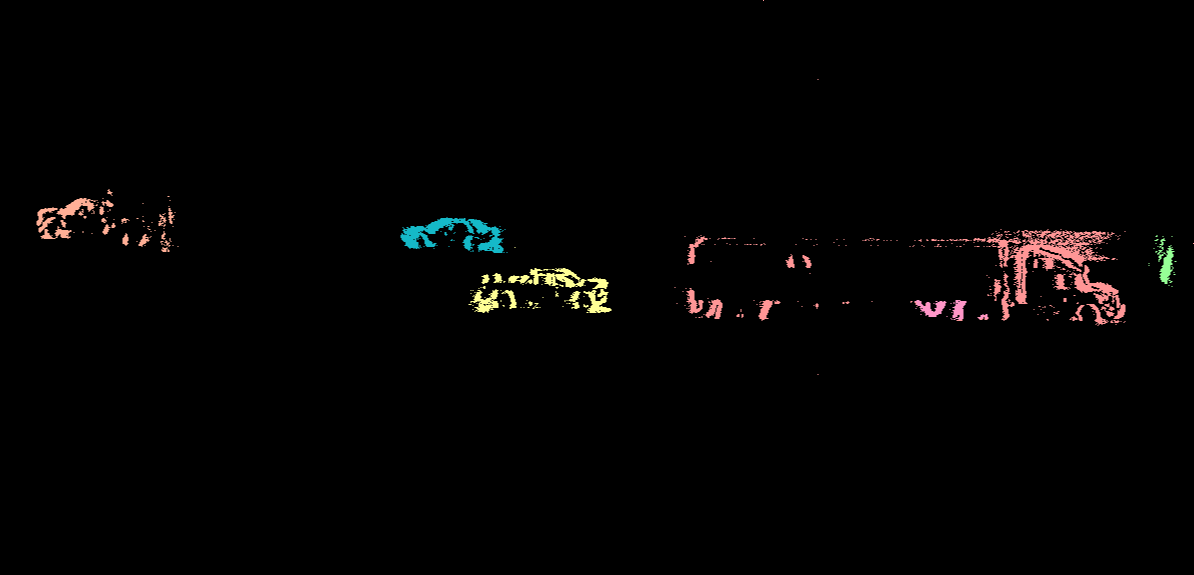} &
    \includegraphics[height=0.9in,width=1.35in]{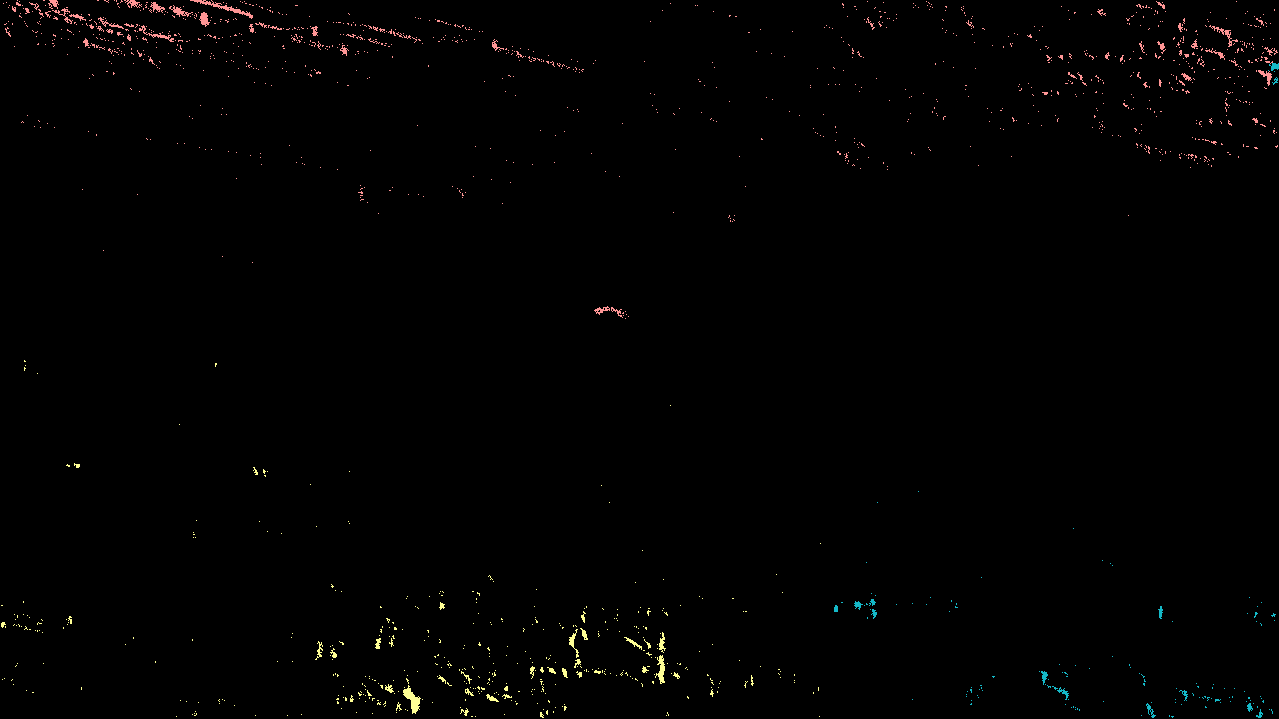} \\
    \rotatebox{90}{\hspace{0.6cm}\textbf{\color{red!100}Ours}} & 
    \includegraphics[height=0.9in,width=1.35in]{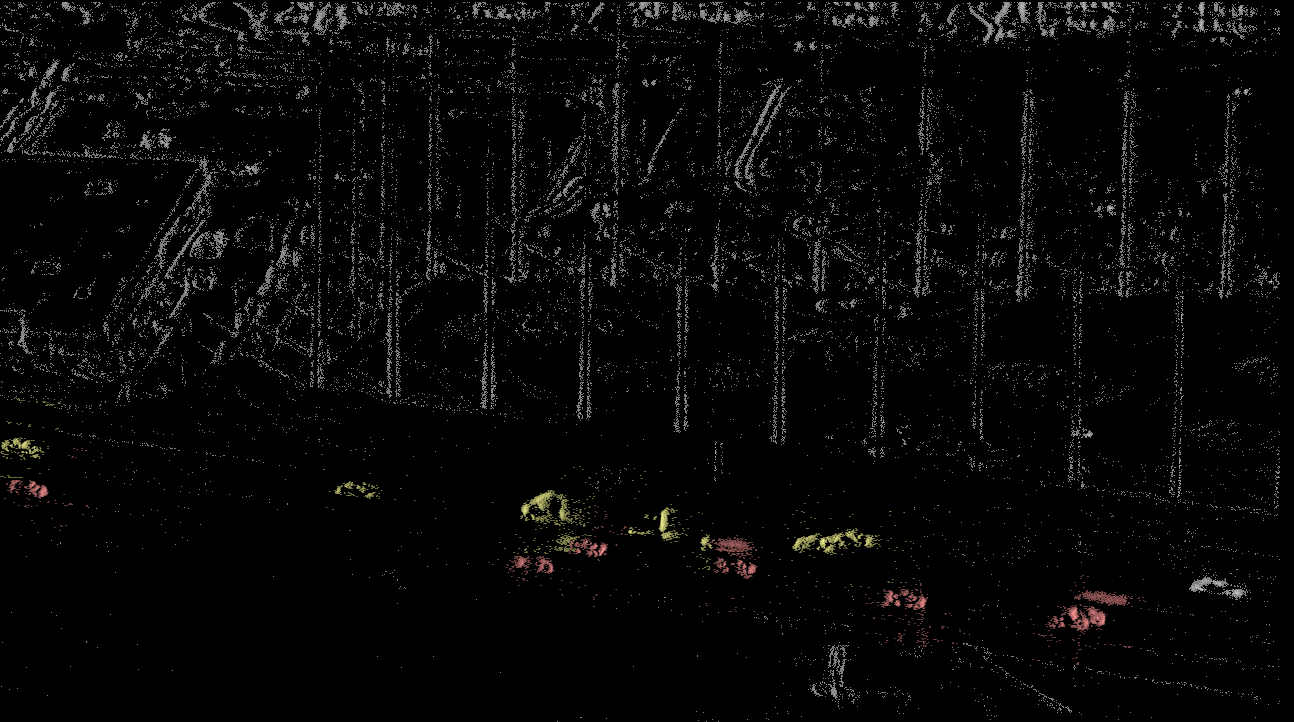} & 
    \includegraphics[height=0.9in,width=1.35in]{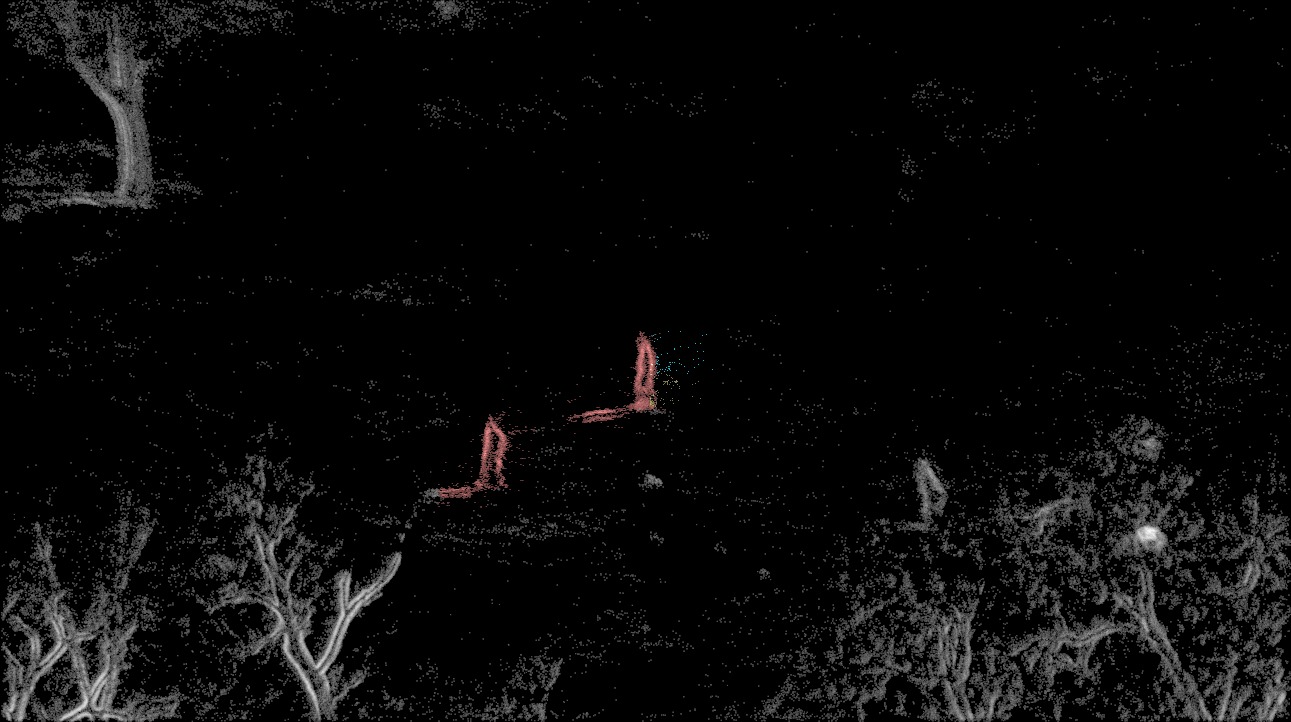} & 
    \includegraphics[height=0.9in,width=1.35in]{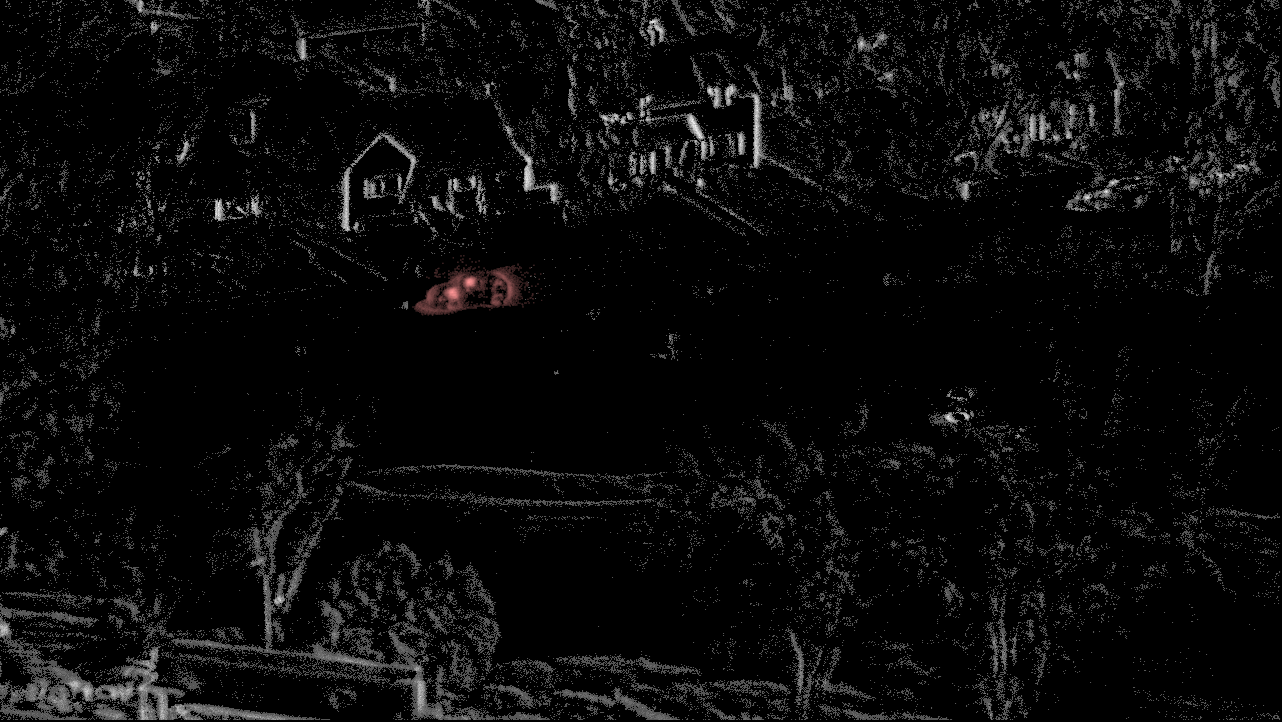} &
    \includegraphics[height=0.9in,width=1.35in]{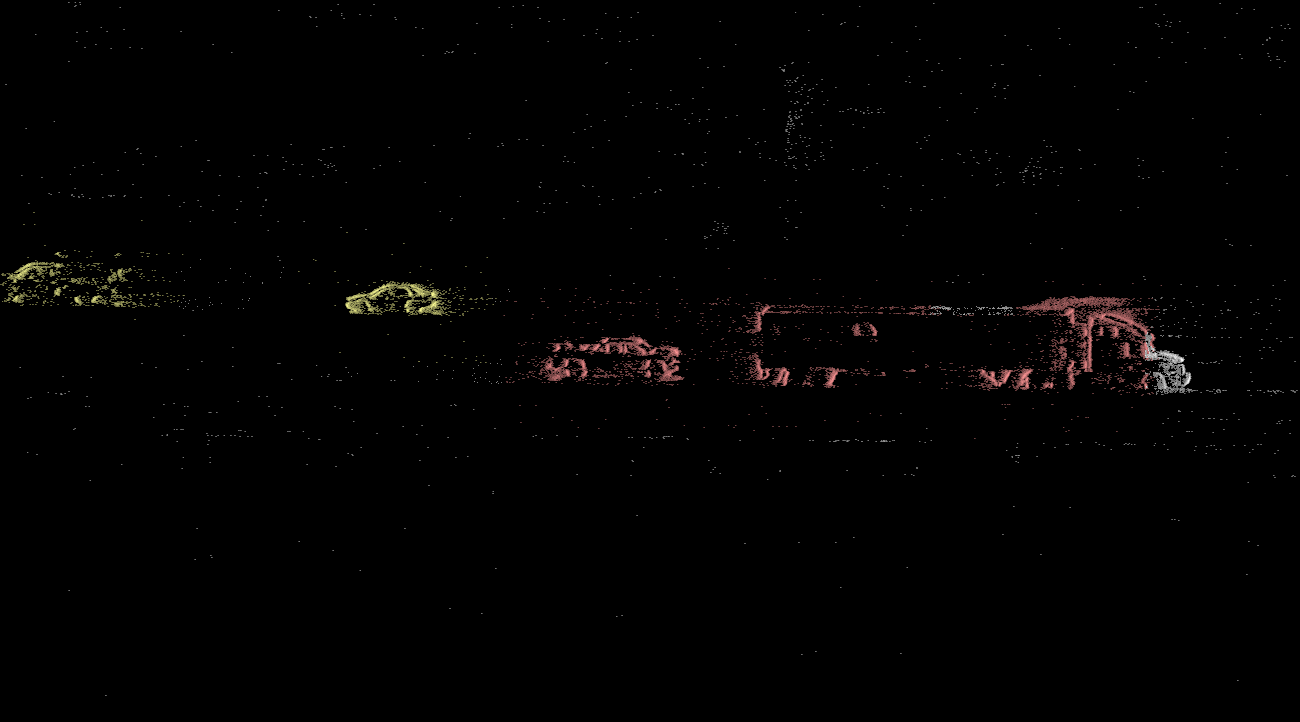} &
    \includegraphics[height=0.9in,width=1.35in]{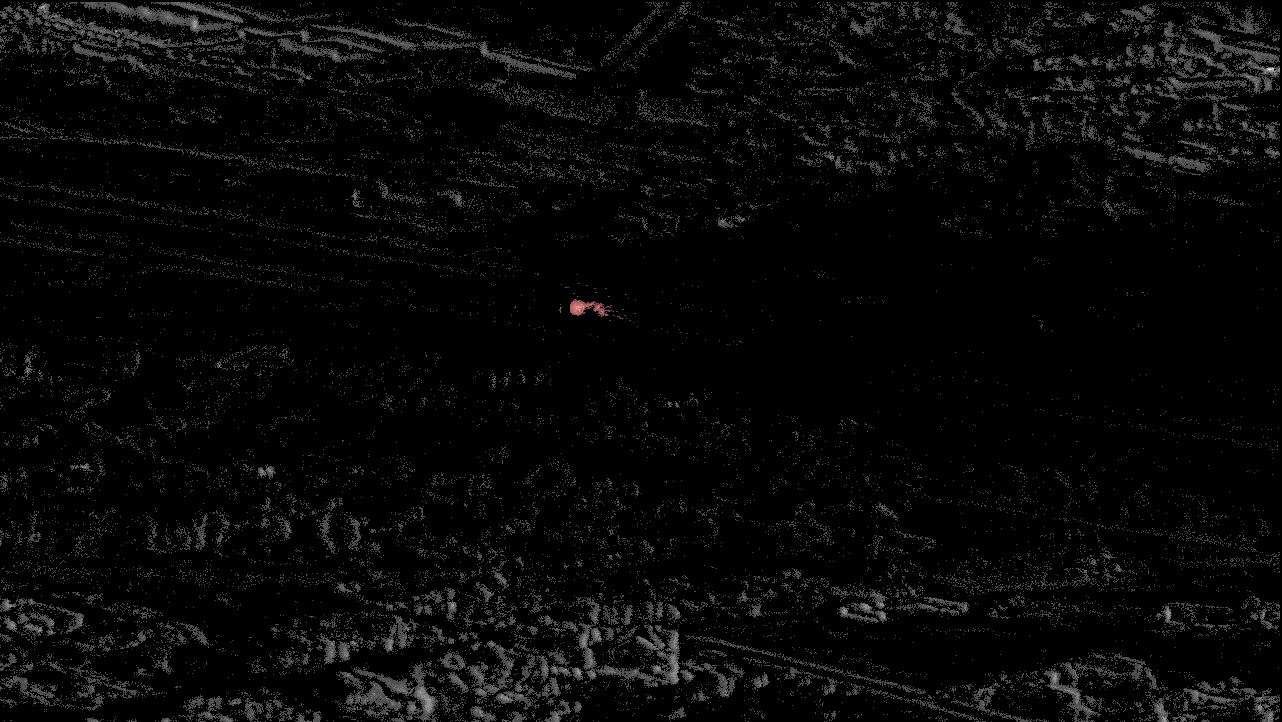} \\
\end{tabular}

\caption{Comparing the motion segmentation output of our method against EMSGC \cite{zhou_event-based_2021} on the Ev-Airborne dataset. From left to right, Rows 1,2: Moving SUV low oblique, Moving SUV/Dome, Golf car high oblique, small cars high oblique, airplane takeoff1. Rows 3,4: Small car high oblique, Pedestrians, moving car low oblique, big cars low oblique, airplane takeoff 2. This demonstrates that our approach does not over-segment the scene and it assigns the correct labels for the most salient objects with a single label for the background.}
\label{tb:evairborne1}
\end{figure*}

Our algorithm, applied to the real-world Ev-Airborne dataset without any modifications to the network architecture, demonstrates strong generalization capabilities. We chose scenes with varied motions, parallax effects, and small IMOs to evaluate our algorithm's robustness, comparing it against the EMSGC~\cite{zhou_event-based_2021} algorithm, leveraging its open-source implementation. The qualitative results are presented in Figs.~\ref{tb:evairborne1}. Our algorithm effectively identifies the correct number of IMOs and camera ego-motion, avoiding over-segmentation. It also maintains a better focus on moving objects, benefiting from the self-supervised features of DINO~\cite{caron_emerging_2021} and RAFT~\cite{teed2020raft}. Notably, the segmented classes remain consistent over time; for example, a moving vehicle initially labelled as "red" retains this label until it exits the field of view, making our method suitable for aerial persistent tracking~\cite{prokaj_persistent_2014}.

Additional qualitative results are provided in Supplementary Materials Section C, demonstrating how our approach addresses the oversegmentation problem, handles small IMOS, and maintains consistency over time.

\section{Ablation Studies and Analysis}
\label{sec:ablations}

\begin{figure}[h] 
\centering
\setlength{\fboxrule}{1.0pt}
\setlength{\fboxsep}{0pt}
\setlength{\tabcolsep}{1pt} %
\renewcommand{\arraystretch}{0.5} %
\textbf{Ev-Airborne: Airplane takeoff}
\begin{tabular}{c c c c c c c c c c c c}
    & \rotatebox{90}{\hspace{0.2cm}\color{gray!90}Input} \hspace{-2mm}\includegraphics[height=0.45in,width=0.5in]{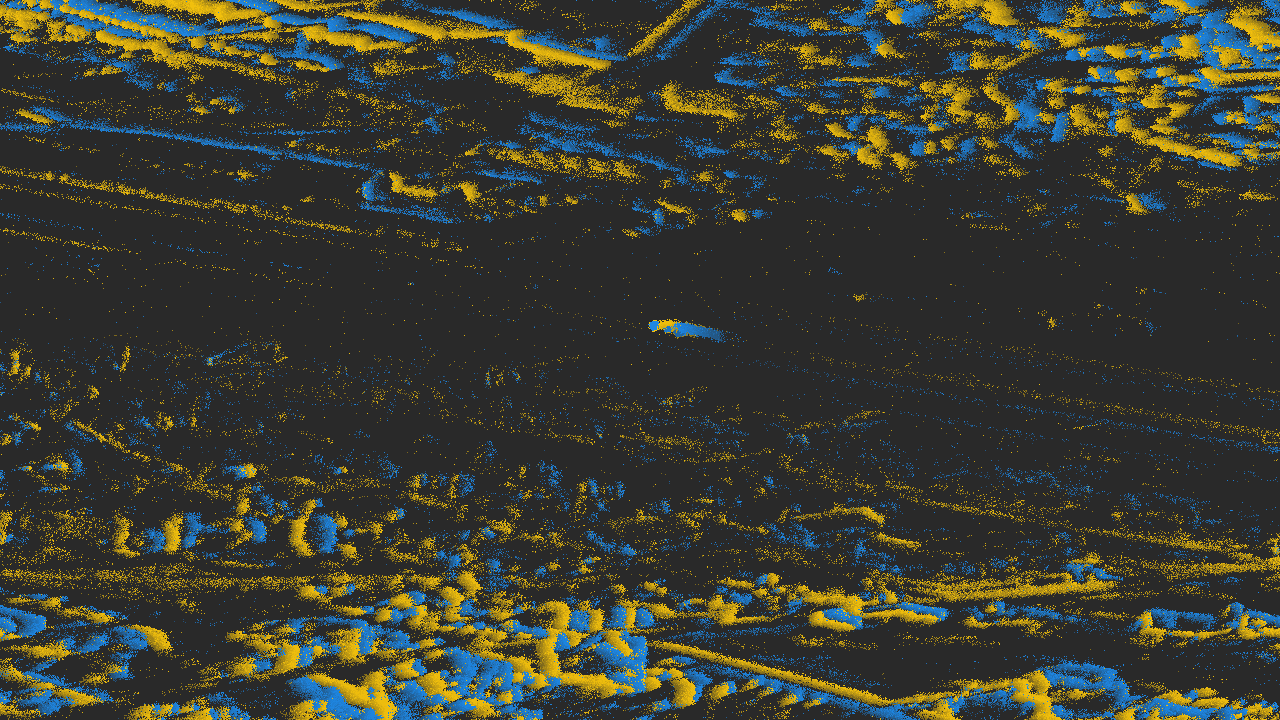}
    & \includegraphics[height=0.45in,width=0.5in]{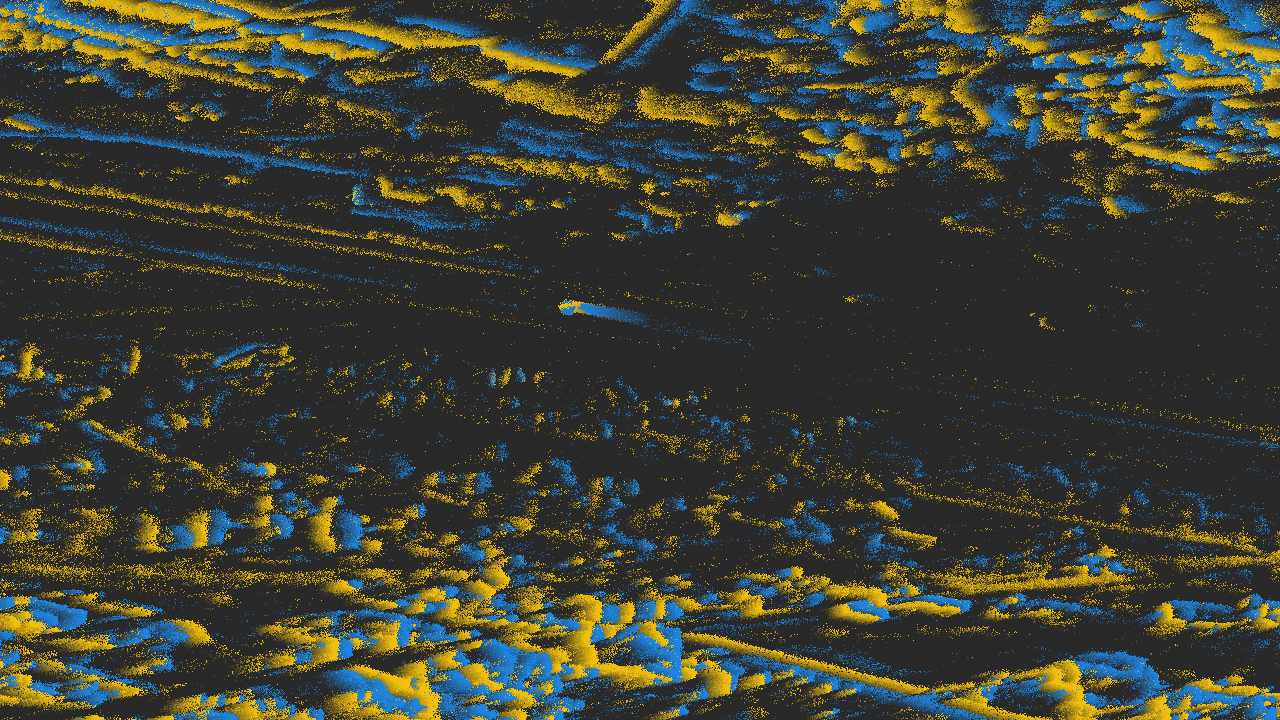}
    & \includegraphics[height=0.45in,width=0.5in]{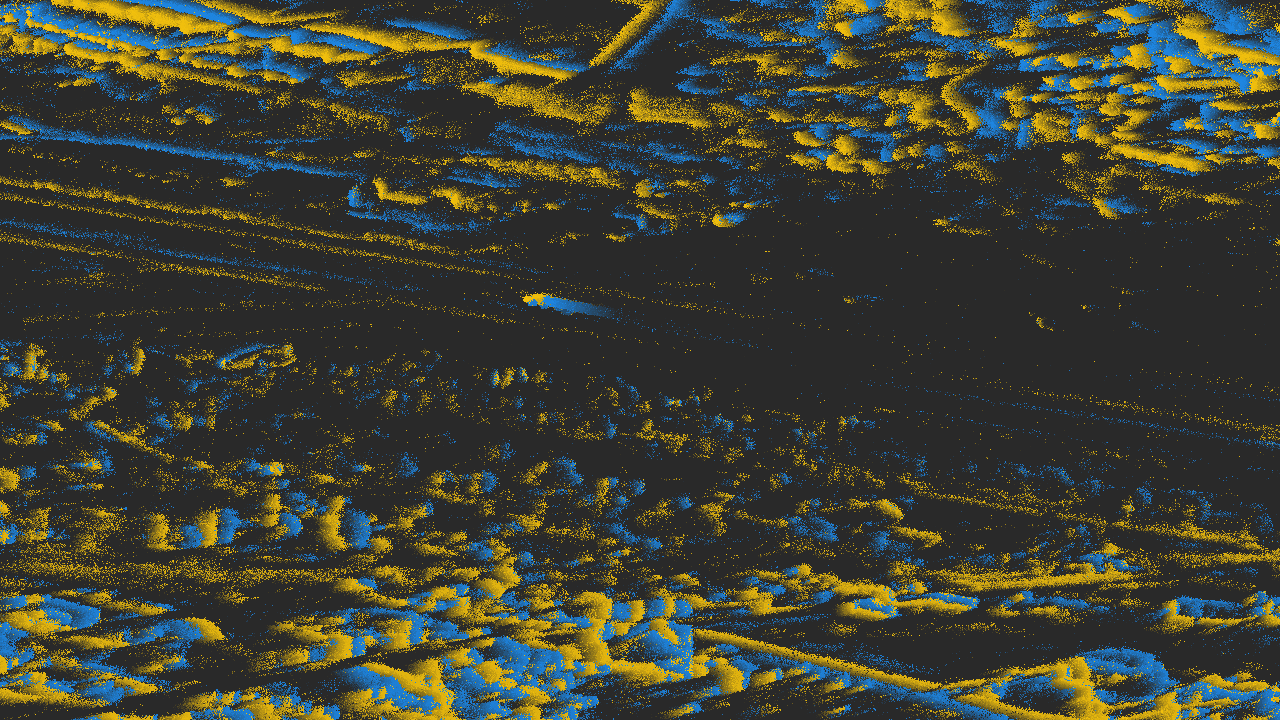}
    & \includegraphics[height=0.45in,width=0.5in]{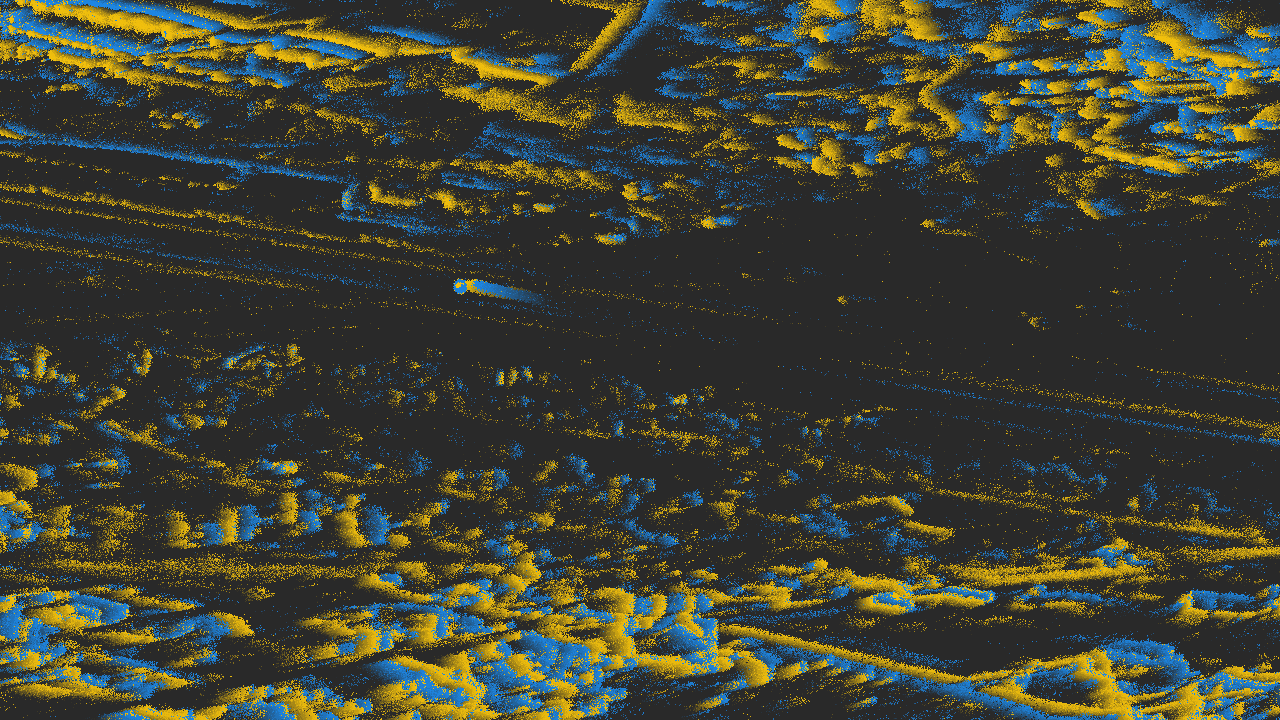}
    & \includegraphics[height=0.45in,width=0.5in]{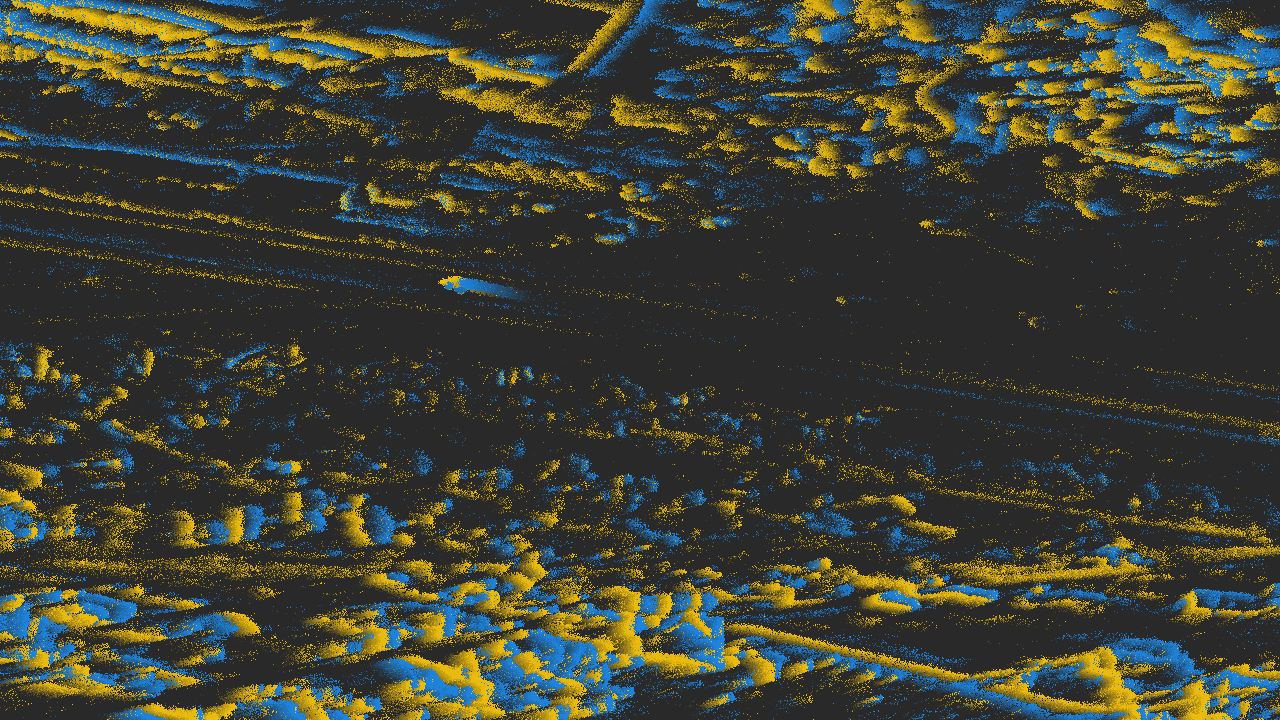}
    & \includegraphics[height=0.45in,width=0.5in]{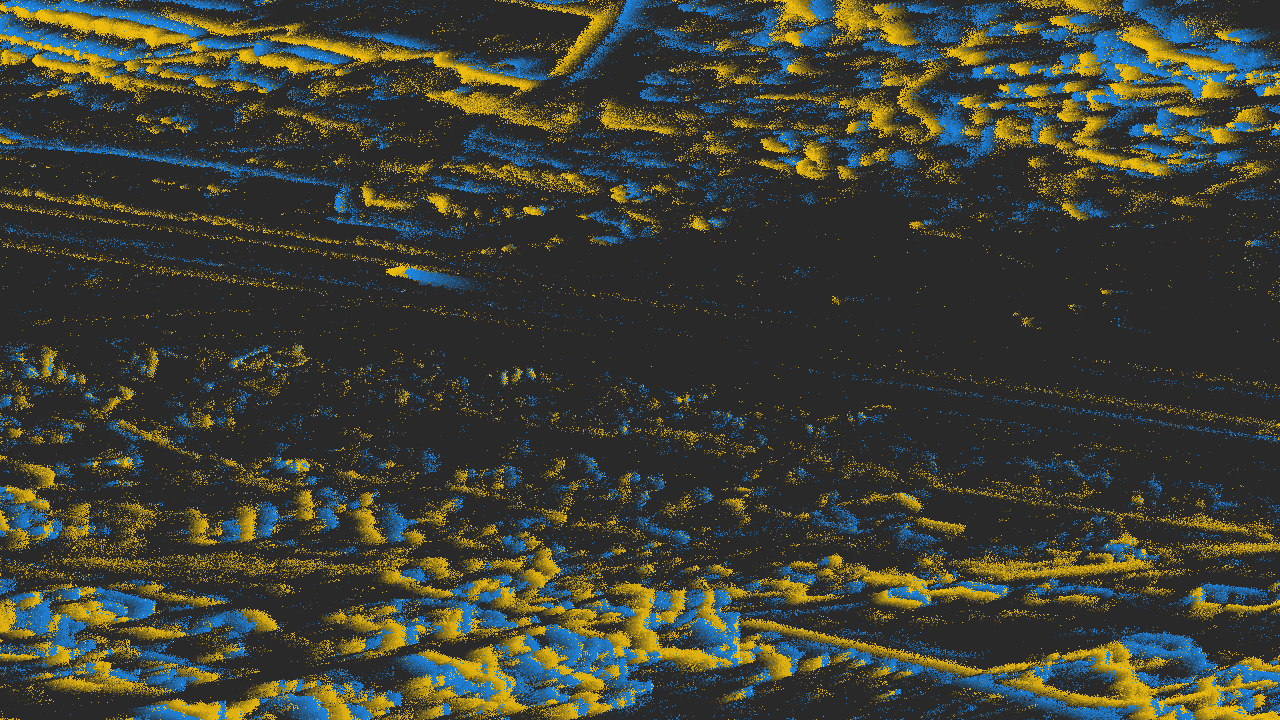}\\

    & \rotatebox{90}{\hspace{0.2cm}\color{gray!90}SM}\includegraphics[height=0.45in,width=0.5in]{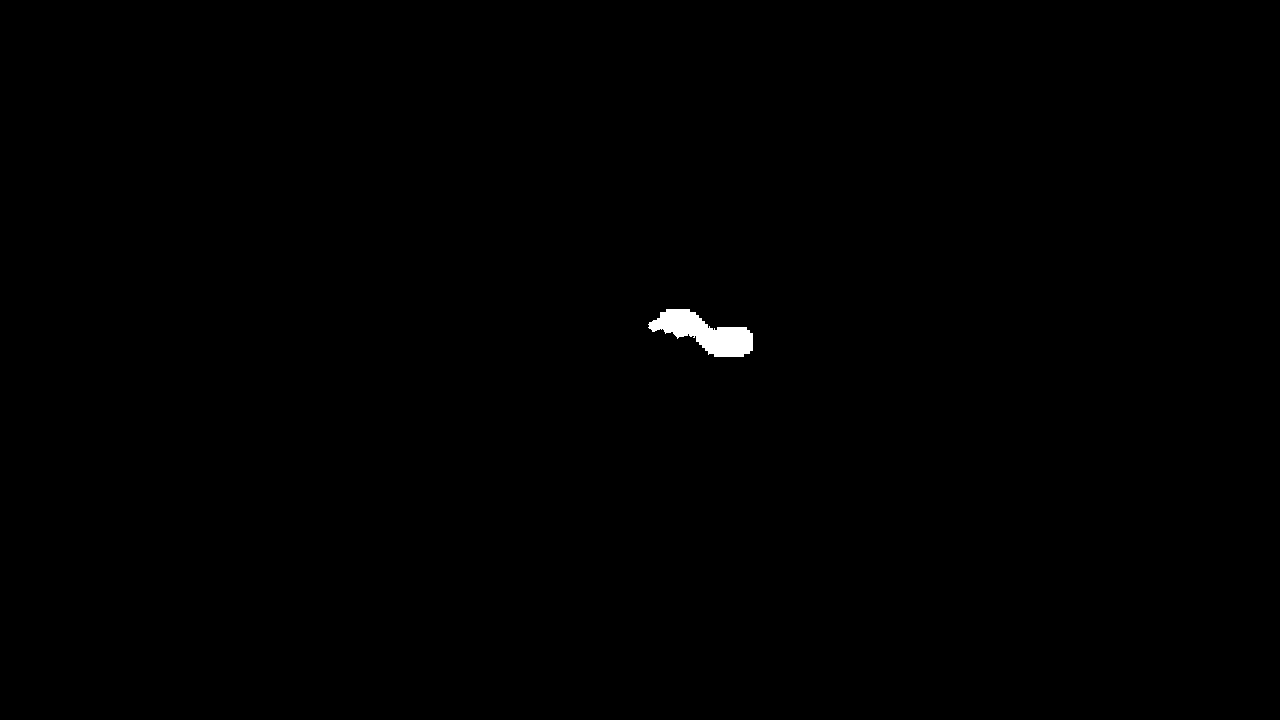}
    & \includegraphics[height=0.45in,width=0.5in]{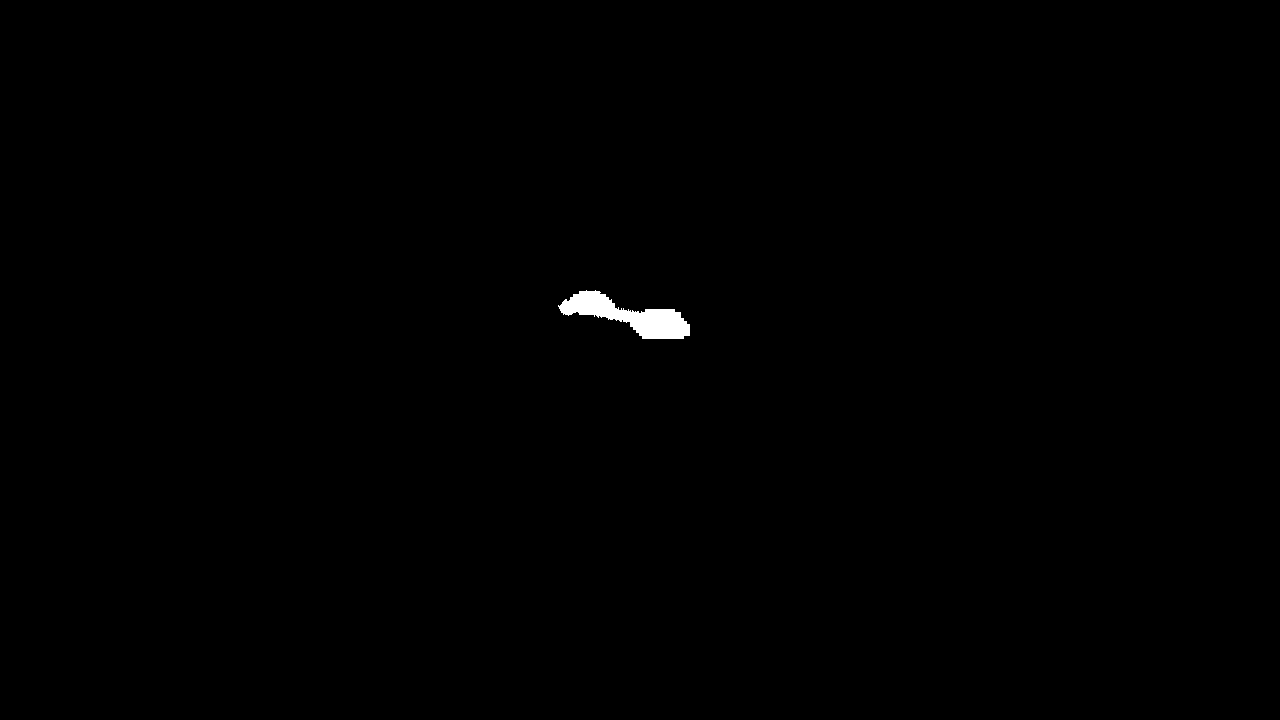}
    & \includegraphics[height=0.45in,width=0.5in]{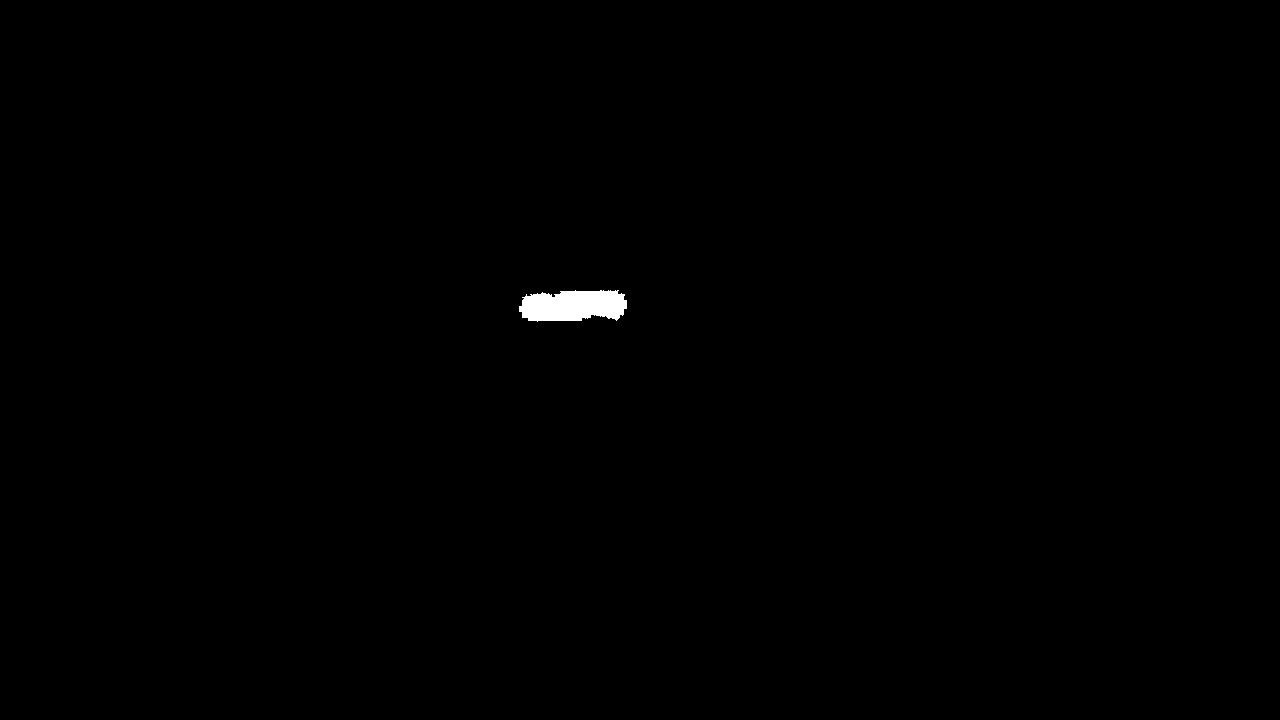}
    & \includegraphics[height=0.45in,width=0.5in]{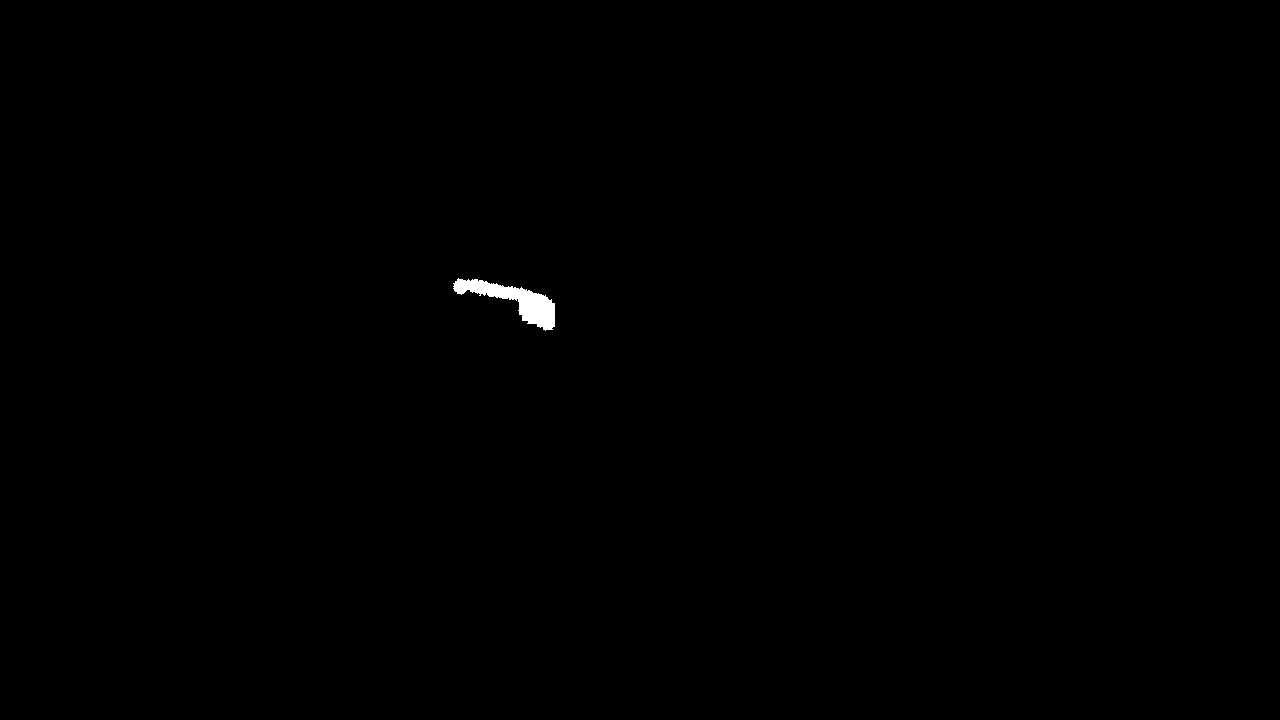}
    & \fcolorbox{red}{white}{\includegraphics[height=0.45in,width=0.5in]{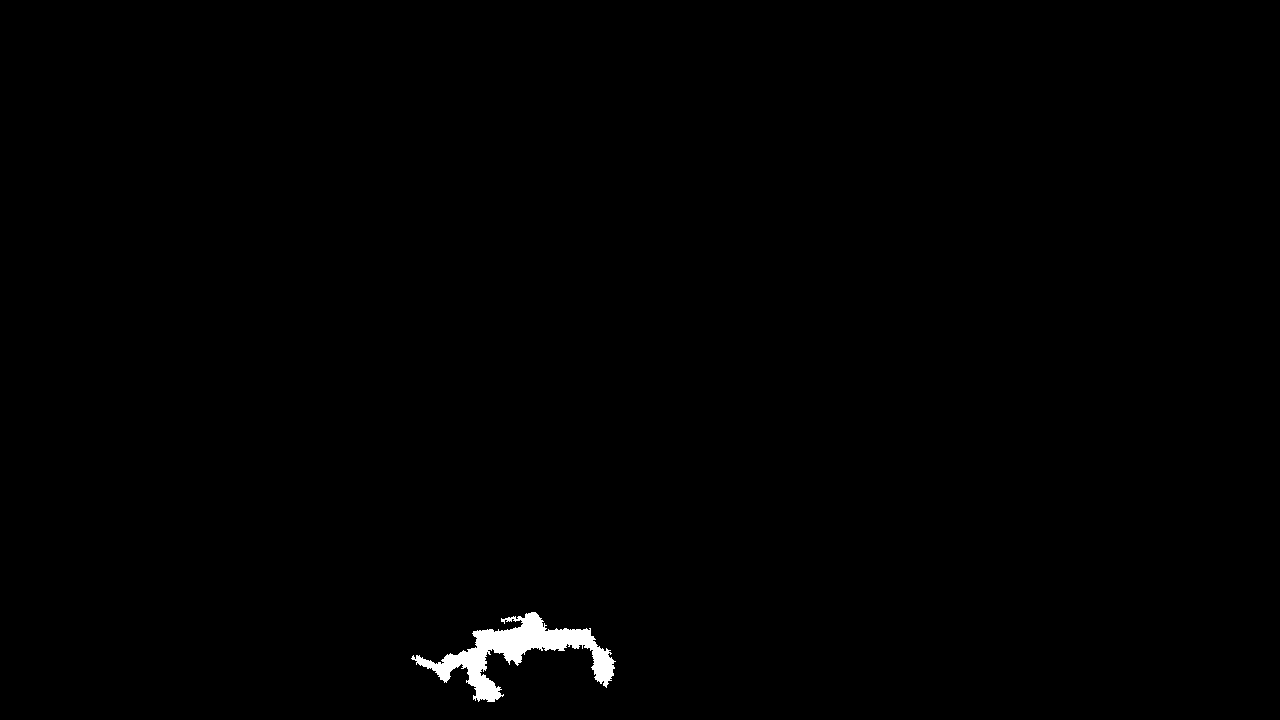}}
    & \fcolorbox{red}{white}{\includegraphics[height=0.45in,width=0.5in]{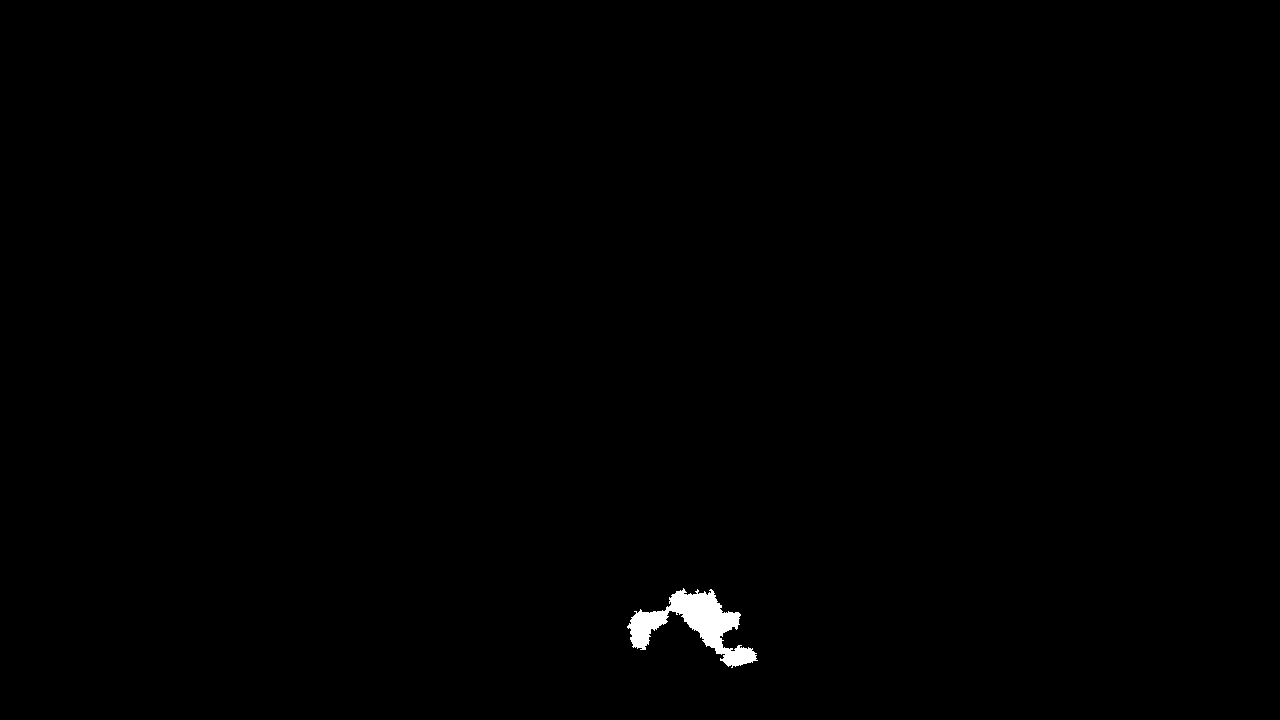}}\\

    & \rotatebox{90}{\hspace{0.1cm}\color{red!100}DMR}\includegraphics[height=0.45in,width=0.5in]{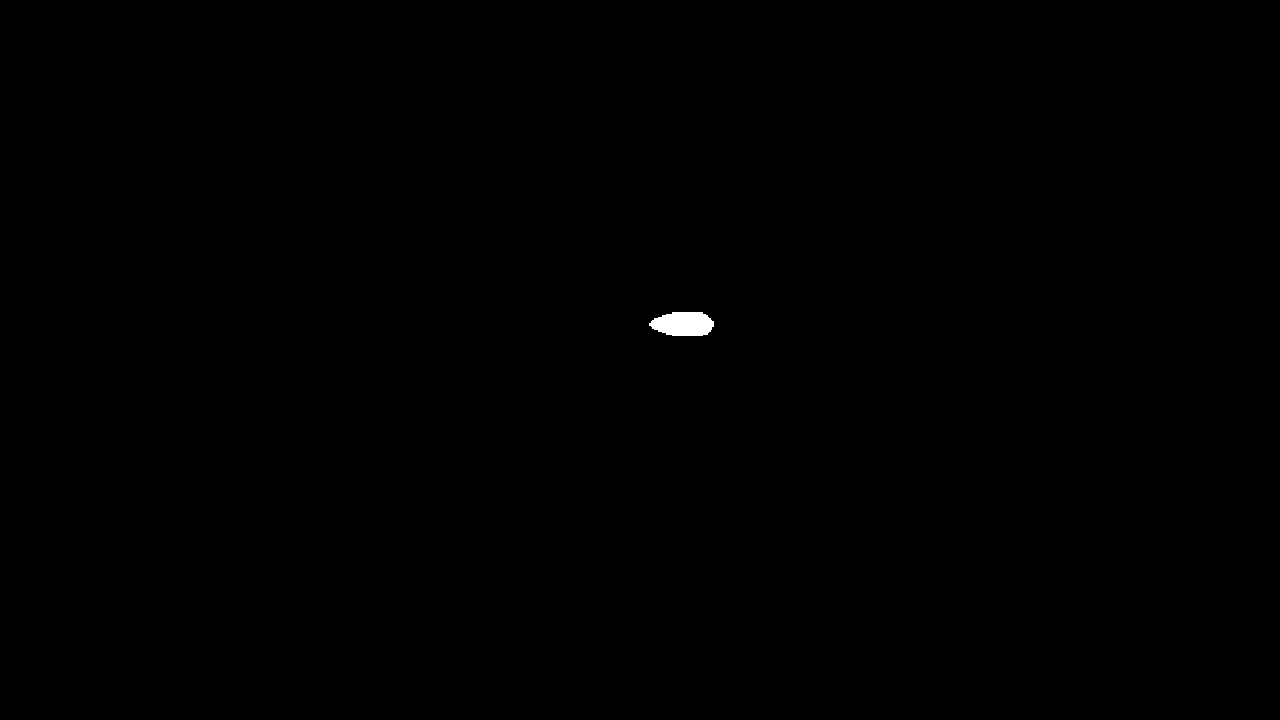}
    & \includegraphics[height=0.45in,width=0.5in]{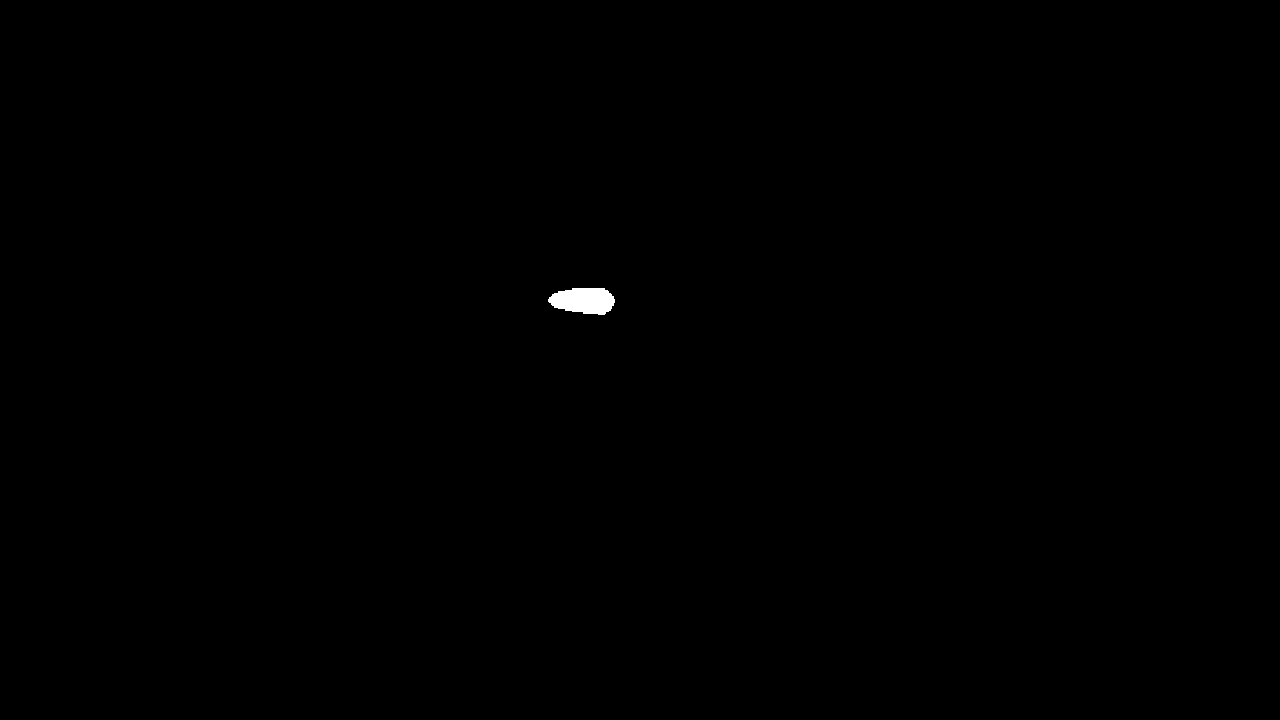}
    & \includegraphics[height=0.45in,width=0.5in]{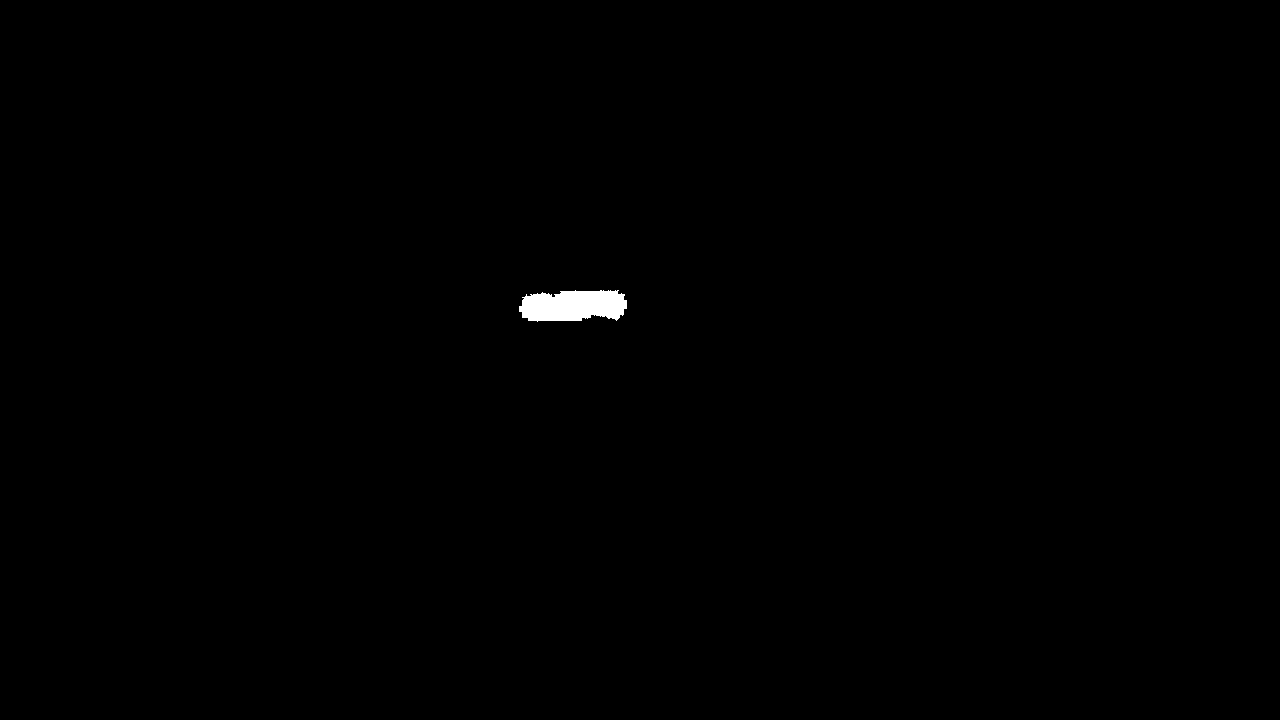}
    & \includegraphics[height=0.45in,width=0.5in]{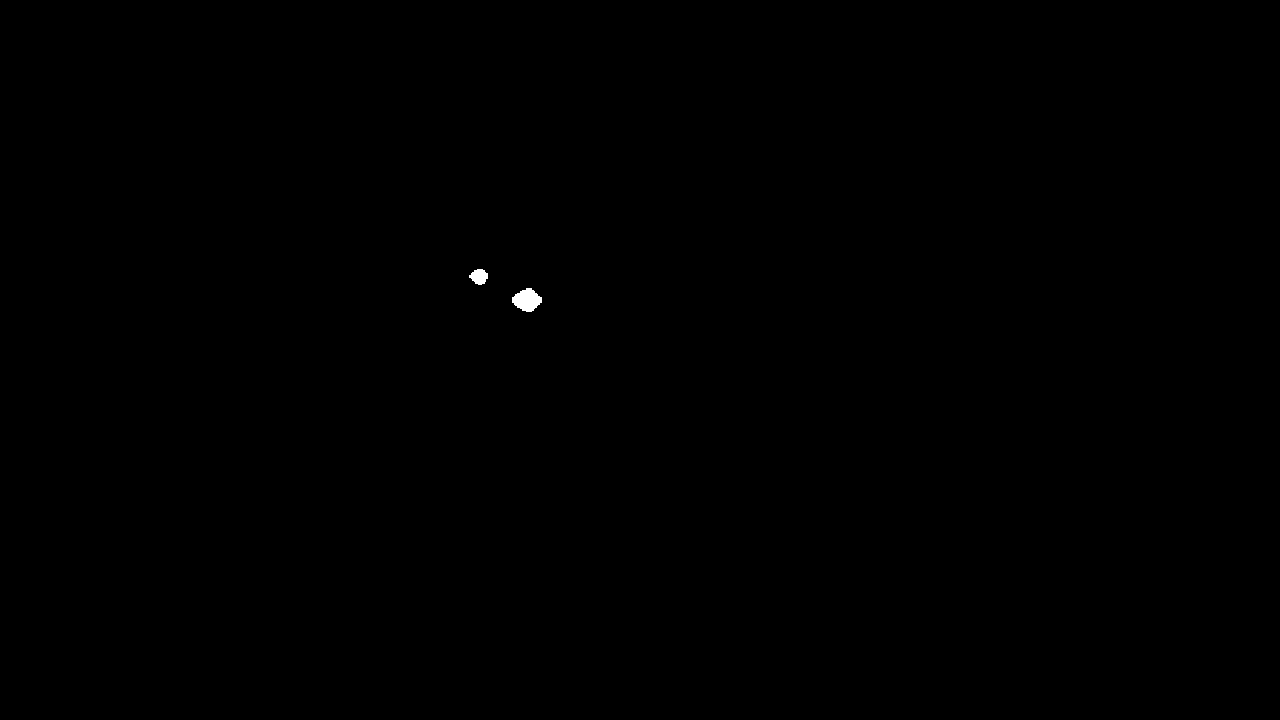}
    & \fcolorbox{red}{white}{\includegraphics[height=0.45in,width=0.5in]{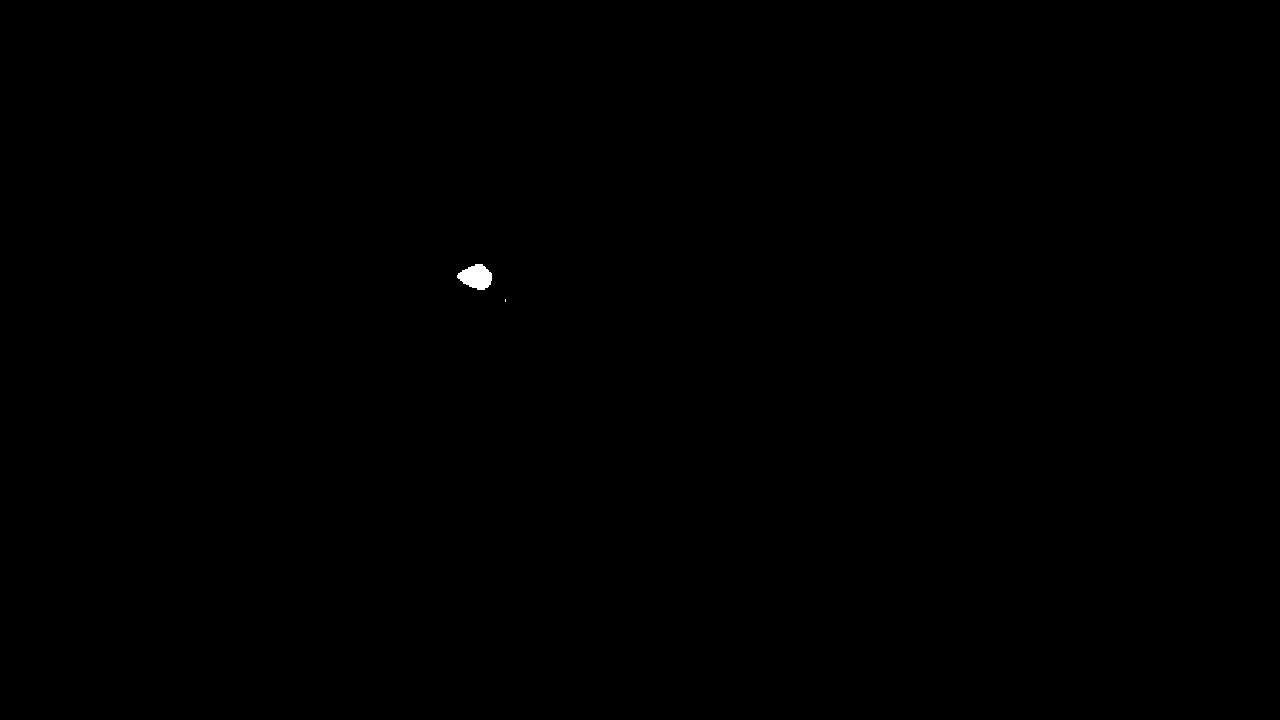}}
    & \fcolorbox{red}{white}{\includegraphics[height=0.45in,width=0.5in]{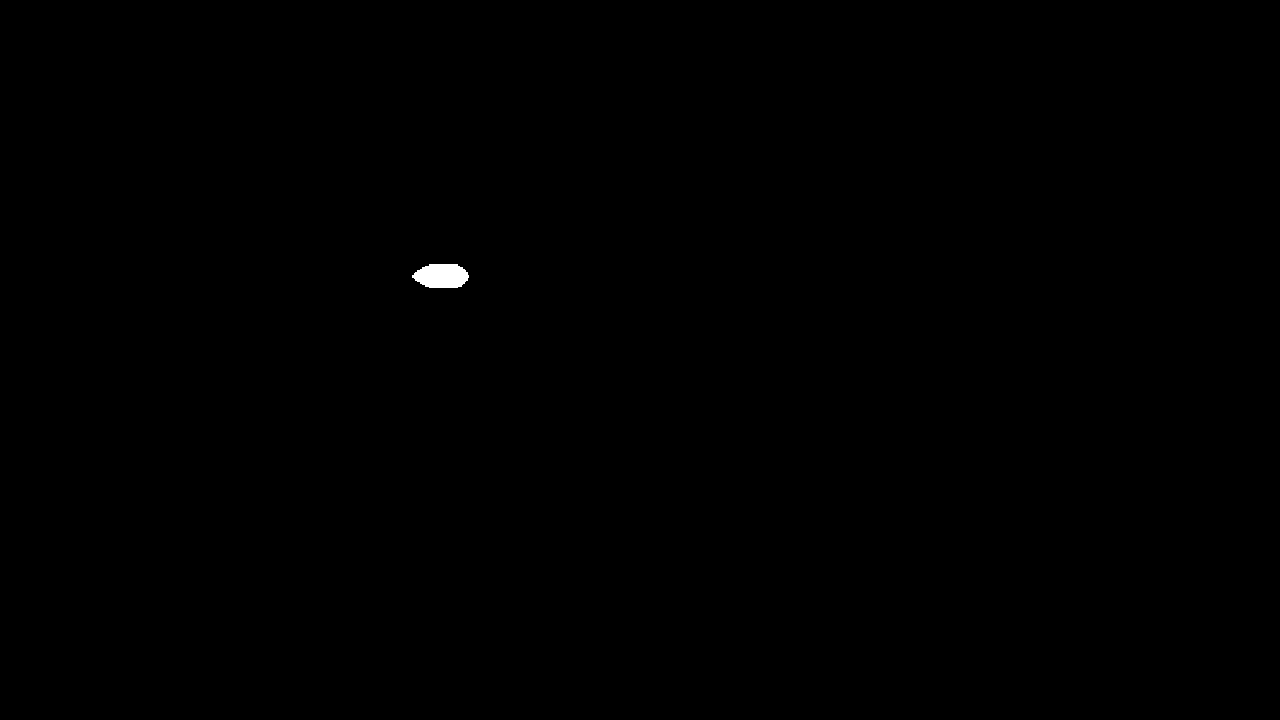}}\\
\end{tabular}
\caption{Qualitative results from the DMR process showing its temporal consistency overtime. The red bounding boxes highlight the frames where the DMR successfully recover the salient mask of the moving objects. SM refers to salient masks.}
\label{tb:evairbornepedestrians}
\end{figure}

\begin{figure}[h] 
\centering
\setlength{\fboxrule}{1.0pt}
\setlength{\fboxsep}{0pt}
\setlength{\tabcolsep}{1pt} %
\renewcommand{\arraystretch}{0.5} %

\textbf{Ev-Airborne: Golf car high oblique}
\begin{tabular}{c c c c c c c c c c c c}
    & \rotatebox{90}{\hspace{0.5cm}\color{gray!90}Input} \hspace{-2mm}\includegraphics[height=0.6in,width=0.65in]{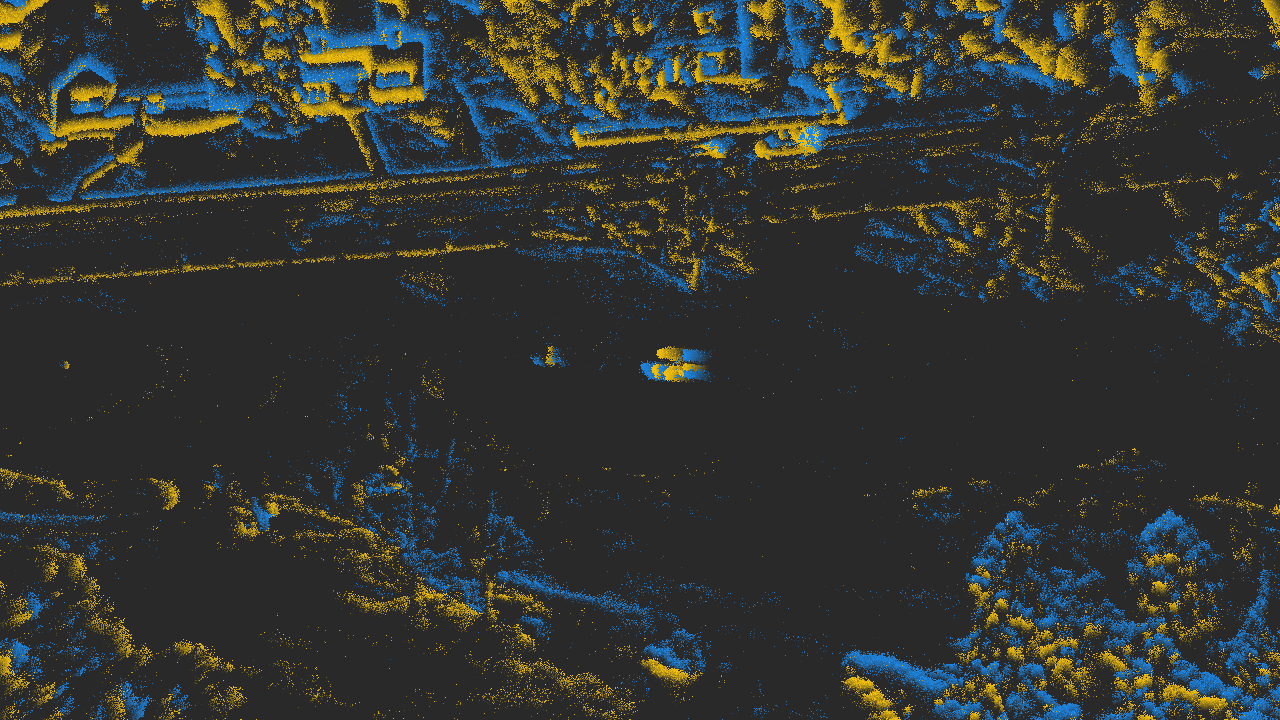}
    & \includegraphics[height=0.6in,width=0.65in]{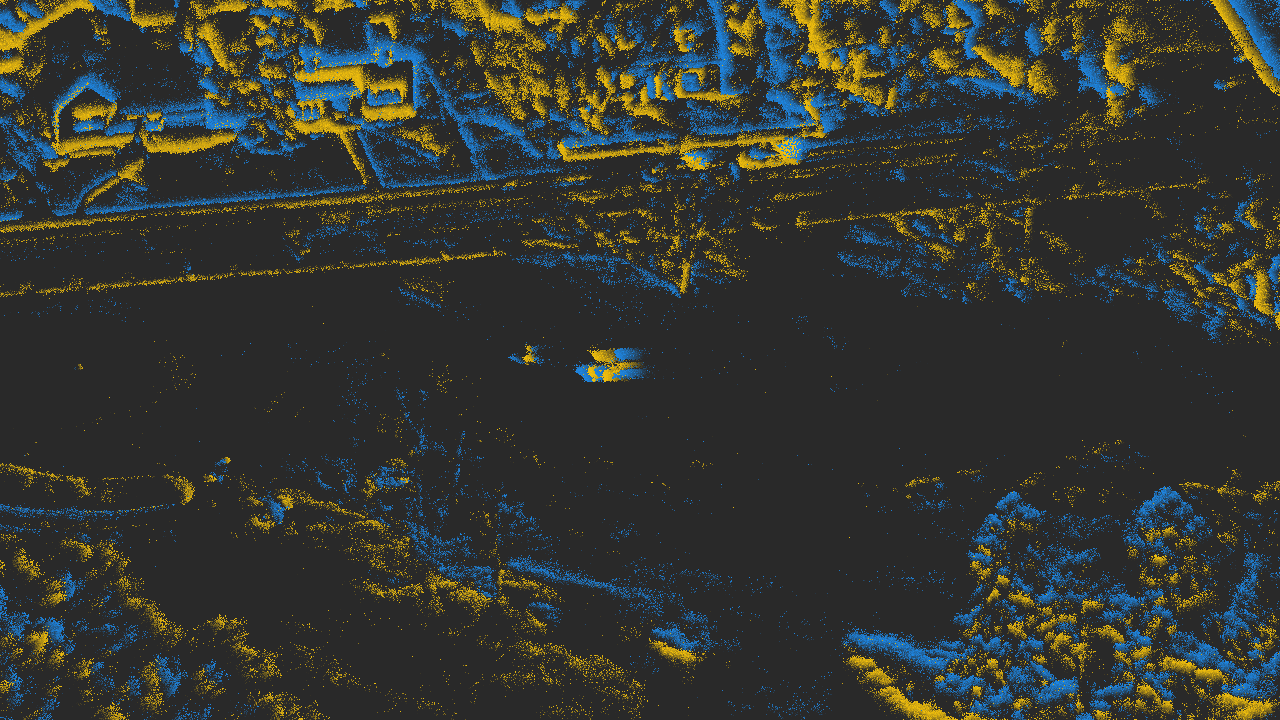}
    & \includegraphics[height=0.6in,width=0.65in]{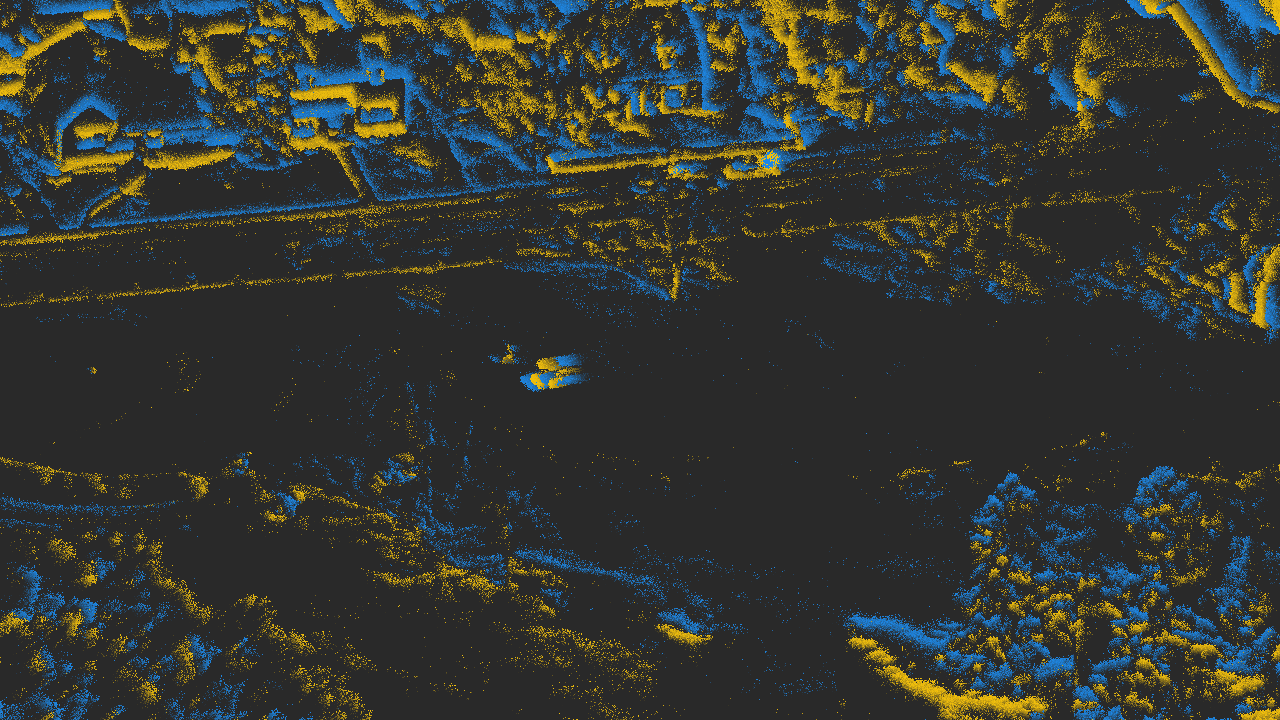}
    & \includegraphics[height=0.6in,width=0.65in]{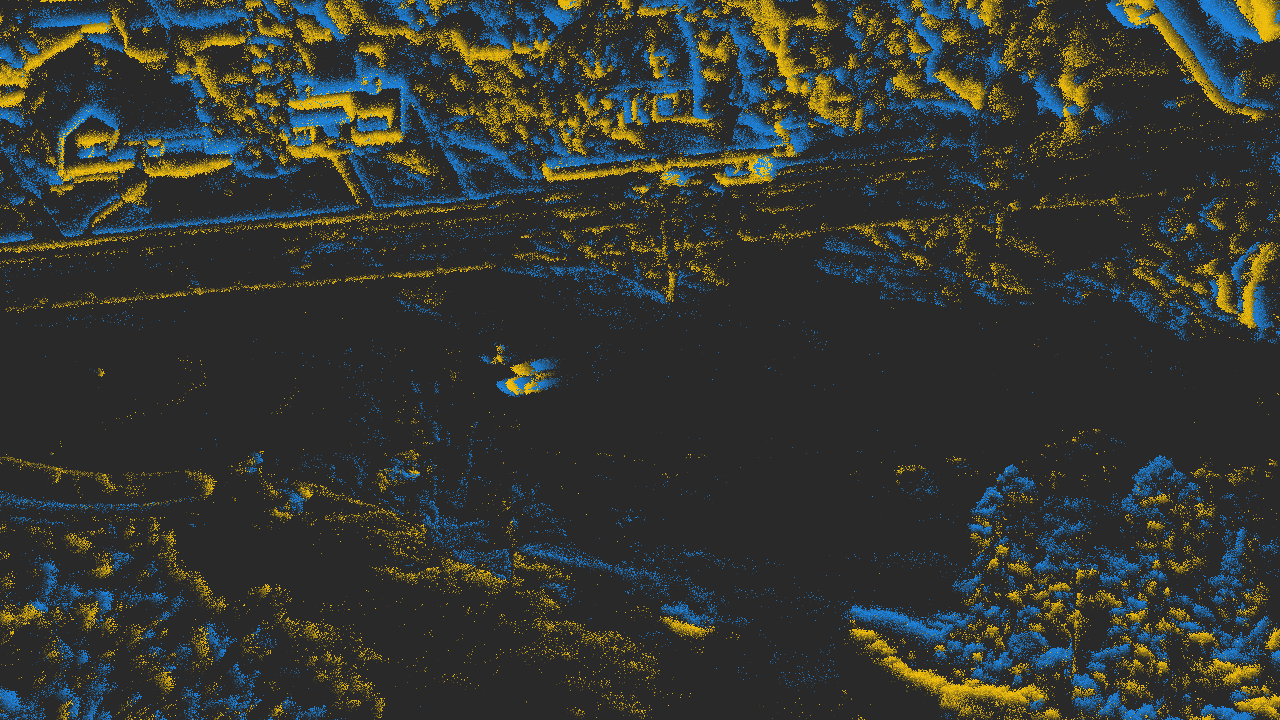}
    & \includegraphics[height=0.6in,width=0.65in]{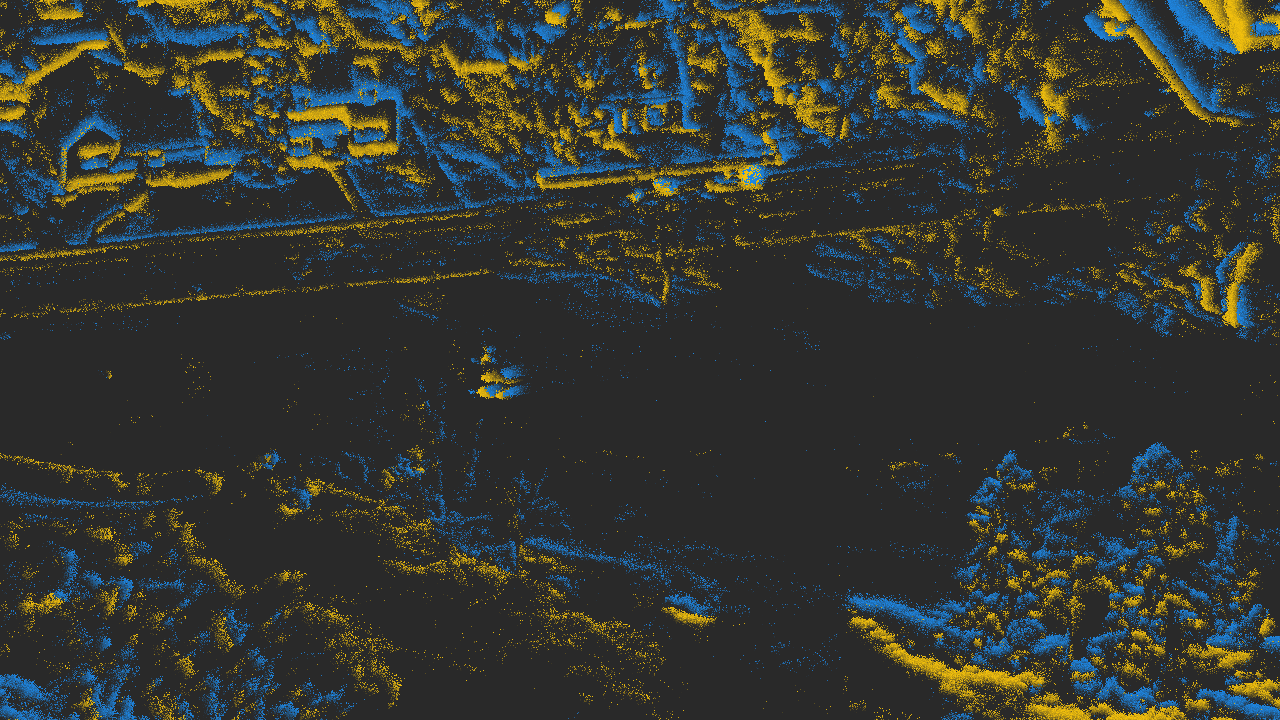}\\

    & \rotatebox{90}{\hspace{0.50cm}\color{gray!90}CM}\includegraphics[height=0.6in,width=0.65in]{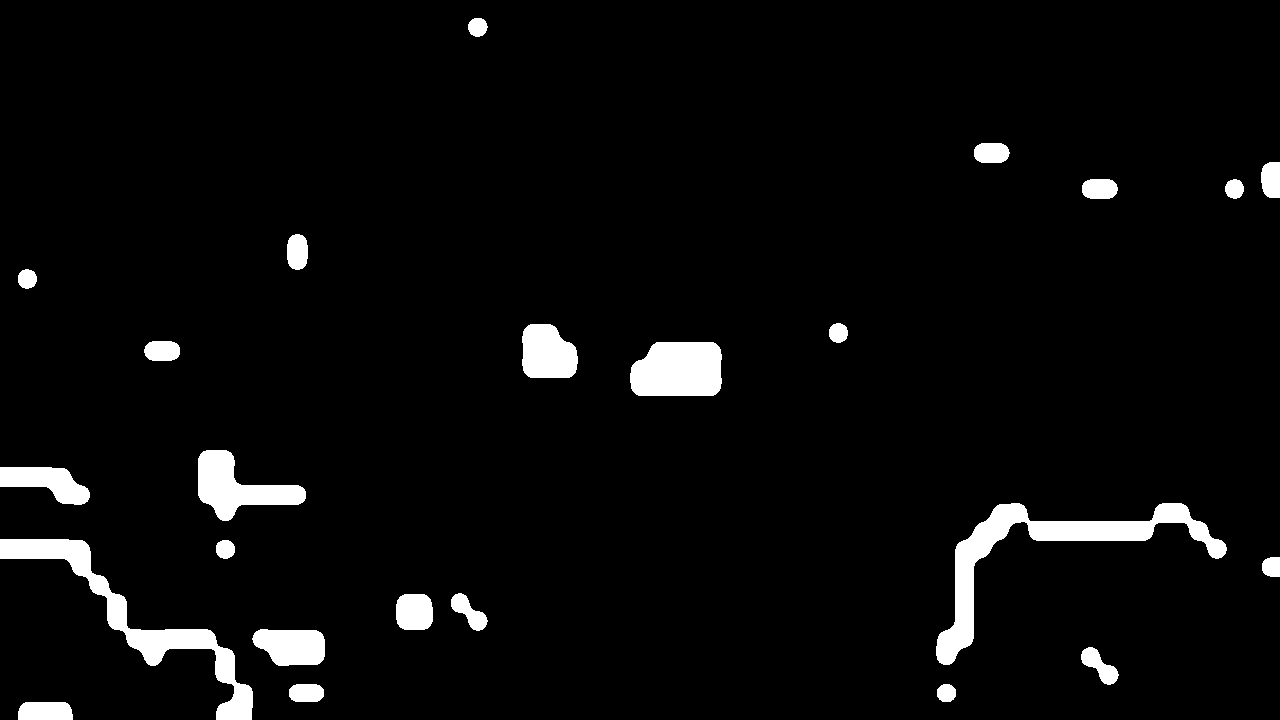}
    & \includegraphics[height=0.6in,width=0.65in]{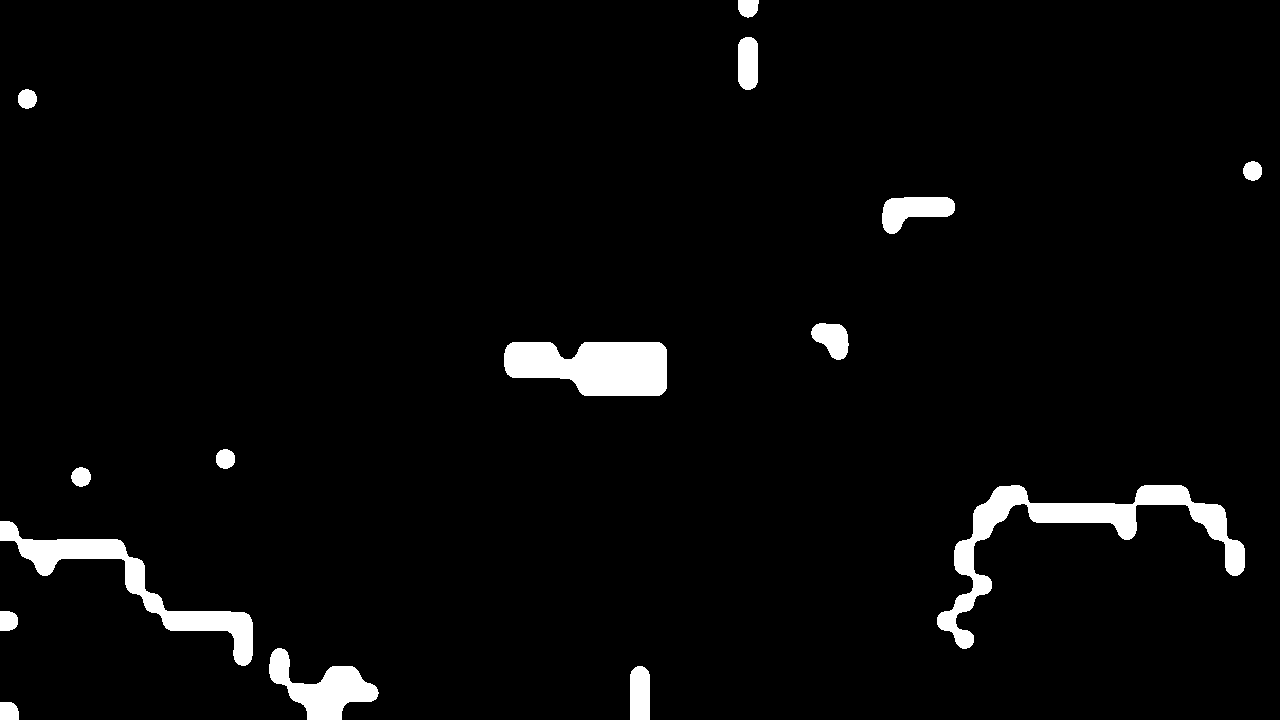}
    & \includegraphics[height=0.6in,width=0.65in]{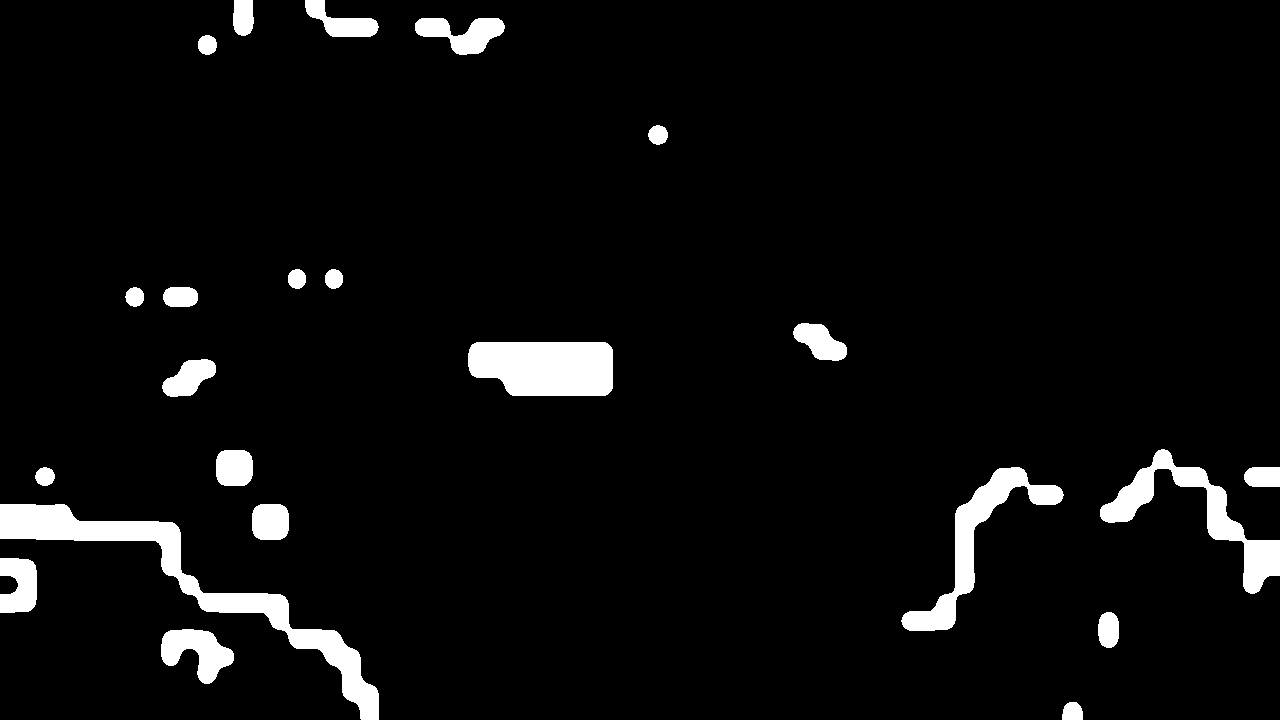}
    & \includegraphics[height=0.6in,width=0.65in]{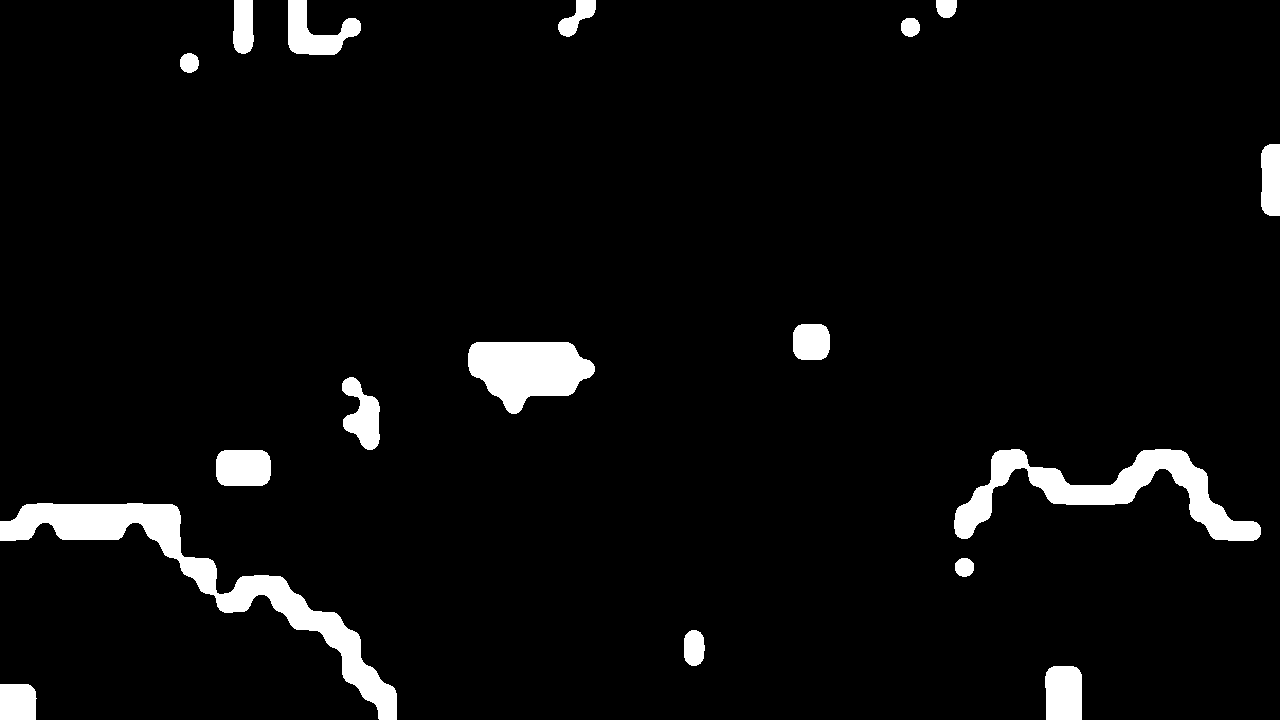}
    & \includegraphics[height=0.6in,width=0.65in]{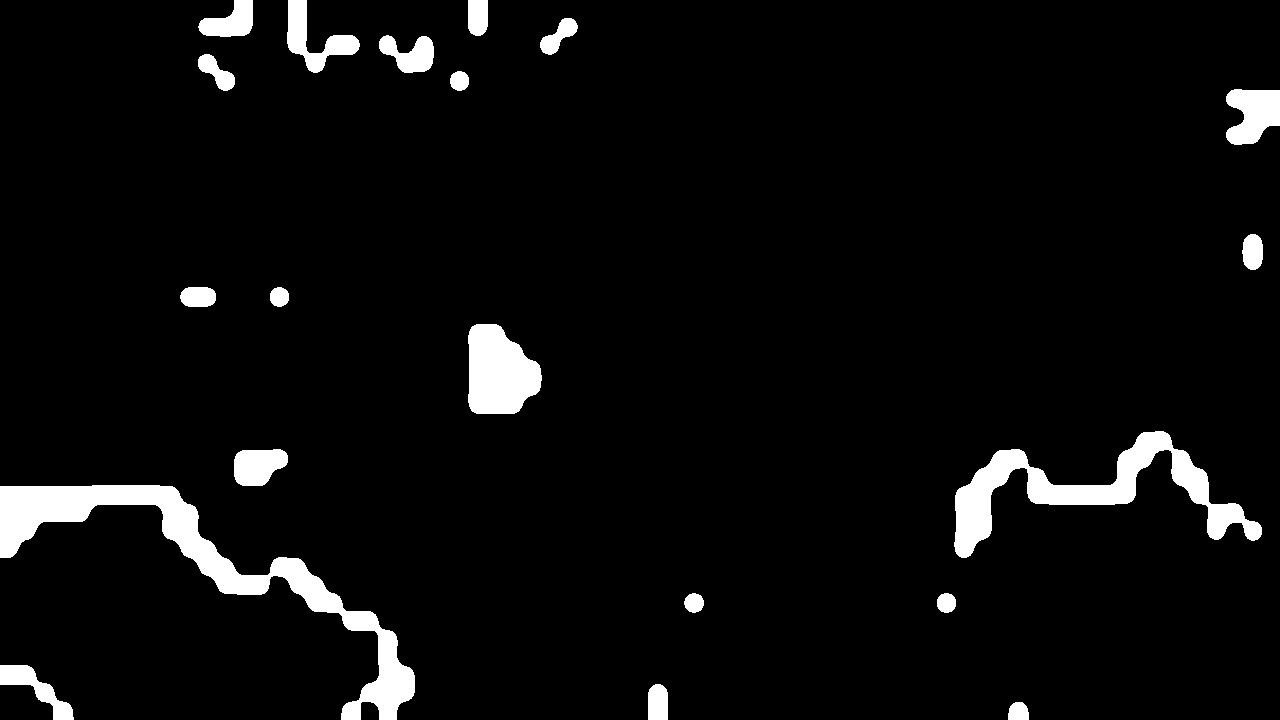}\\

    & \rotatebox{90}{\hspace{0.5cm}\color{gray!90}CRF}\includegraphics[height=0.6in,width=0.65in]{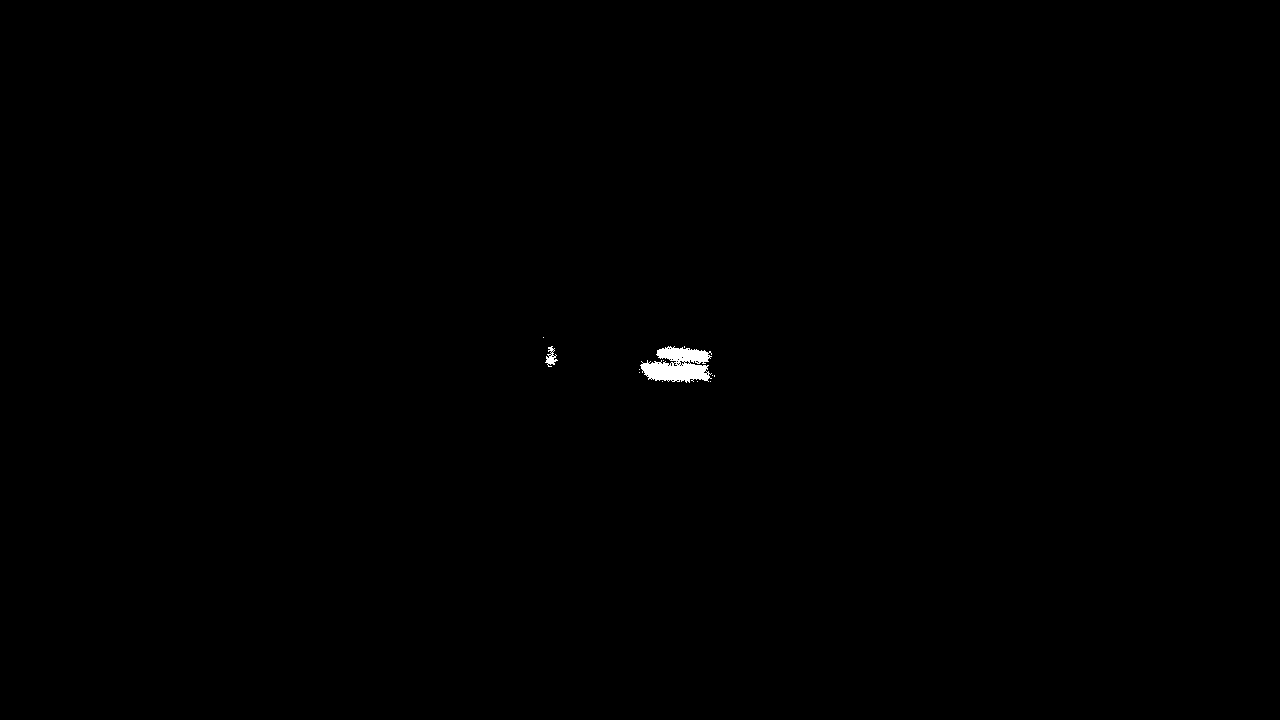}
    & \includegraphics[height=0.6in,width=0.65in]{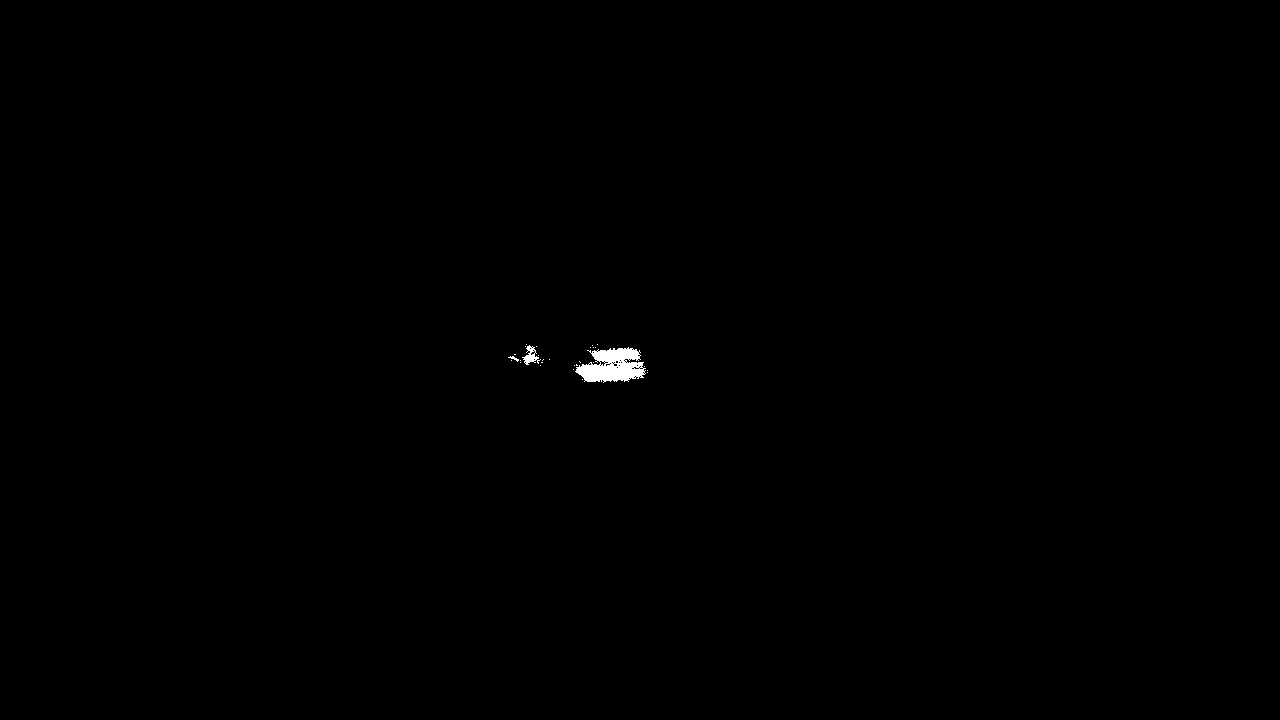}
    & \includegraphics[height=0.6in,width=0.65in]{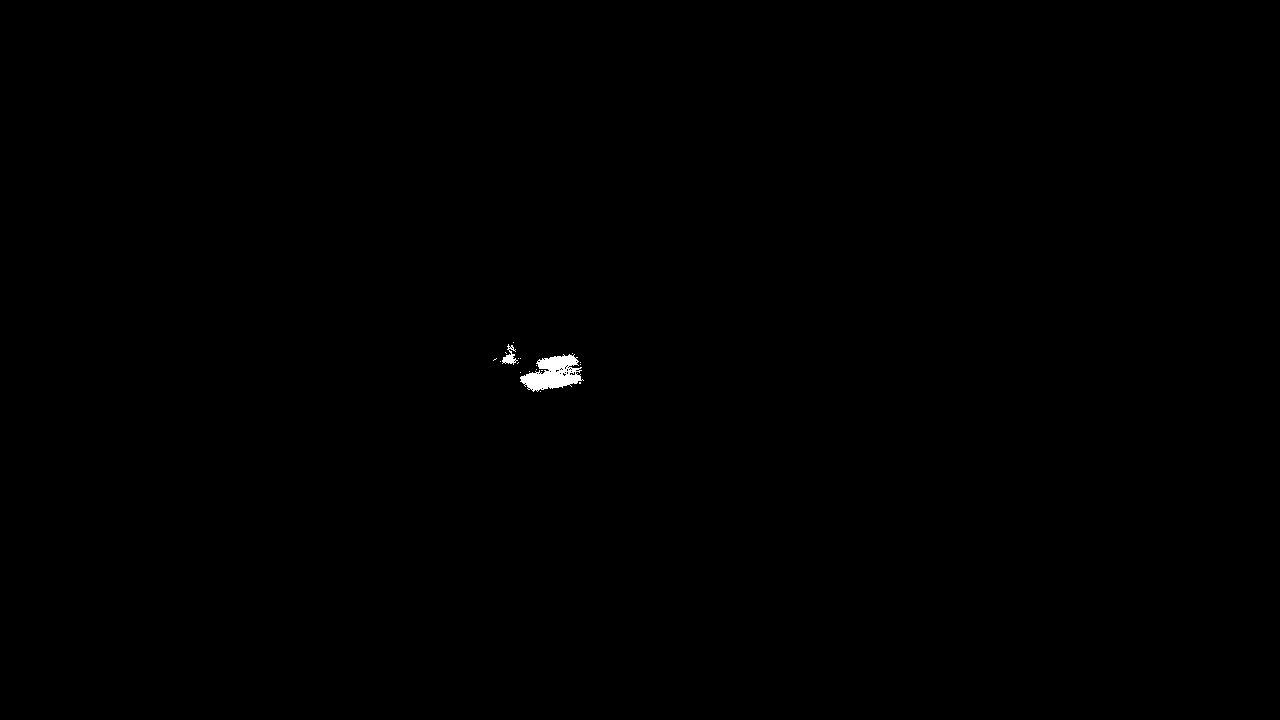}
    & \includegraphics[height=0.6in,width=0.65in]{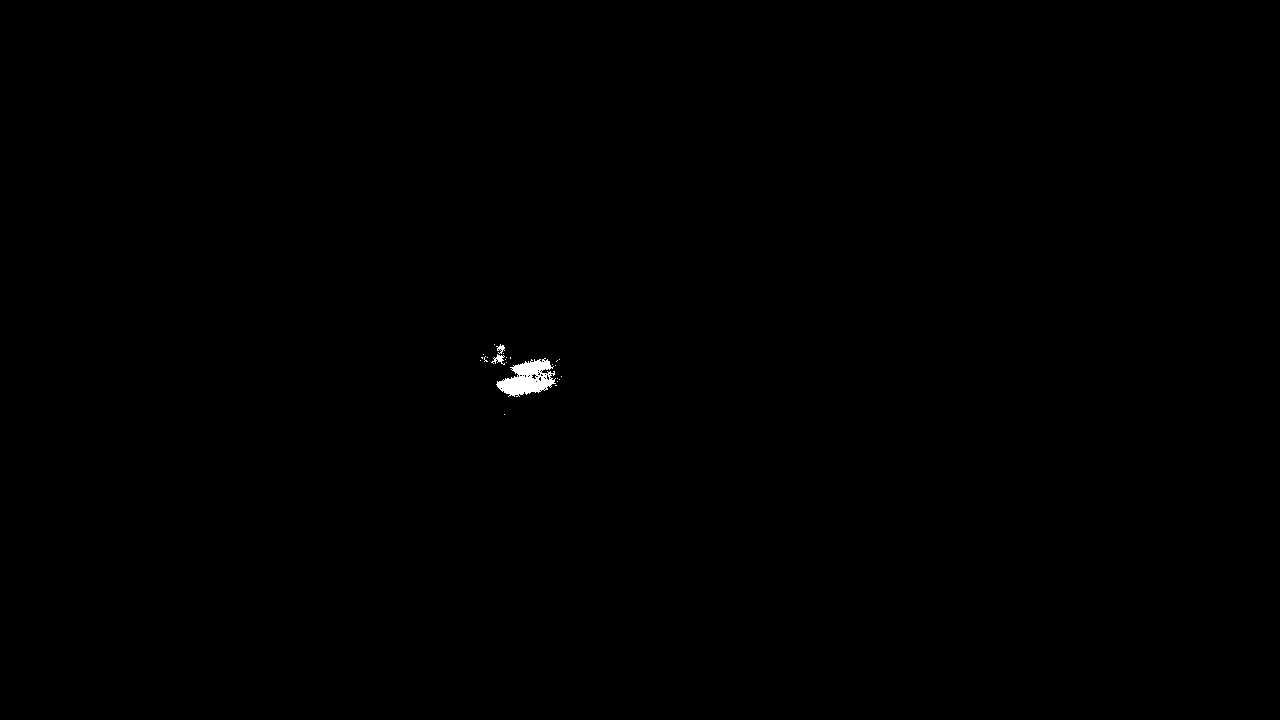}
    & \includegraphics[height=0.6in,width=0.65in]{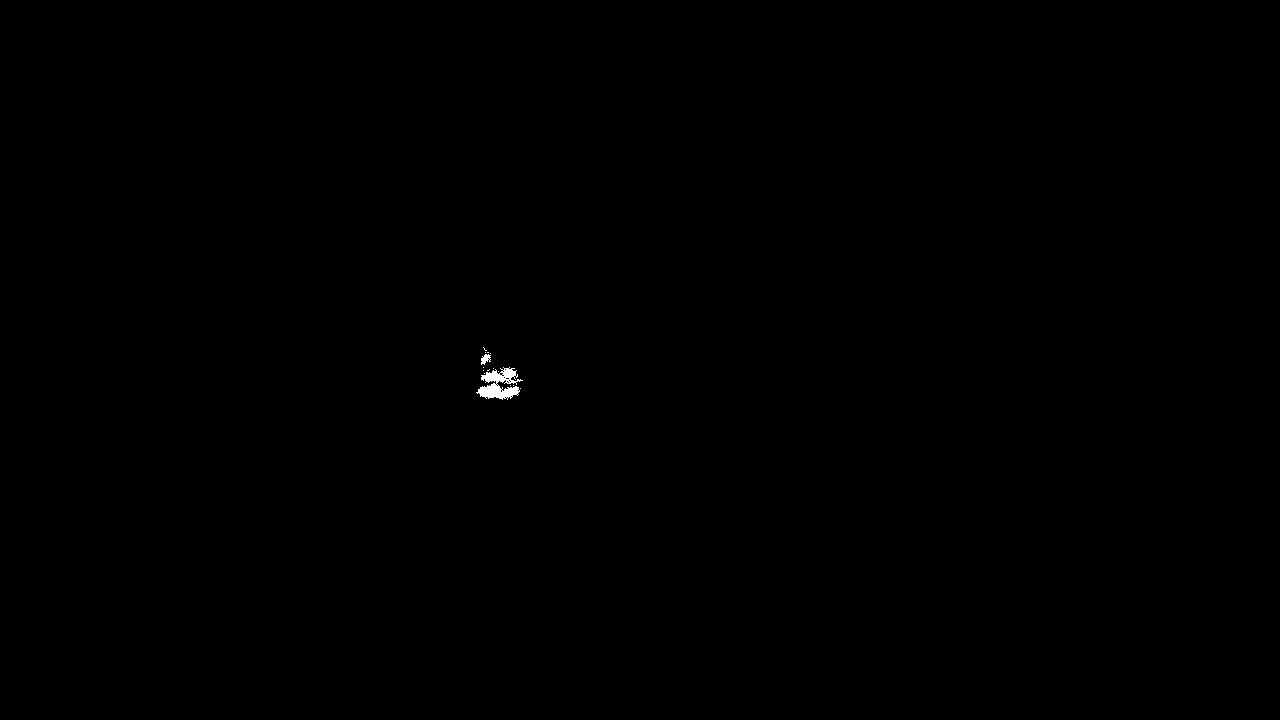}\\

    & \rotatebox{90}{\hspace{0.5cm}\color{gray!90}BS}\includegraphics[height=0.6in,width=0.65in]{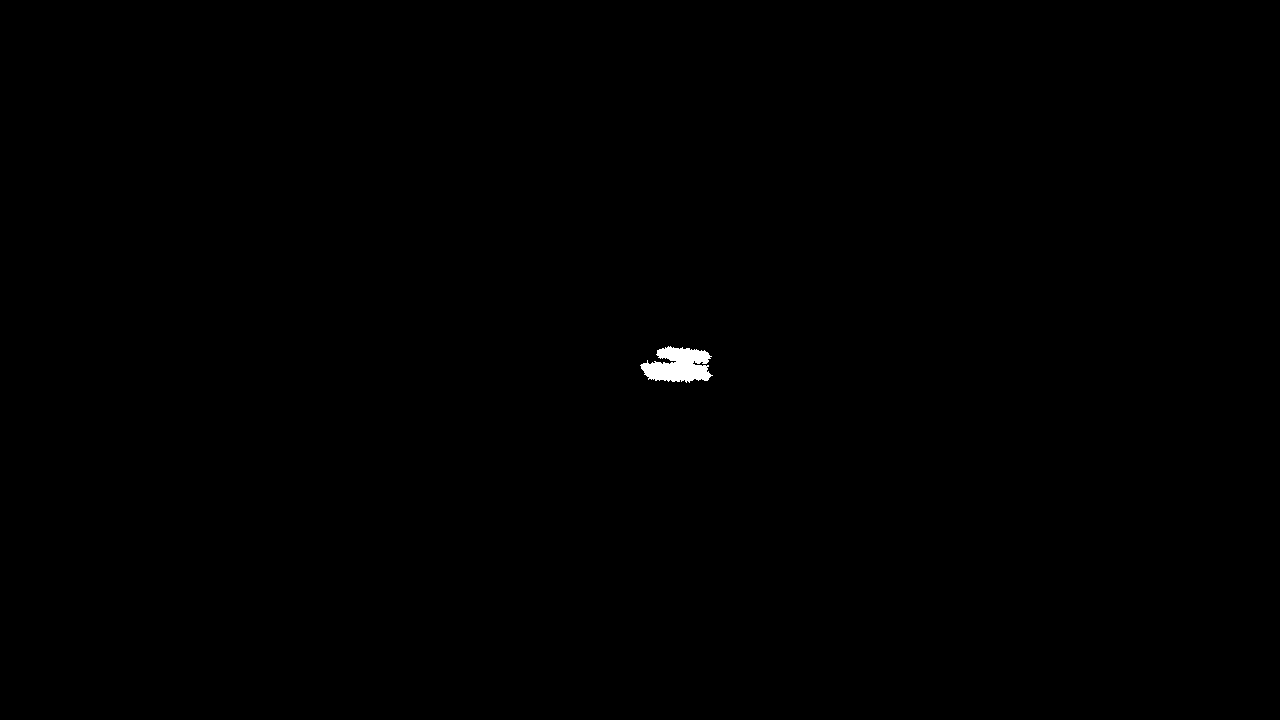}
    & \includegraphics[height=0.6in,width=0.65in]{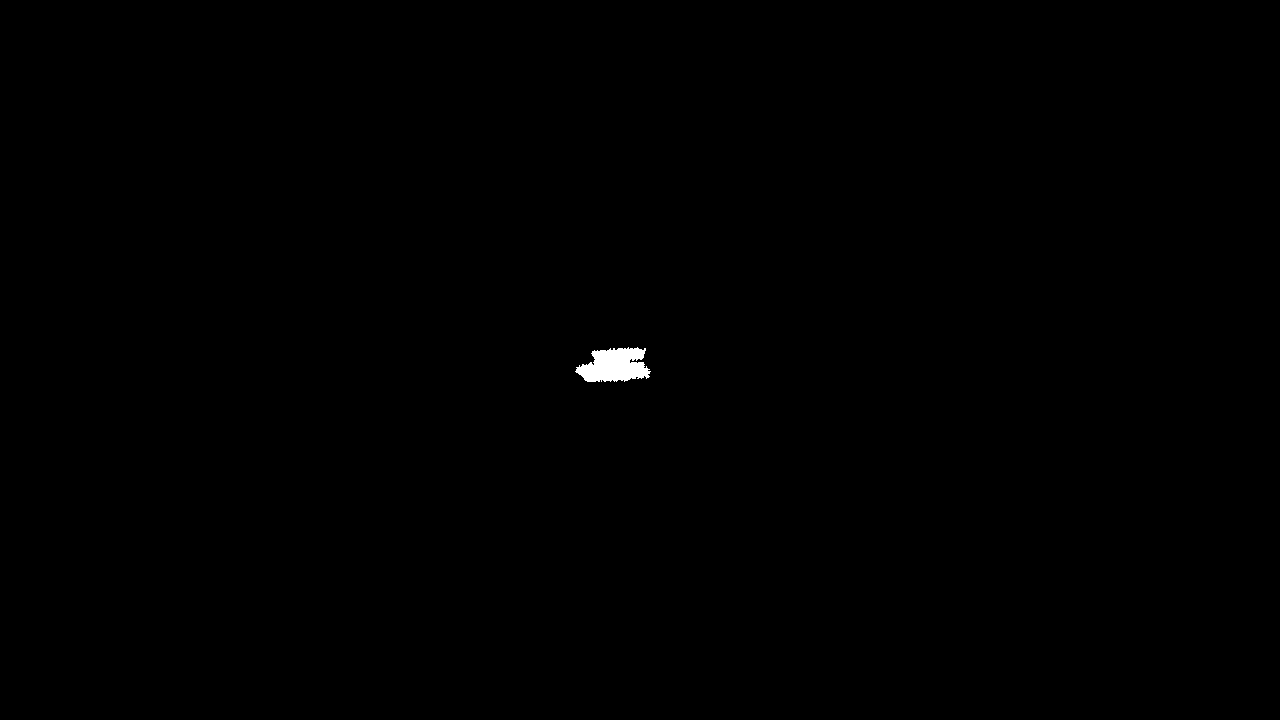}
    & \includegraphics[height=0.6in,width=0.65in]{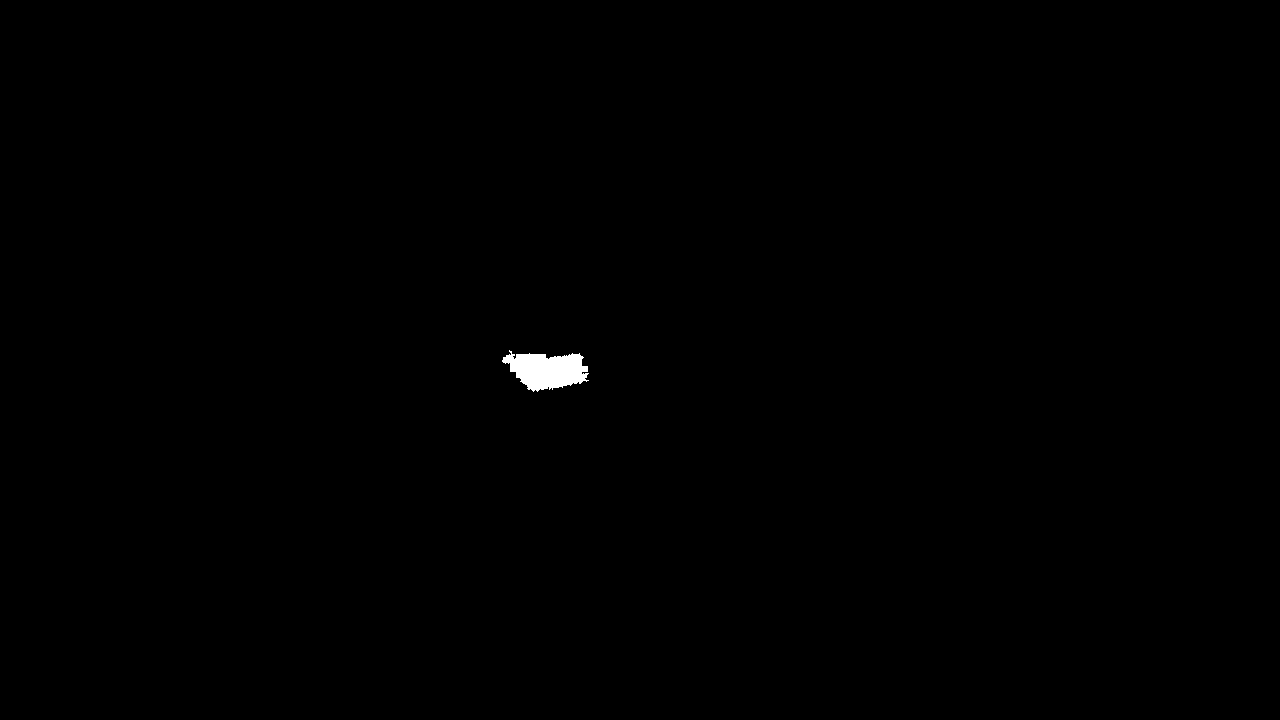}
    & \includegraphics[height=0.6in,width=0.65in]{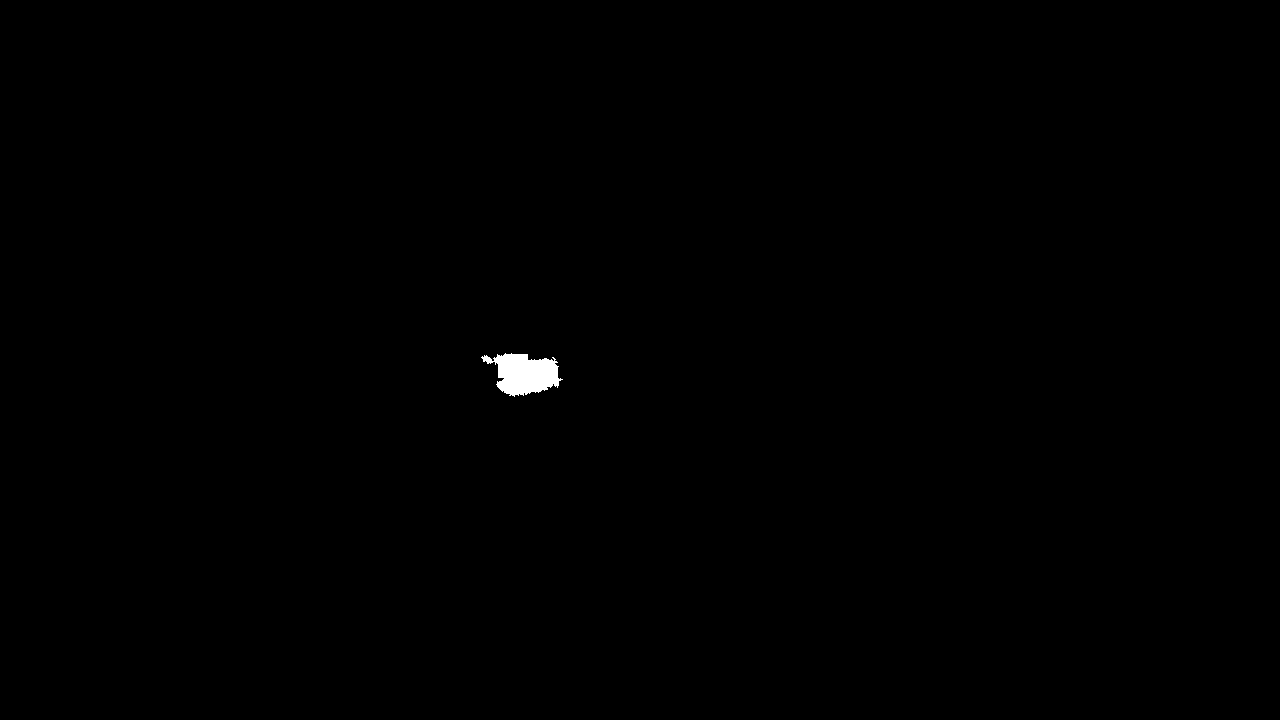}
    & \includegraphics[height=0.6in,width=0.65in]{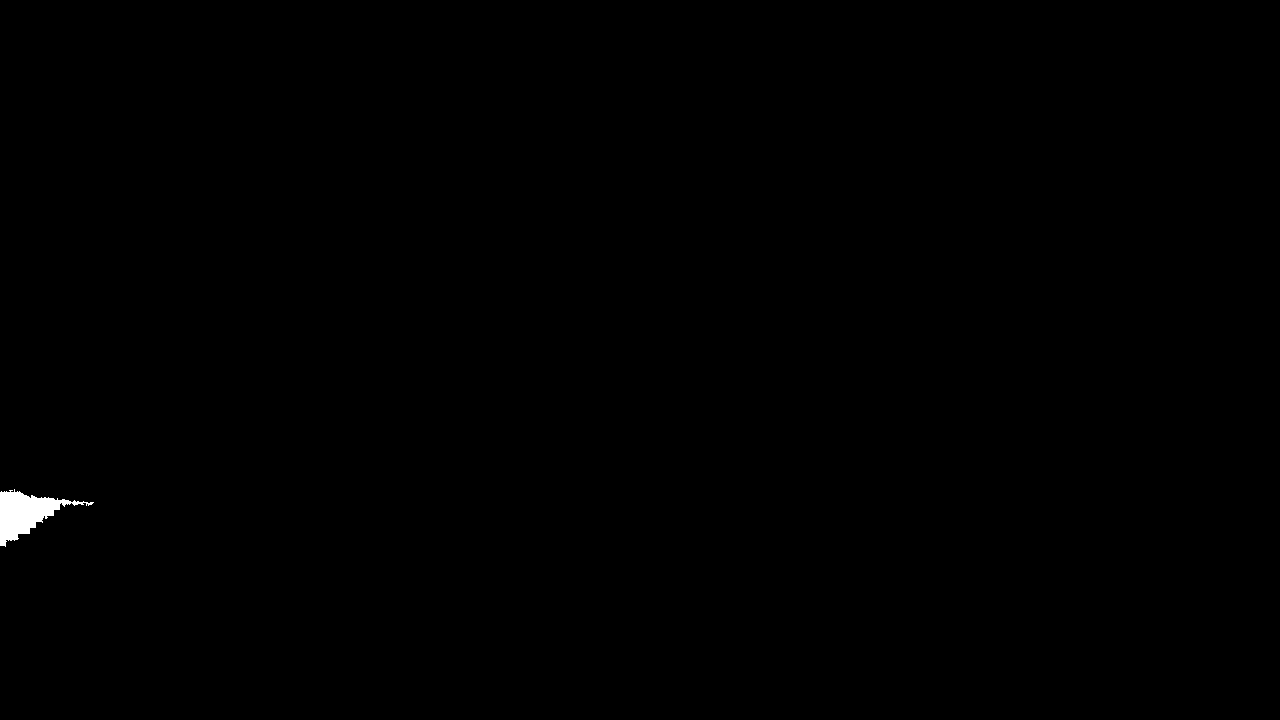}\\

    & \rotatebox{90}{\hspace{0.5cm}\color{red!100}DMR}\includegraphics[height=0.6in,width=0.65in]{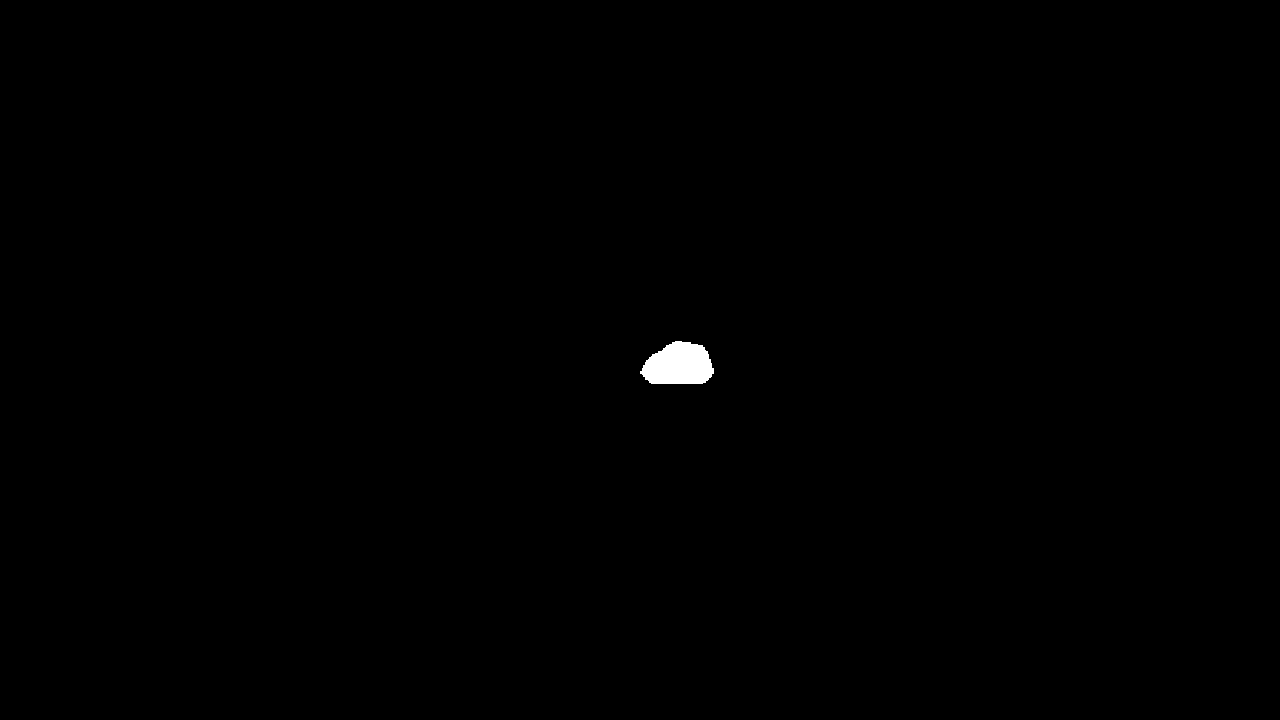}
    & \includegraphics[height=0.6in,width=0.65in]{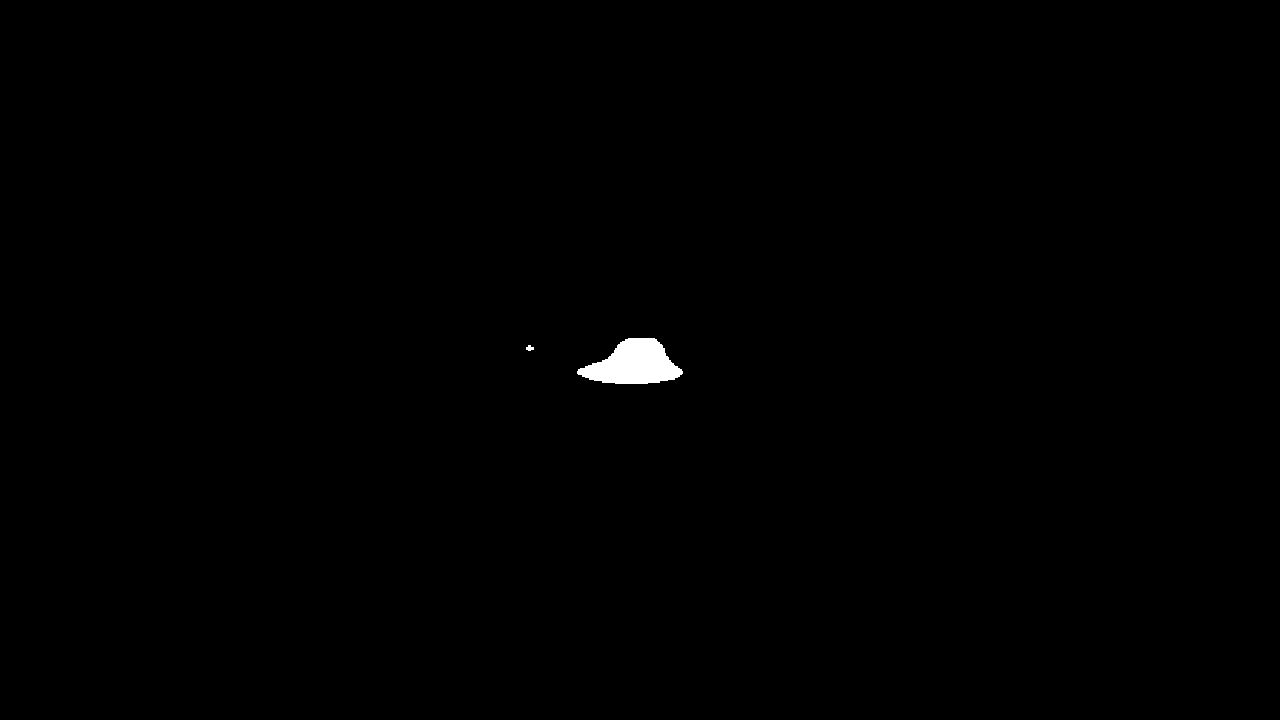}
    & \includegraphics[height=0.6in,width=0.65in]{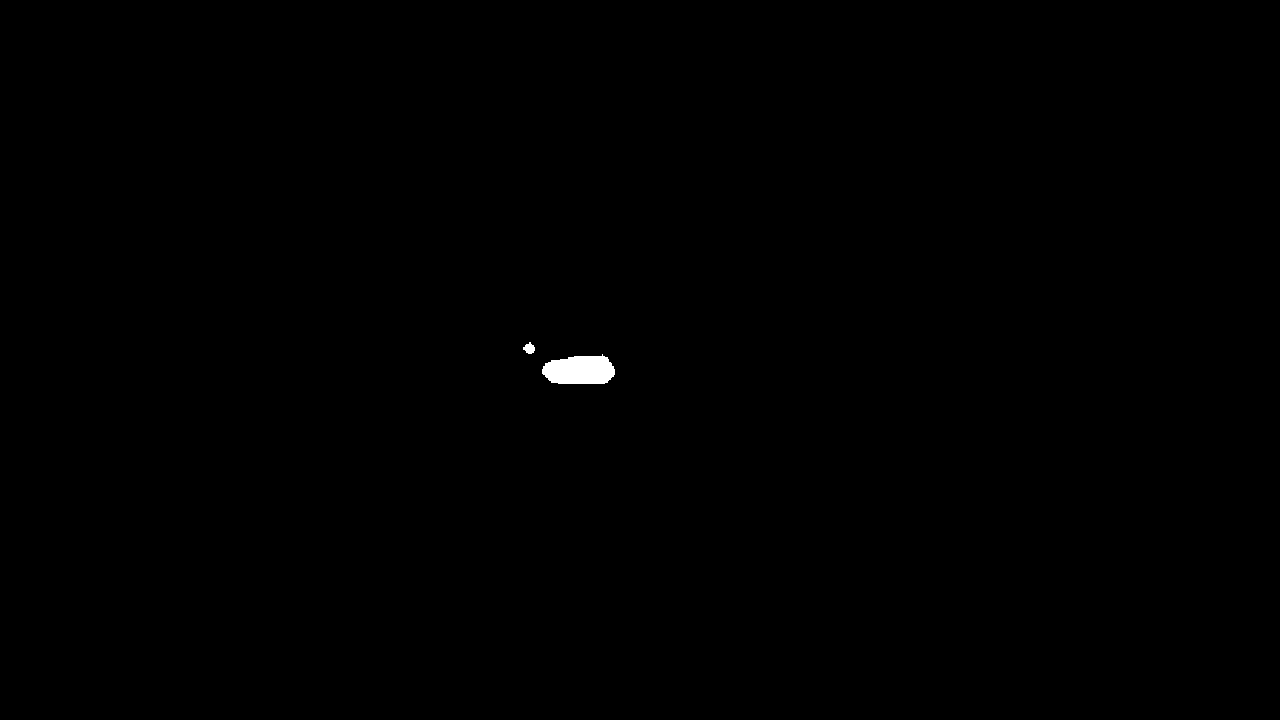}
    & \includegraphics[height=0.6in,width=0.65in]{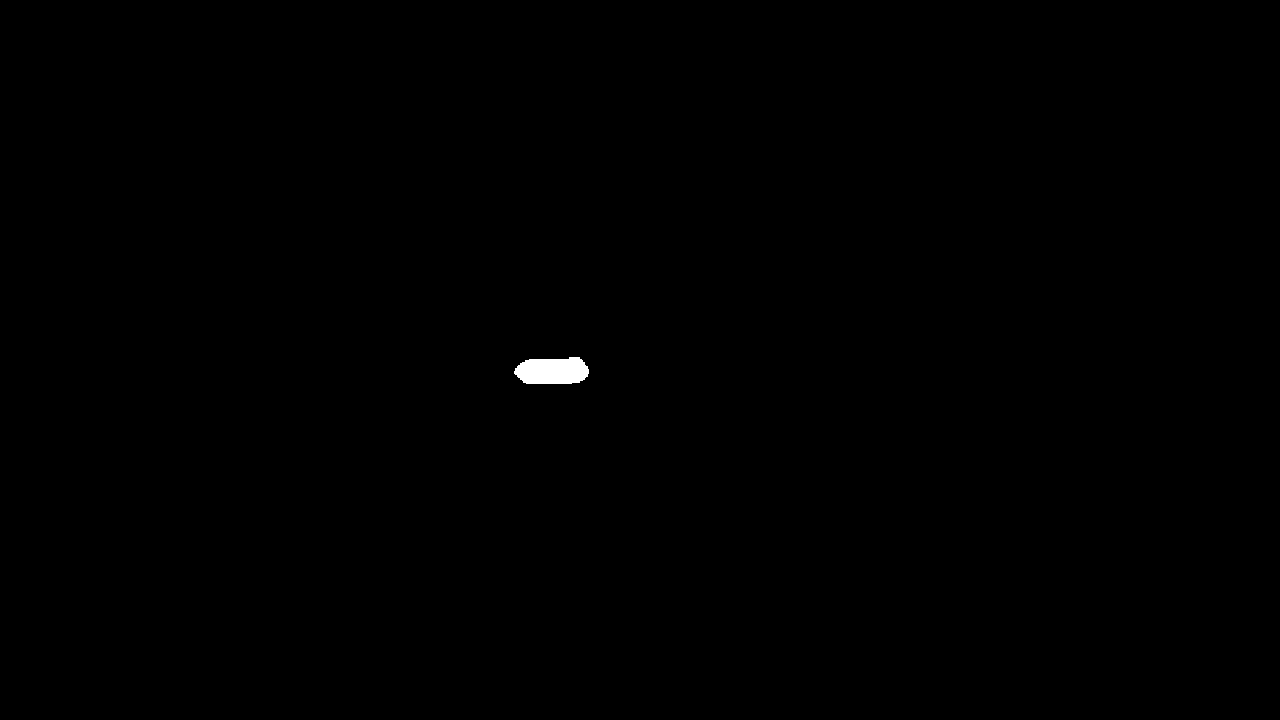}
    & \includegraphics[height=0.6in,width=0.65in]{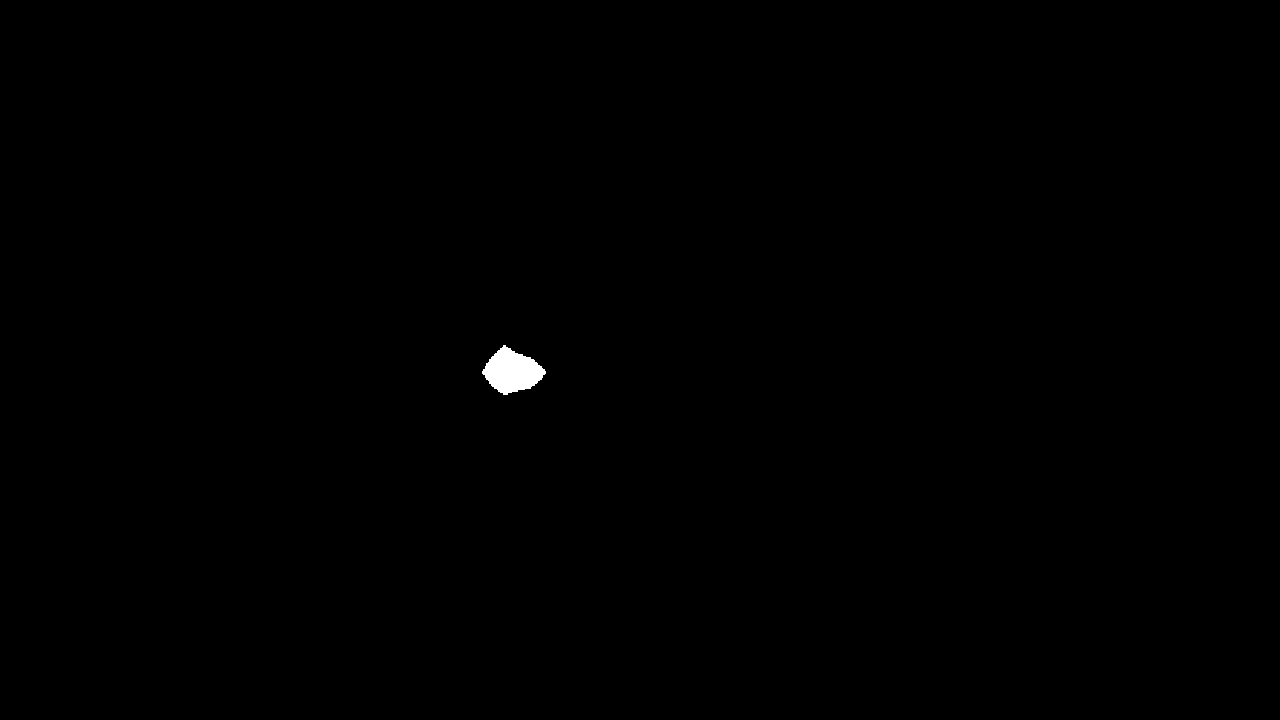}\\
\end{tabular}
\caption{Comparison of the saliency mask prediction using different mask refinement techniques. CM refers to coarse mask, CRF refers to the conditional random fields, and BS refers to the bilateral solver.}
\label{tb:refinementtechniques}
\end{figure}

\textbf{Importance of Dynamic Mask Refinement}. We studied the impact of the DMR method with quantitative results shown in Tab.~\ref{tab:ablation_morph_dyn}. Several insights can be found: (1) DMR significantly enhances the performance, showing its critical role in the network, and (2) the choice of intermediate mask refinement techniques such as BS~\cite{BarronPoole2016} and CRF~\cite{krahenbuhl2011efficient} is also pivotal for improved results. Our findings suggest that combining these refinements with DMR, as in our approach, gives maximum performance with $90.16\%$ detection rate. CRF alone was more effective than just using BS with $81.45\%$ and $78.50\%$ respectively, but surprisingly, completely omitting BS/CRF and DMR gave slightly better results achieving $82.47\%$. This outcome suggests that the coarse masks might be sufficient for the Ev-Airborne data without additional refinement, but to maximize performance even further a combination of CRF and DMR is needed. Fig.~\ref{tb:evairbornepedestrians} demonstrates the effectiveness and temporal consistency of the DMR process overtime. The object masks are accurately recovered, ensuring segmentation consistency and covering the most salient objects throughout the sequence. Fig.~\ref{tb:refinementtechniques} shows an additional example of the output masks for all mask refinement techniques applied on the initial coarse mask. For the \texttt{Golf car high oblique} case, while all techniques manage to isolate masks representing moving objects, only DMR could recover the mask across the entire sequence, unlike CRF and BS which fall short.

\begin{table}[h]
\centering
\caption{Impact of DMR on the Ev-Airborne data.}
\label{tab:ablation_morph_dyn}
\begin{tabular}{lc}
\hline
Method & Detection rate [\%] $\uparrow$ \\ \hline
w/o. DMR             &        82.47        \\ \hline
w/o. DMR + BS        &           78.50     \\ \hline
w/o. DMR + CRF       &         81.45       \\ \hline
\textbf{Ours}       &        \textbf{90.16}        \\ \hline
\end{tabular}
\end{table}

\begin{figure}[h] 
\centering
\setlength{\fboxrule}{1.0pt}
\setlength{\fboxsep}{0pt}
\setlength{\tabcolsep}{1pt} %
\renewcommand{\arraystretch}{0.5} %
\textbf{Ev-Airborne dataset}
\begin{tabular}{c c c c c c c c c}
    & \small\rotatebox{90}{\hspace{0.2cm}\color{gray!90}Input} \hspace{-2mm}\includegraphics[height=0.4in,width=0.6in]{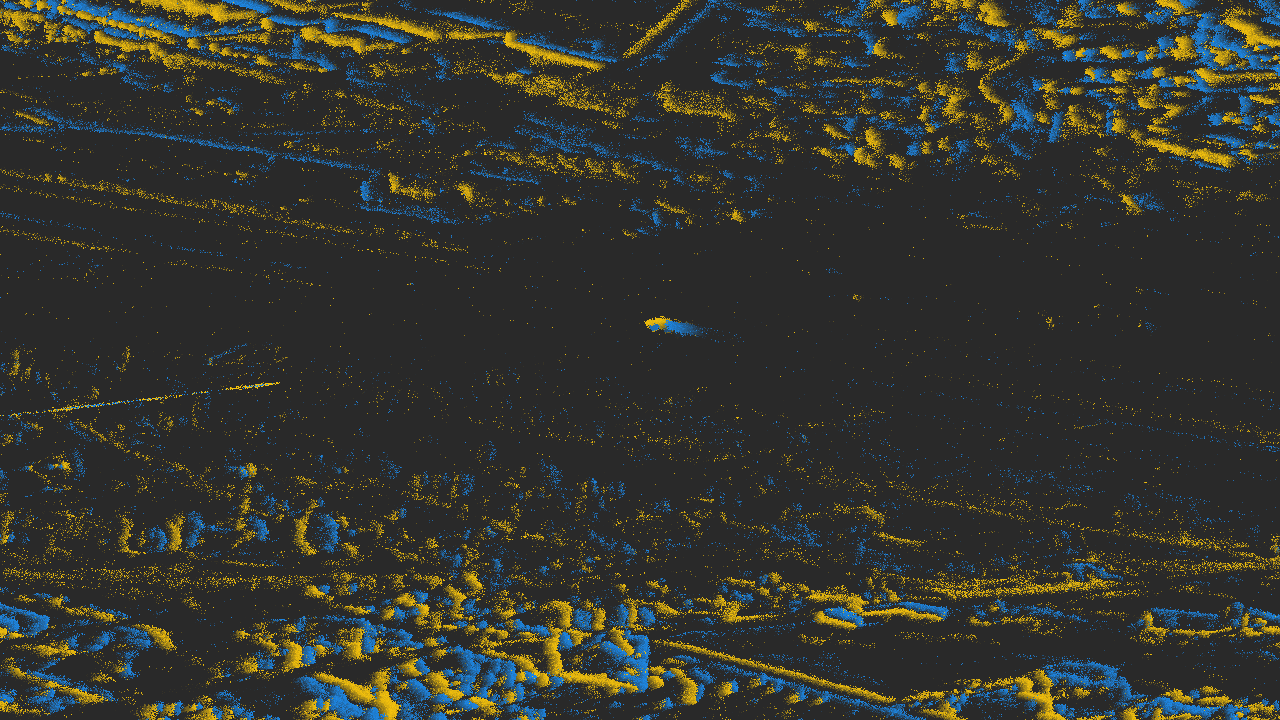}
    & \includegraphics[height=0.4in,width=0.6in]{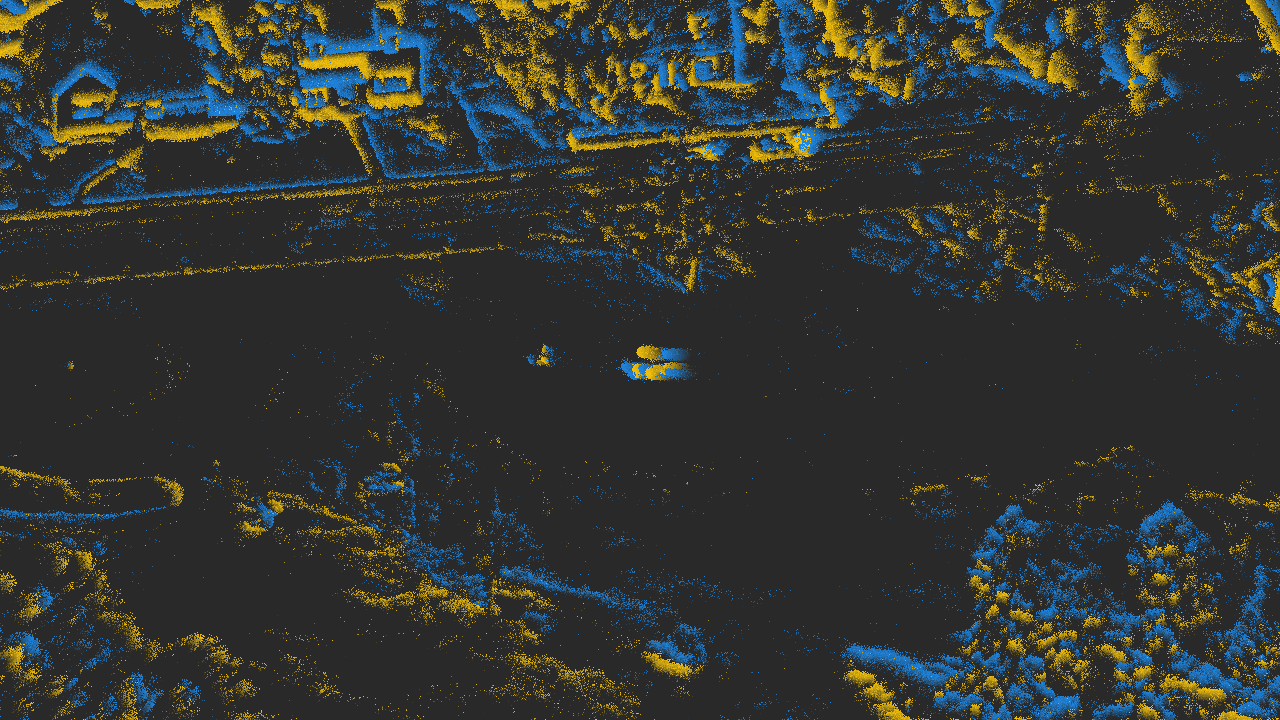}
    & \includegraphics[height=0.4in,width=0.6in]{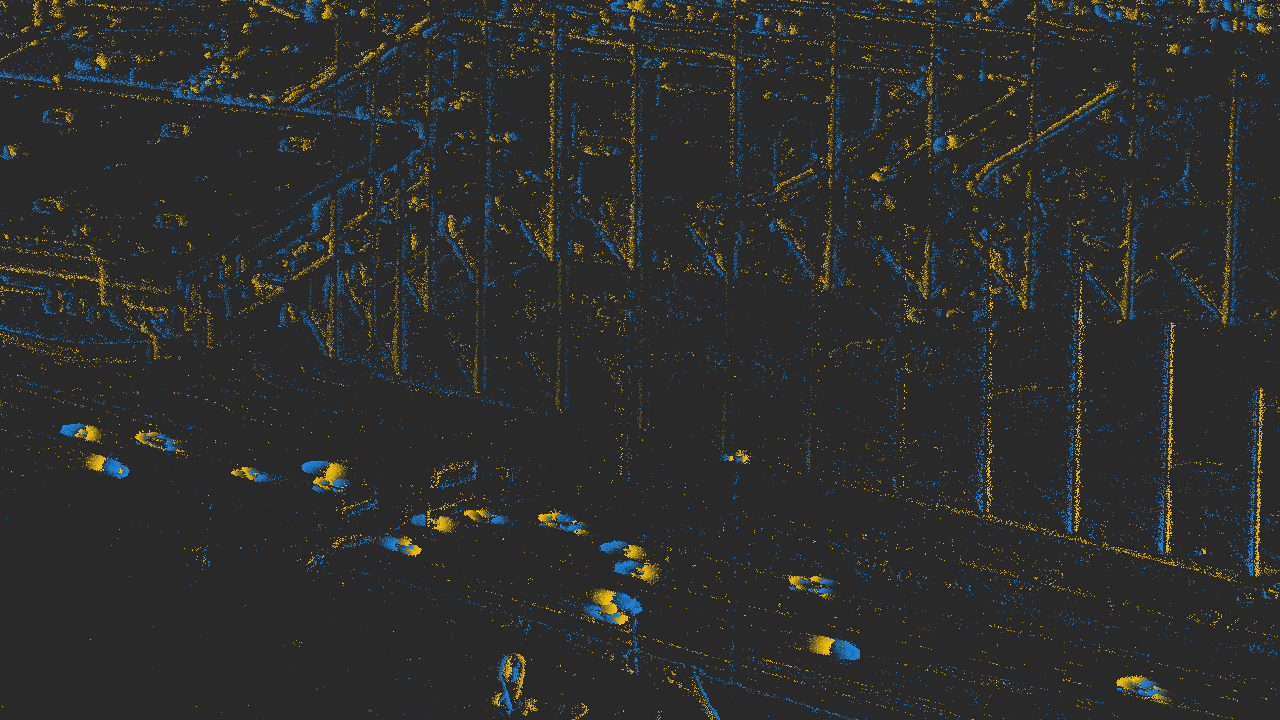}
    & \includegraphics[height=0.4in,width=0.6in]{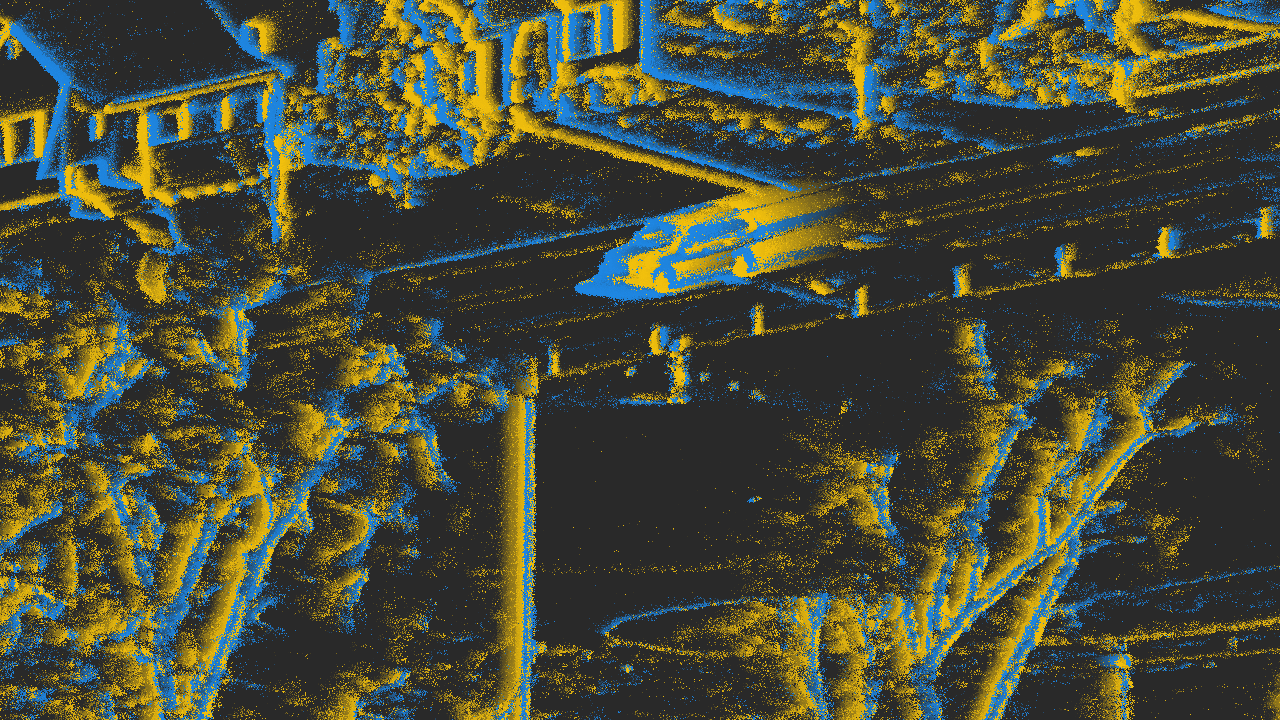}
    & \includegraphics[height=0.4in,width=0.6in]{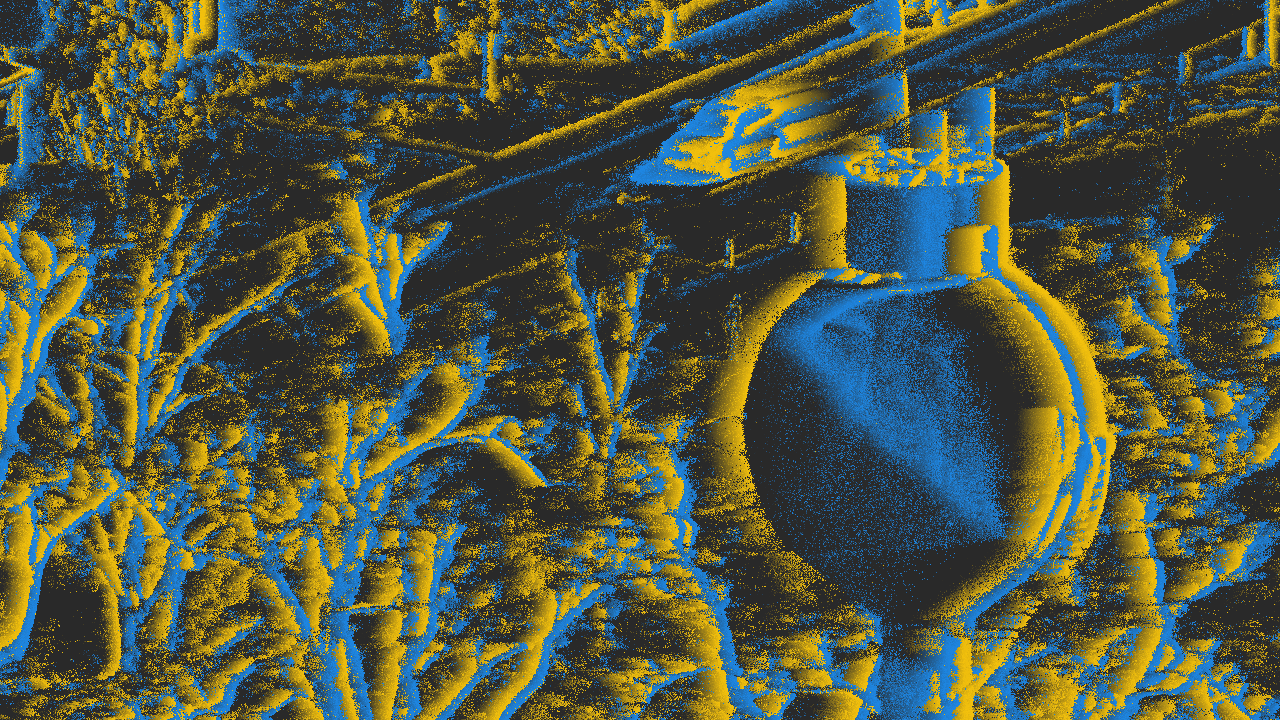}
    \\

    & \tiny\rotatebox{90}{\hspace{0.2cm}\color{gray!90}ViT-B}\fcolorbox{red}{white}{\includegraphics[height=0.4in,width=0.6in]{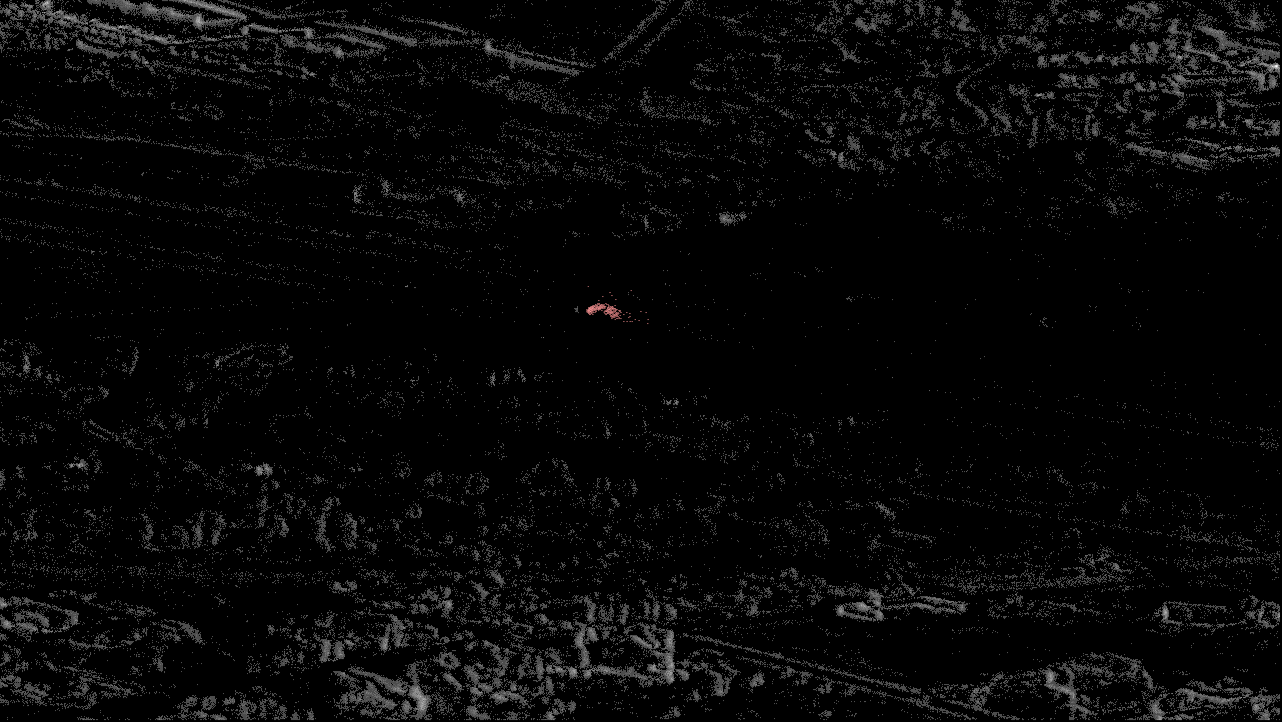}}
    & \includegraphics[height=0.4in,width=0.6in]{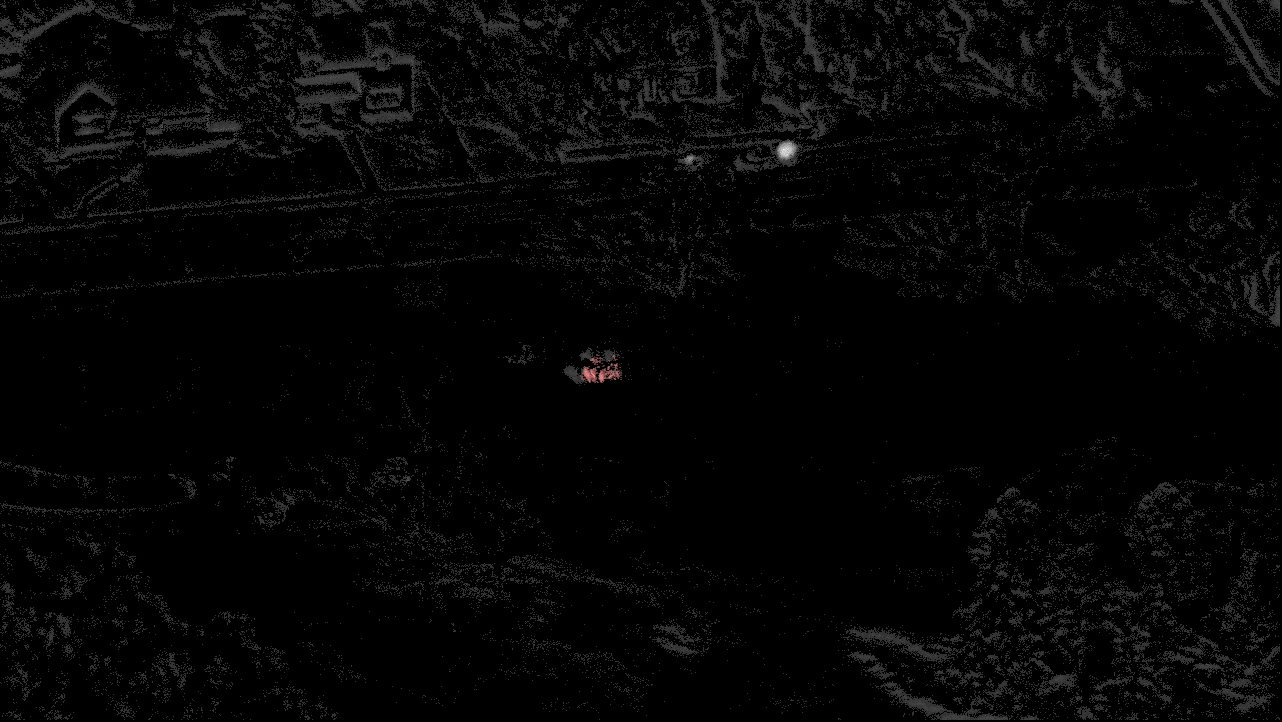}
    & \includegraphics[height=0.4in,width=0.6in]{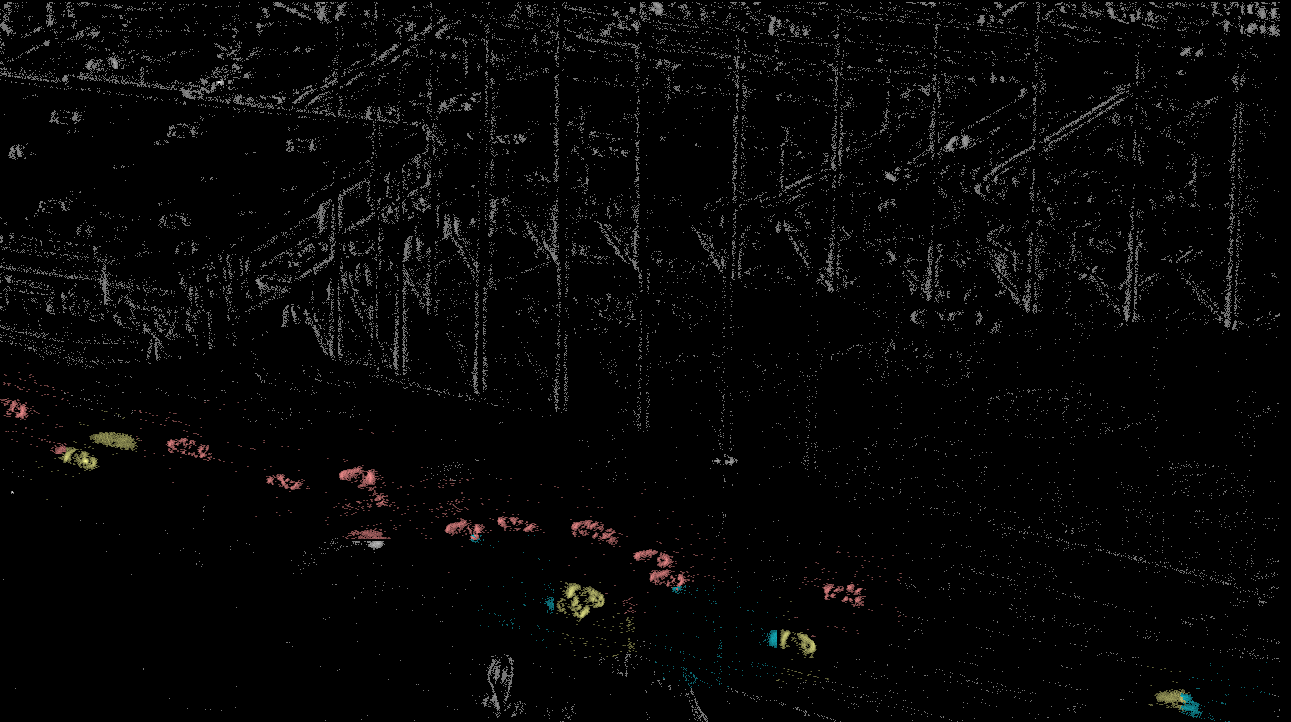}
    & \fcolorbox{red}{white}{\includegraphics[height=0.4in,width=0.6in]{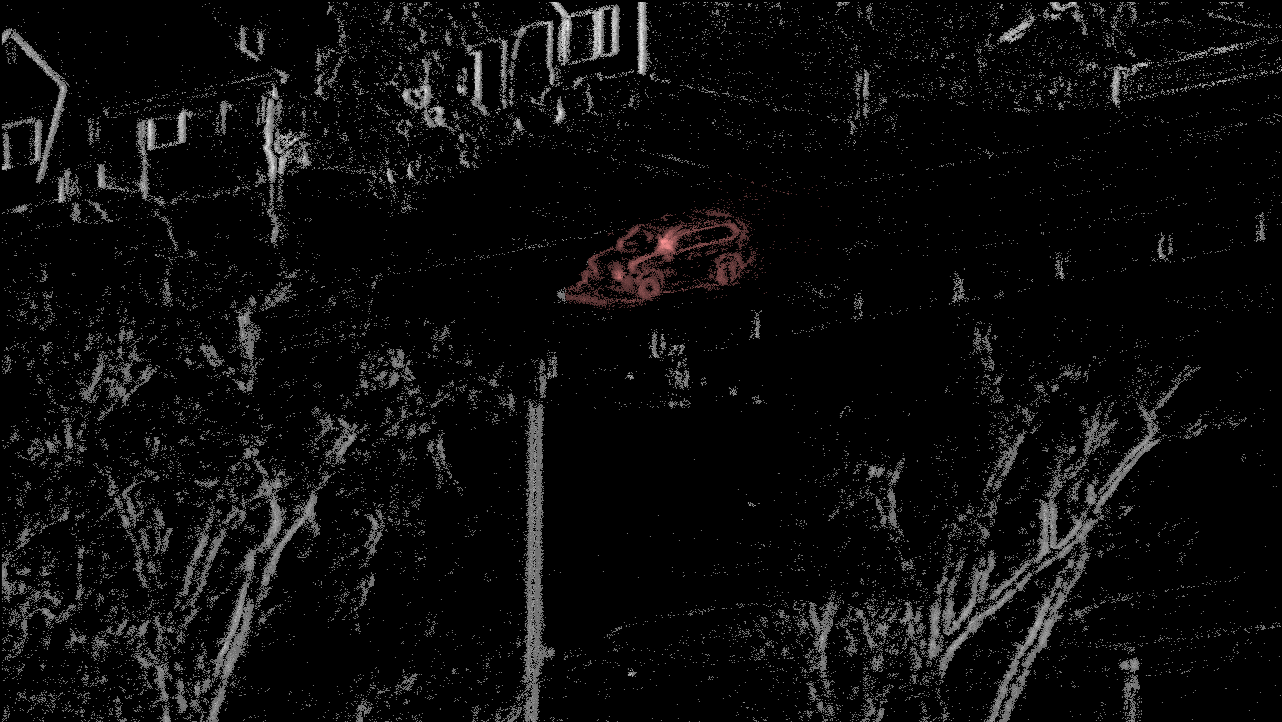}}
    & \fcolorbox{red}{white}{\includegraphics[height=0.4in,width=0.6in]{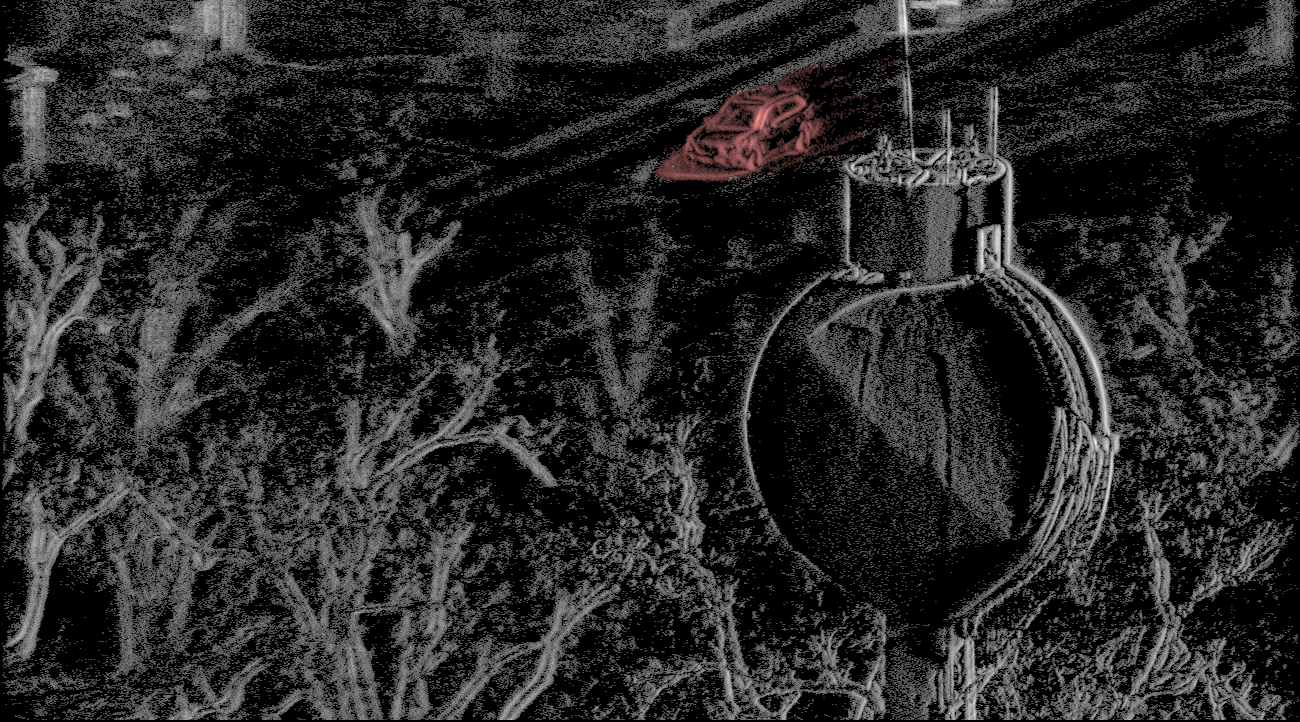}}
    \\
    
    & \tiny\rotatebox{90}{\hspace{0.0cm}\color{gray!90}MoCoV3}\fcolorbox{red}{white}{\includegraphics[height=0.4in,width=0.6in]{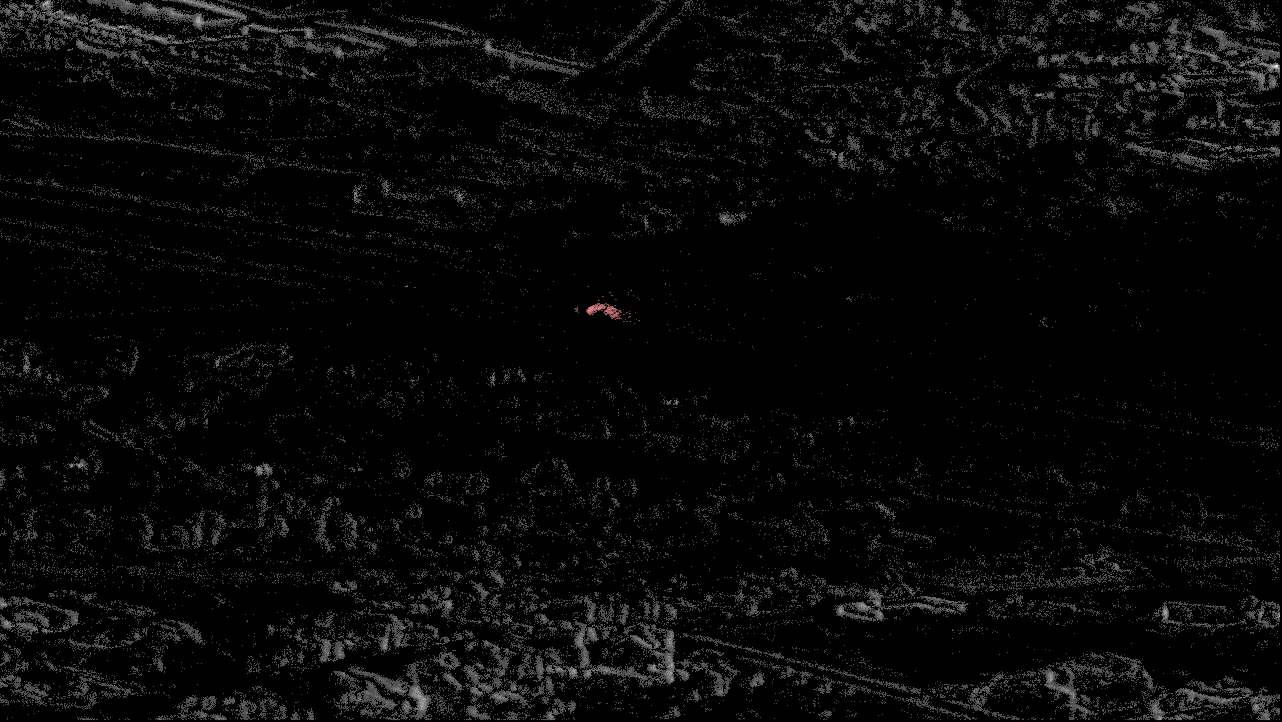}}
    & \includegraphics[height=0.4in,width=0.6in]{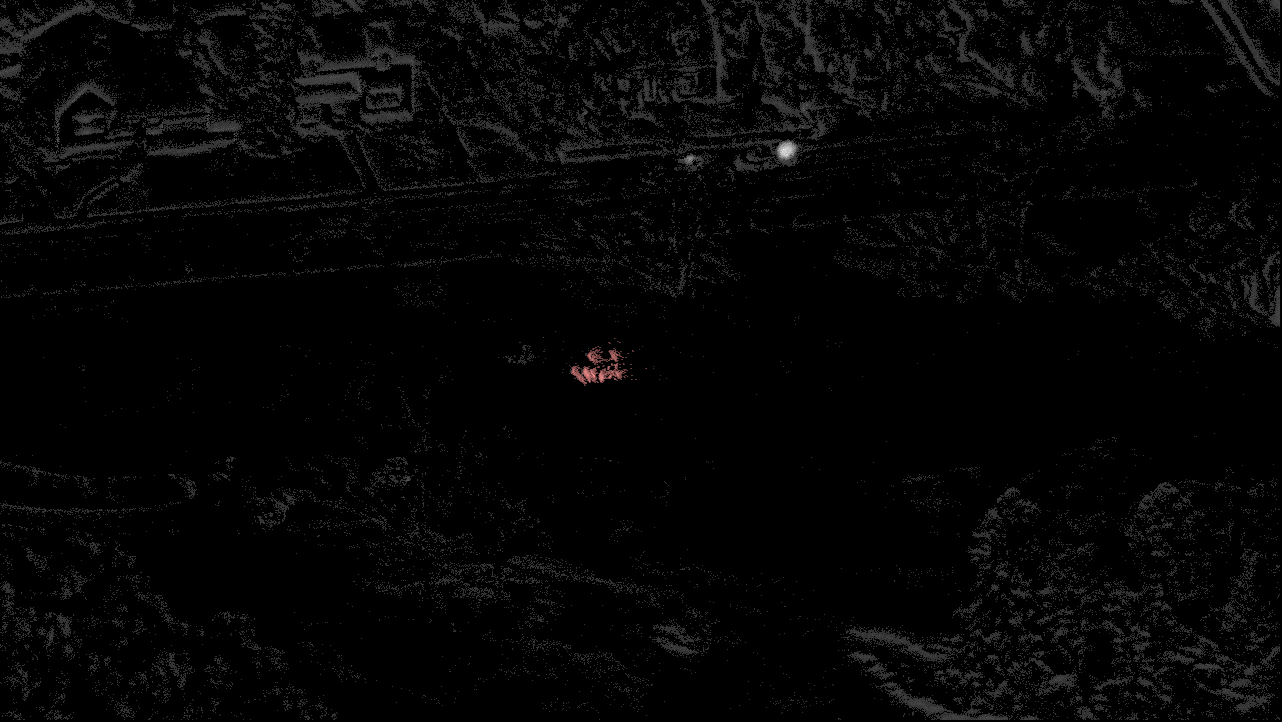}
    & \includegraphics[height=0.4in,width=0.6in]{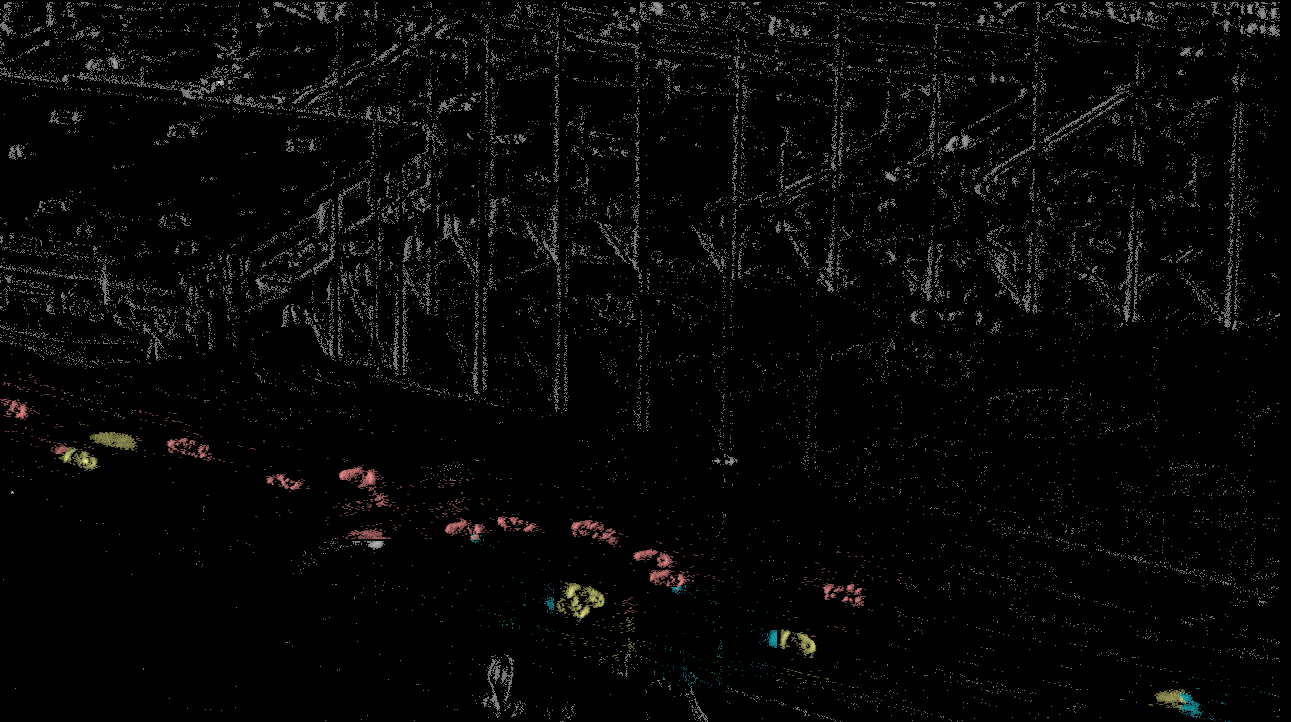}
    & \fcolorbox{red}{white}{\includegraphics[height=0.4in,width=0.6in]{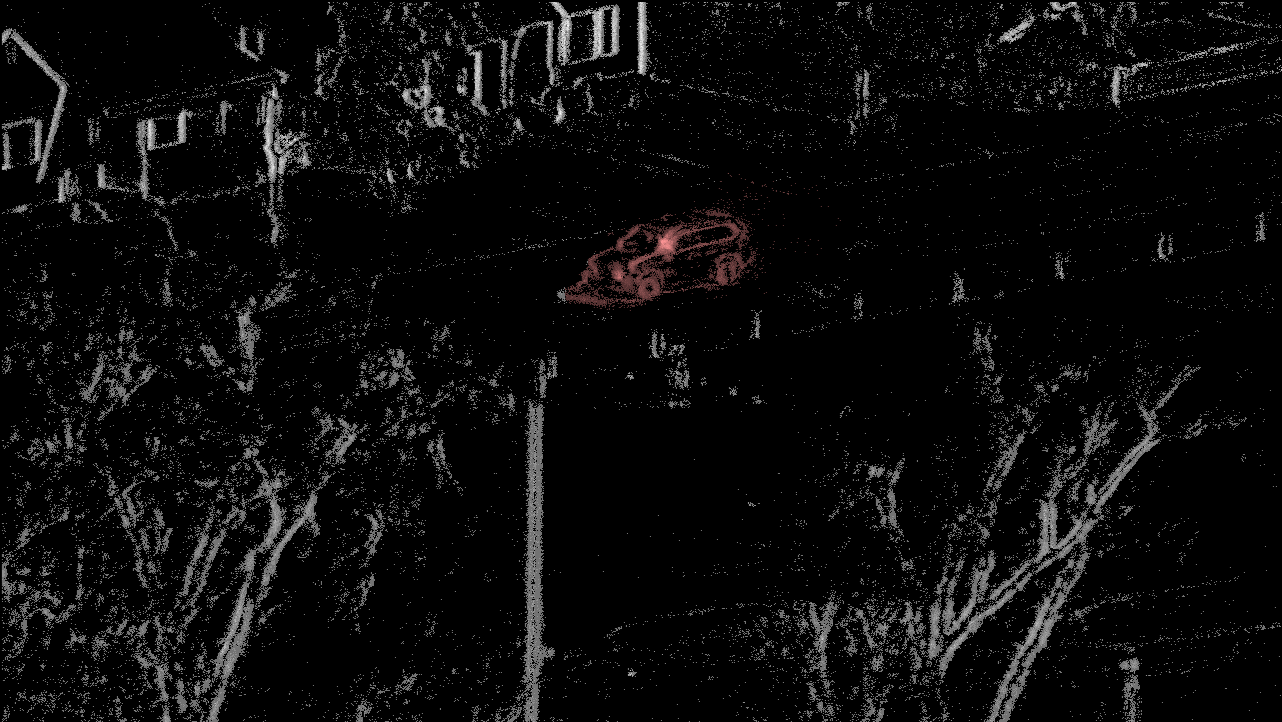}}
    & \includegraphics[height=0.4in,width=0.6in]{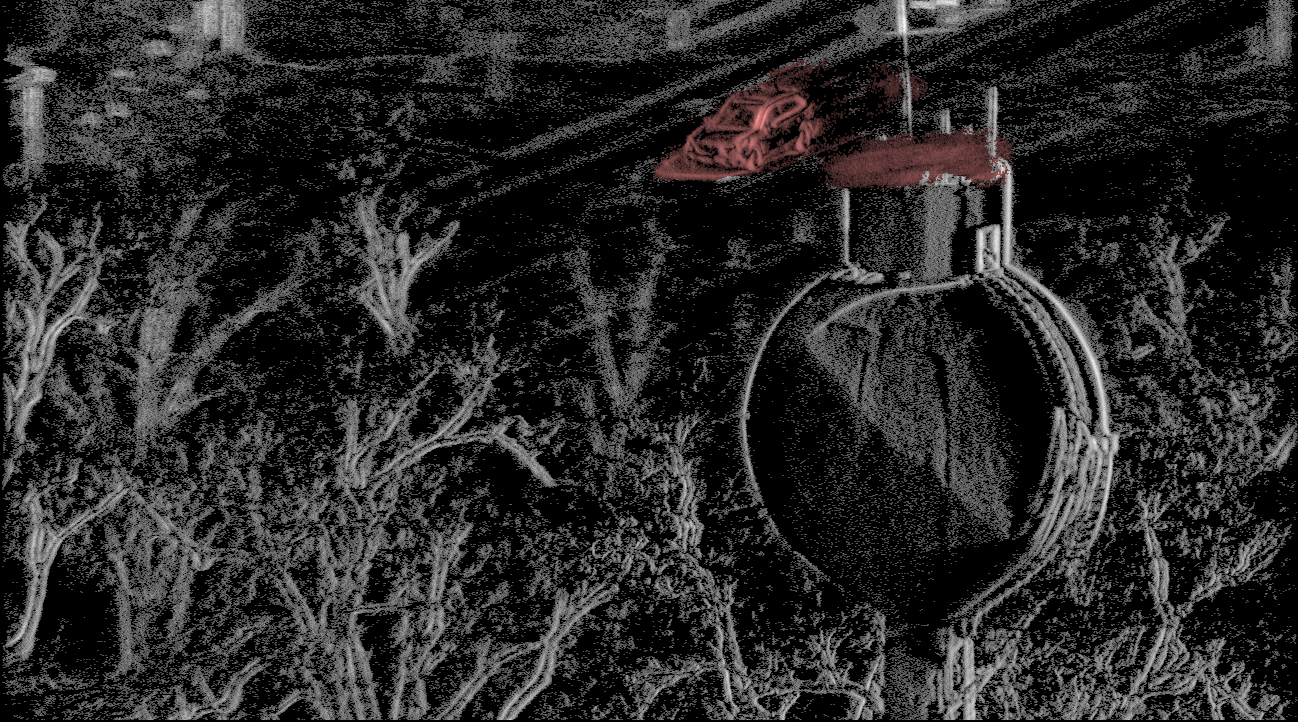}
    \\

    & \tiny\rotatebox{90}{\hspace{0.2cm}\color{gray!90}MAE}\includegraphics[height=0.4in,width=0.6in]{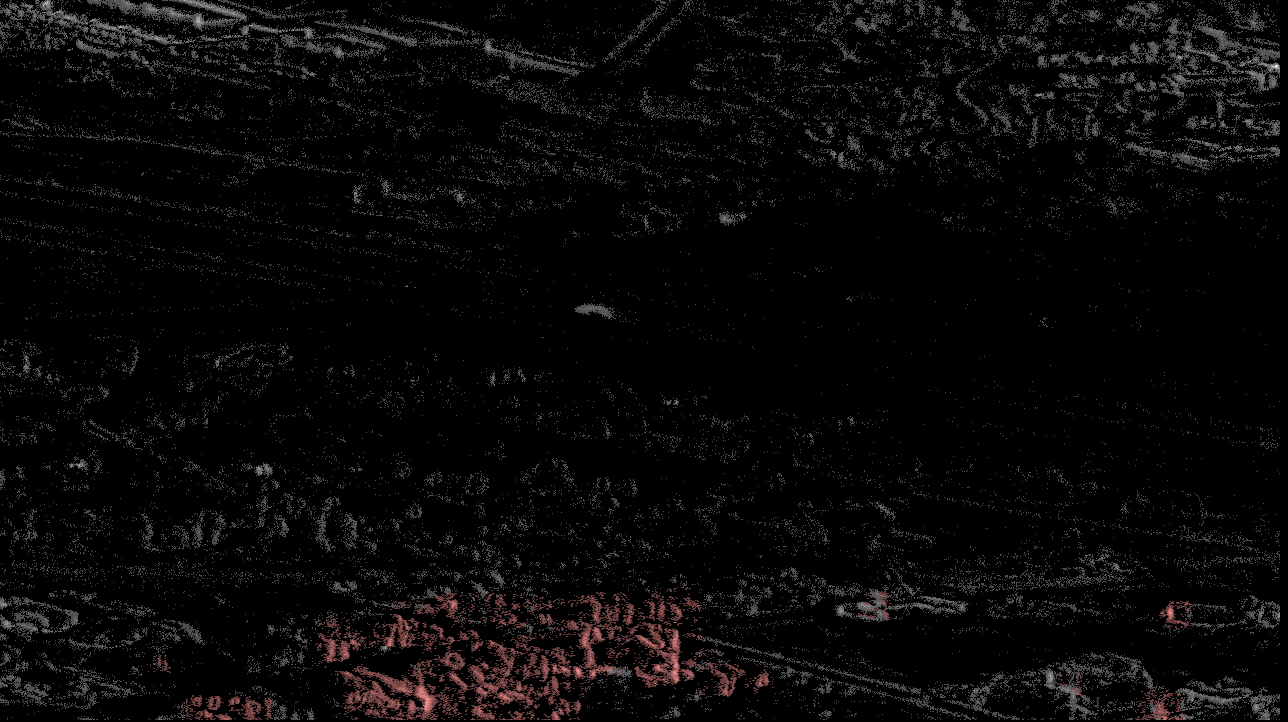}
    & \includegraphics[height=0.4in,width=0.6in]{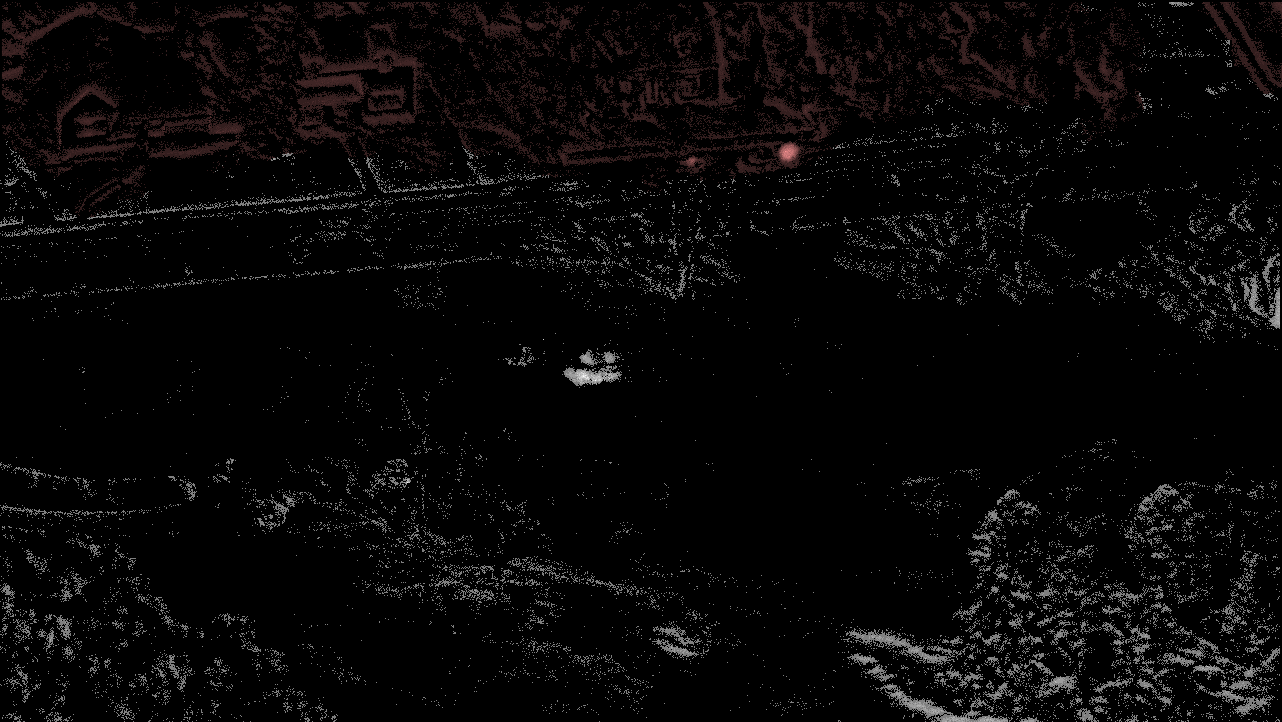}
    & \includegraphics[height=0.4in,width=0.6in]{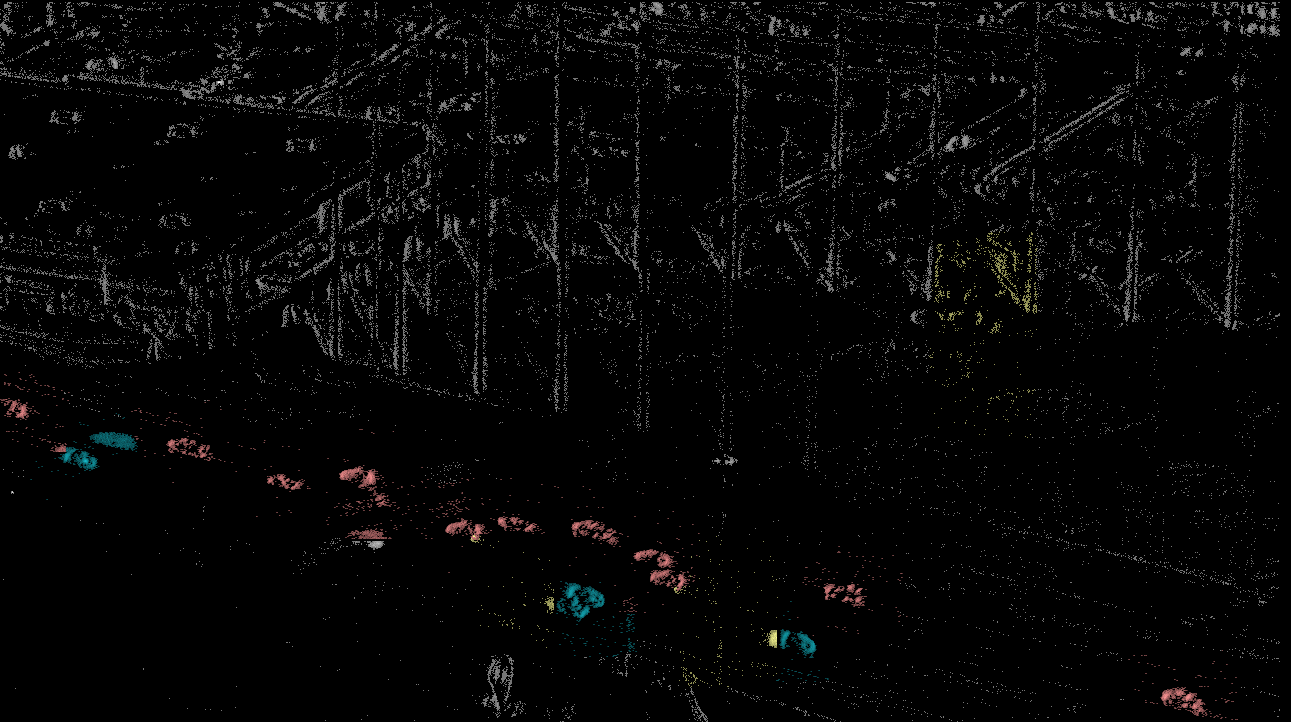}
    & \includegraphics[height=0.4in,width=0.6in]{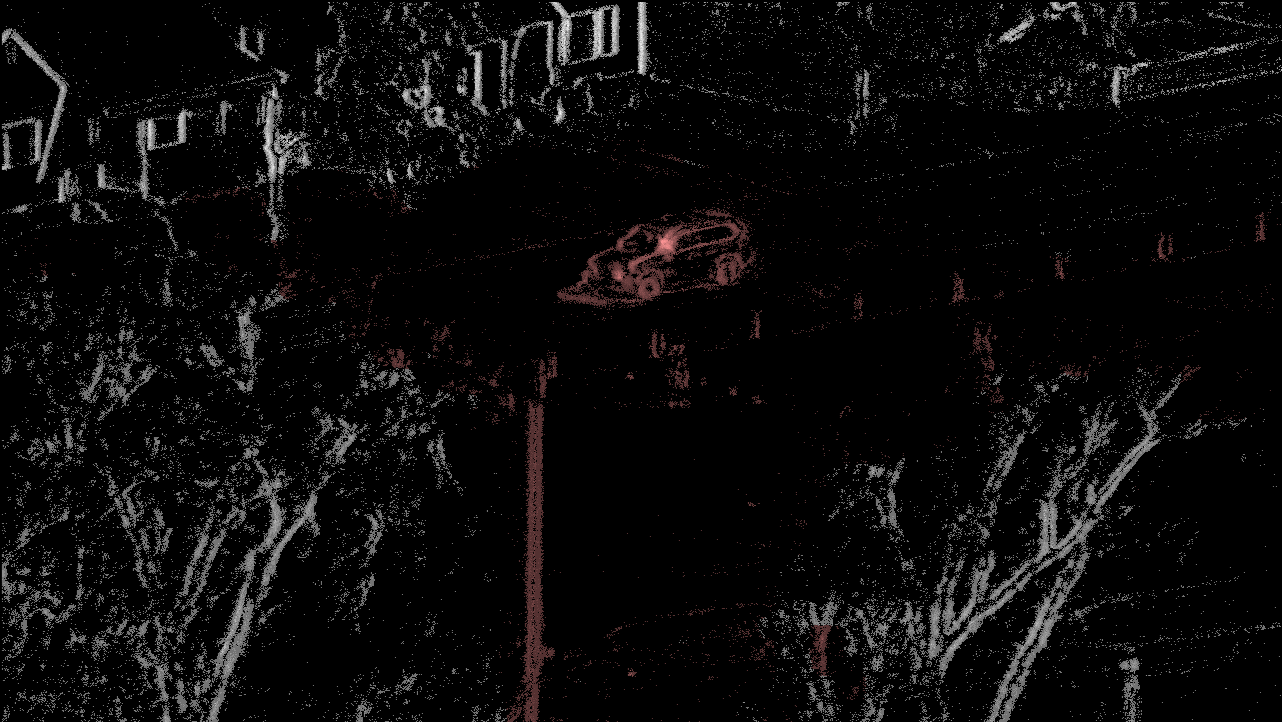}
    & \includegraphics[height=0.4in,width=0.6in]{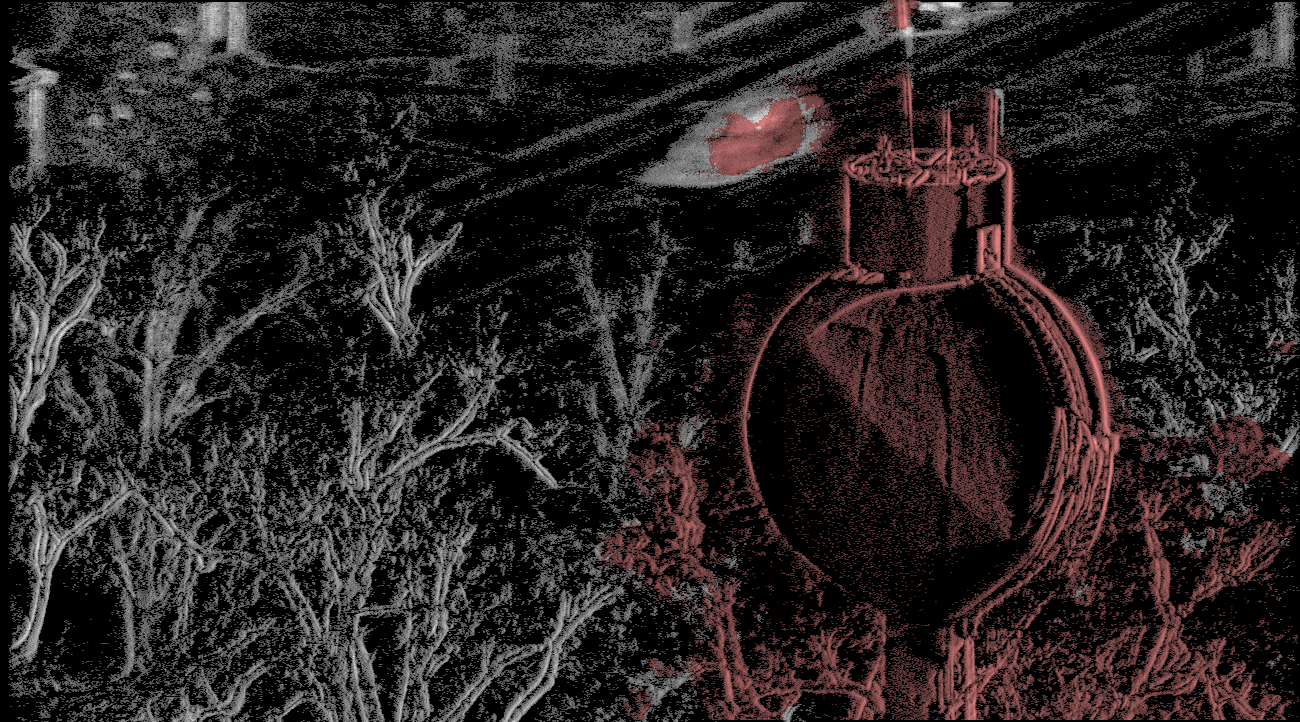}
    \\

    & \tiny\rotatebox{90}{\hspace{0.2cm}\color{gray!90}ViT-S}\fcolorbox{red}{white}{\includegraphics[height=0.4in,width=0.6in]{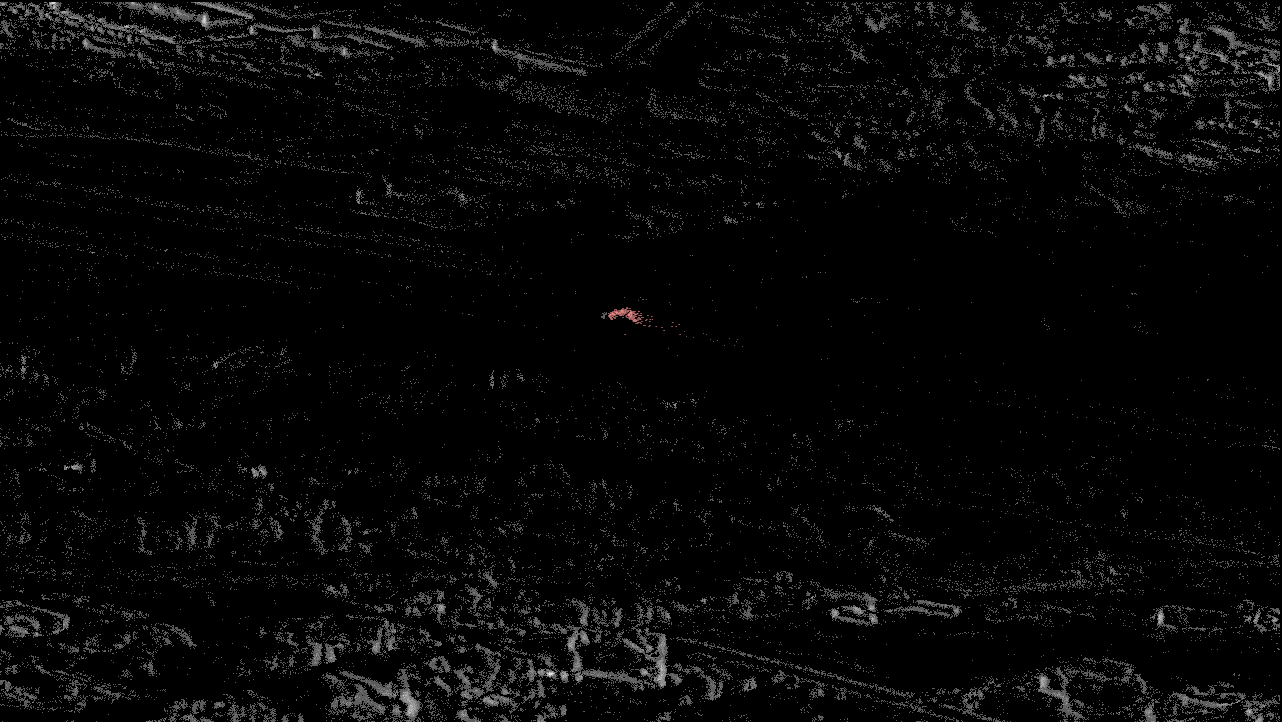}}
    & \fcolorbox{red}{white}{\includegraphics[height=0.4in,width=0.6in]{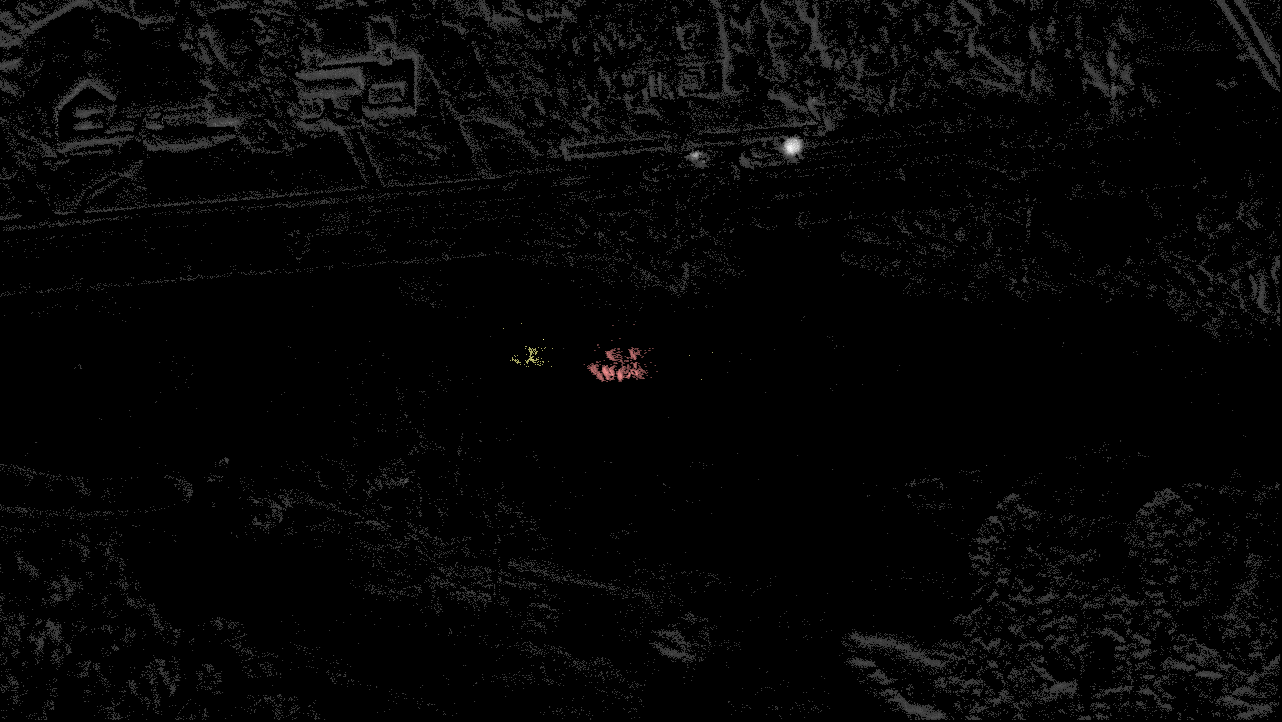}}
    & \fcolorbox{red}{white}{\includegraphics[height=0.4in,width=0.6in]{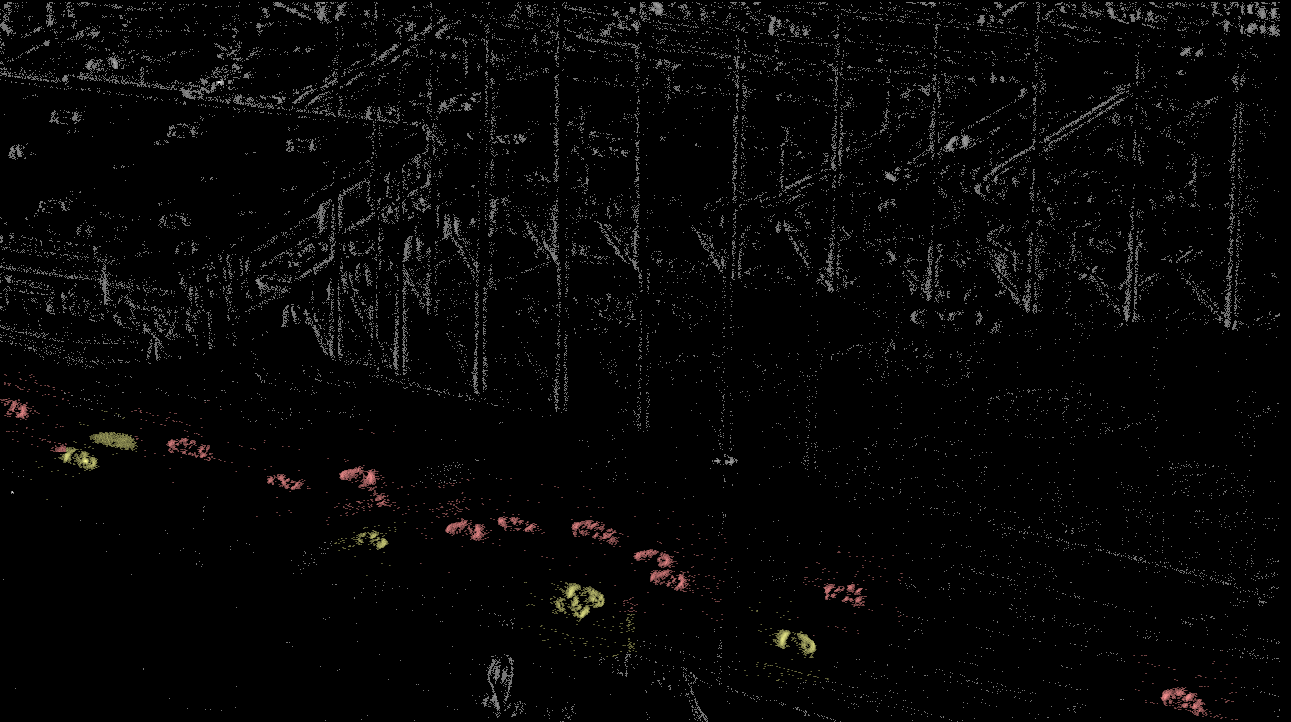}}
    & \fcolorbox{red}{white}{\includegraphics[height=0.4in,width=0.6in]{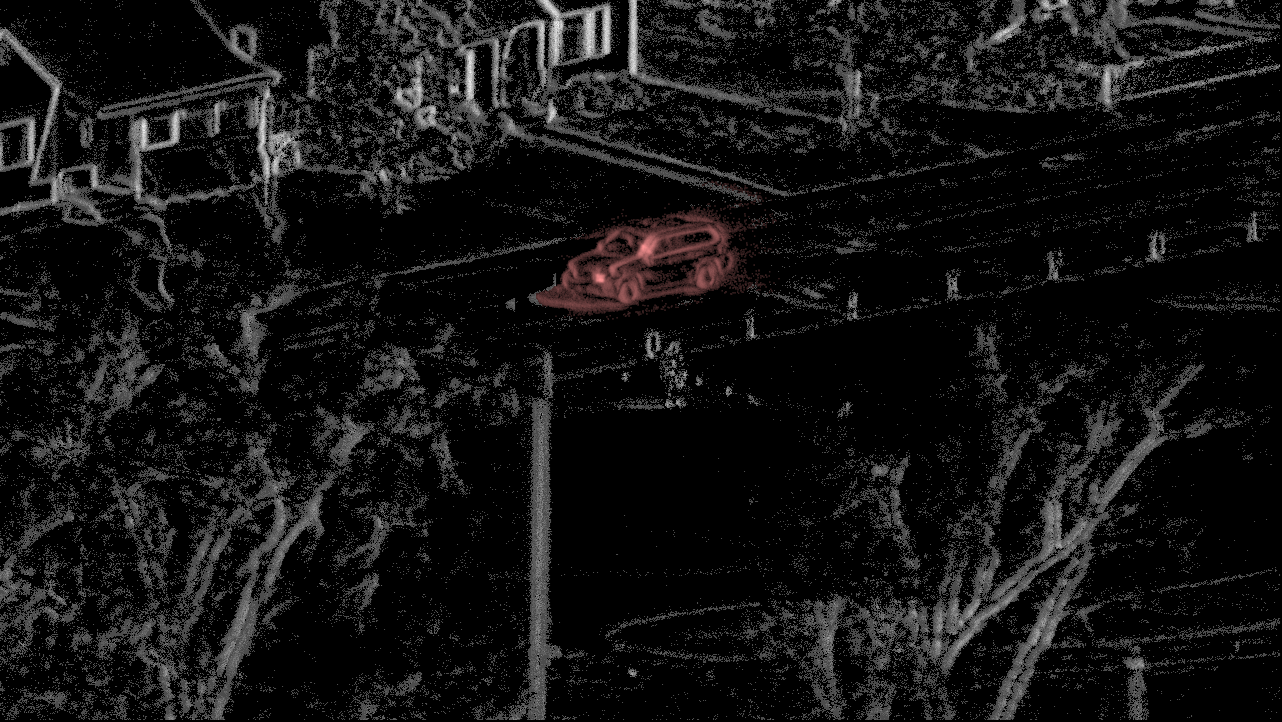}}
    & \fcolorbox{red}{white}{\includegraphics[height=0.4in,width=0.6in]{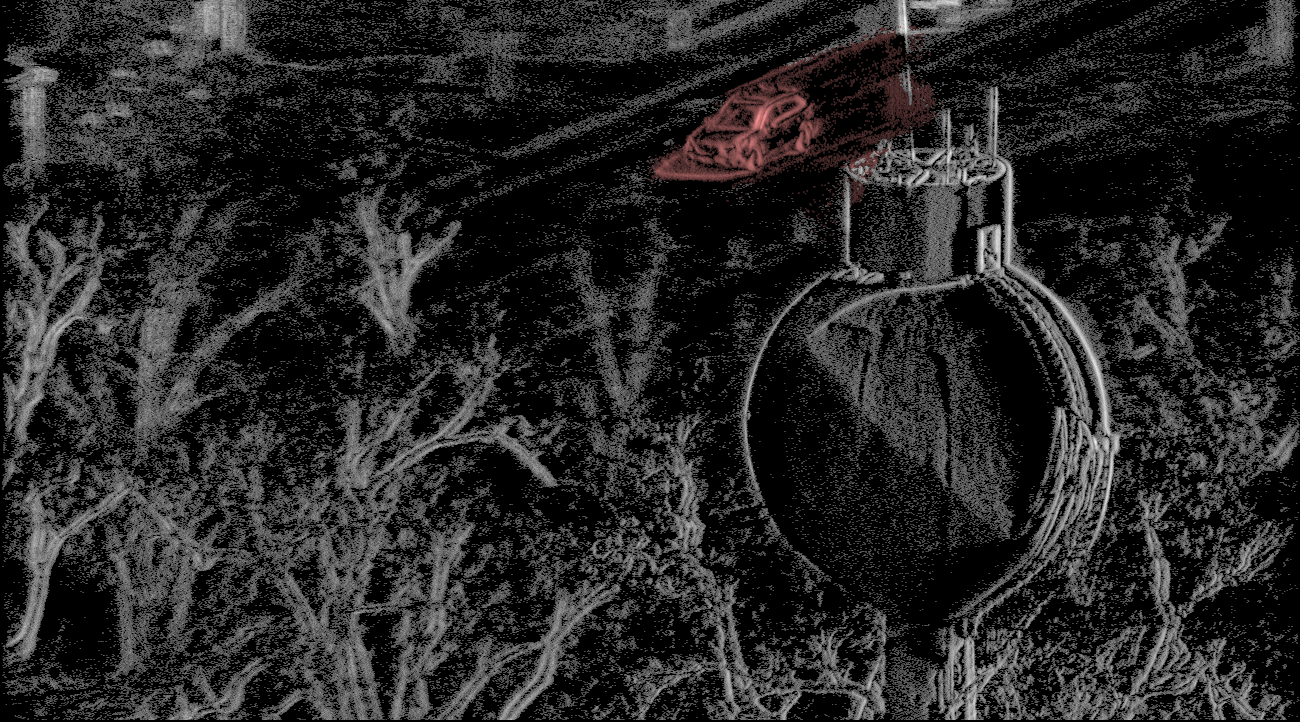}}
    \\
\end{tabular}
\caption{Comparison of the segmentation output on different backbones. Red bounding boxes show which model produces high-quality segmentation with ViT-S outperforming all other backbones.}
\label{tb:backbonesqualitative}
\end{figure}

\textbf{Generalization to different backbones}. In Tab.~\ref{tab:backbones}, we present an ablation analysis of various pre-trained self-supervised transformer backbones, including ViT-small and ViT-base~\cite{caron_emerging_2021,dosovitskiy2020vit}, MoCoV3~\cite{chen_empirical_2021}, and MAE~\cite{he_masked_2022} with $16\times 16$ patch size, applied to the Ev-Airborne dataset using the same parameters. Our findings indicate that ViT-small surpasses the performance of the other models for this specific application. This superiority is likely due to the relatively small size of the Ev-Airborne dataset compared to larger-scale vision datasets. Thus, for this task, ViT-small proves to be sufficiently effective. Fig.~\ref{tb:backbonesqualitative} shows qualitative results on various sequences from Ev-Airborne. It can be seen that the ViT-S/16 outperforms all other models by producing segmentation output that exactly describes the number of motions in the scenes. Among the models evaluated, MAE exhibits the least effective performance.

\begin{table}[h]
\centering
\caption{Performance of different backbones on the Ev-Airborne data.}
\label{tab:backbones}
\begin{tabular}{lc}
\hline
Backbone & Detection rate [\%] $\uparrow$ \\ \hline
\texttt{ViT-B/16}      & 84.94  \\ \hline
\texttt{MoCoV3-S/16}   & 86.90  \\ \hline
\texttt{MAE-B/16}      & 40.24  \\ \hline
\texttt{ViT-S/16}      & \textbf{90.16} \\ \hline
\end{tabular}
\end{table}

\begin{figure}[h] 
\centering
\setlength{\fboxrule}{1.0pt}
\setlength{\fboxsep}{0pt}
\setlength{\tabcolsep}{1pt} %
\renewcommand{\arraystretch}{0.5} %
\textbf{Ev-Airborne dataset}
\begin{tabular}{c c c c c c c c}
    & \small\rotatebox{90}{\hspace{0.2cm}\color{gray!90}Input} \hspace{-2mm}\includegraphics[height=0.45in,width=0.65in]{Figures/000006_700001_airplane_ablation_input.png}
    & \includegraphics[height=0.45in,width=0.65in]{Figures/000008_900000_golf_ablation_input.png}
    & \includegraphics[height=0.45in,width=0.65in]{Figures/000008_900000_highway_ablation_input.png}
    & \includegraphics[height=0.45in,width=0.65in]{Figures/000020_2100000_moving_suv_ablation_input.png}
    & \includegraphics[height=0.45in,width=0.65in]{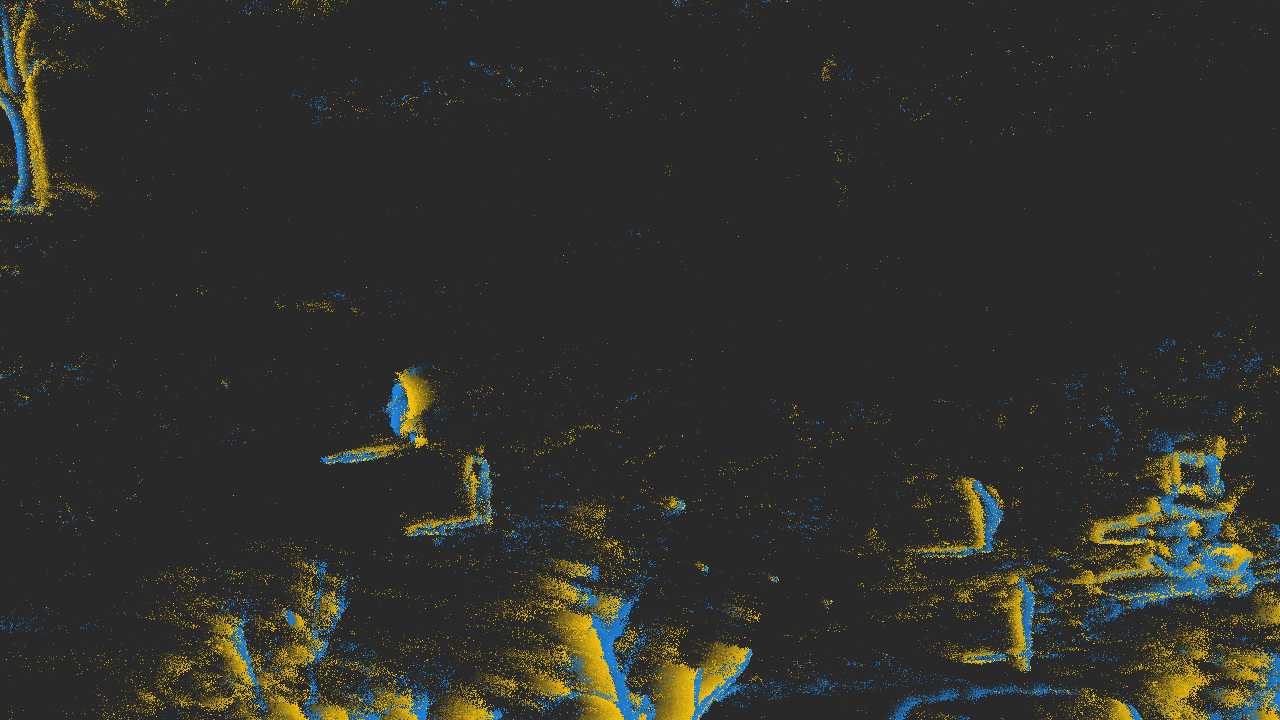}\\

    & \small\rotatebox{90}{\hspace{0.0cm}\color{gray!90}\texttt{chairs}}\includegraphics[height=0.45in,width=0.65in]{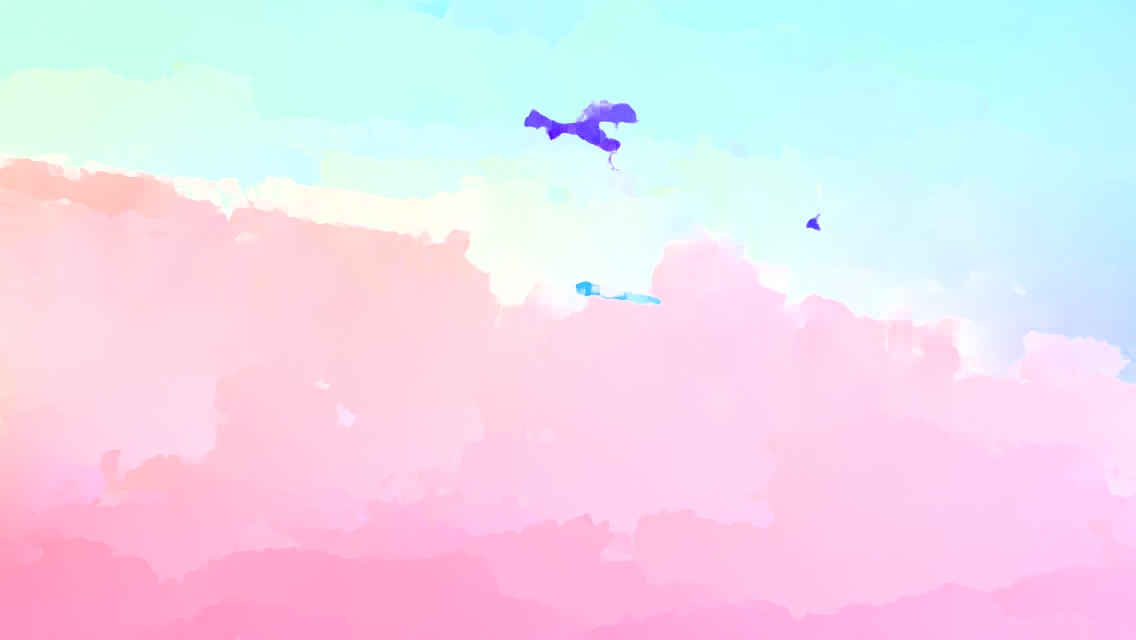}
    & \includegraphics[height=0.45in,width=0.65in]{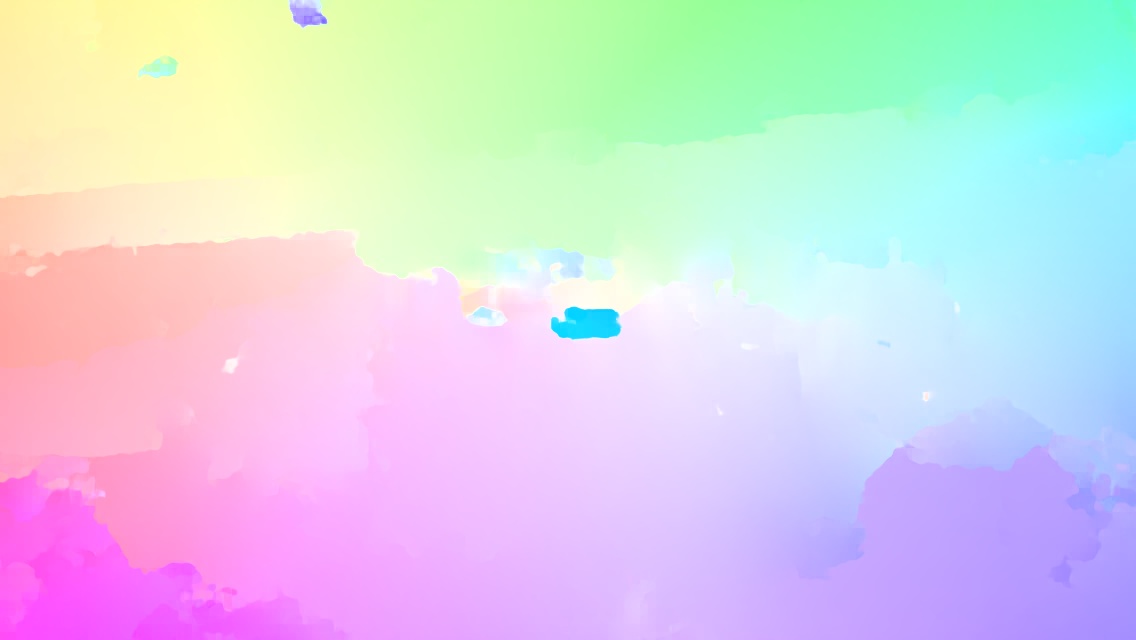}
    & \includegraphics[height=0.45in,width=0.65in]{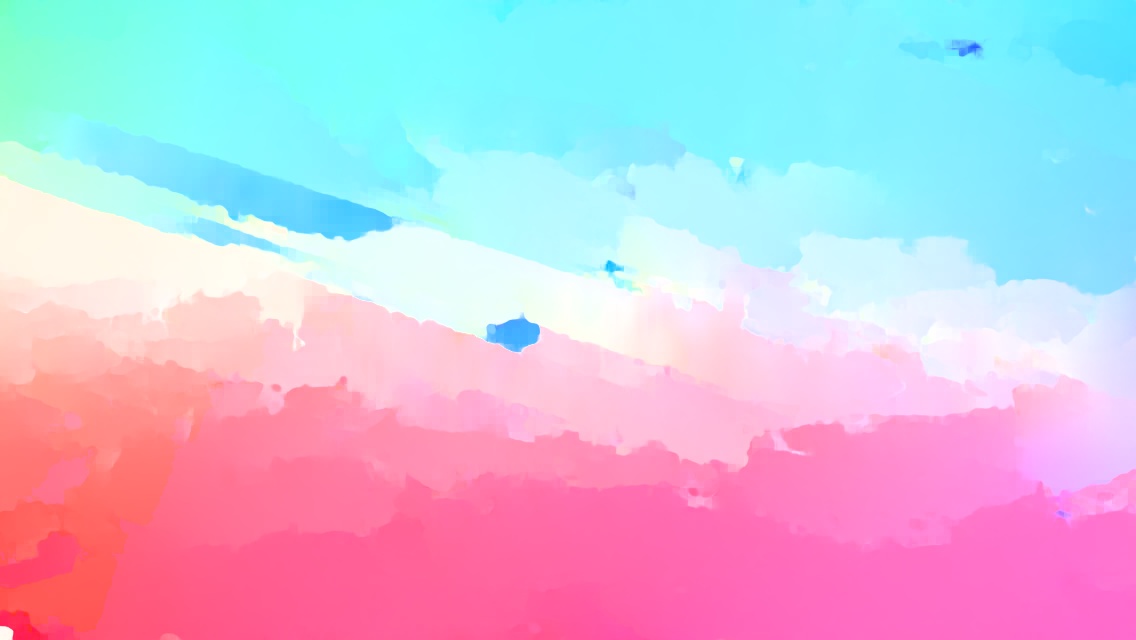}
    & \includegraphics[height=0.45in,width=0.65in]{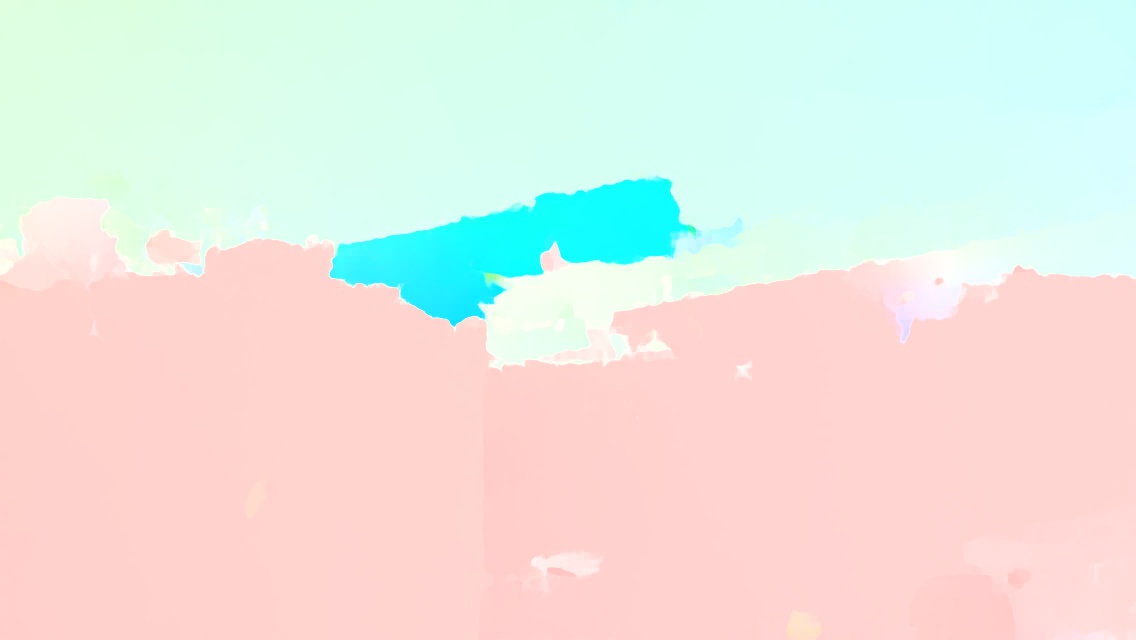}
    & \includegraphics[height=0.45in,width=0.65in]{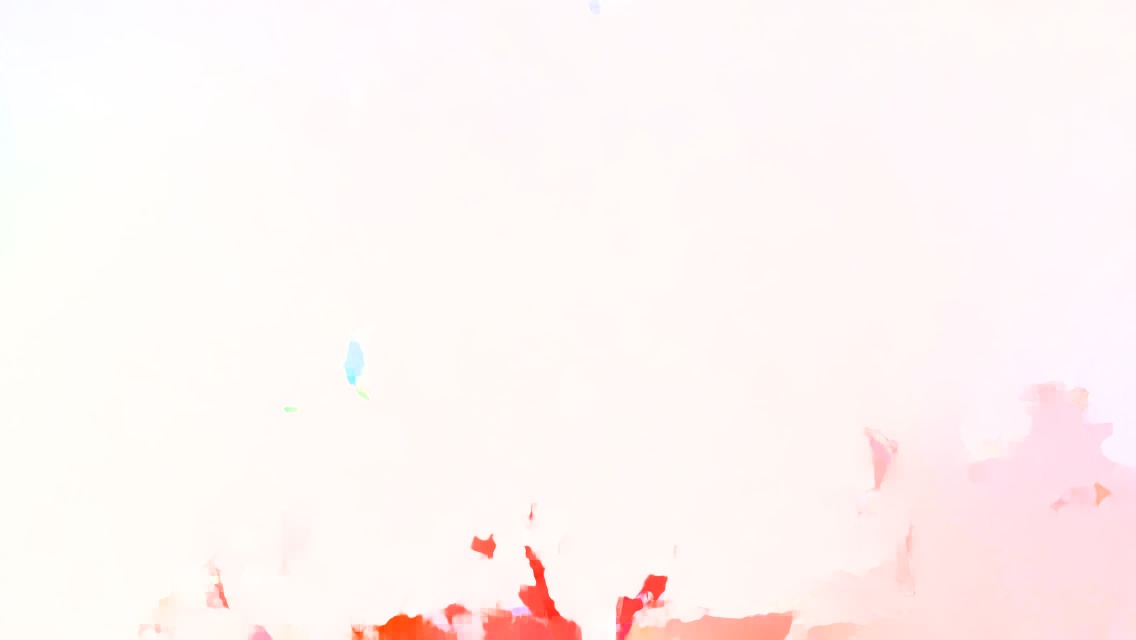}\\

    & \small\rotatebox{90}{\hspace{0.0cm}\color{gray!90}\texttt{KITTI}}\includegraphics[height=0.45in,width=0.65in]{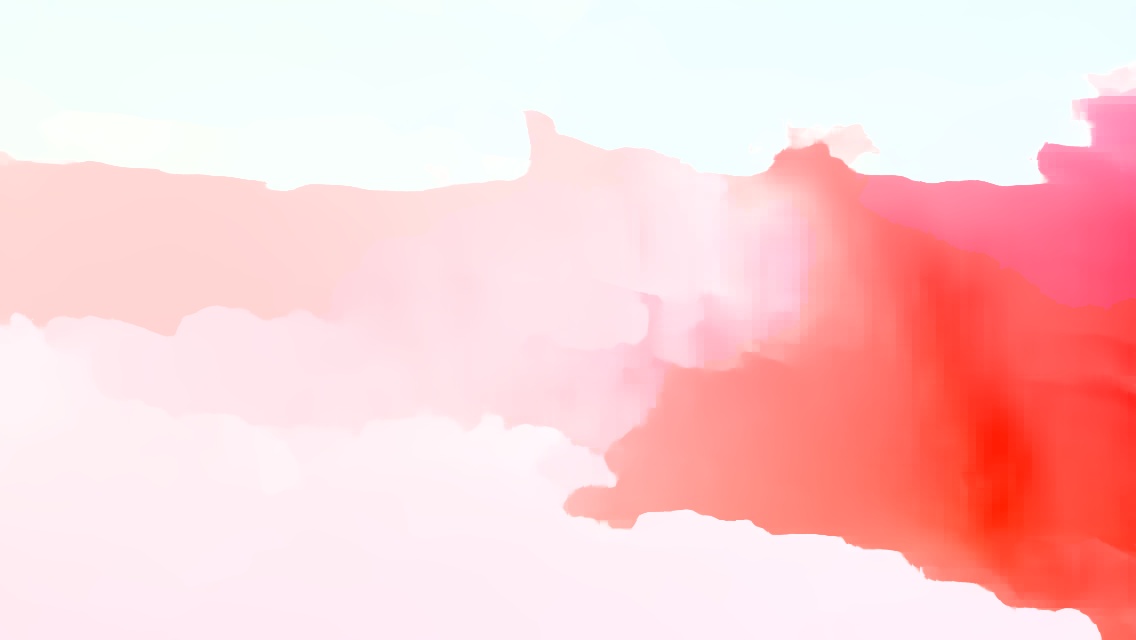}
    & \includegraphics[height=0.45in,width=0.65in]{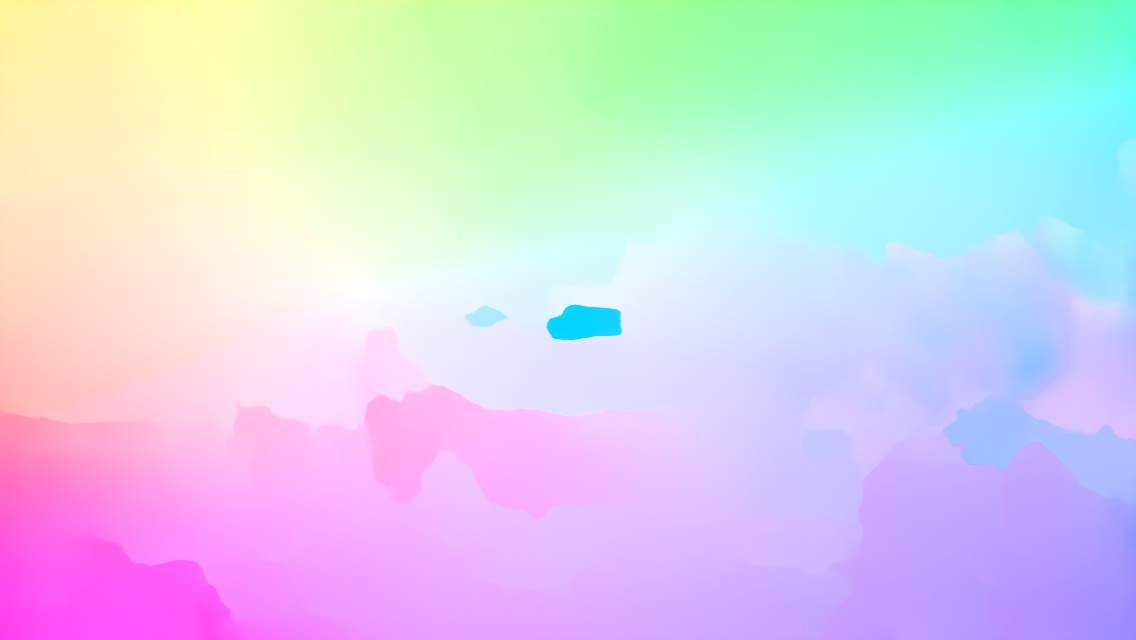}
    & \includegraphics[height=0.45in,width=0.65in]{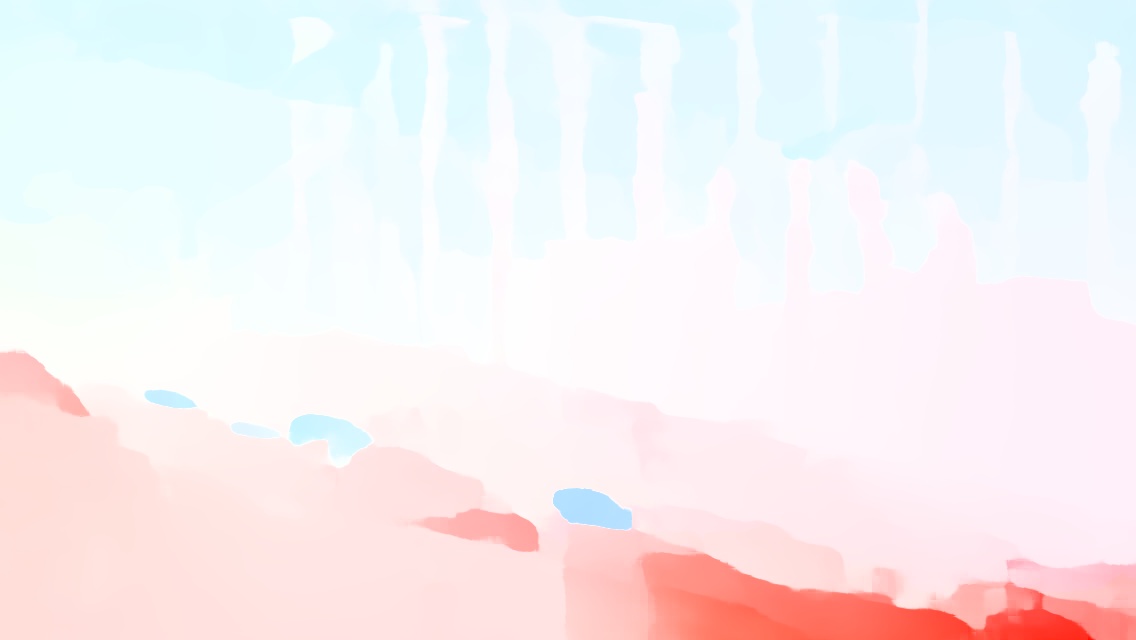}
    & \includegraphics[height=0.45in,width=0.65in]{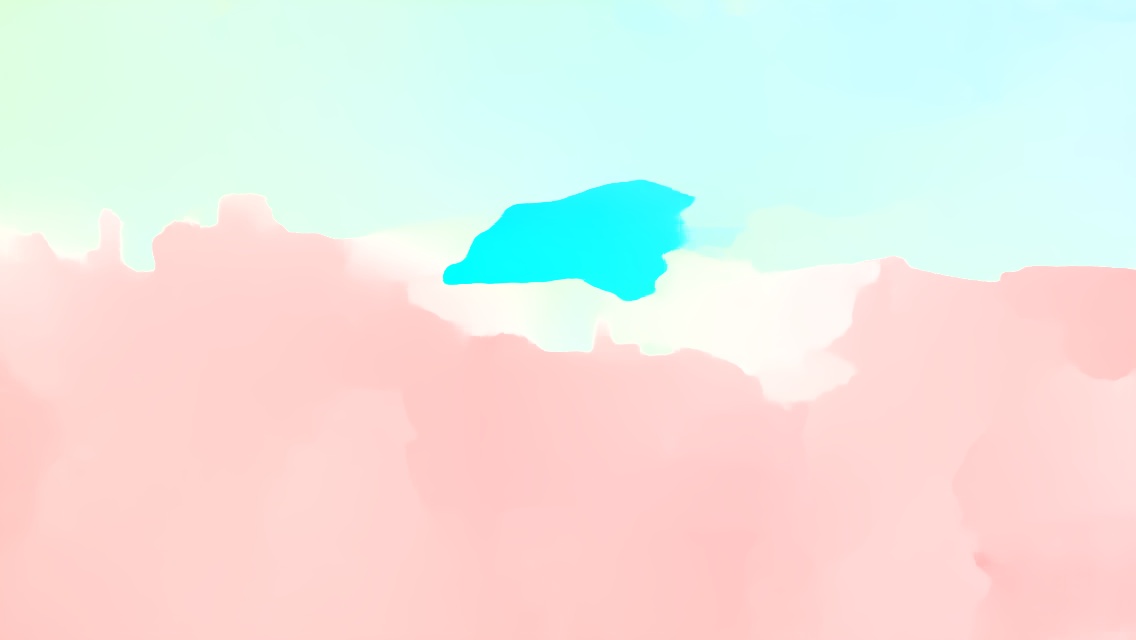}
    & \includegraphics[height=0.45in,width=0.65in]{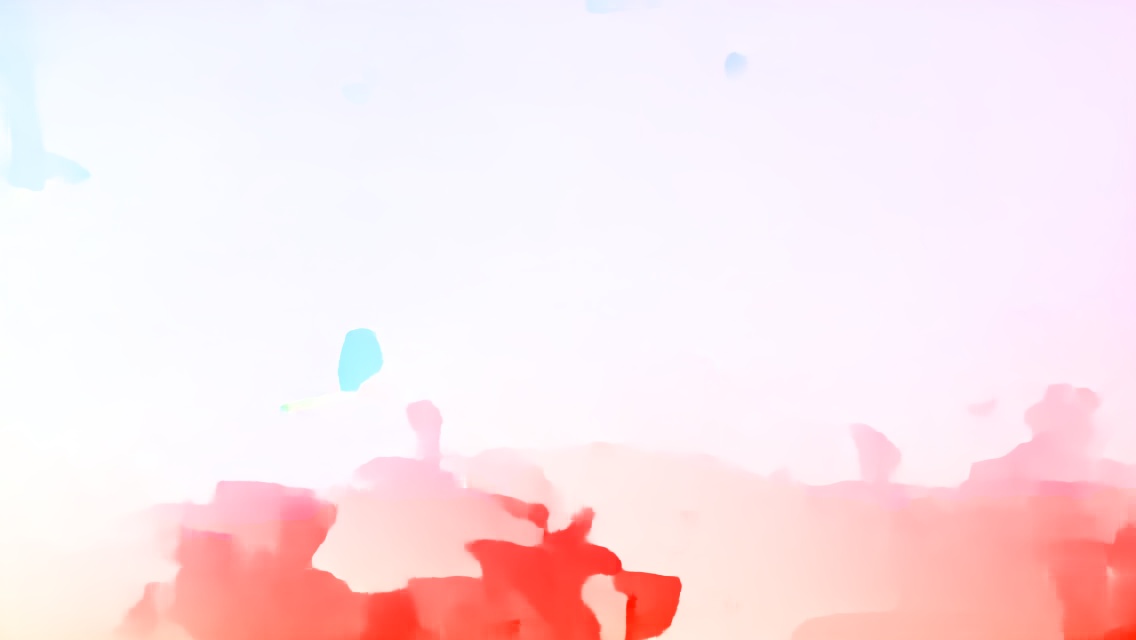}\\

    & \small\rotatebox{90}{\hspace{0.0cm}\color{gray!90}\texttt{things}}\includegraphics[height=0.45in,width=0.65in]{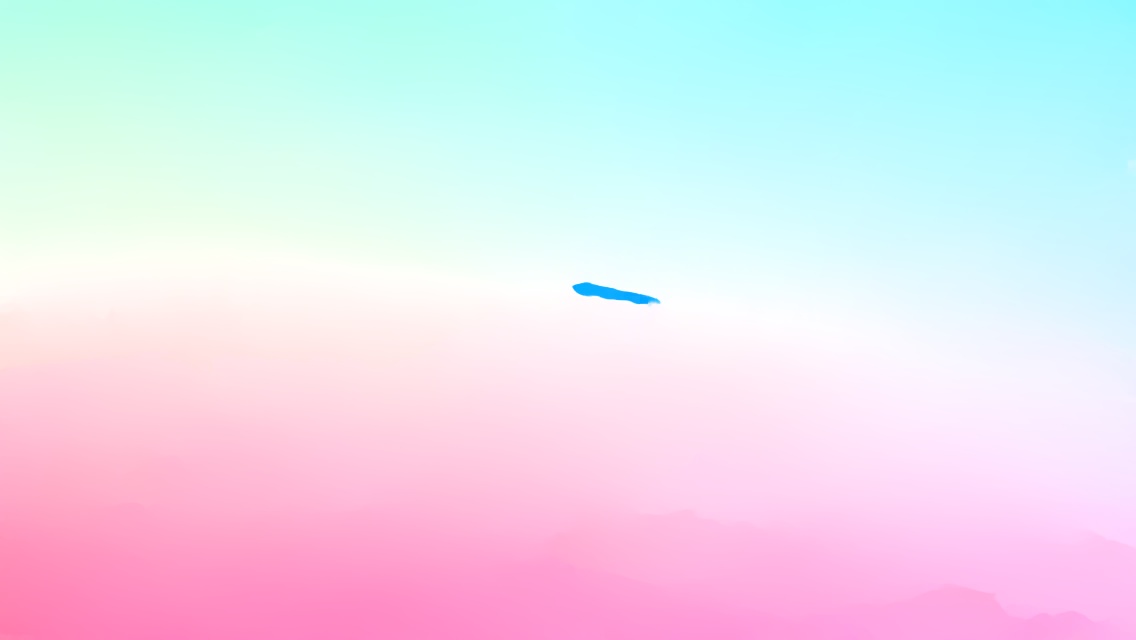}
    & \includegraphics[height=0.45in,width=0.65in]{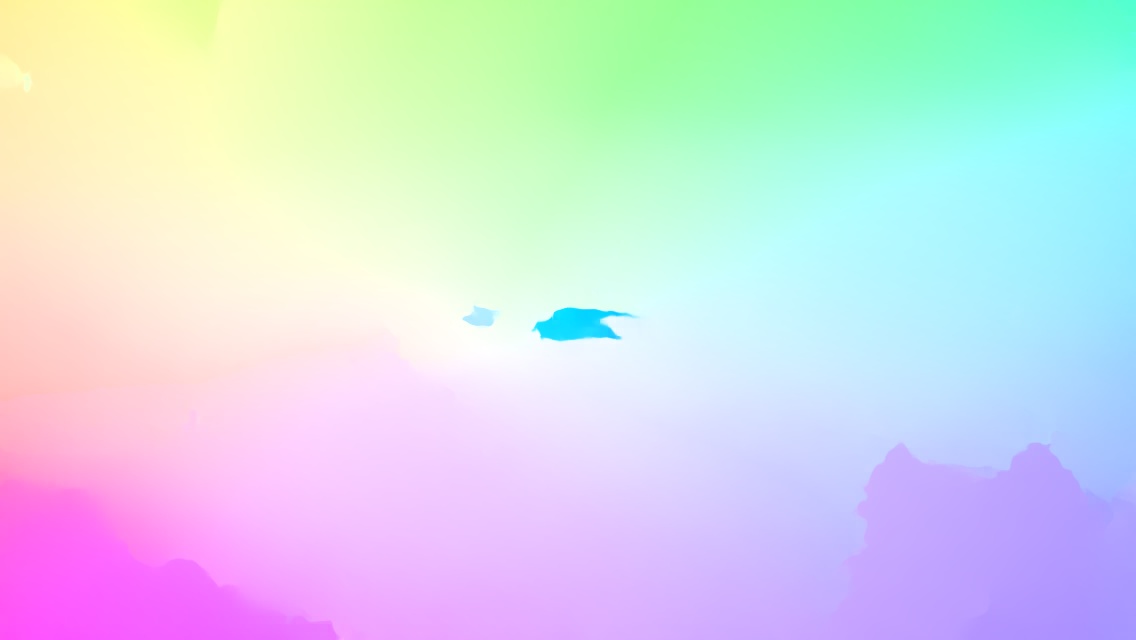}
    & \includegraphics[height=0.45in,width=0.65in]{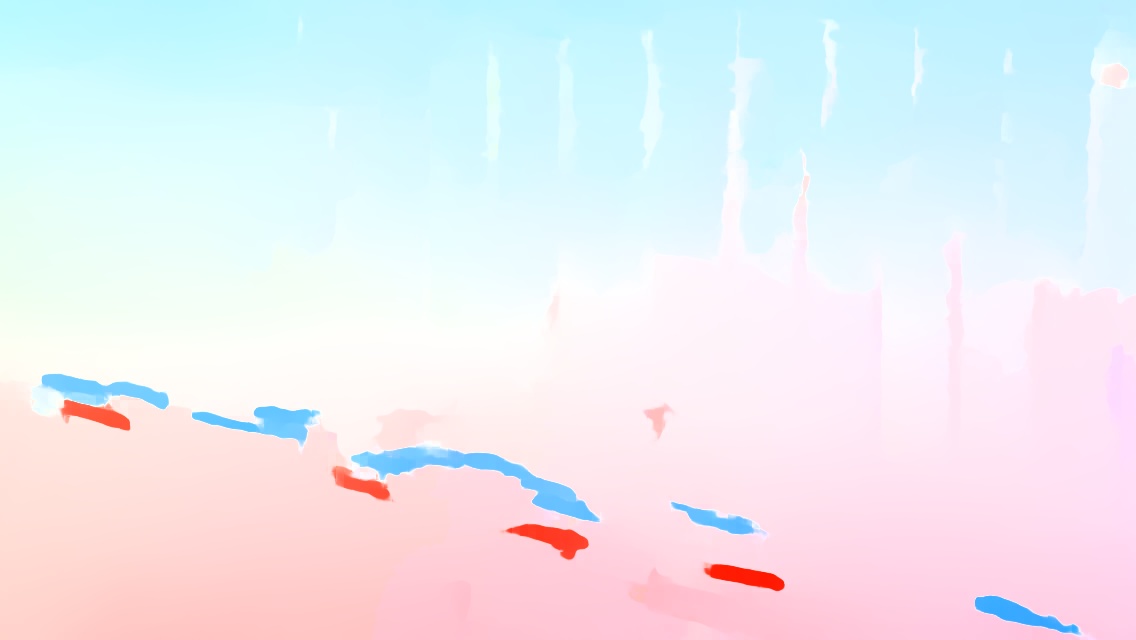}
    & \includegraphics[height=0.45in,width=0.65in]{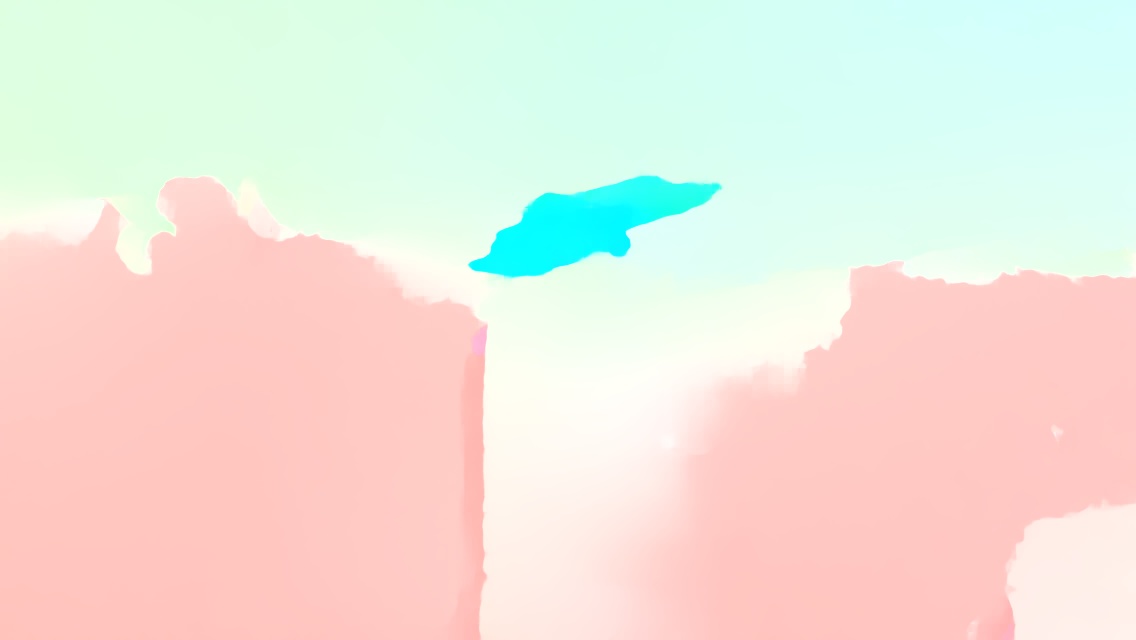}
    & \includegraphics[height=0.45in,width=0.65in]{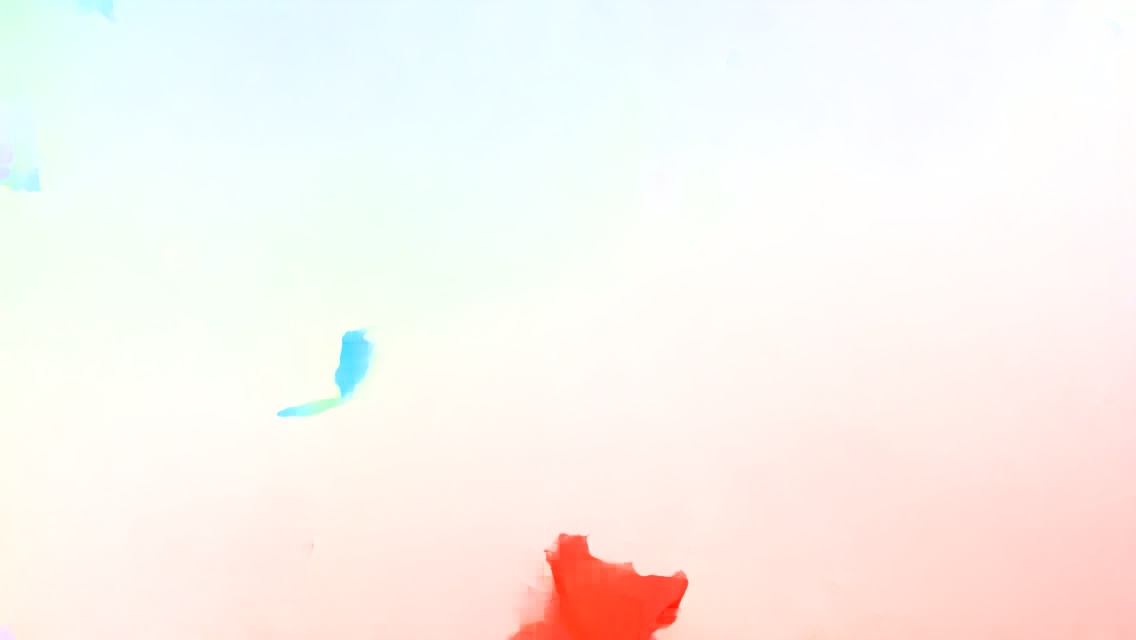} \\

    & \small\rotatebox{90}{\hspace{-0.1cm}\color{gray!90}\texttt{sintel}}\includegraphics[height=0.45in,width=0.65in]{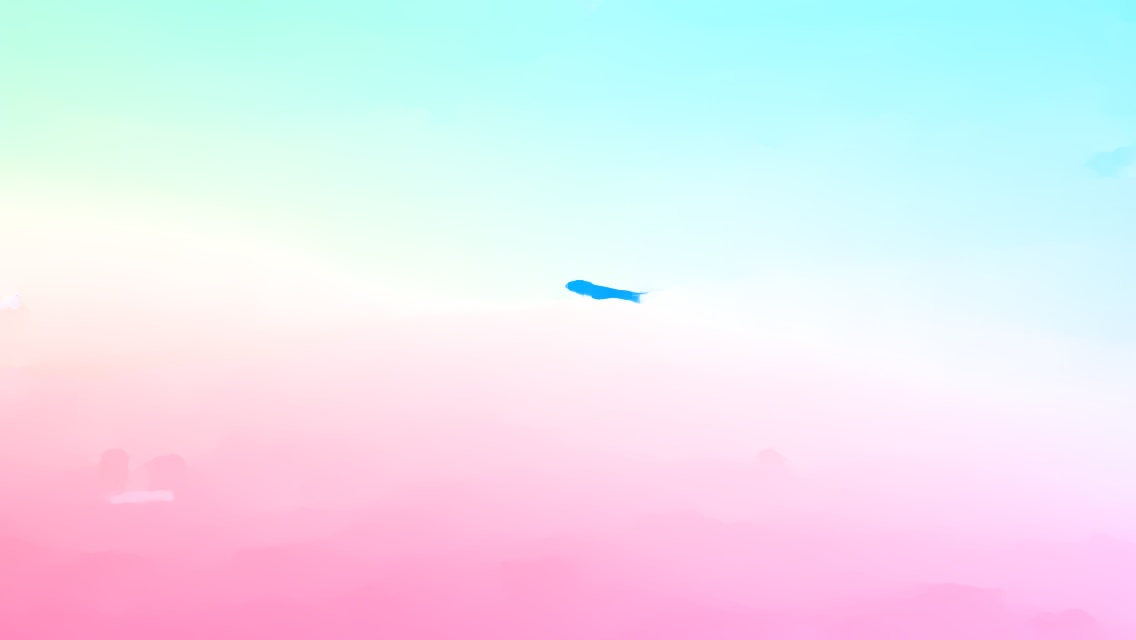}
    & \includegraphics[height=0.45in,width=0.65in]{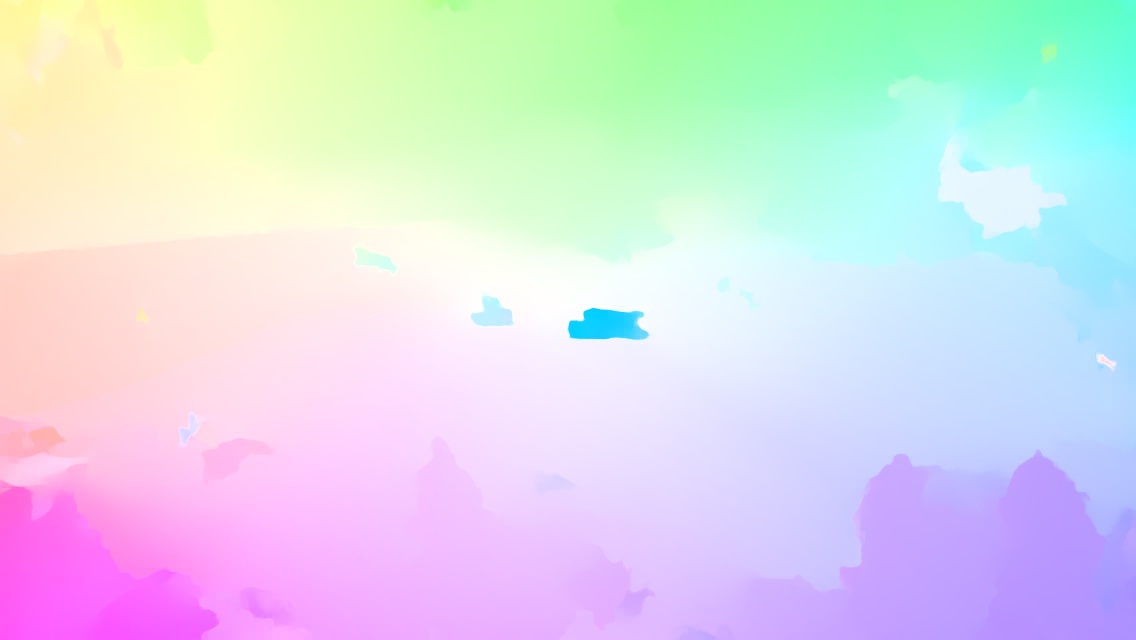}
    & \includegraphics[height=0.45in,width=0.65in]{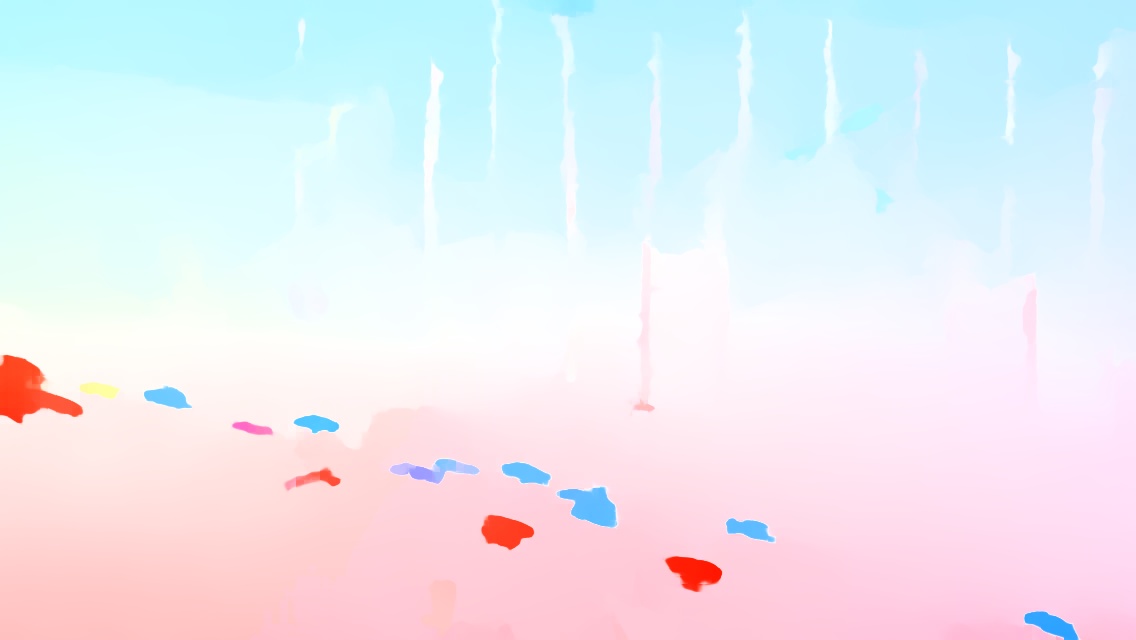}
    & \includegraphics[height=0.45in,width=0.65in]{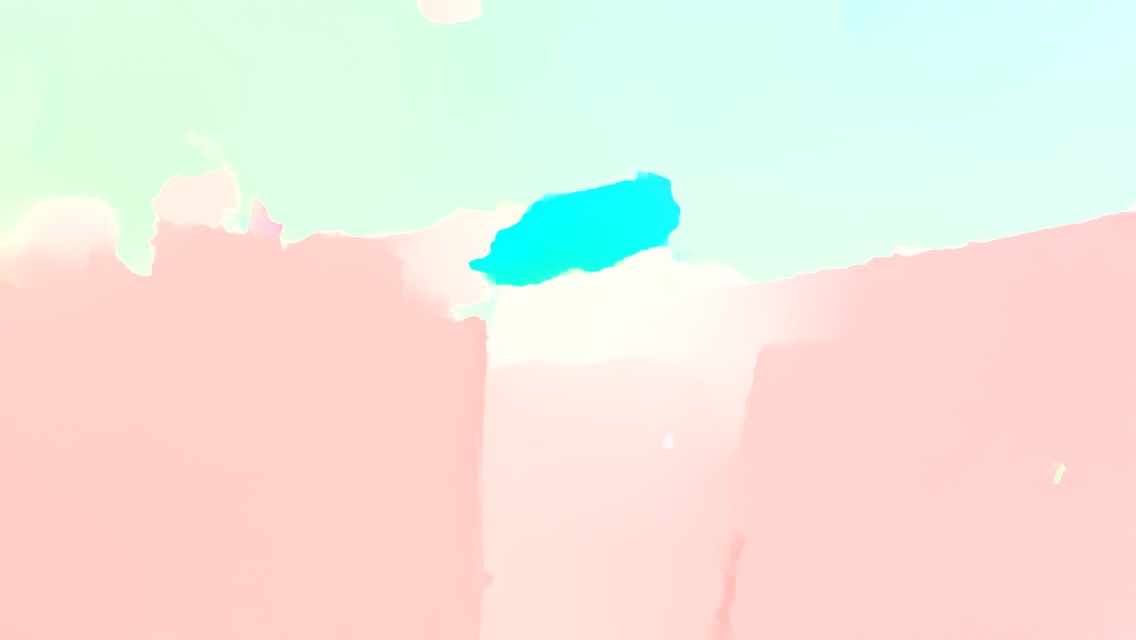}
    & \includegraphics[height=0.45in,width=0.65in]{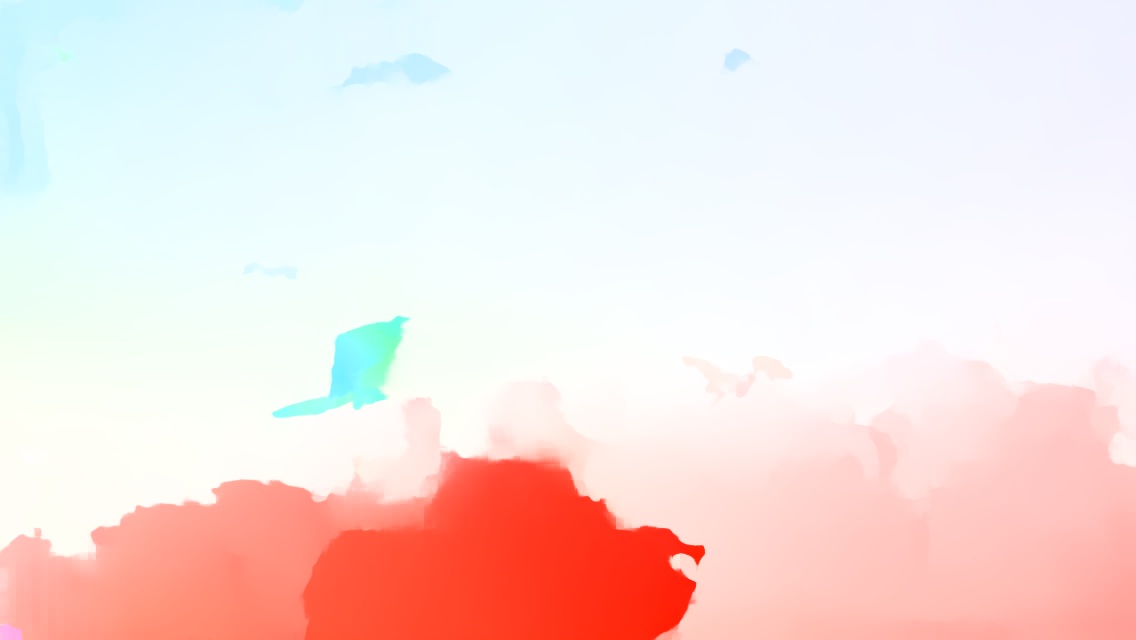} \\
    
\end{tabular}
\caption{Qualitative results of the optical flow output on different RAFT pretrained models with \texttt{sintel} and \texttt{things} models show reliable optical flow across various scenarios.}
\label{tb:opticalflowqualitative}
\end{figure}

\textbf{Impact of RAFT's pretrained models}. To assess the effect of optical flow and to show its relative importance, we ran the algorithm with the same parameter on the Ev-Airborne dataset using different RAFT's pretrained models. Detection rate results are reported in Tab.~\ref{tab:pre-trained-model}. Results indicate that the \texttt{sintel}~\cite{butler2012naturalistic} model is well suited for detecting flow information from event data, followed by \texttt{things}~\cite{mayer_large_2016}. Both models are general-purpose models and trained on a large scale of synthetic datasets of various shapes. Even though the majority of the IMOs in Ev-Airborne consist of moving cars, \texttt{KITTI}~\cite{geiger_vision_2013} and \texttt{chairs}~\cite{dosovitskiy2015flownet} models perform poorly because they focus on a particular task. \texttt{KITTI}~\cite{geiger_vision_2013} model captures cars' motion well but fails at detecting the flow of small IMOs. As can be seen in Fig.~\ref{tb:opticalflowqualitative}, \texttt{sintel} model outperforms the other pretrained models qualitatively on all sequences.

\begin{table}[h]
\centering
\caption{Performance of RAFT's pretrained models on the Ev-Airborne data.}
\label{tab:pre-trained-model}
\begin{tabular}{lc}
\hline
Model & Detection rate [\%] $\uparrow$ \\ \hline
\texttt{KITTI~\cite{geiger_vision_2013}} & 77.83 \\ \hline
\texttt{chairs~\cite{dosovitskiy2015flownet}} & 72.93 \\ \hline
\texttt{things~\cite{mayer_large_2016}} & 79.48 \\ \hline
\texttt{sintel~\cite{butler2012naturalistic}} & \textbf{90.16} \\ \hline
\end{tabular}
\end{table}

\section{Discussion and Limitations}
\label{sec:limitations}

While our method shows remarkable performance, it has limitations. The segmentation accuracy is dependent on the assumption of a constant motion of IMOs and ego-motion within a fixed observation window, limiting it to a constant frame rate. Adopting a dynamic event selection~\cite{huang2023progressive,nunesAdaptiveGlobalDecay2023} or a non-linear motion model~\cite{Seok_2020_WACV} could boost both performance and speed. Without RGB frame data, comparisons to frame-based approaches are restricted, though future work may incorporate this aspect, to show the advantages of the high temporal resolution and dynamic range of event cameras. Like~\cite{wang_-evmoseg_2023,zhou_event-based_2021}, our method segments only moving IMOs, not static ones or those moving at camera speed, incorporating temporal consistency for these scenarios could enhance segmentation. Furthermore, in low signal-to-noise environments, challenges like multiple extrema~\cite{arjaDensityInvariantCMax2023} or event collapse~\cite{shiba_event_2022} may arise, and integrating the solutions from these work could strengthen our algorithm's resilience against noise. Finally, optimizing the pipeline to reduce the complexity and latency of the network is another promising future work, especially the coarse mask prediction and the B-CMax process.

\section{Conclusion}
\label{sec:conclusion}

This paper proposes an unsupervised architecture to segment any moving objects from an aerial platform with a high-resolution event camera. This allows us to perform motion segmentation on challenging scenes, with HDR and/or complex backgrounds and leverages the unique capabilities of event cameras. Our method removes the need for labelled segmentation masks by harnessing features from self-supervised transformers on both the event data and optical flow and adopting dynamic mask refinement to enhance spatial consistency. Our method not only segments objects effectively but also produces motion-compensated images with sharp edges for both IMOs and the background, which can be valuable for further tasks such as recognition and tracking. We have rigorously tested our method, demonstrating its superiority and generalization capabilities over existing state-of-the-art methods, quantitatively and qualitatively, on well-known public motion segmentation datasets. Additionally, we are making our Ev-Airborne dataset available to the research community to encourage further progress in this area.

\section*{Acknowledgments}
We thank Shotover Systems for recording the aerial event-based dataset, which was instrumental in the development of the Ev-Airborne dataset. This work was supported by the Air Force Office of Scientific Research (AFOSR) under grant FA2386-23-1-4005.

\bibliographystyle{IEEEtran}
\bibliography{egbib}

\vspace{11pt}

\vfill

\end{document}


\title{Motion Segmentation for Neuromorphic Aerial Surveillance \\ (Supplementary Materials)} 

\titlerunning{Neuromorphic Aerial Surveillance}

\author{Sami Arja\inst{*}\orcidlink{0000-0001-5852-5828} \and
Alexandre Marcireau \inst{}\orcidlink{0000-0003-4488-033X}\and
Saeed Afshar\inst{}\orcidlink{0000-0002-2695-3745}\and
Bharath Ramesh\inst{}\orcidlink{0000-0001-8230-3803}\and
Gregory Cohen\inst{}\orcidlink{0000-0003-0738-7589}}

\authorrunning{Arja et al.}

\institute{Western Sydney University 
\\ \email{$^{*}$s.elarja@westernsydney.edu.au}\\
}

\maketitle

\appendix

\section{Ev-Airborne dataset}

The Ev-Airborne dataset, summarized in Table~\ref{tb:datasetsummary}, was captured from on a small airplane using a Prophesee Gen 4 event camera with a resolution of 1280x720 (HD). The event camera was mounted on a gyro-stabilized gimbal with an adjustable field-of-view, able to look both downwards and at the horizon described as "low oblique" and "high oblique" in the Table. As a result, the dominant motion in the dataset is primarily translational. The gimbal is also capable of moving at the same speed as the platform, effectively cancelling out the camera's ego-motion, allowing only moving objects to trigger events.

The airplane flew over various areas in the United States, primarily urban locations such as highways, airports, golf courses, and residential areas. During these flights, the camera recorded various moving objects of different sizes, shapes, distances, and speeds, where the main moving objects captured were cars, humans, and an airplane taking off from an airport. To create a suitable dataset for motion segmentation tasks, we selected sequences with diverse and visible motion, trimming them to durations of 5 to 10 seconds. This ensures that there are enough moving objects in the field of view, with and without camera ego-motion, to test the robustness and reliability of the proposed method.

\begin{table}[h]
\caption{Characteristics of the Ev-Airborne dataset}
\centering
\begin{tabular}{lcccc}
\hline
Date-of-recording & Duration(s) & \#Events & Objects type & Field-of-View \\ \hline
 2023-04-26 14:54:52    &    10.7      &   2796448            &  Cars    &   low oblique   \\ \hline
 2023-04-26 15:14:21    &    10.7      &   13010984           &  Cars    &   high oblique   \\ \hline
 2023-04-26 14:53:22    &    3.9       &   2825230                &  Airplane    &   high oblique   \\ \hline
  2023-04-26 15:45:10   &    5.9  &   3676541              &  Cars    &   low oblique   \\ \hline
  2023-04-26 15:30:21   &    10.7 &   6205372               &  Humans    &   low oblique   \\ \hline
  2023-04-26 15:30:21   &    10.7  &   31636630                 &  Cars    &   low oblique   \\ \hline
  2023-04-26 15:30:21   &    8.9      &  5348006                &  Cars    &   low oblique   \\ \hline
  2023-04-26 15:30:21   &    5.7      &  7343909                &  Golf car    &   high oblique   \\ \hline
  2023-04-26 15:30:21   &    4.9      & 3714492                 &  Cars    &   low oblique   \\ \hline

\end{tabular}
\label{tb:datasetsummary}

\end{table}

\section{The B-CMax algorithm}

\begin{figure}[h] %
  \centering
  \includegraphics[width=5in]{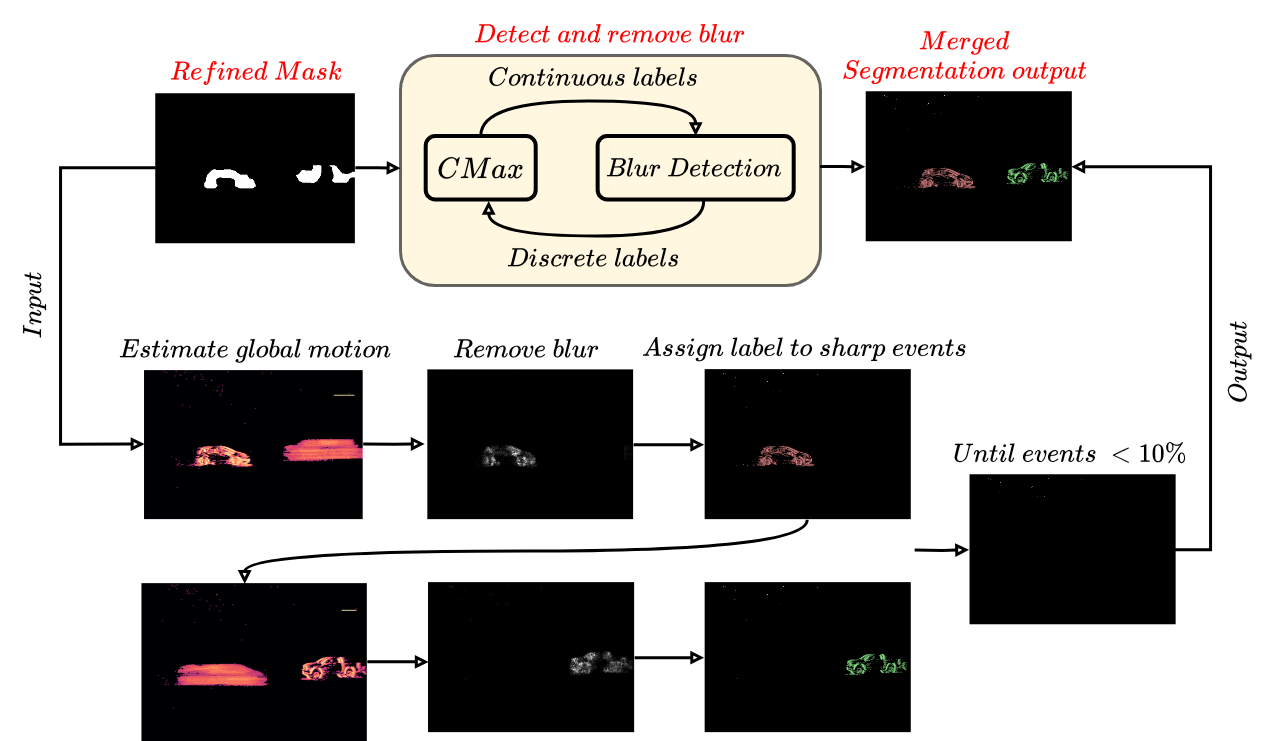}
  \caption{Illustrating the iterative process of the B-CMax algorithm. It demonstrates how CMax is employed to assign a continuous label, while blur detection is utilized to assign a discrete label to each event in an iterative manner.}
   \label{fig:bcmax_process}
\end{figure}

In this section, we provide more details on the B-CMax algorithm, a crucial component of our architecture. Figure~\ref{fig:bcmax_process} demonstrates the application of CMax and blur detection for iterative assignment of discrete and continuous labels to produce a fully segmented scene. This process involves the selection and removal of events. A refined mask from the DMR is provided as input, processing only events within this mask. CMax estimates the global motion, producing a motion-compensated image where sharp edges indicate objects with dominant motion. Subsequently, blur detection generates an attention map highlighting sharp edges with high-intensity pixels, while assigning very low values to blurry events. Consequently, discrete labels are assigned to events at sharp edges, while blurry events are passed to the next processing stage, where the same iterative assignment repeats on the remaining events. This cycle repeats until less than 10\% of the events remain, primarily consisting of noise events, which are then categorized as background.

\section{Additional Qualitative Results}
\setlength{\parskip}{10pt}

\begin{figure}[t] 
\centering
\renewcommand*{\arraystretch}{0.3}
\setlength{\tabcolsep}{0.5pt} %
\textbf{Oversegmentation}
\begin{tabular}{c c c c c c c c c}
    \tiny\rotatebox{90}{\hspace{0.1cm}\color{gray!90}EMSGC\cite{zhou_event-based_2021}} & \includegraphics[height=0.56in,width=0.67in]{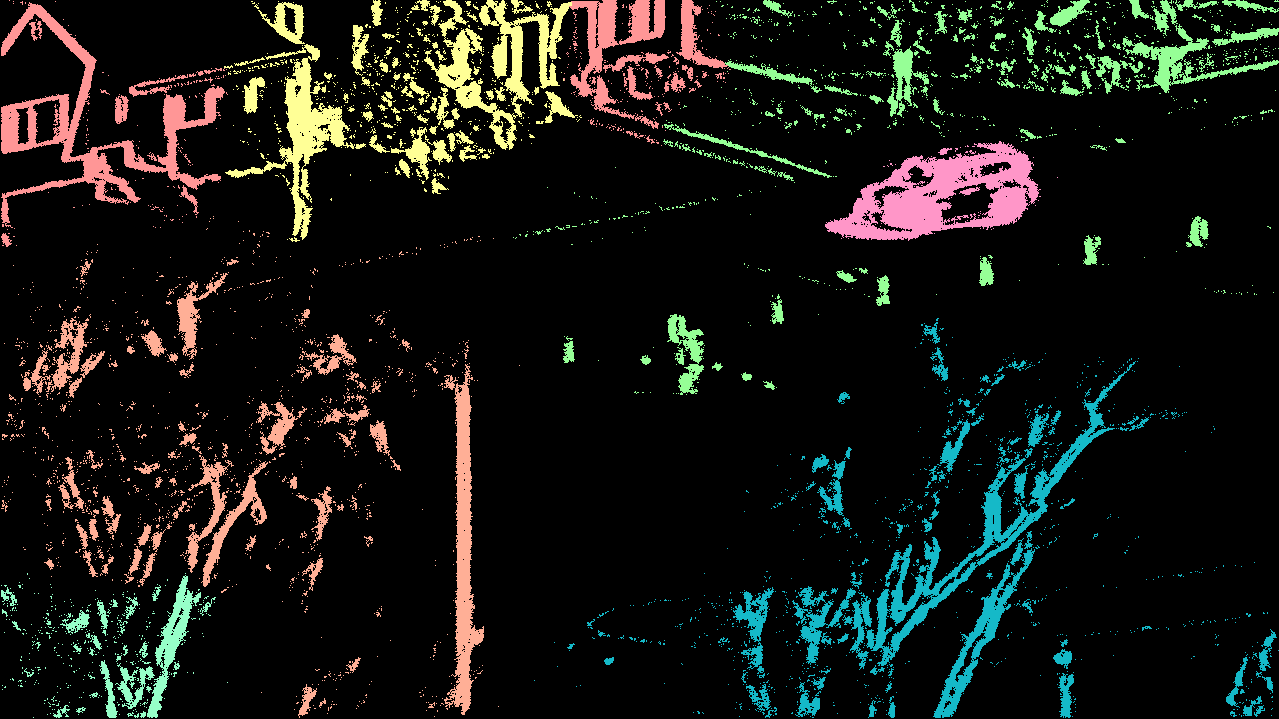}
    & \includegraphics[height=0.56in,width=0.67in]{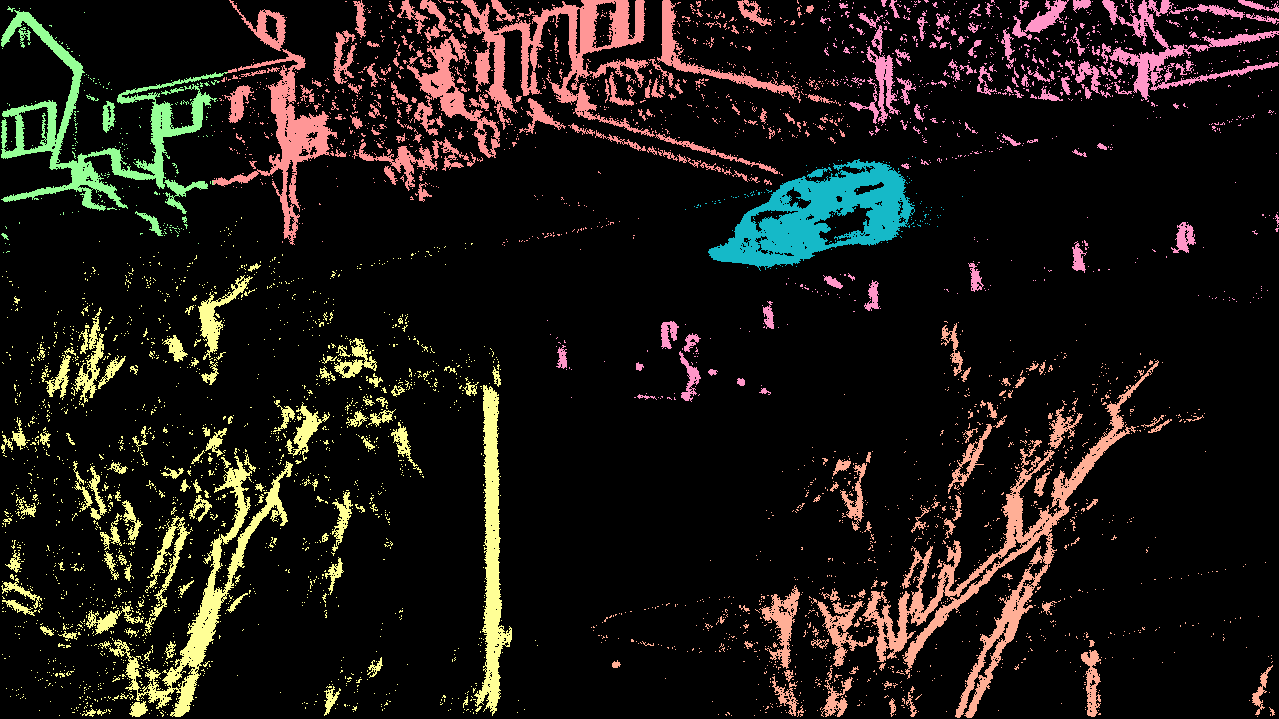}
    & \includegraphics[height=0.56in,width=0.67in]{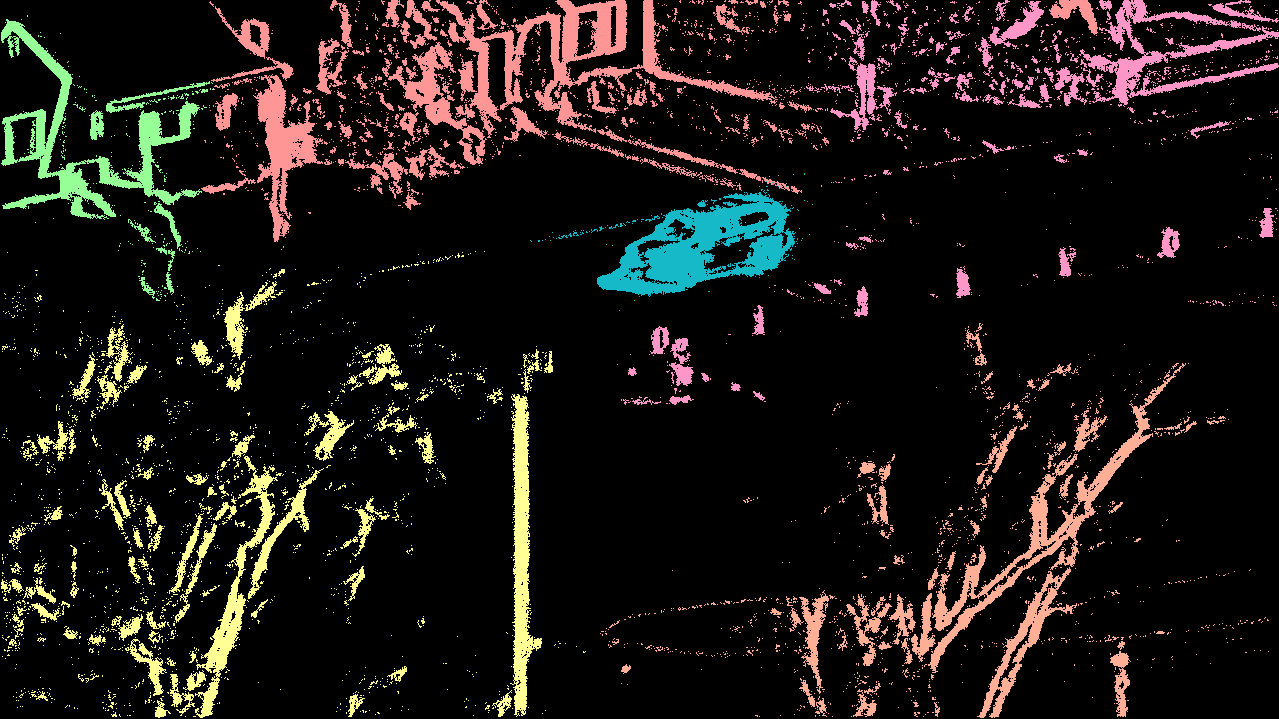}
    & \includegraphics[height=0.56in,width=0.67in]{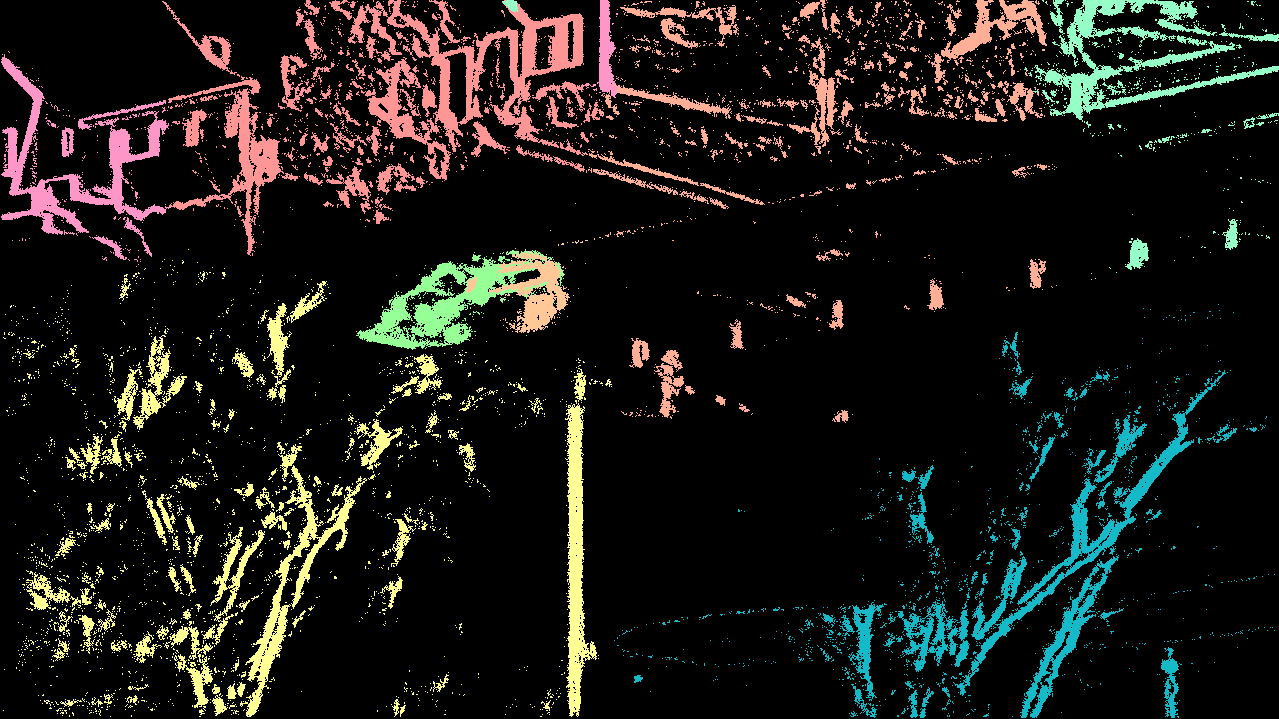}
    & \includegraphics[height=0.56in,width=0.67in]{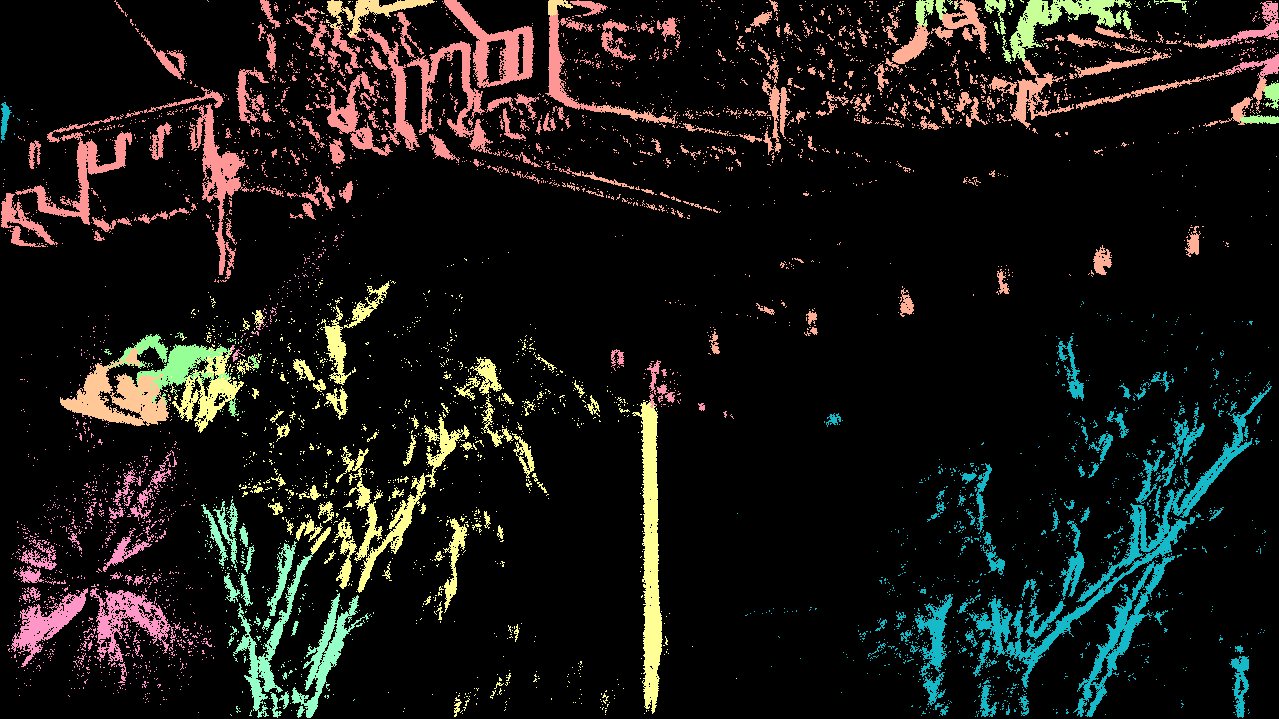} & 
    \includegraphics[height=0.56in,width=0.67in]{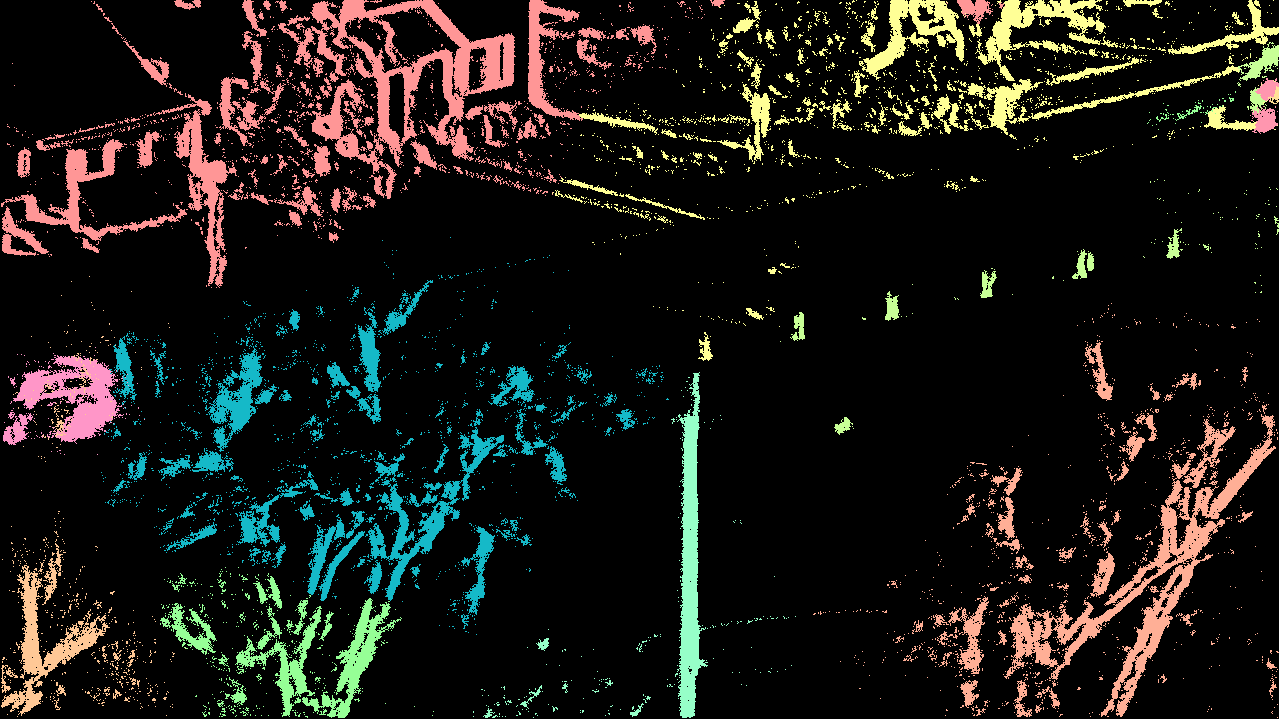} &
    \includegraphics[height=0.56in,width=0.67in]{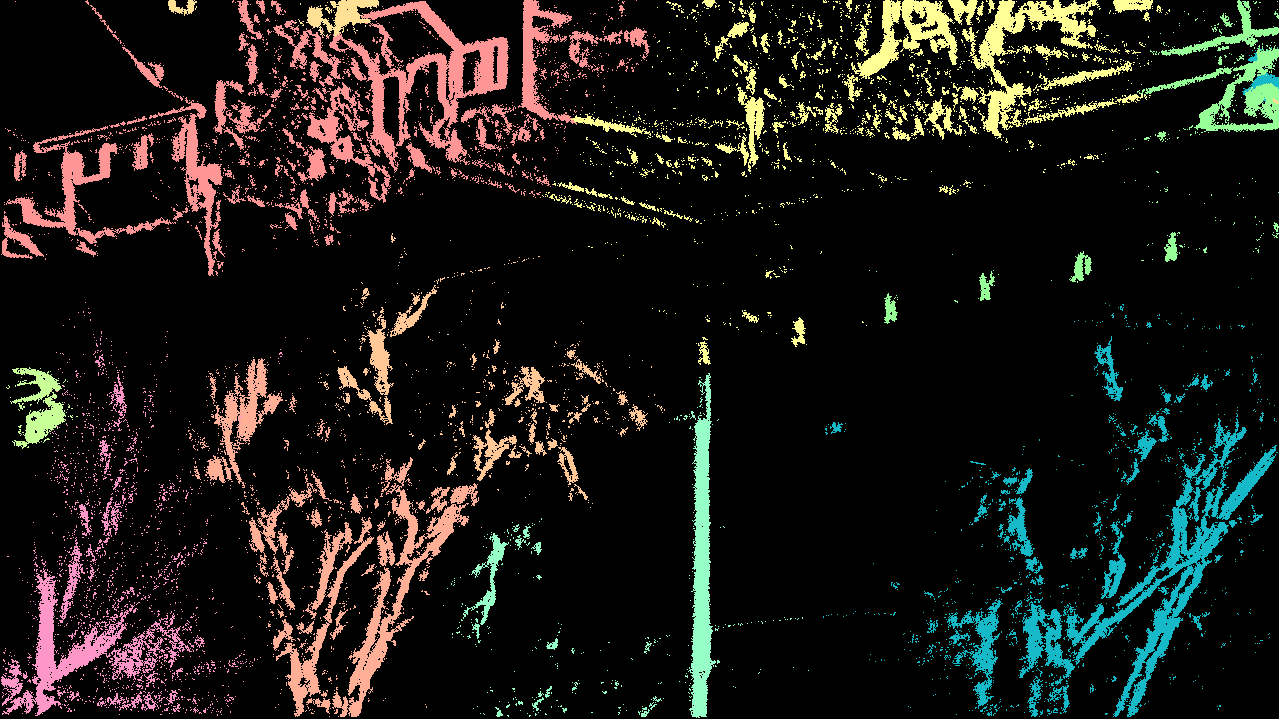} \\
    
    \rotatebox{90}{\hspace{0.4cm}\color{gray!90}Ours} & \includegraphics[height=0.56in,width=0.67in]{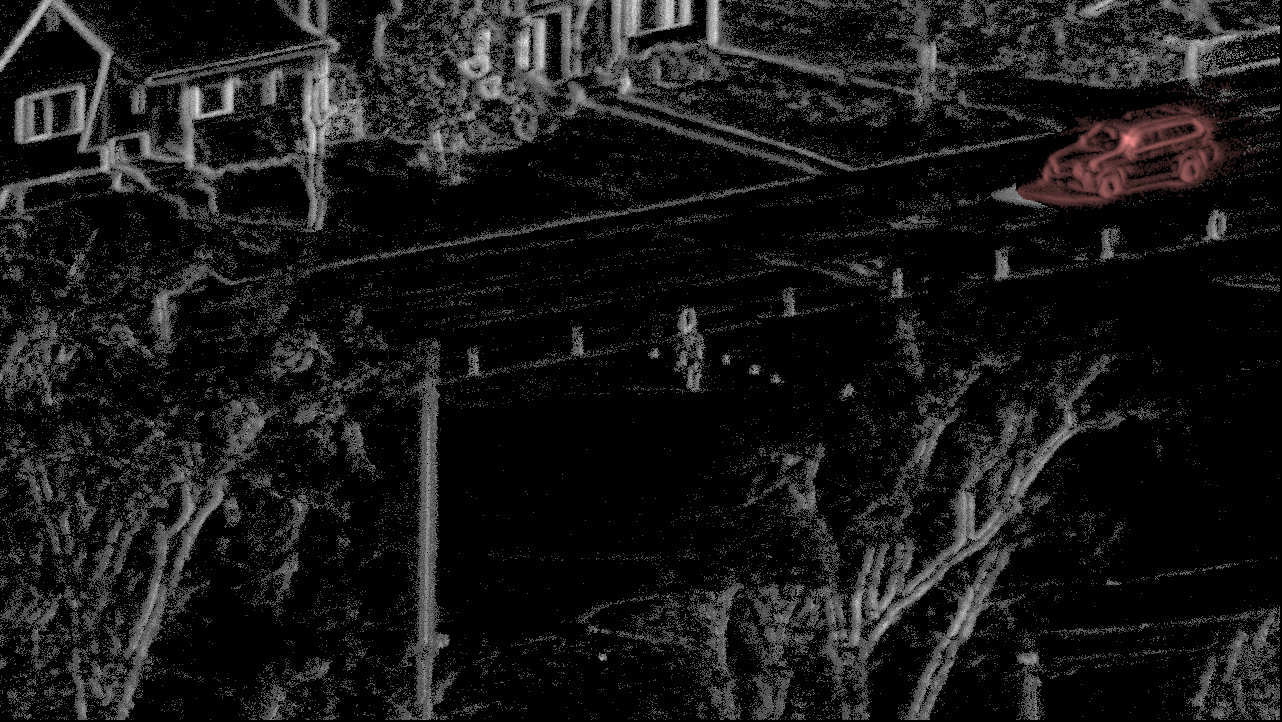}
    & \includegraphics[height=0.56in,width=0.67in]{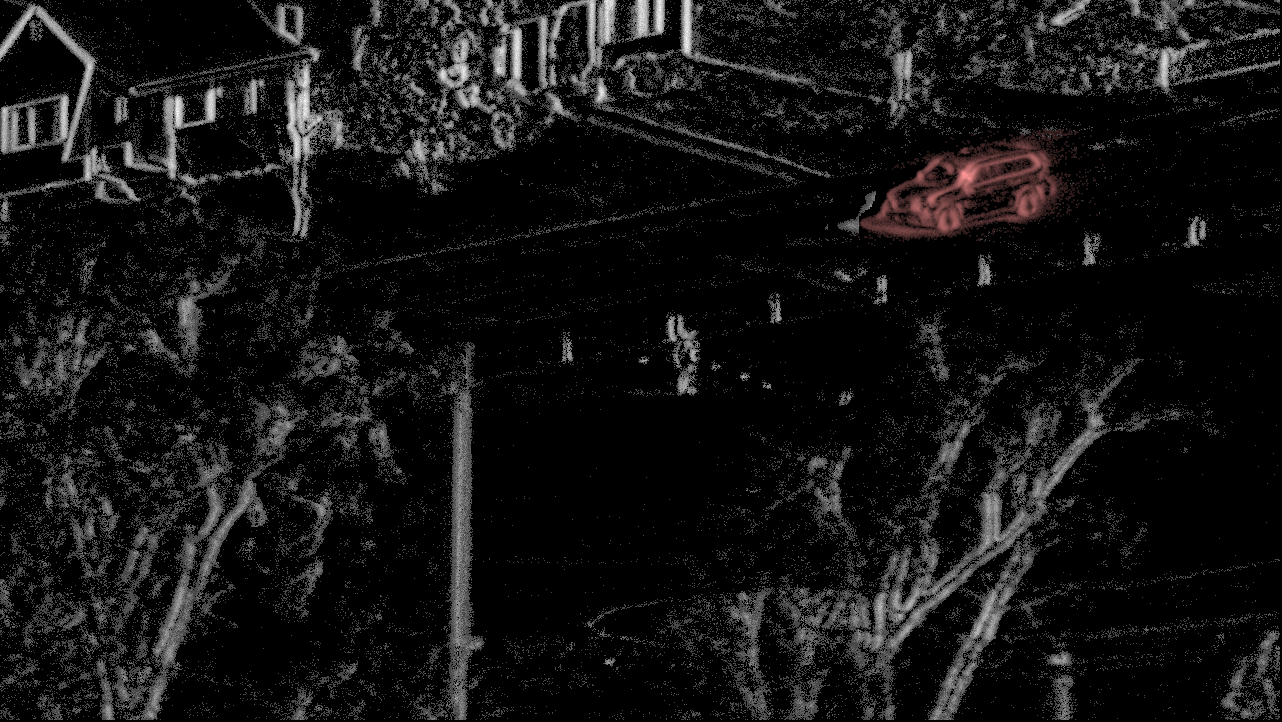}
    & \includegraphics[height=0.56in,width=0.67in]{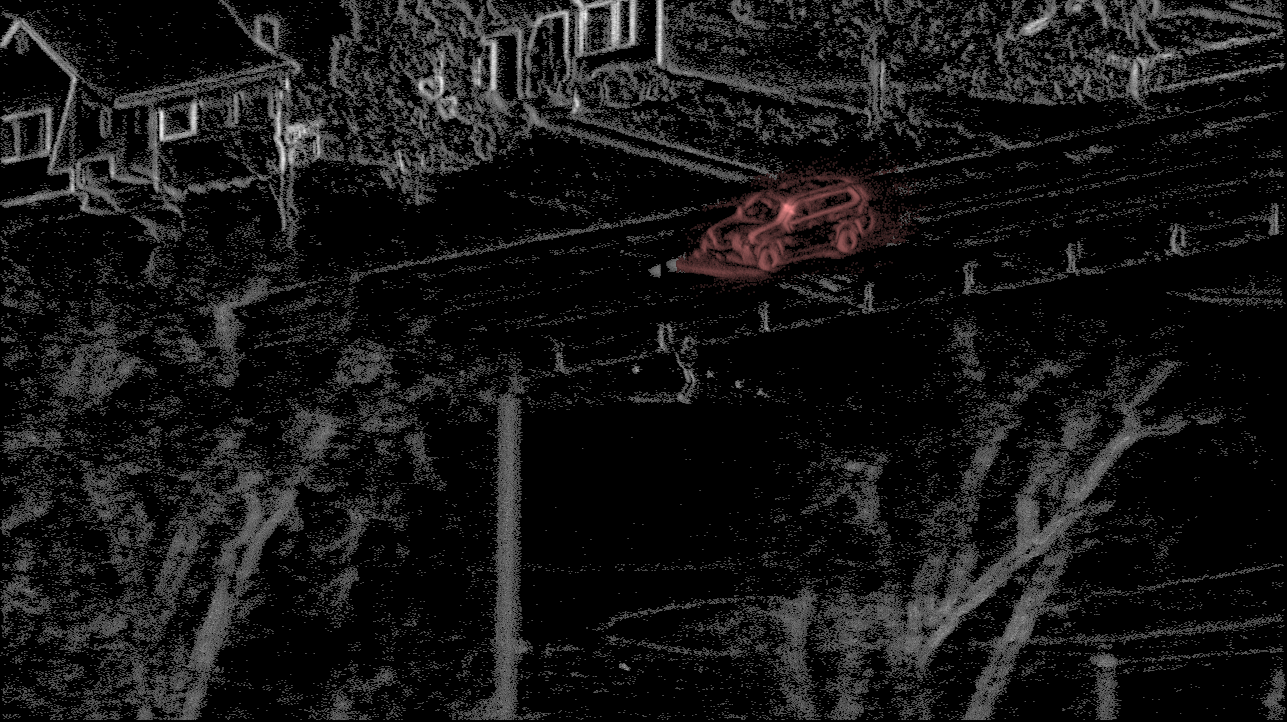}
    & \includegraphics[height=0.56in,width=0.67in]{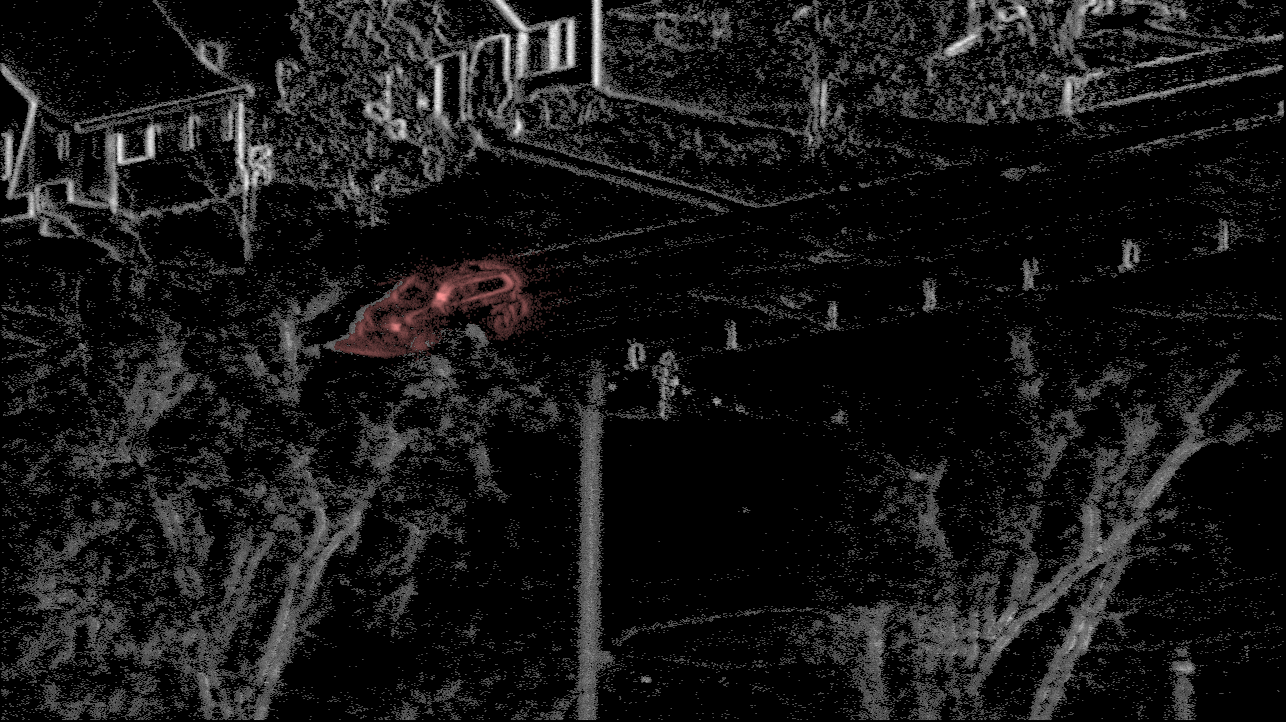}
    & \includegraphics[height=0.56in,width=0.67in]{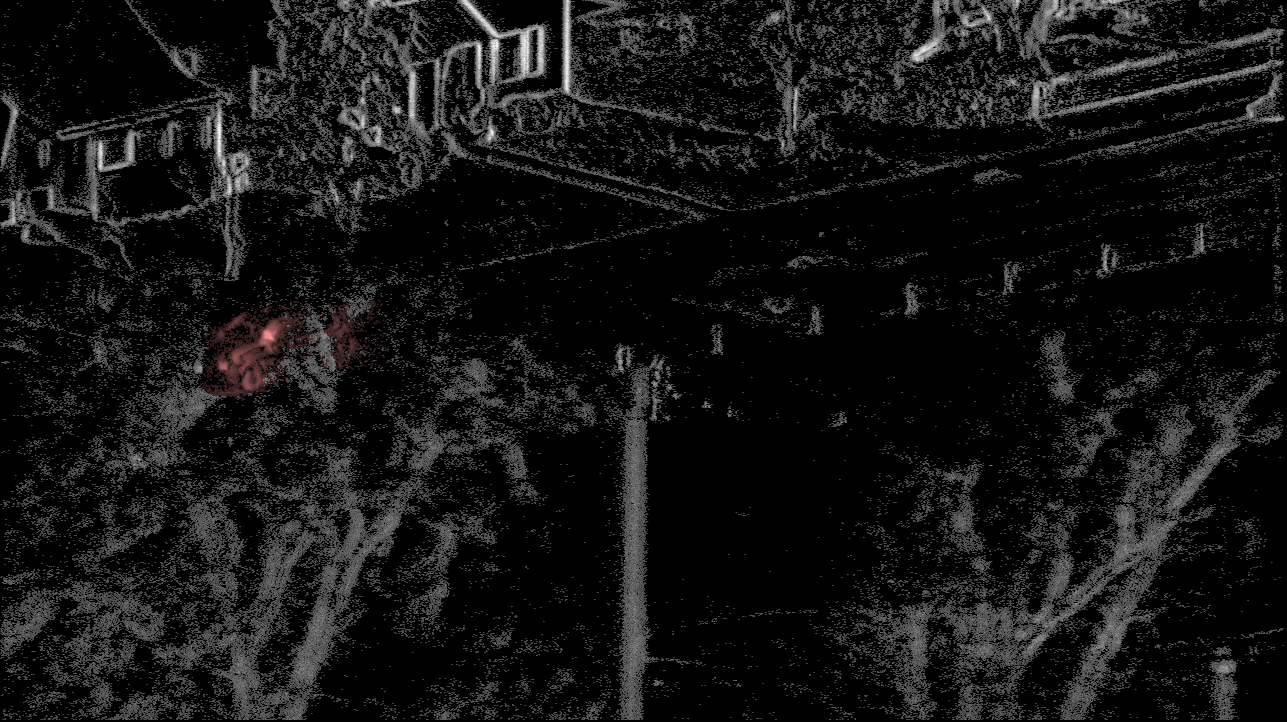} &
    \includegraphics[height=0.56in,width=0.67in]{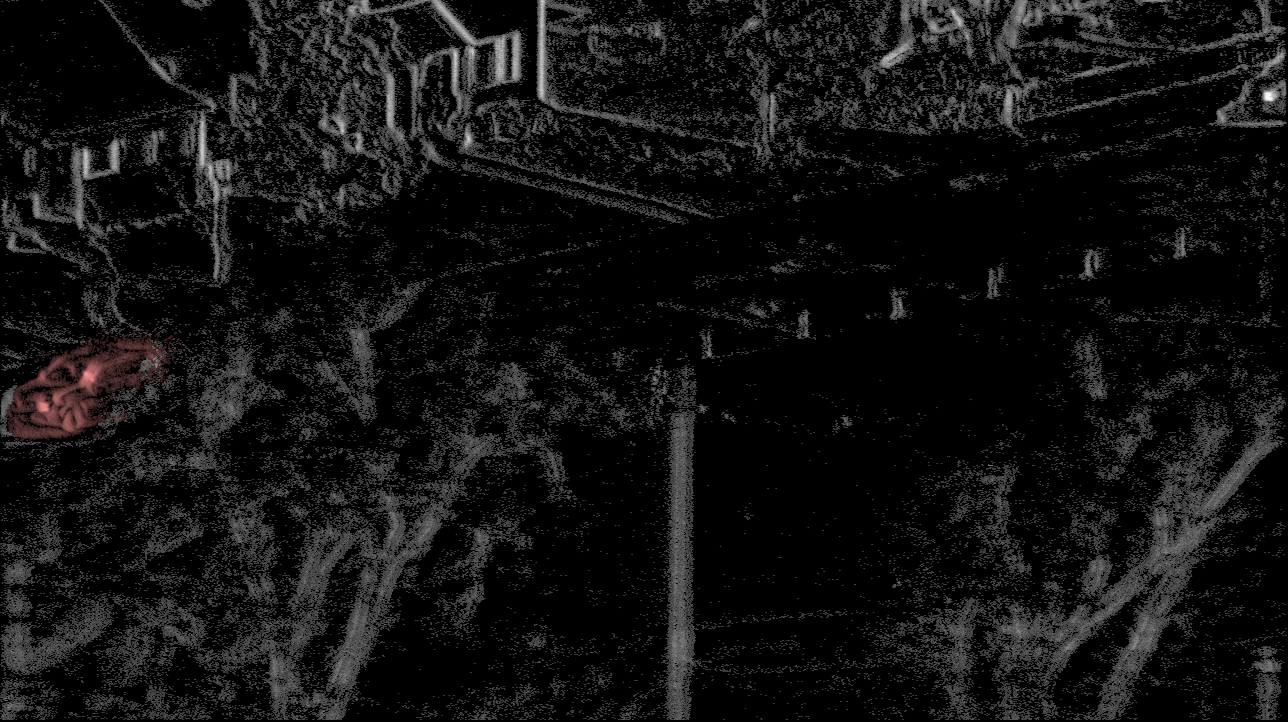}  &
    \includegraphics[height=0.56in,width=0.67in]{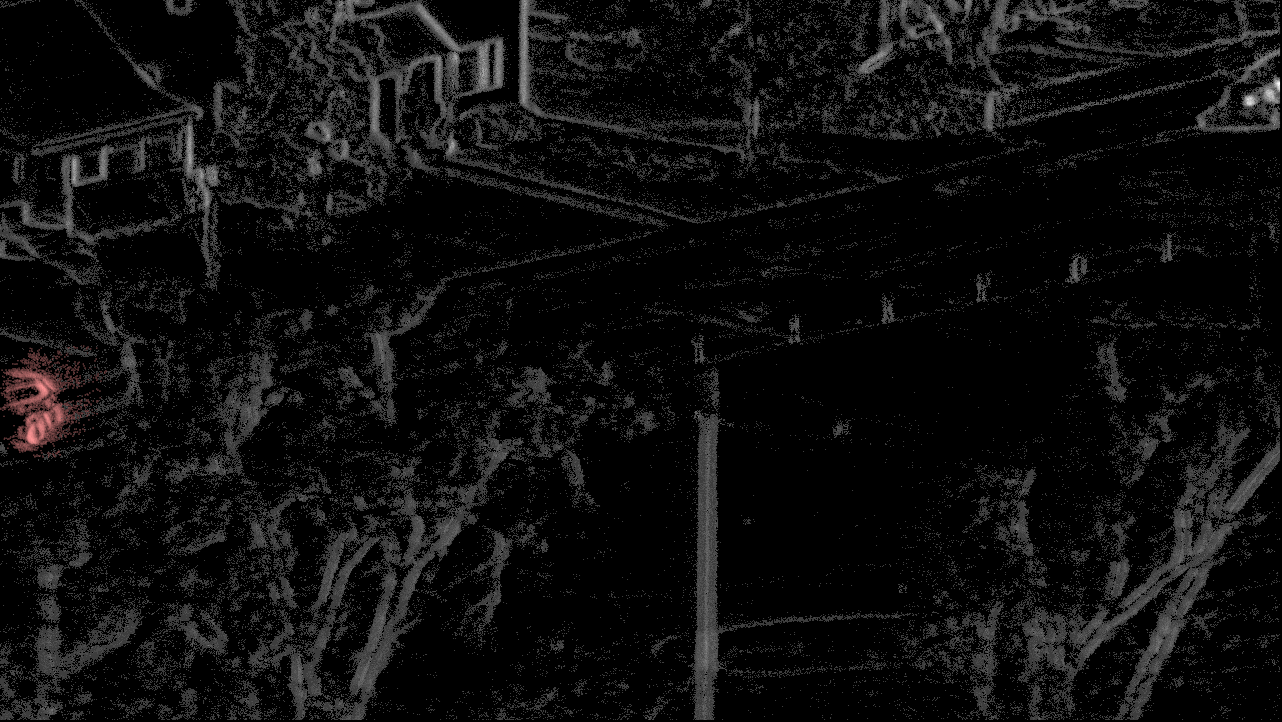} \\
\end{tabular}
\textbf{Dealing with small IMOs}
\begin{tabular}{c c c c c c c c c}
  \tiny\rotatebox{90}{\hspace{0.1cm}\color{gray!90}EMSGC\cite{zhou_event-based_2021}} & \includegraphics[height=0.56in,width=0.67in]{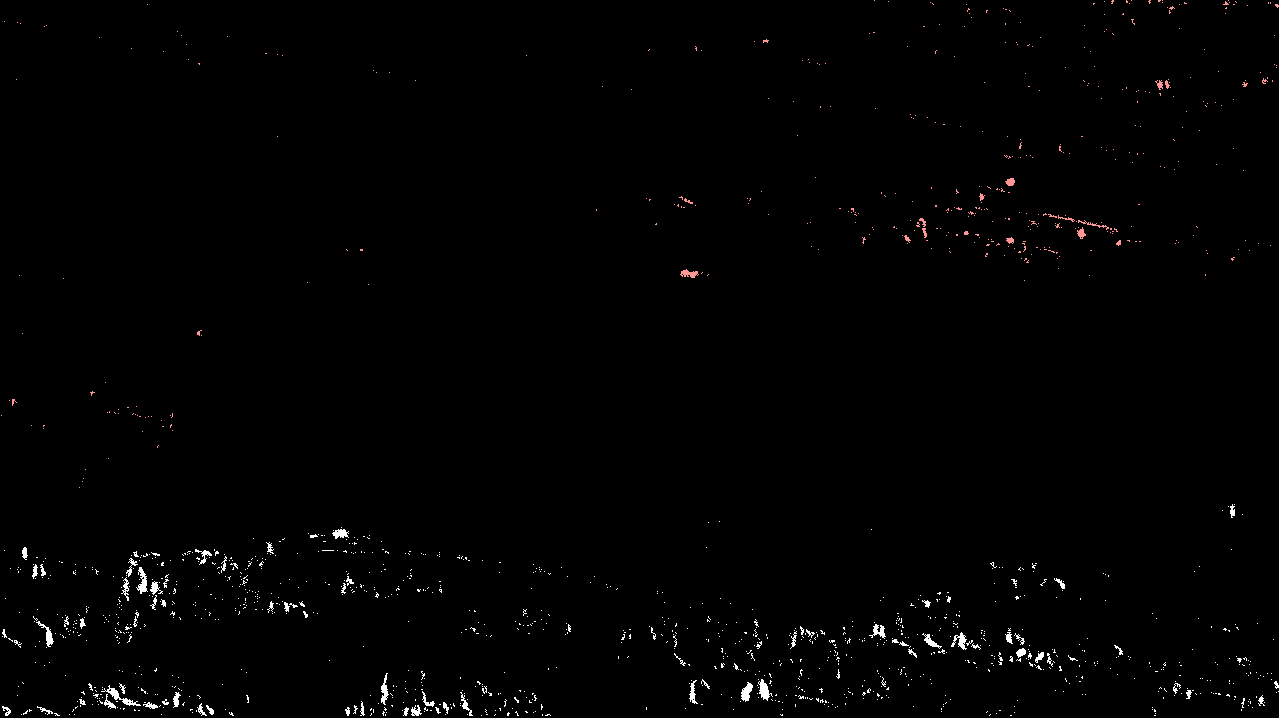}
    & \includegraphics[height=0.56in,width=0.67in]{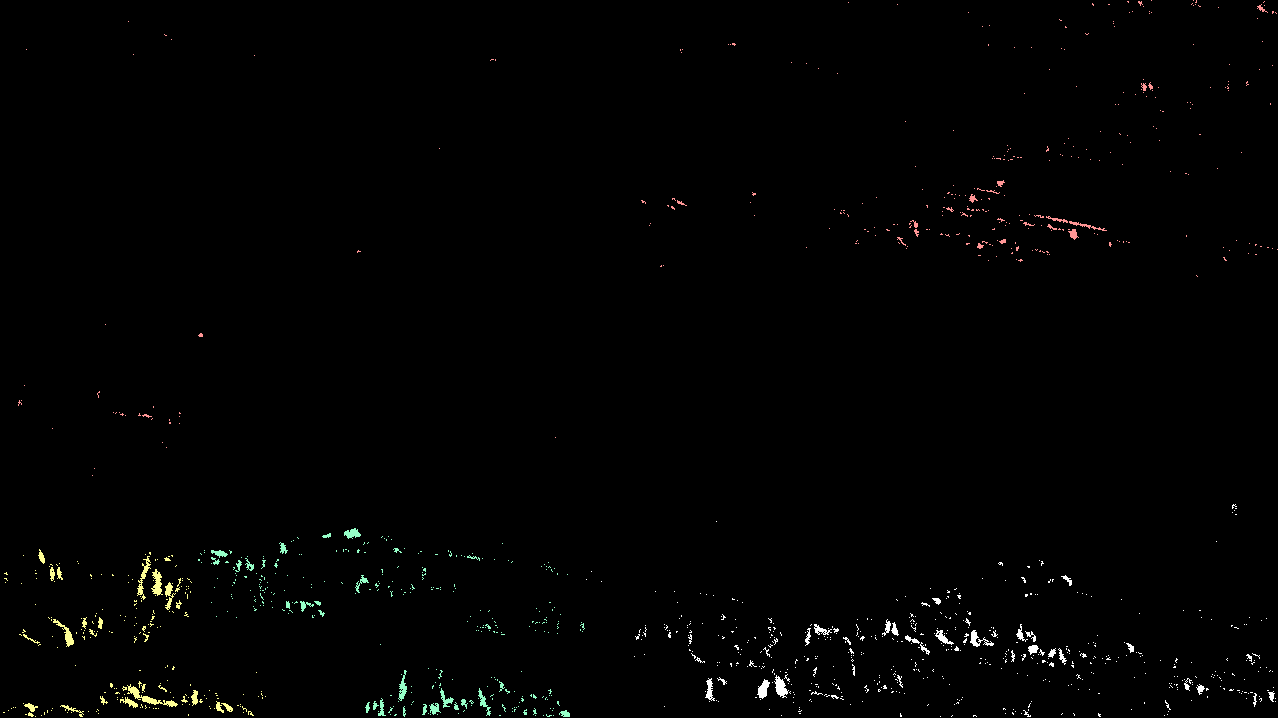}
    & \includegraphics[height=0.56in,width=0.67in]{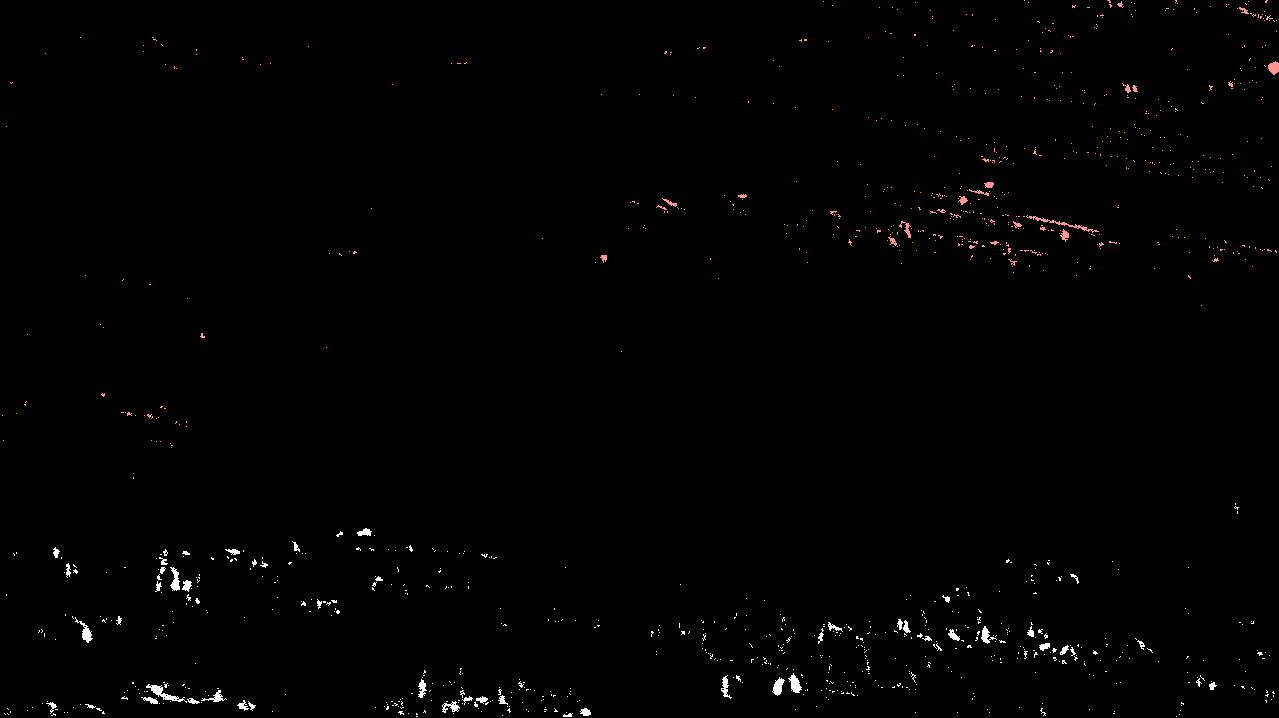}
    & \includegraphics[height=0.56in,width=0.67in]{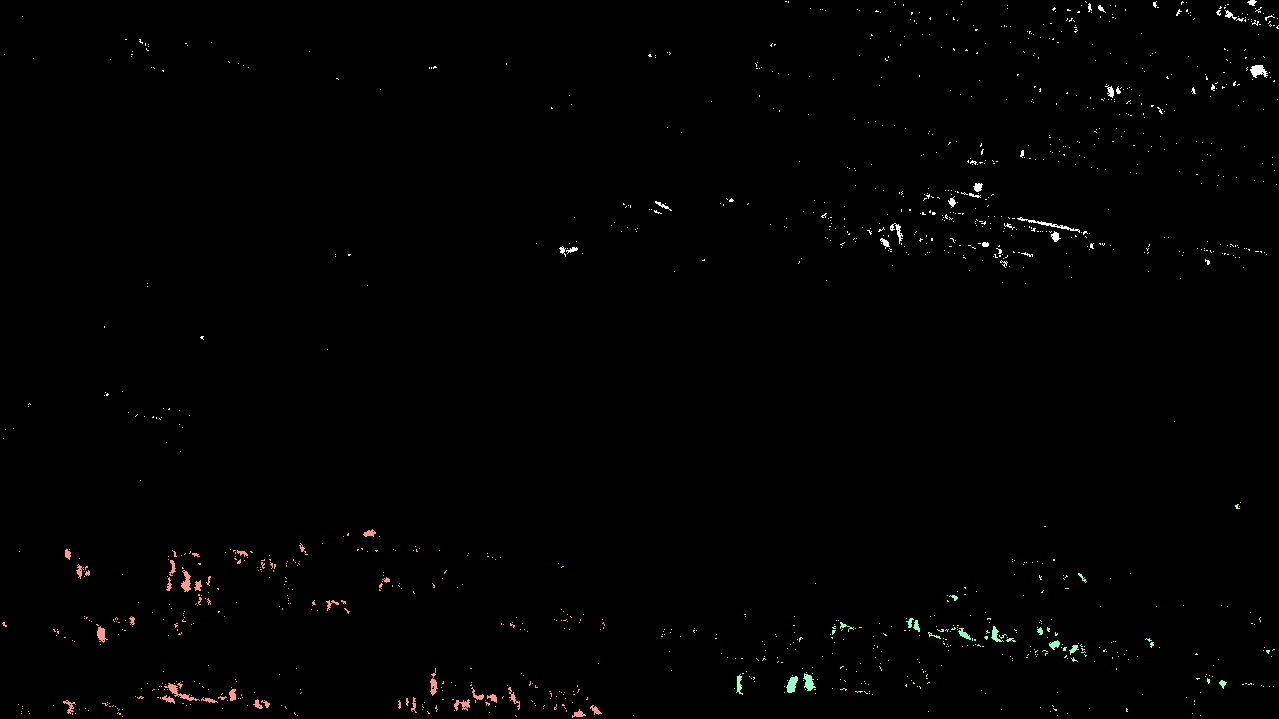}
    & \includegraphics[height=0.56in,width=0.67in]{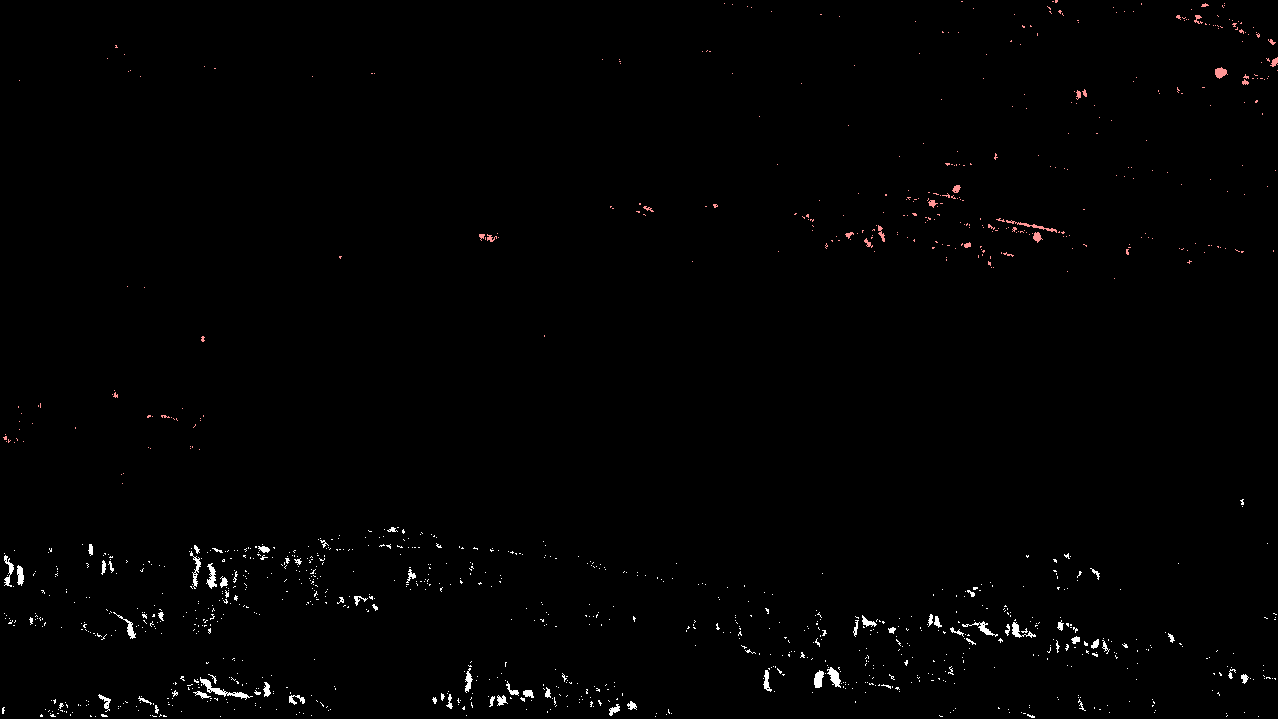} &
    \includegraphics[height=0.56in,width=0.67in]{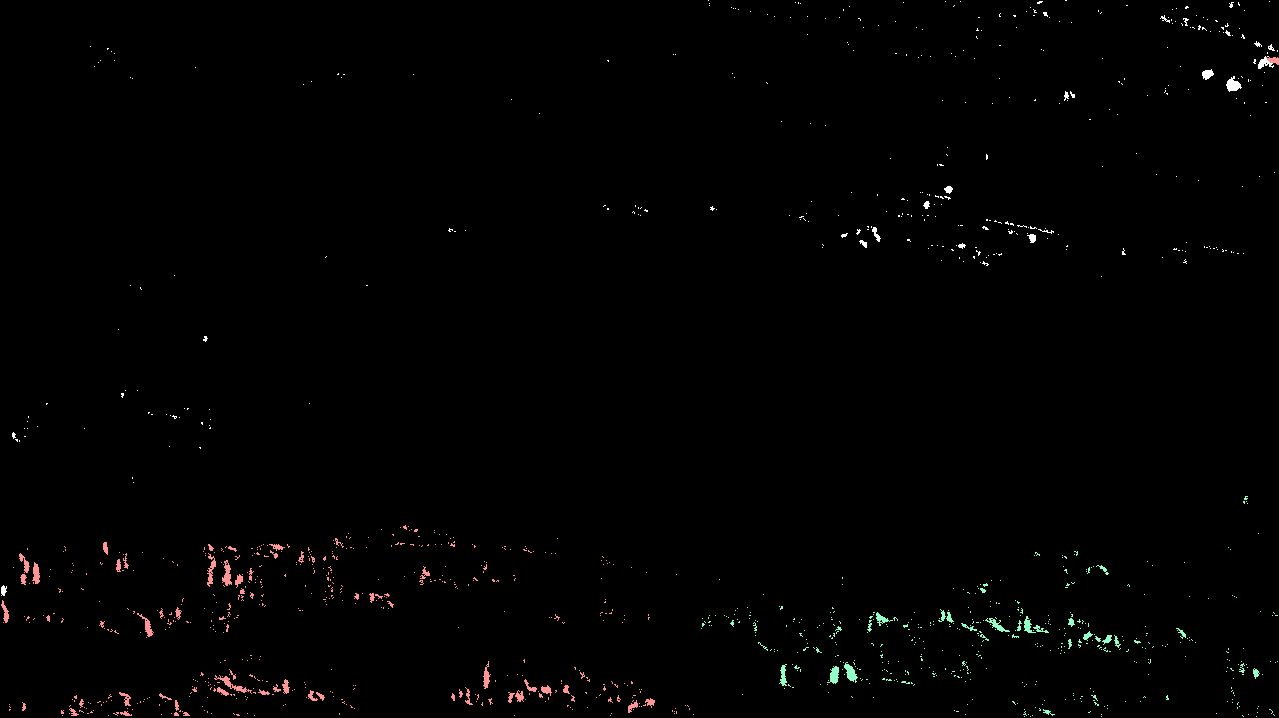}  &
    \includegraphics[height=0.56in,width=0.67in]{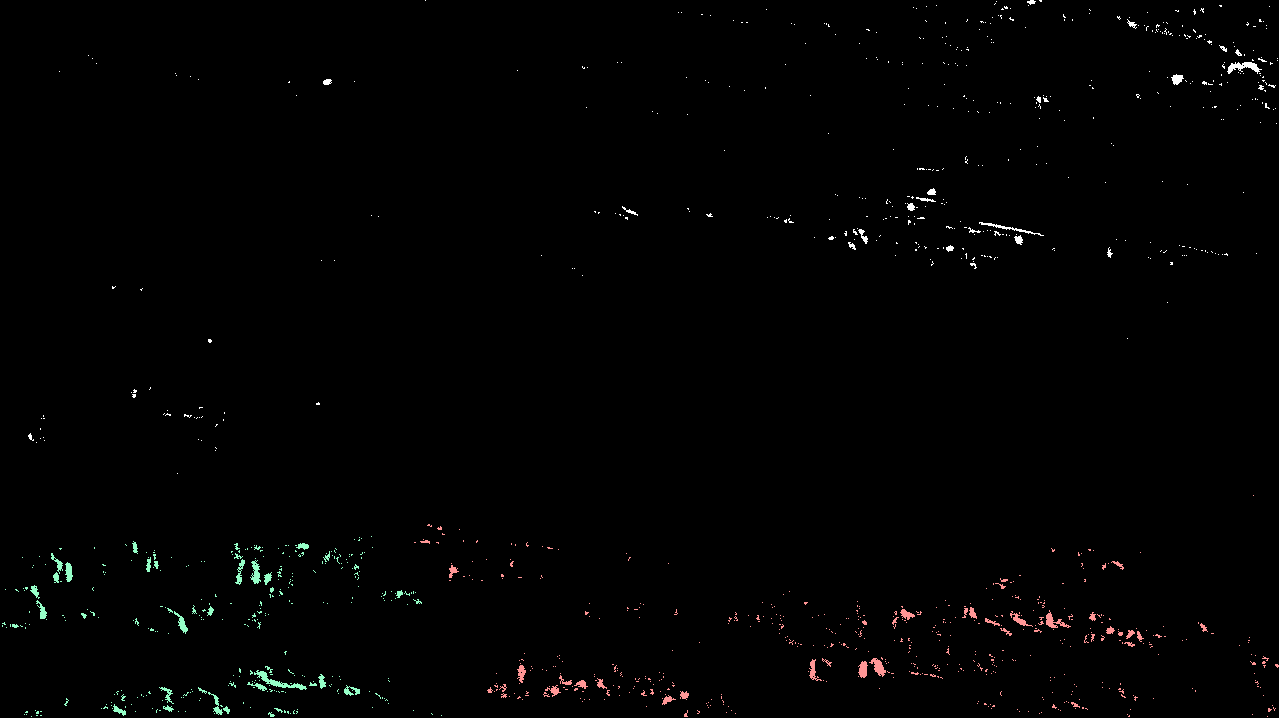} \\
  
    \rotatebox{90}{\hspace{0.4cm}\color{gray!90}Ours} & \includegraphics[height=0.56in,width=0.67in]{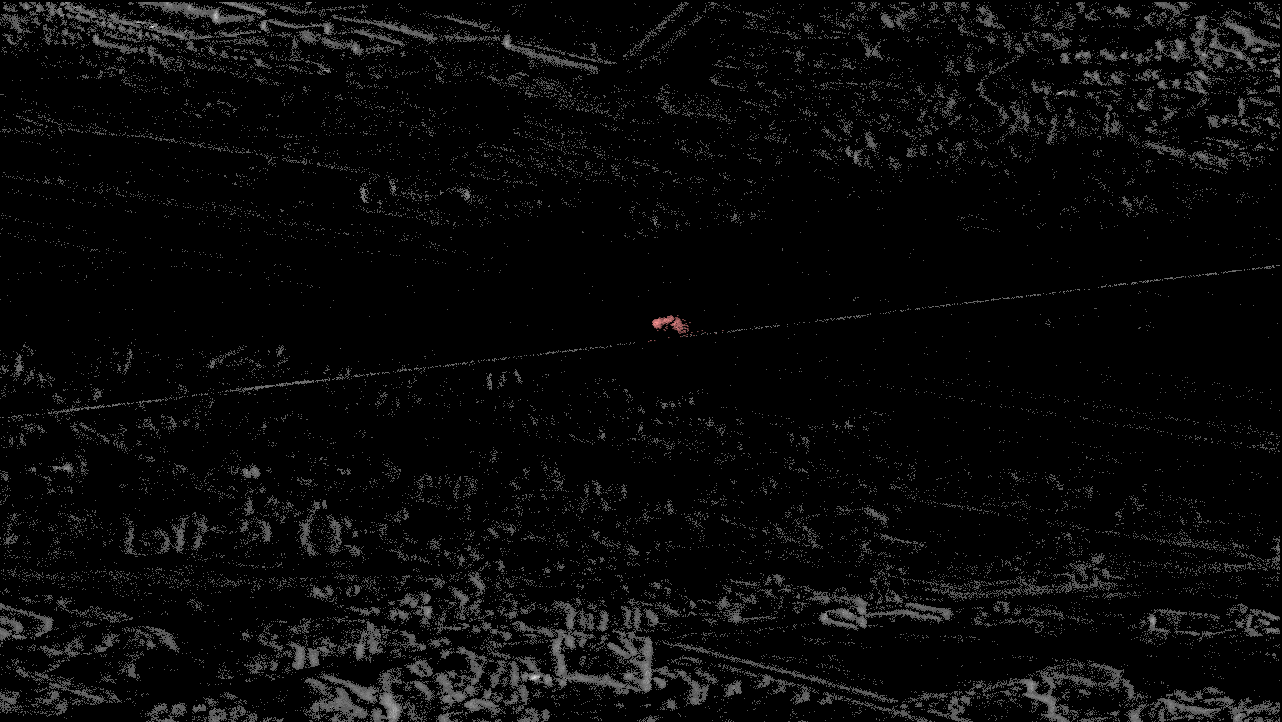}
    & \includegraphics[height=0.56in,width=0.67in]{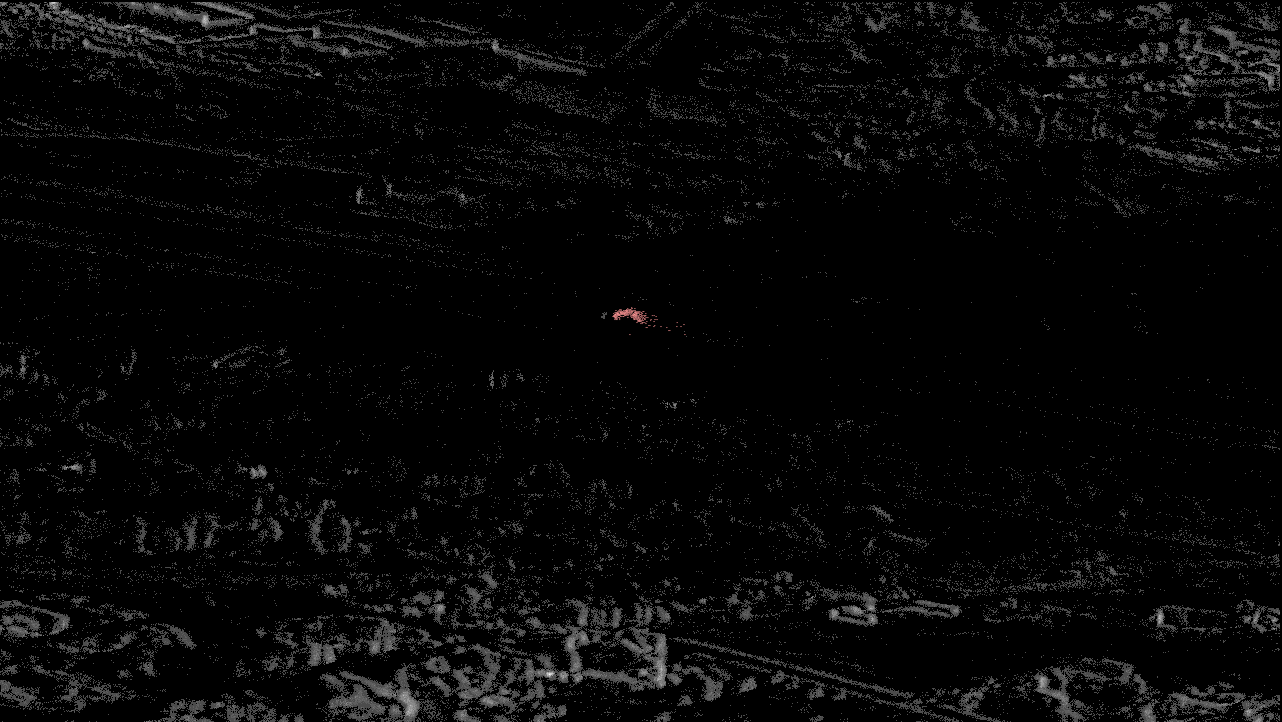}
    & \includegraphics[height=0.56in,width=0.67in]{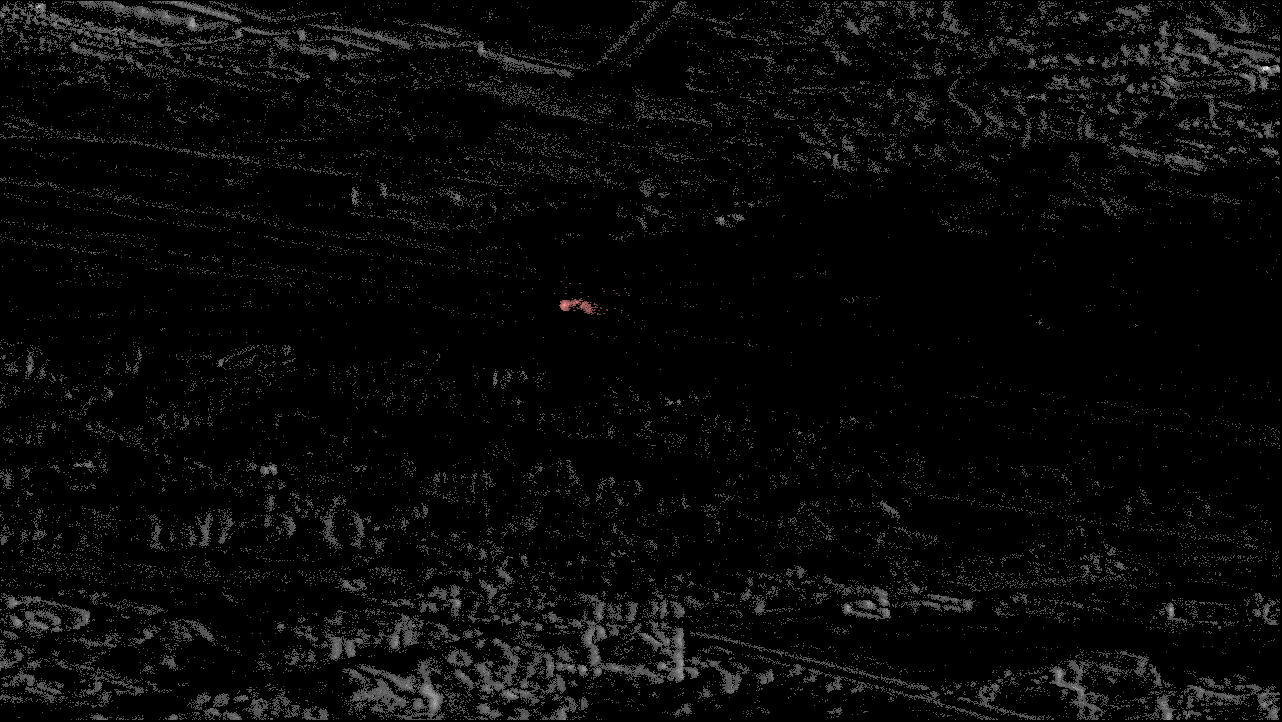}
    & \includegraphics[height=0.56in,width=0.67in]{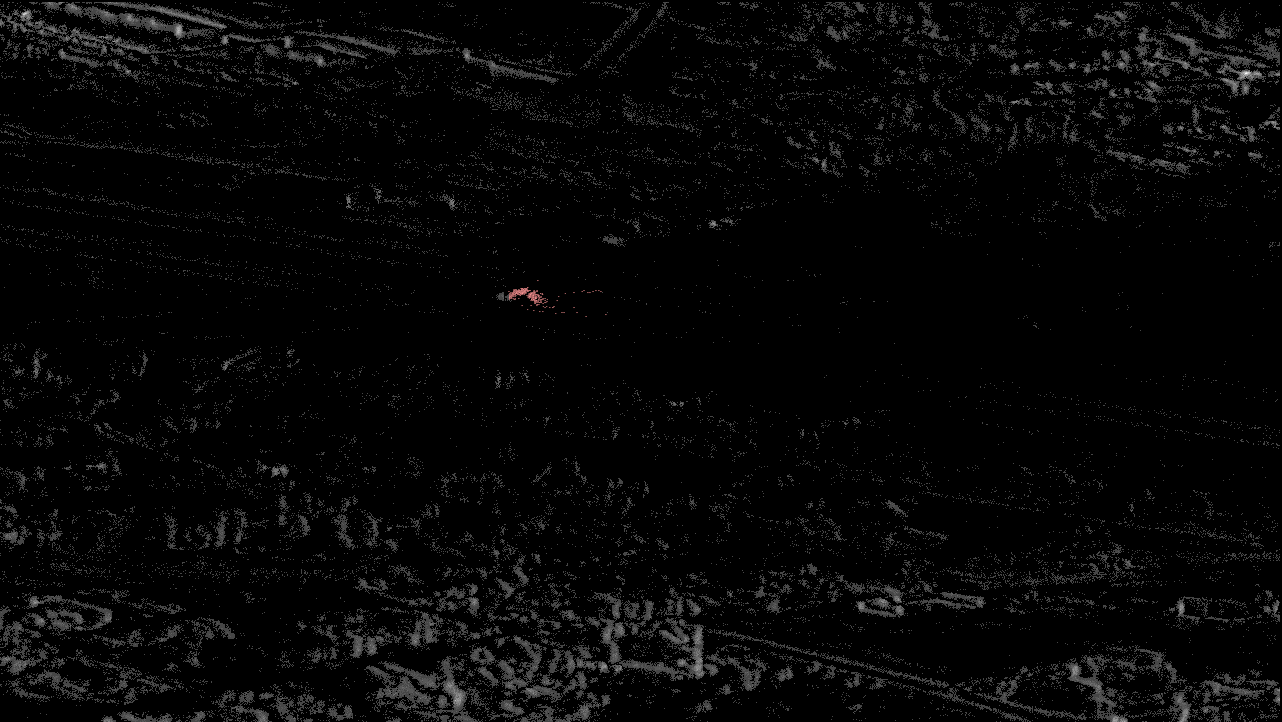}
    & \includegraphics[height=0.56in,width=0.67in]{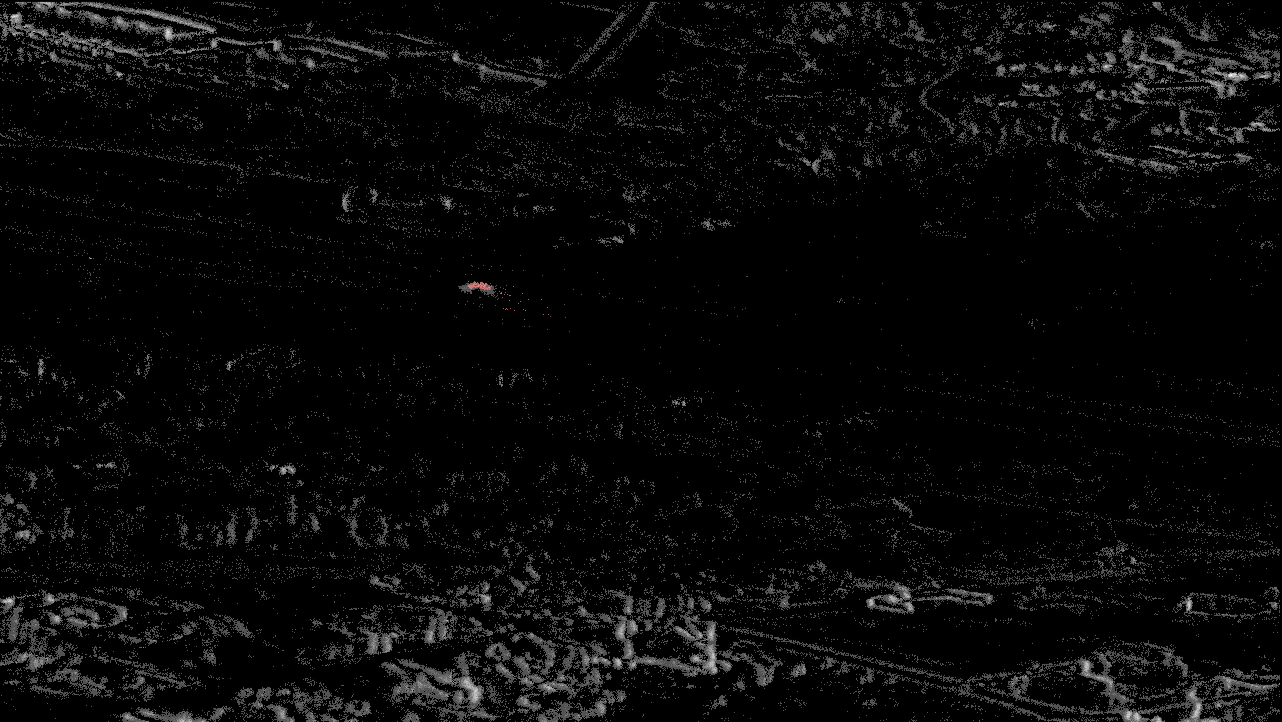} & 
    \includegraphics[height=0.56in,width=0.67in]{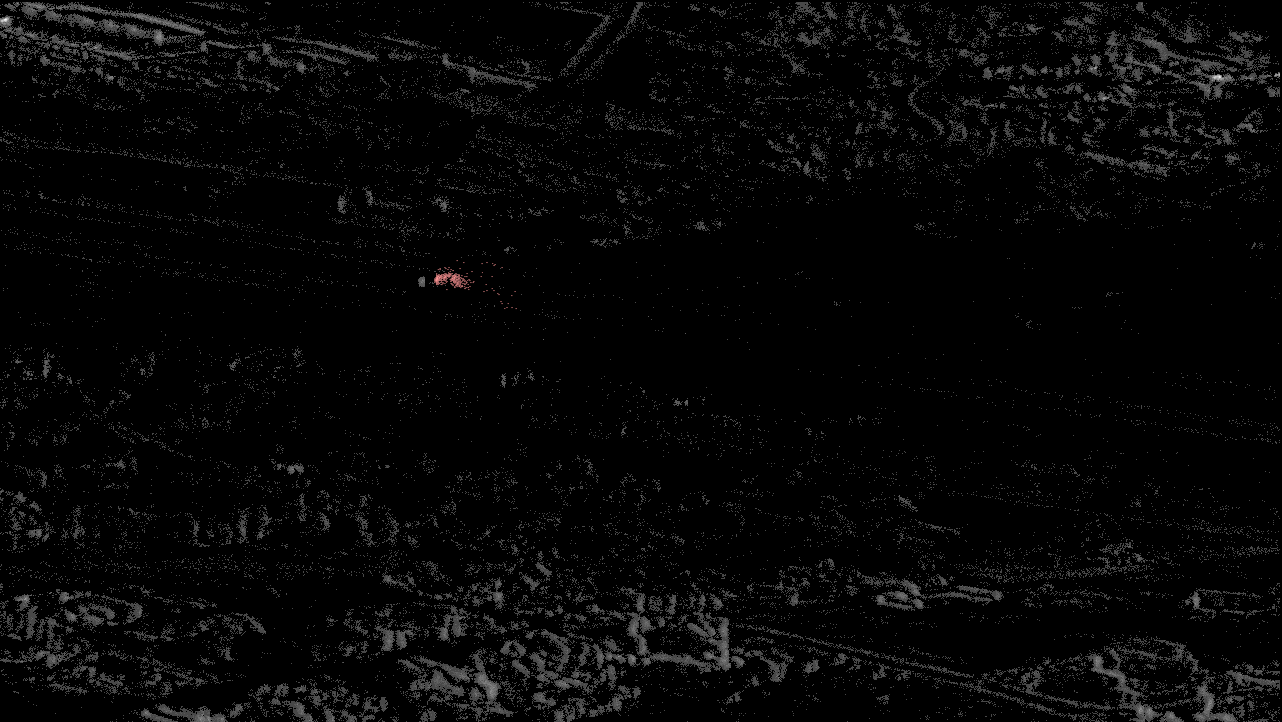} &
    \includegraphics[height=0.56in,width=0.67in]{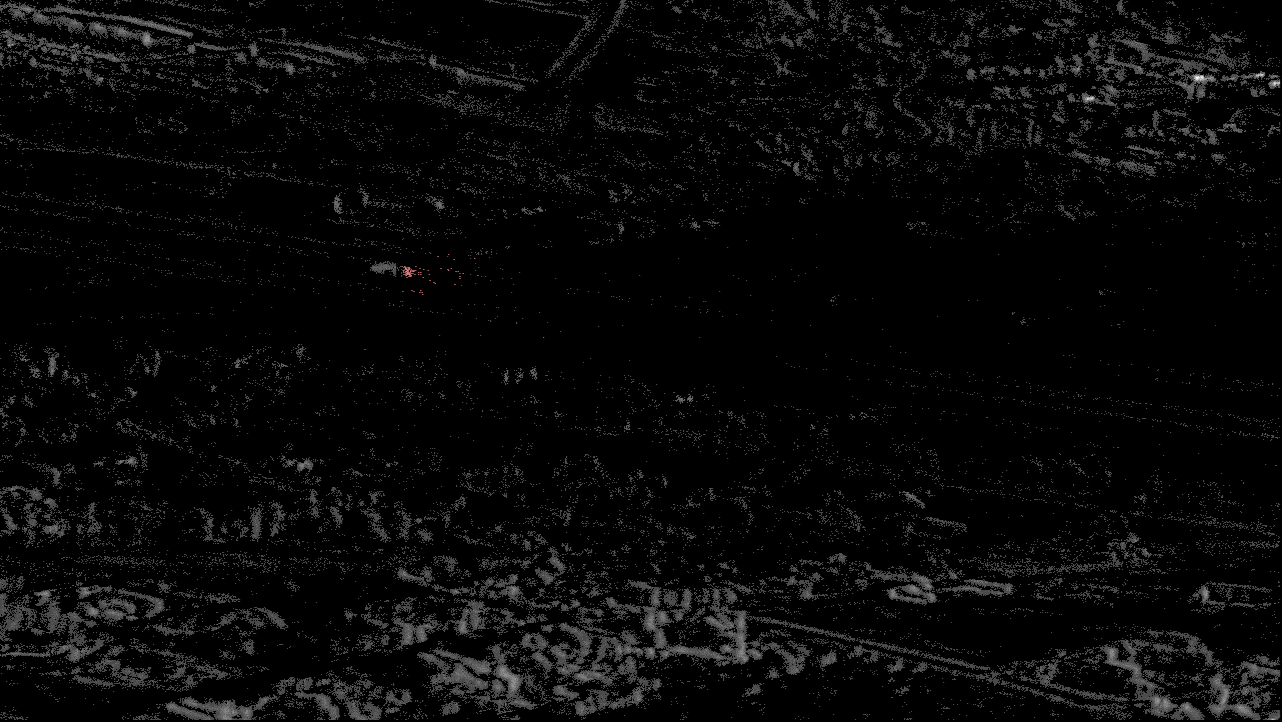} \\

\end{tabular}
\textbf{Segmentation Consistency}
\begin{tabular}{c c c c c c c c c}
  \tiny\rotatebox{90}{\hspace{0.1cm}\color{gray!90}EMSGC\cite{zhou_event-based_2021}} & \includegraphics[height=0.56in,width=0.67in]{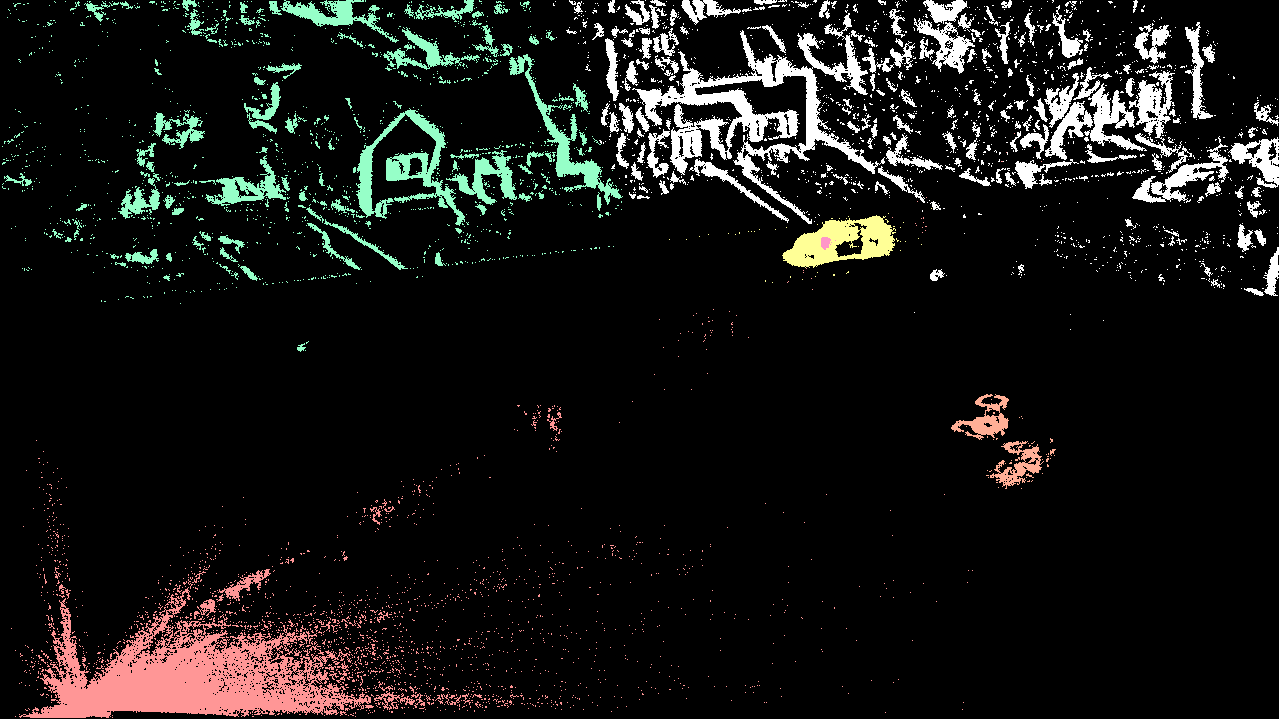}
    & \includegraphics[height=0.56in,width=0.67in]{Figures/0000_continuity.png}
    & \includegraphics[height=0.56in,width=0.67in]{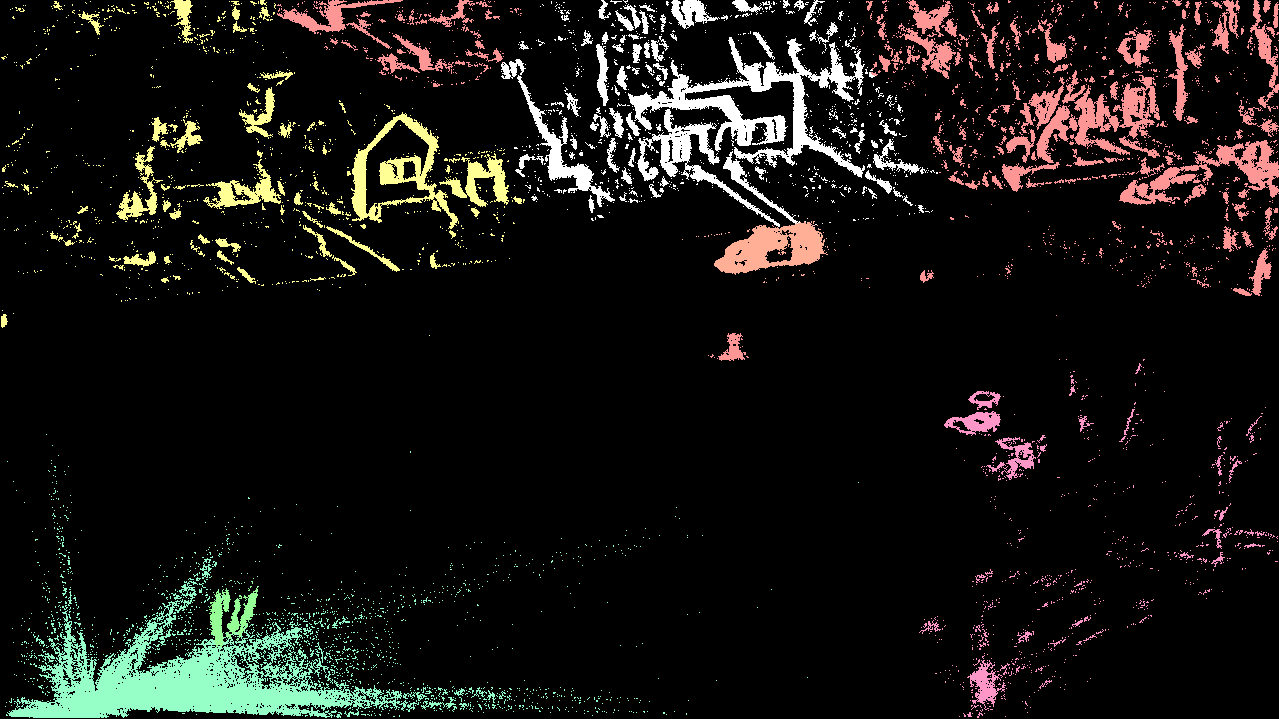}
    & \includegraphics[height=0.56in,width=0.67in]{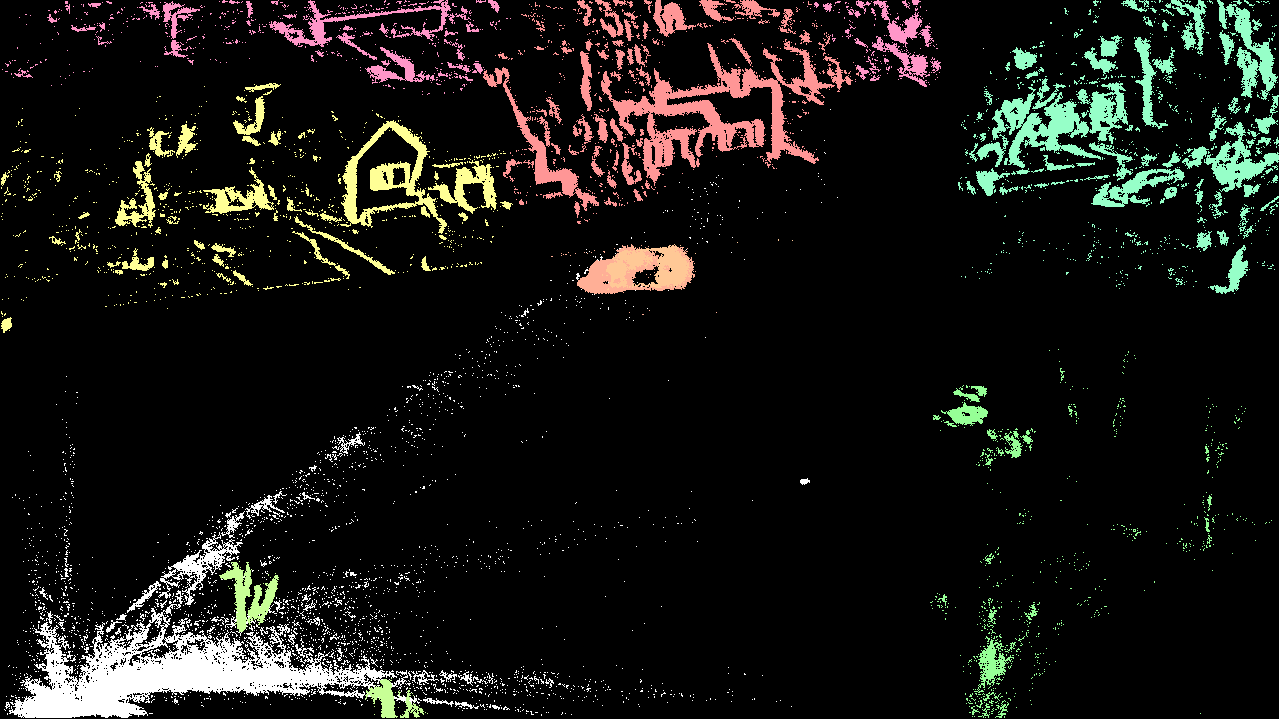}
    & \includegraphics[height=0.56in,width=0.67in]{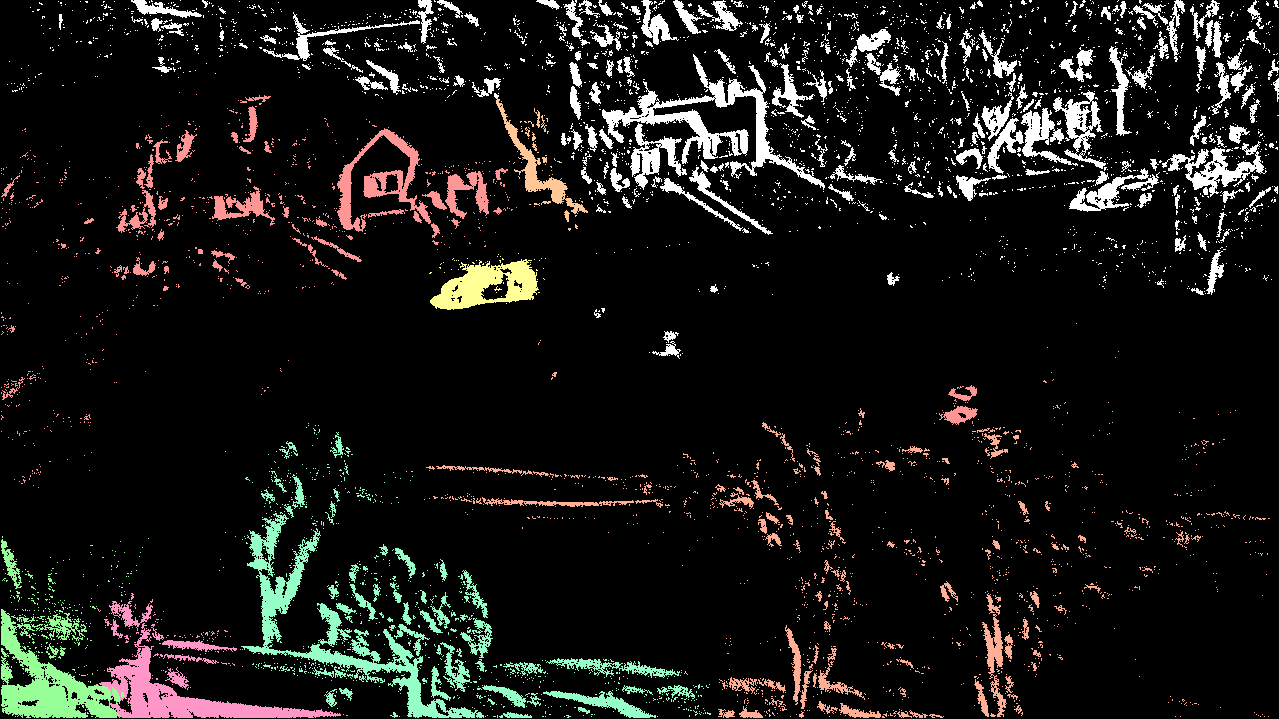} & 
    \includegraphics[height=0.56in,width=0.67in]{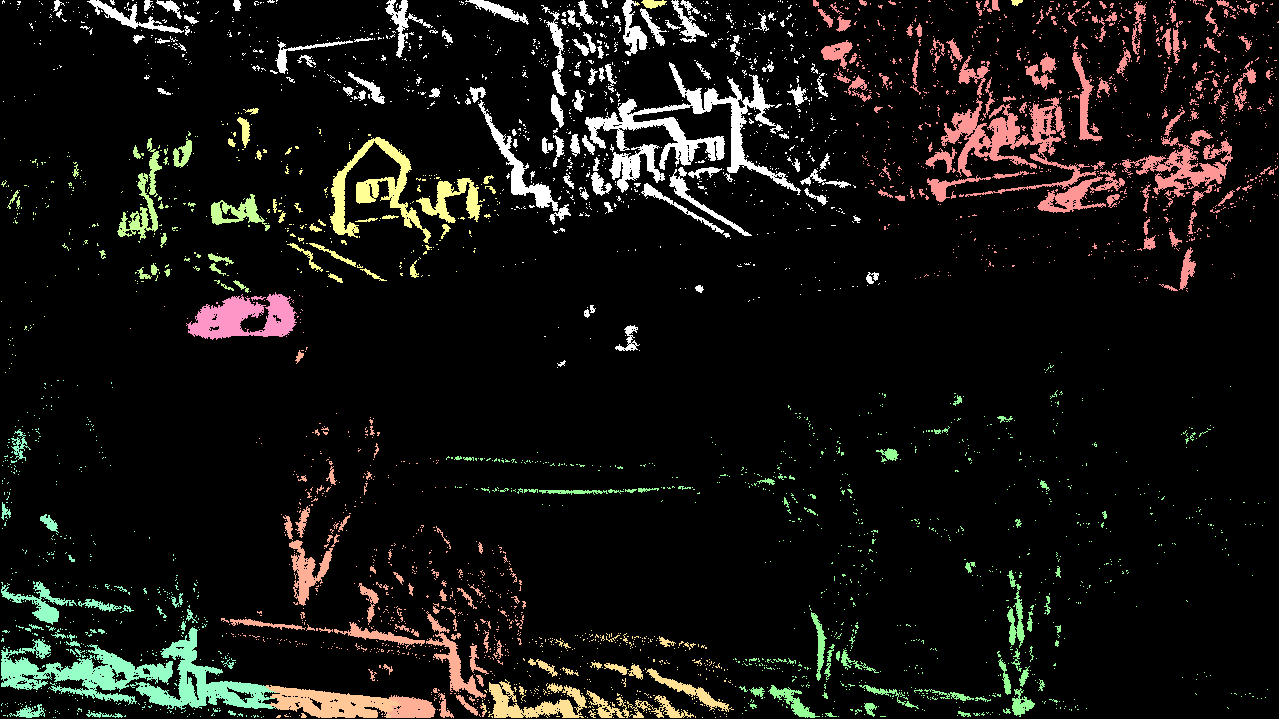} &
    \includegraphics[height=0.56in,width=0.67in]{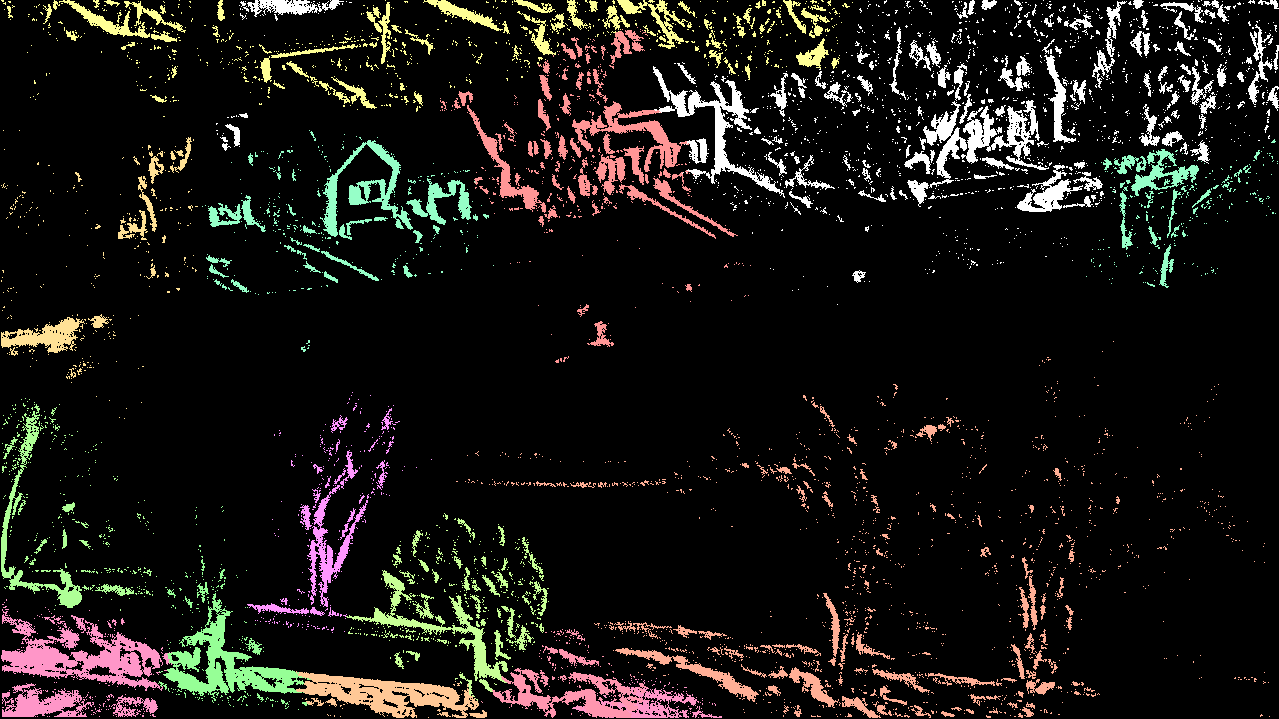} \\
    
    \rotatebox{90}{\hspace{0.4cm}\color{gray!90}Ours} & \includegraphics[height=0.56in,width=0.67in]{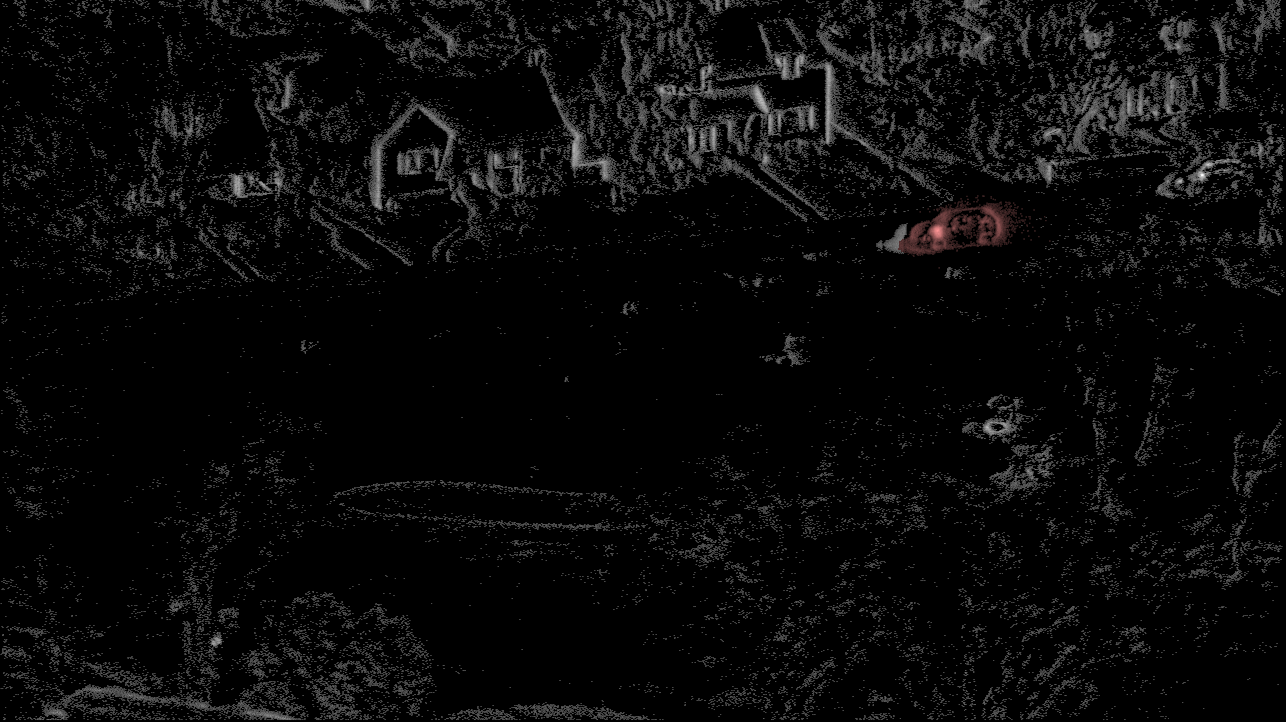}
    & \includegraphics[height=0.56in,width=0.67in]{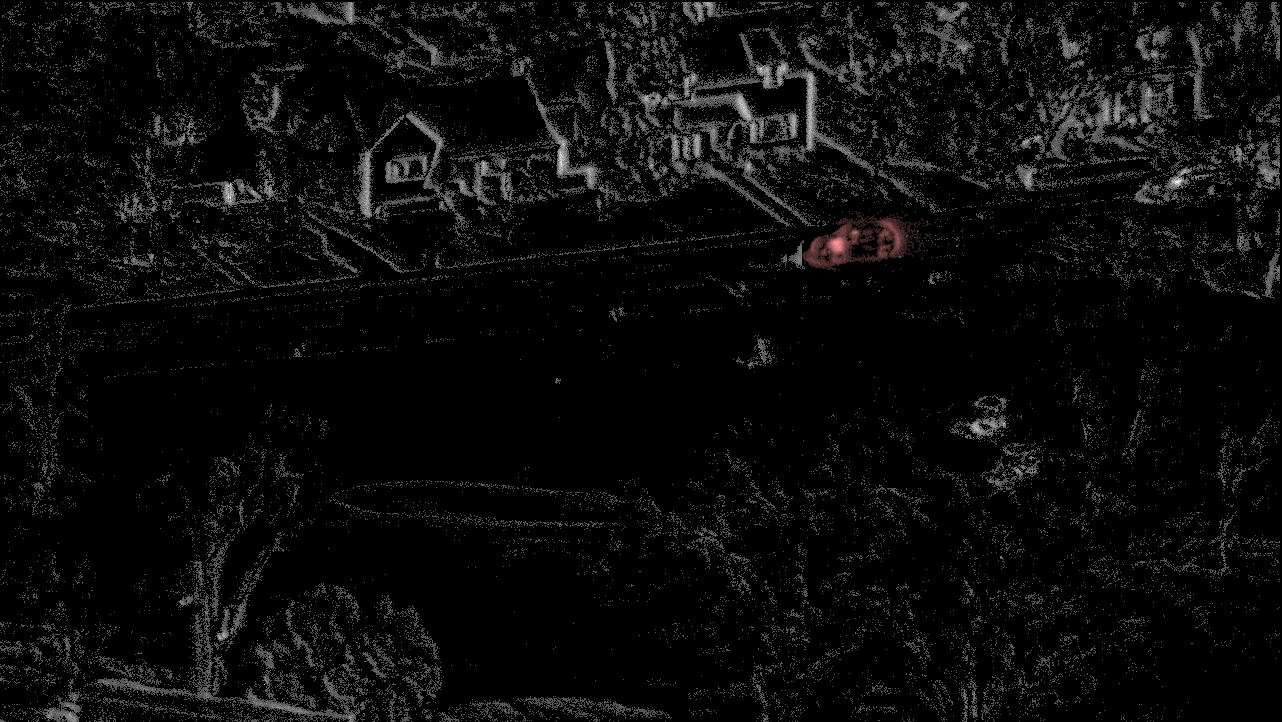}
    & \includegraphics[height=0.56in,width=0.67in]{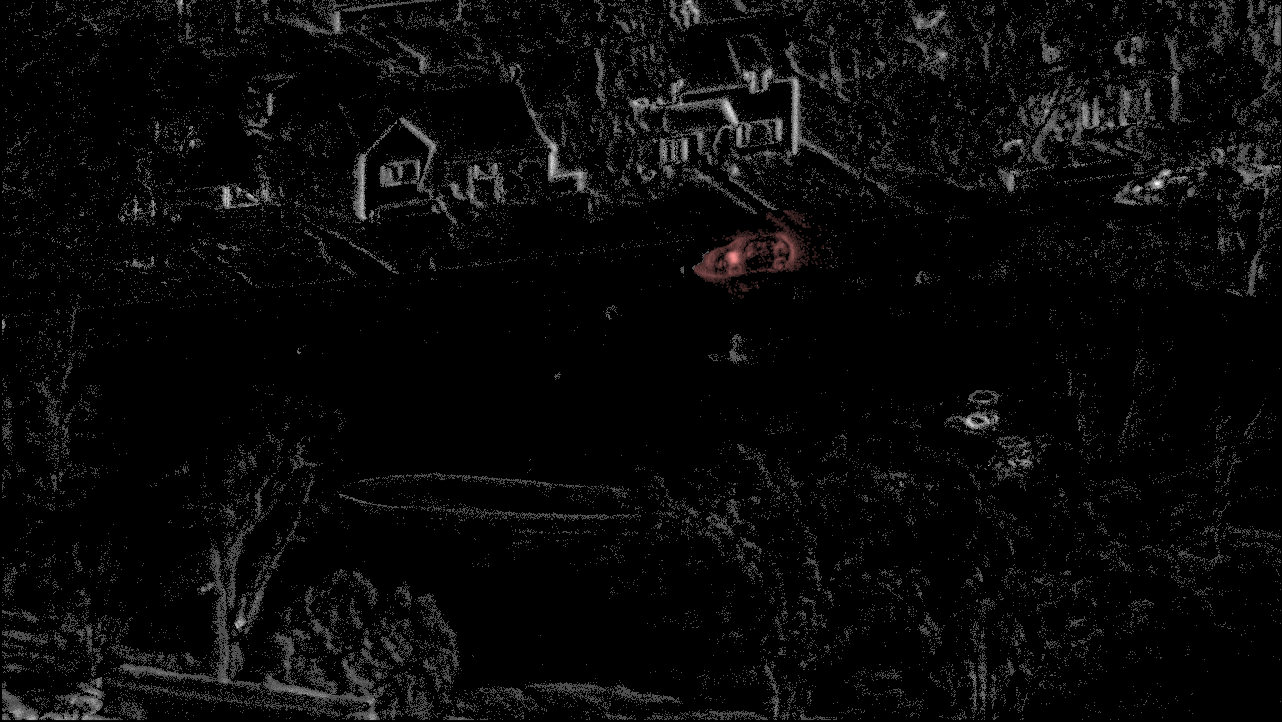}
    & \includegraphics[height=0.56in,width=0.67in]{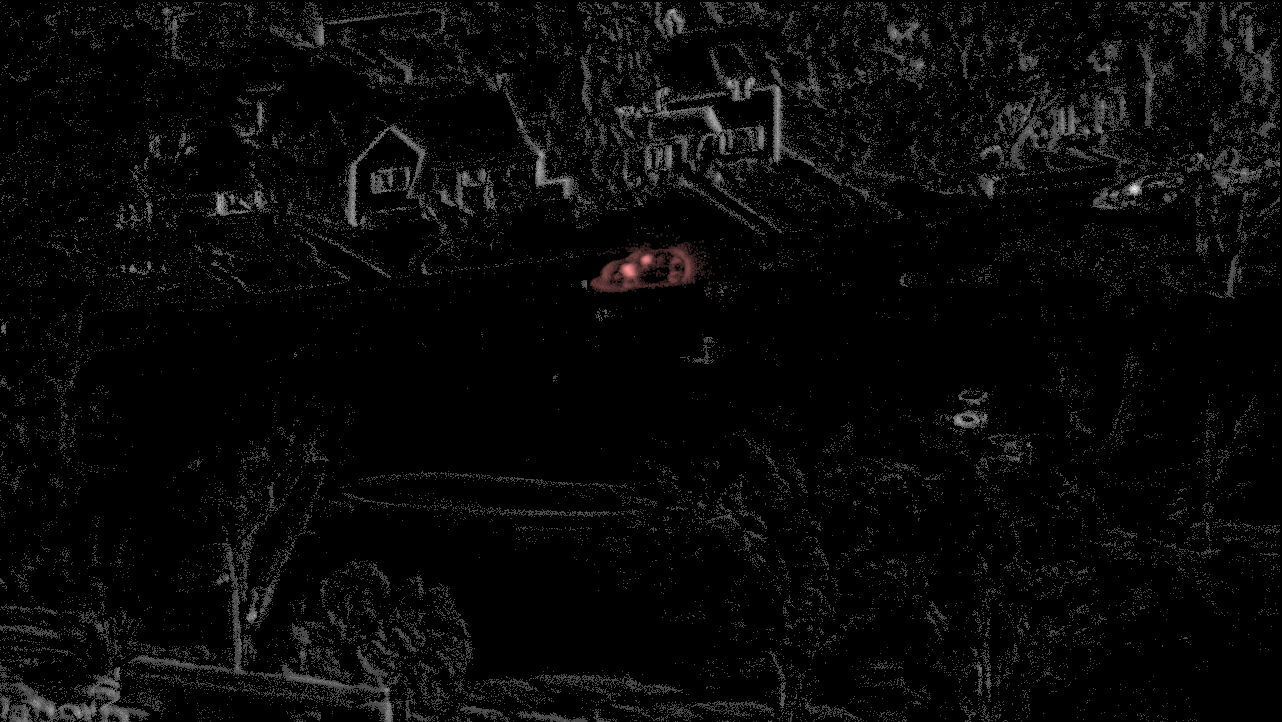}
    & \includegraphics[height=0.56in,width=0.67in]{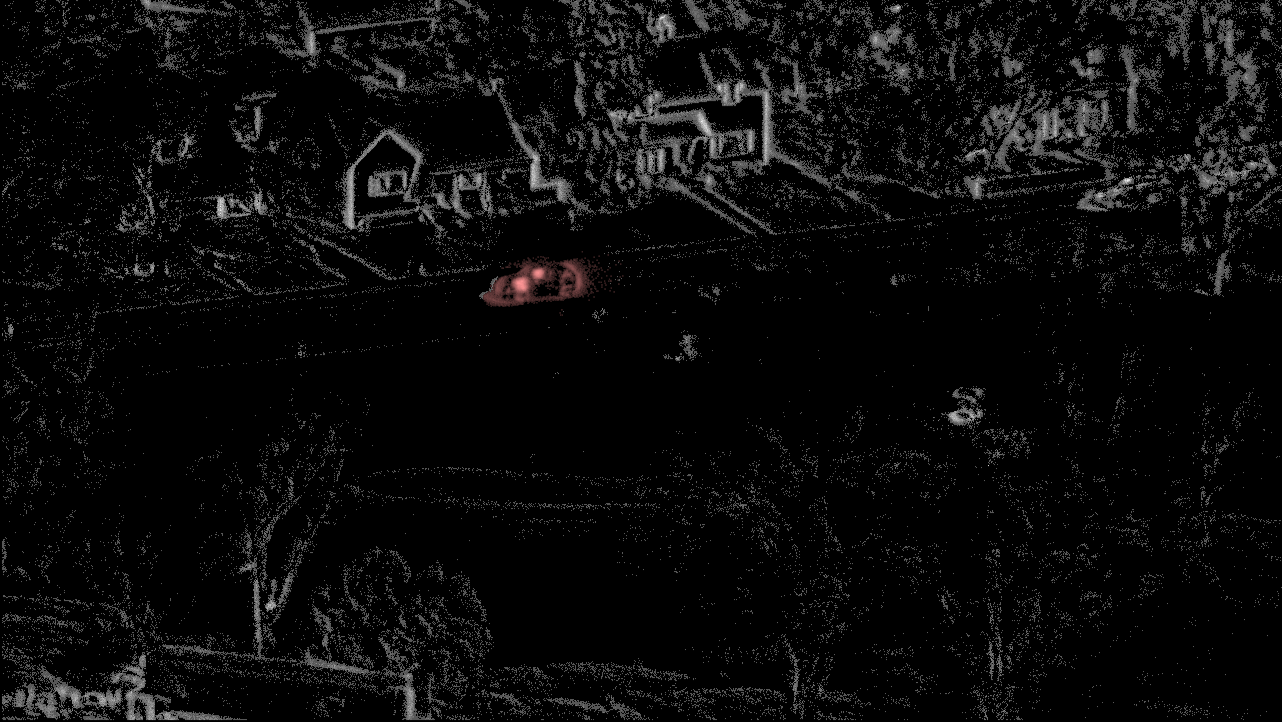} &
    \includegraphics[height=0.56in,width=0.67in]{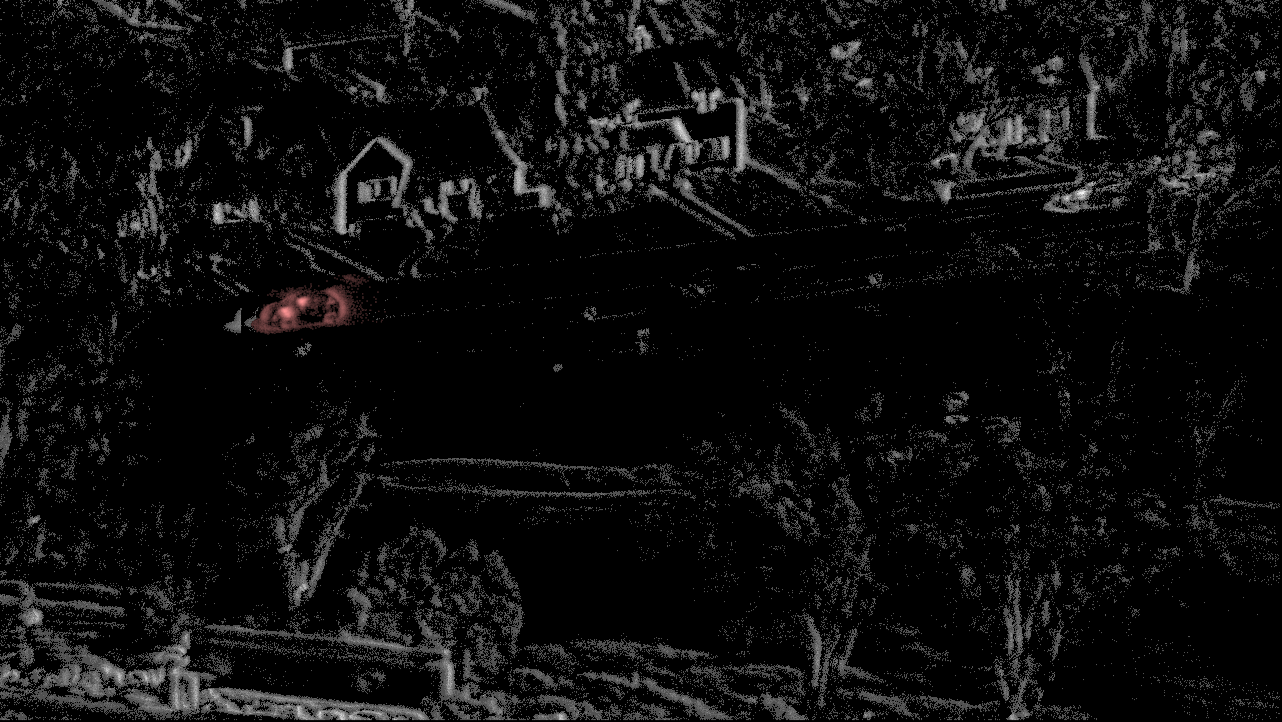}  &
    \includegraphics[height=0.56in,width=0.67in]{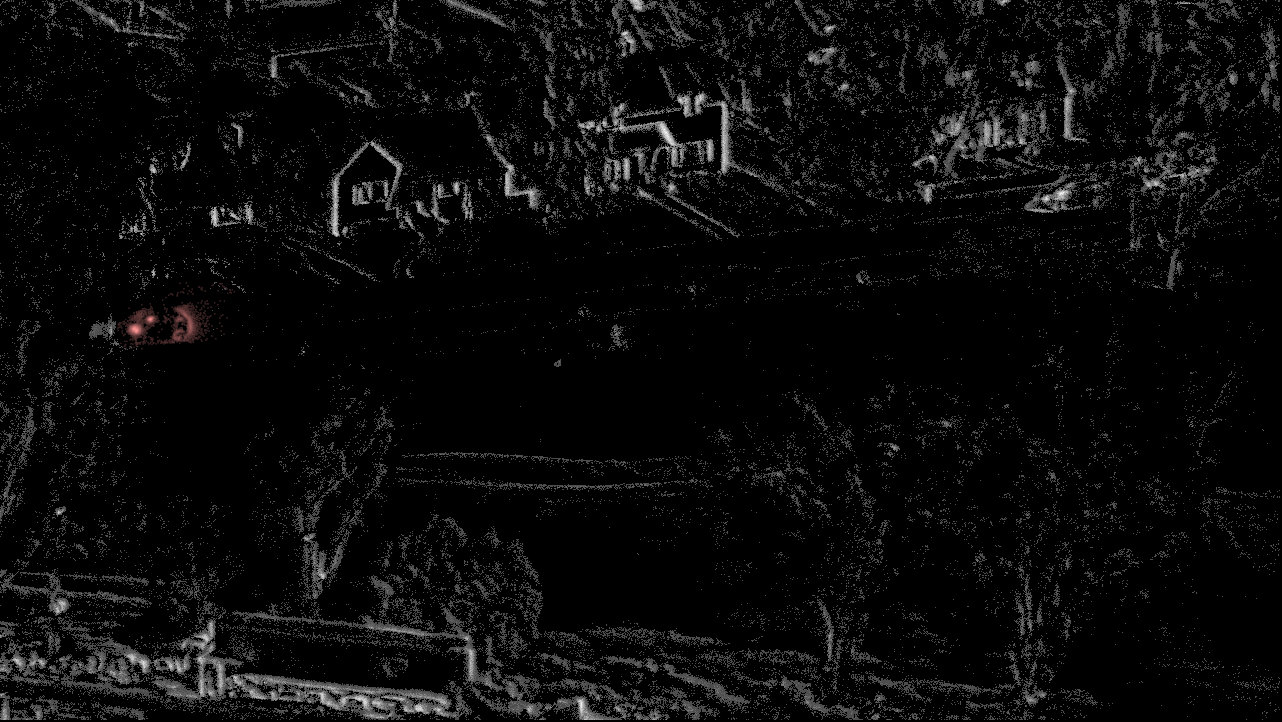} \\
\end{tabular}
\caption{Illustrating the effectiveness of our model in addressing various challenges in motion segmentation. The top panels showcase our model's ability to overcome the issue of oversegmentation, presenting a clear comparison with the baseline method EMSGC~\cite{zhou_event-based_2021}. The middle panels showcase the capability of our model in segmenting small IMOs within dynamic environments, demonstrating its superiority over the baseline. The bottom panels showcase the model's capability to maintain segmentation consistency over time, illustrating its continuous performance and reliability.}
\label{fig:motion_segmentation_problem_solution}
\end{figure}

This section showcases the effectiveness of our proposed method across different scenarios, emphasizing its ability to handle the challenges in motion segmentation on aerial platforms. We specifically highlight how our approach addresses the issue of oversegmentation, efficiently segments small IMOs within dynamic settings, and ensures consistent segmentation over time. Detailed visual representation of these capabilities and comparisons are presented in Fig.~\ref{fig:motion_segmentation_problem_solution}.

\textbf{Oversegmentation}. As shown in the top panels in Fig.~\ref{fig:motion_segmentation_problem_solution}, the segmentation output of our method accurately describes the number of moving objects and the background. In contrast, EMSGC~\cite{zhou_event-based_2021} leads to an excessive number of segmentation labels being incorrectly assigned to the background, often as a result of the parallax effect within the scene. The tendency towards oversegmentation in EMSGC is influenced by its use of a subvolume division parameter that segments the image into uniform subvolumes. A smaller subvolume parameter tends to induce oversegmentation, whereas a larger parameter may cause undersegmentation. Unlike EMSGC, our method effectively avoids both oversegmentation and undersegmentation by distinctly separating and then estimating the motions of background and foreground.

\textbf{Dealing with small IMOs}. As shown in the middle panels in Fig.~\ref{fig:motion_segmentation_problem_solution}, our method was able to estimate the motion of the small airplane during its takeoff while the whole is moving (due to the airplane consistent motion). This was possible due to the small difference between the moving object in the scene and the camera ego-motion which was detected by RAFT~\cite{teed2020raft}. EMSGC~\cite{zhou_event-based_2021} struggles to detect the motion of the airplane taking off even with a small subvolume division parameter.

\textbf{Segmentation consistency}. As shown in the bottom panels in Fig.~\ref{fig:motion_segmentation_problem_solution}, our method demonstrates the superior segmentation consistency across an entire sequence. Our approach ensures that the segmentation of moving objects remains stable and accurate throughout the scene. In comparison, while the EMSGC~\cite{zhou_event-based_2021} manages to recognize and label a moving car, it struggles with maintaining consistent segmentation over time. This inconsistency is particularly evident in the changing segmentation colour of the moving car, highlighting a lack of stability in tracking the object's motion. The consistency observed in our method is attributed to the DMR algorithm, which effectively preserves the object's mask shape across different frames, ensuring reliable and uniform segmentation results.

\bibliographystyle{splncs04}
\bibliography{egbib}